\documentclass[twoside,11pt]{article}

\usepackage{blindtext}
\usepackage{cancel}

\usepackage[preprint]{jmlr2e_without_decorations}

\PassOptionsToPackage{prologue,dvipsnames}{xcolor}

\usepackage{amsmath,yhmath,graphicx,bm,bbm,natbib,tikz}
\usetikzlibrary{calc}
\usepackage[colorlinks=true, allcolors=blue]{hyperref}
\usepackage[all]{xy}
\usepackage{algorithm}
\usepackage{algpseudocode}
\usepackage{mathtools}
\usepackage[dvipsnames]{xcolor}
\usetikzlibrary{positioning,matrix,backgrounds,arrows,automata}
\usepackage{rotating}
\usepackage{makecell}

\allowdisplaybreaks % allow equation page break

\usepackage{enumitem}
\newcommand{\R}{\mathbb{R}}   % reals
\newcommand{\Rp}{\mathbb{R}^+}   % positive reals
\newcommand{\Rnn}{\mathbb{R}^{\ge 0}}   % non-negative reals
\renewcommand{\b}{\mathbf}    % bold maths
\newcommand{\tb}{\textbf}     % bold text
\newcommand{\T}{\top}         % transpose
\newcommand{\X}{\mathcal{X}}  % domain of attributes
\newcommand{\Y}{\mathcal{Y}} % domain of attributes
\renewcommand{\t}{\text} % text
\renewcommand{\P}{\mathbb{P}} % probability 
\newcommand{\Q}{\mathbb{Q}} 
\newcommand{\J}{\mathbb{J}} 
\newcommand{\E}{\mathbb{E}} 

\newcommand{\mH}{\mathcal{H}} % RKHS
\renewcommand{\d}{\mathrm{d}}%dx

\graphicspath{{figs/}}

\newcommand{\Span}{\mathrm{span}} % linear hull
\newcommand{\Z}{\mathbb{Z}} % integers
\newcommand{\Zp}{\mathbb{Z}^+} % positive integers
\newcommand{\diag}{\mathrm{diag}} % diagonal matrix

\renewcommand{\L}{\mathcal{L}} % bounded lin operators
\renewcommand{\O}{\mathcal{O}} % O(rdo)
\newcommand{\eqd}{\stackrel{\t{distr}}{=}} % equality in distr (NEW)

\newcommand{\mN}{\mathcal{N}} %normal random variables

\newtheorem{assumption}{Assumption}

\setenumerate{labelindent=0em,leftmargin=1.6em,topsep=0cm,partopsep=0cm,parsep=0cm,itemsep=1mm}
\setitemize{labelindent=0em,leftmargin=1.3em,topsep=0cm,partopsep=0cm,parsep=0cm,itemsep=1mm}

\DeclareMathOperator{\tr}{tr}
\DeclareMathOperator{\sign}{sign}
\DeclareMathOperator{\rank}{rank}
\DeclareMathOperator{\Image}{Im}
\DeclareMathOperator{\Kernel}{Ker}
\DeclareMathOperator{\Dim}{dim}

\newcommand{\multiline}[1]{%
    \begin{tabularx}{\dimexpr\linewidth-\ALG@thistlm}[t]{@{}X@{}}
        #1
    \end{tabularx}
}

\makeatletter
\renewcommand*{\@opargbegintheorem}[3]{\trivlist
      \item[\hskip \labelsep{\bfseries #1\ #2}] \textbf{(#3)}\ \itshape}
\makeatother

\mathtoolsset{showonlyrefs}  

\hypersetup{hidelinks}

\newcommand{\colAonei}{\textcolor{CadetBlue}}
\newcommand{\colAoneii}{\textcolor{NavyBlue}}
\newcommand{\colAi}{\textcolor{teal}}
\newcommand{\colAii}{\textcolor{magenta}}
\newcommand{\colLfive}{\textcolor{violet}}
\newcommand{\colLsix}{\textcolor{orange}}
\newcommand{\coleone}{\textcolor{BrickRed}}
\newcommand{\coletwo}{\textcolor{LimeGreen}}

\usepackage{lastpage}
\jmlrheading{X}{Y}{1-\pageref{LastPage}}{10/26; Revised X/Y}{X/Y}{21-0000}{Xiaoyi Wen and Zolt\'{a}n Szab\'{o}}

\ShortHeadings{Joint Independence Testing of LPRGs}{Joint Independence Testing of LPRGs}
\firstpageno{1}

\begin{document}

\title{Joint Independence Testing for Latent Position Random Graphs}

\author{\name Xiaoyi Wen\thanks{Part of the work was carried out while XW was a PhD student at Renmin University of China.}\email xiaoyi.wen@ucsf.edu \\
       \addr Department of Epidemiology and Biostatistics \\
        University of California, San Francisco\\
       San Francisco, USA
       \AND
       \name Zolt\'{a}n Szab\'{o}  \email z.szabo@lse.ac.uk \\
       \addr Department of Statistics\\
       London School of Economics\\
       London, UK}

%\editor{My editor}
%\editor{}

\maketitle

\begin{abstract}%  

Kernel techniques are among the most powerful methods in data science and statistics. One of the key properties of kernels contributing to their numerous successful applications is that the associated reproducing kernel Hilbert space is expressive enough to encode probability distributions without loss of information or to capture the joint independence of $M\ge 2$ random variables, under mild conditions. Computationally, these measures (i) rely on the so-called Gram matrix which contains the pairwise similarity of sample points, and (ii) converge at a rate $\O\left(n^{-\frac{1}{2}}\right)$, where $n$ denotes the sample size.  We focus on measuring the independence of latent position random graphs (LPRGs), a model widely-applied for instance in neuroscience, sociology, and epidemiology. In LPRGs a latent representation is associated to each node. The similarity of these representations (meant via a kernel) determines whether two nodes are connected, making the Gram matrices only \emph{implicitly} observable and the estimation challenging. Despite the fundamental importance of quantifying the  independence of $M$ LPRGs, very little is known about the solution of this problem; existing techniques are restricted to $M=2$ components and only guarantee consistency without rates. In this work, (i) we define an independence measure called graph Hilbert-Schmidt independence criterion (GHSIC) for $M\ge 2$ LPRGs, (ii) we prove that it can be estimated at a rate $\tilde{\O}\left(n^{-\frac{1}{3}}\right)$ relying on spectral decomposition of adjacency matrices, (iii) without the need to choose kernels and their hyperparameters. Numerical experiments demonstrate the effectiveness of the proposed GHSIC estimator both on synthetic examples and in the analysis of brain functional connectivity subnetworks in natural language perception.

\end{abstract}

\begin{keywords}
 reproducing kernel Hilbert space, Hilbert-Schmidt independence criterion, independence testing, latent position random graph, adjacency spectral embedding
\end{keywords}

\section{Introduction}

Kernel methods \citep{wahba90spline,berlinet04reproducing,steinwart08support,saitoh16theory} are omnipresent in data science and statistics with a large number of successful applications; this popularity stems from their multiple favorable properties. (i) At the core of these techniques is the so-called kernel function, which encodes the similarity of points in the domain of interest, via mapping them to a (potentially infinite-dimensional) Hilbert space and computing the inner product of these featurized points (often) implicitly. Kernels have been designed for various data types, including sequences \citep{kiraly19kernel, chevyrev22signature, toth25random}, sets \citep{gartner02multi,buathong10kernels}, probability distributions \citep{berlinet04reproducing, sriperumbudur10hilbert, marienwald21high}, and graphs \citep{kondor16multiscale, schulz22graph, nikolentzos23graph, cosmo2024graph}. (ii) Kernels give rise to a reproducing kernel Hilbert space (RKHS; \citealt{aronszajn50theory}), which thanks to its Hilbert structure facilitates the design of theoretically-grounded approaches  \citep{cortes95support, vapnik00nature, caponnetto07optimal,  rakotomanonjy08simplemkl, zhou08derivative, steinwart09optimal, vishwanathan10graph, kadri12multiple,sejdinovic2013equivalence, rudi2015less, sangnier16joint, kadri16operator,brouard16input, jitkrittum17linear, rudi17falkon, sriperumbudur17density, gilles18optimal, oglic19scalable, fang20optimal,sterge2020gain, balasubramanian21optimality, mucke21stochastic, korba21kernel, muandet21counterfactual, vankadara21recovery, huang22kernel, kohler22total, chen22online, aubin22handling, chatalic22nystrom, kanagawa23kernel, schrab23mmd, fernandez2023kernel, haddouche2024generalisation, alquier24universal, barp24targeted, yu24distributed, wacker24improved, ziegel24characteristic, ahmad24sketch, kanagawa25controlling, he25regularized, rudi25finding, steinert25integration, xiong25support, chen25deregularized, martinez25sequential, wolfer25variance, lin2025adaptive, matabuena26kernel, bitzer26spectral, chattejee26martingale, hagrass26minimax, hashimoto26spectral} leading to principled learning algorithms. (iii) RKHSs are expressive under mild conditions; for instance, characteristic kernels \citep{fukumizu08kernel, sriperumbudur10hilbert} encode probability measures without loss of information, $\mathcal I$-characteristic kernels \citep{szabo2017characteristic}  capture the independence of $M\ge 2$ random variables on kernel-equipped domains, universal kernels \citep{steinwart2002influence, micchelli06universal} lead to RKHSs dense in the space of continuous bounded functions on compact metric spaces. (iv) The impressive modeling power of RKHSs is accompanied by computational tractability for objectives which can be expressed based on values of functions \citep{steinwart08support, yu13characterizing} and function derivatives \citep{zhou08derivative}; in this case, it is sufficient to solve a finite-dimensional optimization problem.

Kernel mean embedding \citep{berlinet04reproducing, smola07hilbert} is a popular approach based on the flexibility of kernels, mapping Borel probability measures into an RKHS in a lossless fashion for characteristic kernels. Taking the RKHS distance of the resulting embeddings of distributions gives rise to maximum mean discrepancy (MMD; \citealt{gretton12kernel}) which is 
(i) a metric for characteristic kernels, (ii) known to be equivalent \citep{sejdinovic2013equivalence} to energy distance \citep{baringhaus04new, szekely04testing, szekely05new}---also referred to as $N$-distance \citep{zinger92characterization, klebanov05ndistance}---, (iii) a specific integral probability metric \citep{zolotarev83probability, muller97integral}. Applying MMD on product spaces to compare the joint distribution with the product of its marginals yields the Hilbert-Schmidt independence criterion (HSIC). HSIC was initially proposed for $M = 2$ components \citep{gretton05measuring, gretton05kernel} and was subsequently generalized to the case of $M \ge 2$ \citep{quadrianto2008kernelized, sejdinovic13kernel, pfister18kernel}. For $M = 2$, HSIC captures independence when the kernel components underlying HSIC are characteristic \citep{lyons13distance}; for $M \ge 2$, $c_0$-universality \citep{carmeli10vector, sriperumbudur10hilbert} of the respective kernel components is sufficient \citep{szabo2017characteristic}. HSIC has been leveraged in numerous applications including independence testing \citep{gretton2007kernel, chwialkowski14kernel, pfister18kernel, zhou25dual}, feature selection \citep{song12feature, li12feature}, representation learning \citep{li2021self, miklautz25hsplid}, domain adaptation \citep{greenfeld2020robust, huang2020domain}, and causal discovery \citep{mooij16distinguishing, kalinke2023nystr, zhang2023differentially}. The flexibility and computational tractability of HSIC makes it an ideal candidate to be used in the context of independence testing of random graphs, which is the focus of the current paper.

To formulate independence testing for random graphs, it is instrumental to recall the definitions of graphs and random graphs. A graph $G=(V,E)$ consists of a set of nodes $V$ and a set of edges $E$ encoding pairwise relationships between nodes \citep{wilson1979introduction}. For a graph with $n$ nodes ($|V|=n$; in the current manuscript, we consider graphs with finite many nodes), the edge structure (the presence/absence of edges) is commonly represented by an adjacency matrix, which for unweighted graphs (which is the focus of our paper) takes the form $\b A=\left[ A_{i,j}\right]_{i,j=1}^n \in \{0,1\}^{n \times n}$, where the order of the nodes is assumed to be fixed. Specifically, $A_{i,j}=1$ indicates the presence of an edge between nodes $i$ and $j$; otherwise, $A_{i,j}=0$. Intuitively, one can think of edges as pairwise interactions or similarities between nodes; for instance, in a social network the nodes are individuals, and the edges describe whether they know each other. In a random graph, these adjacency entries collected in $\b A$ are random rather than fixed, with their distributions determined by an underlying probability model \citep{janson2000random}. In many applications, an observed graph is more naturally viewed as a realization from an underlying random graph. For example, functional brain networks estimated from repeated scans of the same set of brain regions may exhibit varying  edge patterns because of biological variability and measurement noise \citep{noble2019decade}.

There are multiple applications in science which can be phrased as independency testing between two ($M=2$) random graphs and are of fundamental importance. In computational biology, statistical associations between molecular interaction networks can reveal biological mechanisms and inform applications such as drug-target prediction, adverse-effect identification, and drug-interaction analysis \citep{klipp2010biochemical, cheng2019network, loscalzo2022molecular, wang2023zerobind}. In neuroscience, independence testing can be used to assess whether two brain networks, such as structural and functional networks or networks obtained under different conditions, exhibit related connectivity patterns, thereby providing evidence of coordinated organization across brain systems \citep{arnatkeviciute2018hub, suarez2020linking, zhao2023heart, fotiadis2024structure}. Independence testing is also used in graph matching for identifying corresponding nodes across networks, with applications in social network analysis \citep{fan2012graph, duan2025multi} and computer vision \citep{leordeanu2012unsupervised, haller2022comparative}.

There exist various important examples involving $M\ge 2$ network layers with each layer representing a distinct type of interaction or modality among a common (as in multiplex networks) or related (as in multilayer networks) set of objects:\footnote{Multilayer networks (graphs) represent nodes and edges across multiple layers, where edges can connect nodes both  within and between layers. Multiplex networks are a special case in which the same set of nodes appears in every layer, with each layer representing a different type of relationship among them.} multilayer ecological networks integrating different types of species interactions \citep{pilosof2017multilayer,hervias2024structure}, multilayer transportation networks connecting subnetworks associated to different transportation modes \citep{zhou2022network,xie2024analyzing}, multiplex brain networks in which structural and functional connectivity define different relations among the same brain regions \citep{jauny2024linking} and multiplex social networks in which different layers encode distinct interpersonal relationships among the same individuals \citep{wang2019modeling}.

Although associations among  layers in multiplex networks have been widely investigated, formal methods for testing their joint independence remain comparatively limited. In applications, relationships among network layers have been characterized using measures of layer similarity or correlation, for example in multiplex transportation and social networks \citep{brodka2018quantifying} and multiplex brain networks \citep{de2016mapping,jauny2024linking}. Model-based approaches have also been developed to characterize cross-layer associations in multilayer networks \citep{krivitsky2020exponential,aguiar2024tensor}. These approaches focus on specific forms of layer similarity or shared structural organization, rather than on testing the mutual independence of the underlying random graphs. Consequently, a general framework for testing the joint independence of multiple ($M \ge 2$) random graphs is though highly motivated, vastly open.

A natural idea using HSIC with graph kernels is not applicable to the types of problems we are motivated by. Indeed, standard joint independence testing for graph-valued data assumes a sample of graph tuples $\left( G_i \right)_{i=1}^n$, where each $G_i^{m}$ is an observed graph and the graph tuple $G_i=\left( G_i^m \right)_{m=1}^M$ constitutes one observation. Graph kernels can then be used to form Gram matrices and thus HSIC can be applied. In contrast, our setting provides only a single graph tuple $G=(G^m)_{m=1}^M$ defined on a common set of $n$ nodes, so the Gram matrices cannot be formed from repeated graph-level observations. Moreover, each $G^m$ is a random graph, and we observe only one realization of its random edge structure rather than repeated realizations. This combination of a single graph realization and random connectivity within the graph makes the problem fundamentally different and more challenging compared to standard independence testing with graph-valued observations. A key step to approach this problem is to understand the random graph through its underlying edge probabilities.

A widely-used model for random graphs is the Bernoulli random graph model \citep{gilbert1959random, erdds1959random, erdHos1960evolution}, in which each edge indicator satisfies $A_{i,j} \sim \t{Bernoulli}\left(p_{i,j} \right)$ with probabilities $p_{i,j} \in [0,1]$. The classical Erd\H{o}s-R\'{e}nyi model (ER; \citealt{erdds1959random}) is the homogeneous special case with $p_{i,j}=p$ for all node pairs ($i,j\in [n]$). Such a homogeneous assumption is often too restrictive for real-world networks, where connectivity can vary substantially across node pairs \citep{barabasi1999emergence, newman2003structure, clauset2009power}. Inhomogeneous Bernoulli random graph models relax this assumption by allowing heterogeneous pair-specific probabilities $p_{i,j}$ \citep{bollobas2007phase, lu2013spectra}. This improved flexibility, however, requires some mechanism to parameterize the probabilities $p_{i,j}$ ($i,j\in [n]$).

Latent space random graph models \citep{hoff2002latent, handcock2007model, raftery2012fast} provide a natural way to parameterize the heterogeneous connection probabilities $p_{i,j}$ ($i,j\in [n]$). These models associate each node with an unobserved latent variable and two nodes are more likely to be connected if their latent features are similar. This assumption is particularly well-suited for applications including social  \citep{hoff2002latent, schweinberger2003settings} and biological networks \citep{raftery2012fast, alanis2018latent}, where characteristics driving the connectivity are often not (completely) directly observable.

\begin{figure}
    \centering
\begin{tikzpicture}[node distance=3.5cm]\tikzstyle{every state}=[thick]
\tikzstyle{accepting}=[accepting by arrow]
\node[] (q_0) {LPRG};
\node[] (q_1) [below left of=q_0] {ER};
\node[] (q_2) [below right of=q_0] {RDPG};
\path[->] (q_0) edge node [below   left, fill=white]  {$k\left(x,\tilde{x}\right):=p$} (q_1)  
                edge node [below right, fill=white] {$k\left(x,\tilde{x}\right):=x^{\top}\tilde{x}$} (q_2)
          ;
\end{tikzpicture}
\caption{Illustration of specific latent position random graph (LPRG) models. ER: Erdős-Rényi model; RDPG: random dot product graph.}
    \label{fig:LPRG_related}
\end{figure}
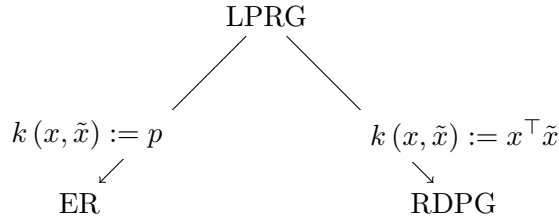

Latent position random graphs (LPRGs; \citealt{tang2013universally}) form a powerful, widely-used subclass of latent space random graph models, with applications in epidemiology \citep{loyal2020statistical}, social network analysis \citep{liu2021social, dilworth2025assessing}, and neuroscience \citep{pedigo2023generative, bridgeford2025network}. Specifically, in the LPRG model $x_i, x_j \in \X$ denote the latent positions associated to nodes $i$ and $j$ $(i,j\in [n]$) and $\X$ stands for the latent space. The connection probability of node $i$ and node $j$ is modeled as $p_{i,j}=k(x_{i},x_{j})$, where $k:\X \times \X \to [0,1]$ is a (reproducing) kernel. Fig.~\ref{fig:LPRG_related} illustrates specific models contained within this framework: a constant kernel yields the Erd\H{o}s-R\'{e}nyi model, whereas a linear  kernel yields the random dot product graph (RDPG; \citealt{young2007random}). By combining heterogeneous connectivity with explicit latent structure, LPRGs provide a flexible framework for modeling complex networks and a natural setting for studying dependence between random graphs.

A substantial literature studies edge correlation (a specific form of dependence) in random graphs in order to solve random graph matching problems \citep{lyzinski2014seeded, fishkind2019seeded}. This task boils down to independence testing between pairs ($M=2$) of random graphs \citep{barak2019nearly, wu2023testing, ding2023detection, oren2026testing}, with an inferential target significantly different from ours. Most works  \citep{barak2019nearly, wu2023testing, ding2023detection, mao2024testing, huang2025testing, feng2026strong} in this area of graph matching focus on the homogeneous setting of Erd\H{o}s-R{\'e}nyi model with the null hypothesis of two independent graphs $H_0: G_1 \perp G_2$, against the alternative ($H_1$) of the two graphs being correlated through a latent vertex correspondence. Namely, for two $n$-node graphs with adjacency matrices $\b A^1 = \left[ A^1_{i,j} \right]_{i,j=1}^n \in \{0,1\}^{n\times n}$ and $\b A^2 = \left[ A^2_{i,j} \right]_{i,j=1}^n \in \{0,1\}^{n\times n}$, one requires that the edge correlations $\t{Corr}\left( A^1_{i,j}, A^2_{\pi(i), \pi(j)} \right) = \rho$ are same for all $i,j \in [n]$ under some permutation $\pi$.\footnote{Correlation refers to the classical Pearson one.} \cite{ding2025efficiently} extend this correlated Erd\H{o}s-R{\'e}nyi setting by considering the inhomogeneous Bernoulli random graph model, which allows the edge correlations to be heterogeneous. However, even in case of $M=2$, focusing only on edge correlations discards non-linear associations.

Capturing these higher-order dependencies for independence testing between $M=2$ random graphs has been in the focus of various works. \citet{fujita2017correlation} represent each graph by the spectral radius of its adjacency matrix and use Spearman correlation to infer dependence between two sequence of random graphs. Another strategy is based on graph embedding \citep{rohe2011spectral, sussman2012consistent}, in which the adjacency or Laplacian matrix of the graph is mapped to a low-dimensional Euclidean representation. Such embeddings have subsequently been used for independence testing \citep{lee2019network, shen2026community, wen2026rkhs} and two-sample testing \citep{tang2017nonparametric, tang2017semiparametric} between random graphs, with the embedding dimension chosen heuristically.

Despite recent progress, several fundamental questions remain open for independence testing between random graphs; these include (i) defining a population dependence measure of random graphs accompanied by consistent estimators, (ii) determining the rate of the designed estimators, (iii) assuming that the estimator is based on graph embeddings, deriving theoretically the optimal embedding dimension. The first question has recently been tackled by \citet{wen2026rkhs} by introducing a graph HSIC measure for $M=2$ LPRGs with a consistent estimator based on adjacency spectral embeddings, without touching the case of $M\ge 2$ and questions (ii)-(iii).

Our goal is to address these severe limitations systematically within the multiplex framework of the LPRG models for $M \ge 2$. Particularly, our \tb{contributions} can be summarized as follows.
\begin{enumerate}
\item We define a dependency measure of $M\ge 2$ LPRGs based on HSIC, construct an estimator for it based on the spectral decomposition of the adjacency matrices, and prove its $\tilde{\mathcal{O}} \left( n^{-\frac{1}{3}}\right)$ convergence.
\item We derive the optimal embedding dimension to be used in the spectral decomposition in case of the commonly-encountered polynomial and exponential spectral decay of the associated integral operators.
\item We demonstrate the applicability of the proposed independence testing approach for multiplex graphs, on both simulated examples and in the analysis of functional brain networks (one of our motivating examples).

\end{enumerate}

The remainder of the paper is structured as follows. In Section~\ref{sec_problem_formulation} we introduce the notations used throughout the paper, followed by problem formulation. Our proposed estimator for the dependence of $M$ LPRGs with its theoretical guarantees are presented in Section~\ref{sec:graph-HSIC-estimation}. The numerical efficiency of the designed estimator is demonstrated in Section~\ref{sec:experiments}. Conclusions are drawn in Section~\ref{sec:conclusions}. Our proofs are gathered in the supplement.

\section{Problem Formulation}\label{sec_problem_formulation}
In this section, after defining a few notations, we gradually formulate our task, the joint independence testing of latent position random graphs.

We introduce below the \tb{notations} $\Z^+$,  $[M]$, $\Rp$, $\mathbbm{1}_A$, $\log$,  $\left \langle \b x, \b y \right \rangle$, $\left\|\b x\right\|_2$, $\bm{1}_d$, $\bm{0}_d$, $\b A^{\top}$, $\b{A}^{\frac{1}{2}}$, $\odot_{m \in[M]} \b A^m$, $\left \| \b A \right \| _F$, $\left \| \b A \right \|$, $\diag(\b A)$, $\diag(\b a)$, $\times_{m=1}^M S^m$, $S^M$, $\mN(\b m, \bm{\Sigma})$, $\O (a_n)$, $\Omega(a_n)$, $\Theta(a_n)$, $o(a_n)$,  $\tilde{\O} (a_n)$, $\O_p(a_n)$, $\left \langle \mathcal{A}, \mathcal{B}\right\rangle$, $\left \| \mathcal{A} \right \|_2$, $\Span(S)$, $\bar{S}$, $\mathcal{M}_{+}^1(\X)$, $\mathrm{supp}(\P)$,  $\phi_k$, $\mH_k$,  $\left \| A \right \|$, $\L(\mH_1,\mH_2)$, $\L(\mH)$, $A^*$, $A^{\frac{1}{2}}$, $|A|$, $\tr(A)$, $\left\|A\right\|_{HS(\mH_1,\mH_2)}$, $HS(\mH_1,\mH_2)$, $HS(\mH_1)$, $\mathcal{L}_2(\X,\nu)$, $S_k$, $T_k$, $C_k$, $\hat{C}_k$, $\mu_k$, $\t{MMD}_k$, $\otimes_{m=1}^M \P^m$, $\otimes_{m=1}^M k^m$, $\otimes_{m=1}^M h^m$, $\otimes_{m=1}^M \mH_{k^m}$ and $\t{HSIC}_k$. 

For a positive integer $M\in \Z^+:=\{1,2,\ldots\}$, let $[M]:=\left \{ 1,\dots,M \right \} $.  The set of positive reals is denoted by $\Rp=:\{x\in \R\,:\, x>0\}$. For a set $A$, $\mathbbm{1}_A$ denotes its indicator function: $\mathbbm{1}_A(x)=1$ if $x \in A$ and $\mathbbm{1}_A(x)=0$ otherwise. 
The logarithm function with base $e$ is denoted by $\log$. The inner product of two vectors $\b x,\b y\in \R^d$ is written as $\left \langle \b x, \b y \right \rangle =\sum_{i \in [d]} x_i y_i$; the induced Euclidean norm is $\left\|\b x \right\|_2 = \sqrt{\langle \b x, \b x \rangle}$. The $d$-dimensional vector of ones and that of zeros is $\b{1}_d$ and $\b{0}_d$, respectively. For matrix $\b A \in \R^{d_1 \times d_2}$,  $\b A^{\top} \in \R^{d_2 \times d_1}$ denotes its transpose. For an $\b A \in \R^{d\times d}$ positive semi-definite matrix (i.e.\ $\b A = \b A^\top$ and $\b z^\top \b A \b z \ge 0$ for all $\b z \in \R^d$), let $\b A^{\frac{1}{2}} \in \R^{d\times d}$ denote its uniquely existing positive semi-definite square root. The Hadamard product of matrices $\b A^m \in \R^{d_1 \times d_2}$ ($m \in [M]$) is written  as $\odot_{m \in[M]} \b A^m = \left [ {\textstyle \prod_{m\in [M]}} (\b A^m)_{i,j} \right ]_{i,j = 1,1}^{d_1,d_2}$. The Frobenius norm of a matrix $\b A \in \R^{d_1 \times d_2}$ is $\left \| \b A \right \| _F := \sqrt{ {\textstyle \sum_{i \in [d_1]}\textstyle \sum_{j \in [d_2]}} A_{i,j}^2 } $. The spectral norm of $\b A \in \R^{d_1 \times d_2}$ is $\left \| \b A \right \|=\sup_{\b x\ne \b 0_{d_2}}\frac{\left\|\b A \b x\right\|_2}{\left\|\b x\right\|_2}$, and it equals to the largest eigenvalue of $\left(\b A^{\top} \b A\right)^{\frac{1}{2}}$. 
For a matrix $\b A \in \R^{n \times n}$, $\diag(\b A)\in \R^n$ represents its diagonal. For a vector $\b a \in \R^n$, $\diag(\b a) \in \R^{n\times n}$ stands for the diagonal matrix with diagonal $\b a$. The Descartes product of sets $(S^m)_{m=1}^M$ is written as $\times_{m=1}^M S^m$; when $S:=S_1=\ldots=S_M$, we use the shorthand $S^M$. Let $\mN(\b m, \bm{\Sigma})$ denote the normal distribution with mean $\b m$ and covariance $\bm{\Sigma}$.
For positive sequences $(a_n)_{n\in\Z^+}$ and $(b_n)_{n\in\Z^+}$, (i) $b_n = \O (a_n)$ if there exist constants $C > 0$ and $n_0 \in \Z^+$ such that $b_n \le C a_n$ for all $n \ge n_0$, (ii) $b_n = \Omega(a_n)$ if $a_n = \O(b_n)$, (iii) $b_n = \Theta(a_n)$ if $b_n = \Omega(a_n)$ and $b_n = \O(a_n)$, (iv) $b_n = o(a_n)$ if $\lim_{n \to \infty} \frac{b_n}{a_n} =0$, (v) $\tilde{\O}(a_n)$ is the same as $\O(a_n)$ apart from discarding polylogarithmic factors of $n$, in other words $\tilde{\O}(a_n)=\O(a_n \mathrm{polylog}(n))$. For a sequence of real-valued random variables $(X_n)_{n\in \Z^+}$, $X_n=\O_p(a_n)$ means that $\left(\frac{X_n}{a_n}\right)_{n\in \Z^+}$ is bounded in probability. For tensors $\mathcal{A}=\left[a_{i_1, \dots, i_M}\right]_{i_1=1,\ldots, i_M=1}^{d^1,\ldots,d^M}, \mathcal{B}=\left[b_{i_1, \dots, i_M}\right]_{i_1=1,\ldots, i_M=1}^{d^1,\ldots,d^M} \in \R^{d^1 \times \dots \times d^M}$, their inner product\footnote{Notice that this definition is compatible with the inner product of vectors ($\langle \b a, \b b \rangle$).} is defined as
\begin{align}
    \left \langle \mathcal{A}, \mathcal{B}\right\rangle &:= \left ( \sum_{i_1 \in [d^1]} \sum_{i_2 \in [d^2]}\dots \sum_{i_M \in [d^M]} a_{i_1, \dots, i_M} b_{i_1, \dots, i_M}  \right )^{1/2};
    \label{eq:inner-product-on-tensors}
\end{align}
accordingly the 2-norm of $\mathcal{A}$ is 
\begin{align}
    \left \| \mathcal{A} \right \|_2 := \sqrt{\left\langle \mathcal{A}, \mathcal{A} \right\rangle} = \left ( \sum_{i_1 \in [d^1]} \sum_{i_2 \in [d^2]}\dots \sum_{i_M \in [d^M]} a^2_{i_1, \dots, i_M}  \right )^{1/2}.
    \label{eq:2-norm-on-tensors}
\end{align}
Let $S \subseteq \mH$ be a subset of a Hilbert space $\mH$; $\Span(S)$ and $\bar{S}$ denotes  the linear hull and the closure of $S$, respectively. Let $\X$ be a measurable space. The set of Borel probability measures  on $\X$ is denoted by $\mathcal{M}_{+}^1(\X)$. The support of a probability measure $\P \in \mathcal{M}_{+}^1(\X)$ denoted by $\mathrm{supp}(\P)$ is defined as the subset in $\X$ for which every open neighborhood of $x \in \X$ has a positive measure.
A function $k: \X \times \X \rightarrow \R$ is called kernel if there exists a feature map $\phi: \X \rightarrow \mH$ with a Hilbert space $\mH$ such that $k(x,x') = \langle \phi(x),\phi(x')\rangle_{\mH}$ for all $x, x' \in \X$.\footnote{We assume throughout the paper that $k$ is measurable.} The canonical feature map of $k$ is denoted by $\phi_k(x) = k(\cdot,x)$, where $k(\cdot,x)$ stands for the function $x' \in \X \mapsto k(x',x) \in \R$ with $x \in \X$ fixed. Let $\mH_k$ be the reproducing kernel Hilbert space (RKHS) associated to a kernel $k: \X \times \X \to \R$; constructively $\mH_k = \overline{\Span(k(\cdot,x)\,:\, x\in \X)}$, with the inner product defined as (the linear extension of) $\langle k(\cdot,x),k(\cdot,x')\rangle_{\mH_k}=k(x,x')$ for all $x,x' \in \X$. Let $\mH_1$ and $\mH_2$ be Hilbert spaces, and $A:\mH_1\rightarrow \mH_2$ a linear operator. Let $\left\| A \right\|=\sup_{x\ne 0} \frac{\left\|Ax\right\|_{\mH_2}}{\left\|x\right\|_{\mH_1}}$ denote the operator norm of $A$. The set of $\mH_1 \to \mH_2$ bounded linear operators is defined as $\L(\mH_1,\mH_2):=\{A:\mH_1 \to \mH_2 \t{ linear operator }\,:\, \left\| A \right\|<\infty\}$; we use the shorthand $\L(\mH):=\L(\mH,\mH)$.
The adjoint $A^* \in \L(\mH_2,\mH_1)$ of an operator $A\in \L(\mH_1,\mH_2)$ is the operator for which  $\left \langle A f, g \right \rangle_{\mH_2} = \left \langle f, A^* g \right \rangle_{\mH_1} $ for all $f \in \mH_1$ and $g \in \mH_2$. An operator $A\in \L(\mH)$ is called positive (shortly $A\ge 0$) if $A$ is self-adjoint ($A^*=A$) and $\langle A f, f\rangle_\mH \ge 0$ for all $f\in \mH$. If $A\ge 0$, then there exists a unique $B \ge 0$ such that $B^2 = A$; we write $B = A^{\frac{1}{2}}$ and call $B$ as the square root of $A$. With this notation, the absolute value of $A \in \L(\mH)$ is defined as $|A|=\left(A^*A\right)^{\frac{1}{2}}$. An operator $A\in \L(\mH)$ is referred to as trace class if $\mH$ is separable and $\sum_{i \in I}\left\langle |A| e_i,e_i \right\rangle_{\mH} <\infty$ for some countable orthonormal basis (ONB)  $(e_i)_{i \in I}$ in $\mH$, and in this case the trace of $A$ is defined as $\tr (A) := \sum_{i \in I} \left\langle A e_i,e_i \right\rangle_{\mH} <\infty $.\footnote{\label{footnote:TR-HS:ONB-independence}The separability of $\mH$ implies the existence of a countable ONB in $\mH$; the definition of an operator being trace-class or Hilbert-Schmidt (and the value of the Hilbert-Schmidt norm) is independent of the specific ONB considered.} An operator $A \in \L(\mH_1,\mH_2)$ is called (i) compact if  $\overline{\{A f: f \in \mH_1, \left \| f \right \|_{\mH_1} \le 1\}}$ is a compact set, (ii) Hilbert-Schmidt if $\mH_1$ is separable and $\sum_{i\in I} \left\|A e_i\right\|_{\mH_2}^2 < \infty$, where $(e_i)_{i\in I}$ is some countable ONB of $\mH_1$ and in this case $\left\|A\right\|_{HS(\mH_1,\mH_2)}:= \sqrt{\sum_{i\in I} \left\|A e_i\right\|_{\mH_2}^2}$. The set of $\mH_1 \rightarrow \mH_2$ Hilbert-Schmidt operators denoted by $HS(\mH_1,\mH_2)$ forms a Hilbert space with the inner product $\langle A,B \rangle_{HS(\mH_1,\mH_2)} := \sum_{i\in I} \langle A e_i,B e_i\rangle_{\mH_2}$; we use the shorthand $HS(\mH_1):=HS(\mH_1,\mH_2)$ when $\mH_1 = \mH_2$.$^{\ref{footnote:TR-HS:ONB-independence}}$ Trace-class and Hilbert-Schmidt operators are compact.

Let $\mathcal{L}_2(\X,\nu)$ denote the space of real-valued square-integrable functions on $\X$ with respect to a probability measure $\nu \in \mathcal{M}_{+}^1(\X)$. 
Let the integral operator $S_k=S_{k,\nu}: \mathcal{L}_2(\X,\nu) \to \mH_k$ associated to the $(k,\nu)$ pair be defined as   
\begin{align}
    (S_k g)(x) &= \int_{\X} k(x,{x}')g({x}') \d \nu({x}'), \quad (g \in \mathcal{L}_2(\X,\nu), x \in \X),
\end{align}
under the assumption that $\int_\X k(x,x)\d \nu(x) < \infty$; this holds for instance when $k$ is bounded (in other words, $\sup_{x,x'\in \X}k(x,x') < \infty$) which we suppose in the sequel for all the kernels appearing. If $\mH_k$ is separable\footnote{\label{footnote:sep-Hk}A continuous kernel $k:\X\times\X\rightarrow \R$ on a separable topological domain $\X$ is sufficient \citep[Lemma~4.33]{steinwart08support} to guarantee the separability of $\mH_k$.} and $\int_{\X}k(x,x)\d \nu(x)<\infty$, then it is known \citep[Theorem~4.27]{steinwart08support} that (i) the operator $T_k = T_{k,\nu} = S_k^* S_k: \mathcal{L}_2(\X,\nu) \to \mathcal{L}_2(\X,\nu)$ is a positive, self-adjoint, trace class operator, (ii) $S_k$ is Hilbert-Schmidt, and (iii) $C_k = C_{k,\nu} = S_k S_k^* = \E_{X\sim \nu}[\phi_k(X) \otimes \phi_k(X)] \in \L(\mH_k)$ has the same eigenvalues as $T_k$. 
Specifically, the properties of $T_k$ imply \citep[Theorem~A.5.13]{steinwart08support} that it has a countable set of non-negative eigenvalues  $\left(\lambda_i \right)_{i\in I}$ associated to an ONS (orthonormal system) $(\psi_i)_{i\in I}\subset \mathcal{L}_2(\X,\nu)$ converging to $0$; w.l.o.g.\ we assume that the eigenvalues are ordered  in a non-increasing fashion $\lambda_1 \ge \lambda_2 \ge \cdots$. The empirical estimator of $C_k$ is defined as $\hat{C}_k=\frac{1}{n} \sum_{i=1}^n \phi_k(X_i) \otimes \phi_k(X_i)$.

The mean embedding of $\P$ is defined as 
$\mu_k (\P) = \int _{\X} \phi_k(x) \d \P(x)$, where the integral is meant in Bochner sense. $\mu_k (\P)$ exists if $\int_\X \sqrt{k(x,x)} \d \P(x) < \infty$; this condition holds for bounded kernels.  The maximum mean discrepancy (MMD) of two probability measures $\P, \Q \in \mathcal{M}_{+}^1(\X)$ is defined as 
\begin{align}
    \mathrm{MMD}_k(\P, \Q) = \left \| \mu_k(\P) - \mu_k(\Q) \right \| _{\mH_k}.
\end{align}

Let $X = (X^m)_{m \in [M]}\sim\P \in \mathcal{M}_{+}^1(\X)$ be a random variable on a product space $\X = \times_{m=1}^M \X^m$ with marginals $X^m \sim \P^m$ ($m\in [M]$); the product of these marginal distributions is written as  $\otimes_{m=1}^M \P^m \in \mathcal{M}_{+}^1(\X)$. The spaces $(\X^m)_{m\in [M]}$  are assumed to be enriched with kernels $k^m : \X^m \times \X^m \to \R$ ($m\in [M]$).  The tensor product kernel $k=\otimes_{m=1}^M k^m:\X \times \X \rightarrow \R$ is defined as 
\begin{align}
    k\left ( (x^m)_{m=1}^M , (\tilde{x}^m)_{m=1}^M \right ) = \prod_{m \in [M]} k^m(x^m,\tilde{x}^m), \quad (x^m,\tilde{x}^m \in \X^m),
\end{align}
and it is known that $\mH_k = \otimes_{m=1}^M \mH_{k^m}$, where the r.h.s.\ stands for the tensor product of the Hilbert spaces $\left(\mH_{k^m}\right)_{m=1}^{M}$. 
Particularly, given Hilbert spaces $(\mH_m)_{m=1}^M$ and $h^m \in \mH_{m}$ $(m\in [M])$, the multi-linear operator $\otimes_{m=1}^M h^m \in \otimes_{m=1}^M \mH_{m}$ is defined as  $\left(\otimes_{m=1}^M h^m\right) \left(g^1, \dots, g^M\right) = \prod_{m \in [M]} \left \langle h^m,g^m \right \rangle_{\mH_{m}} $ for $g^m \in \mH_{m}, m\in [M]$; $\otimes_{m=1}^M \mH_{m}=\overline{\Span(\otimes_{m=1}^M h^m \,:\, h^m \in \mH_{m}, m\in [M])}$ with the inner product defined as (the linear extension of)
\begin{align}
    \left \langle \otimes_{m=1}^M a^m, \otimes_{m=1}^M b^m  \right \rangle_{\otimes_{m=1}^M \mH_{m}} = \prod_{m \in [M]} \left \langle a^m, b^m \right \rangle_{\mH_{m}}.
    \label{eq:def:tensor-product}
\end{align} 
Specifically, choosing $\X^m=\R^{d^m}$ and $k^m$ to be the linear kernel for $m\in [M]$, $\otimes_{m=1}^M \mH_{k^m}$ can be identified with $\R^{d^1\times \ldots \times d^M}$, and $\left\|\cdot\right\|_{\otimes_{m=1}^M \mH_{k^m}}$ simplifies to $\left\|\cdot\right\|_2$ in \eqref{eq:2-norm-on-tensors}.
The Hilbert-Schmidt independence criterion (HSIC) is
\begin{align}
    \t{HSIC}_k(\P) = \t{MMD}_k (\P, \otimes_{m=1}^M \P^m).
\end{align}
Assuming the universality  of $(k^m)_{m=1}^M$-s, one can show \citep{szabo2017characteristic} that $\t{HSIC}_{k}(\P)=0 \Leftrightarrow \P  =  \otimes_{m=1}^M \P^m$; hence one can use HSIC to test the independence of $(X_m)_{m=1}^M$-s:
\begin{align}
    H_{0} &: \P  = \otimes_{m=1}^M \P^m, &     H_{1} &: \P  \ne  \otimes_{m=1}^M \P^m.
\end{align}
Given i.i.d.\ samples $\hat{\P}_n = \left(x_i \right)_{i=1}^n$ drawn from $\P$ with  $x_i=\left(x_i^1,\dots,x_i^M\right)$ ($i\in [n]$), replacing the population means with the sample means, gives rise to the V-statistic based estimator of the squared HSIC:
\begin{equation}\label{sample_HSIC1}
    \t{HSIC}^2_k(\hat{\P}_n) = \frac{1}{n^2} \bm{1}_n^{\top}\left ( \odot_{m\in[M]} \b K^m \right )\bm{1}_n + \frac{1}{n^{2M}}\prod_{m=1}^{M} \bm{1}_n^{\top} \b K^m \bm{1}_n -\frac{2}{n^{M+1}} \bm{1}_n^{\top} \left ( \odot_{m\in[M]} \b K^m \bm{1}_n \right ),
\end{equation}
with the Gram matrices $\b K^m = \left [ k^m\left(x_i^m,x_j^m\right) \right ]_{i,j=1}^n \in \R^{n \times n}$ $(m\in [M])$. 

Our \tb{focus} is to perform independence testing in the context of latent position random graphs (LPRGs). The primary challenge to tackle is that in this case  one does not have access to the Gram matrices $\left(\b K^m\right)_{m=1}^M$, which is critical in the evaluation of the HSIC estimator \eqref{sample_HSIC1}. Instead, in LPRGs the Gram matrices $\left(\b K^m\right)_{m=1}^M$-s associated to latent positions determine the \emph{connection probabilities} in the graphs, and the key question is how to translate this implicit knowledge to a consistent estimator for the dependency of  LPRGs. This is the problem we formulate below, followed by our proposed estimator in Section~\ref{sec:graph-HSIC-estimation}.

Specifically, we consider $M$ latent position random graphs (LPRGs)
\begin{align}
\left(G^m\right)_{m=1}^{M} &= ((V^m, E^m))_{m=1}^{M},
\end{align}
with the graph $G^m$ described by its nodes $V^m=(1,\ldots,n)$ and edges $E^m$. We suppose that the LPRGs are undirected without self-loops. Under these $M$ graphs one assumes the existence of (i) a joint latent random variable $X=(X^m)_{m=1}^{M} \sim \P$ taking values in $\X = \times_{m=1}^M\X^m$, $X^m \sim \P^m$ and $(\X^m,\tau^{m})$ topological spaces, (ii) kernels  $k^m:\X^m \times \X^m \to  \left [ 0,1 \right]$ ($m\in [M]$), and (iii)  $n$ i.i.d.\ samples $X_1=x_1:=\left(x_1^m\right)_{m=1}^M, \ldots, X_n=x_n:=\left(x_n^m\right)_{m=1}^M$ drawn from $\P$---$x_i^m$ is the realized latent position of the $i^{th}$ node in graph $G^m$---inducing the Gram matrix
\begin{align}
\R^{n \times n}  \ni \b K &= \left[K_{i,j}\right]_{i,j=1}^n, & K_{i,j}=
                    \begin{cases}
                        k(x_i,x_j), & \text{ if } i \ne j \\
                        0 & \text{ if } i=j
                     \end{cases}, \quad i,j\in [n].
\end{align} 
The drawn  $m^{th}$ latent positions $(x^m_i)_{i=1}^{n}$ (samples from $\P^m$) with the kernel $k^m$ determine the connection probability matrix in the LPRG $G^m$:
                \begin{align}
                 \R^{n \times n} \ni \b K^m &=\left[K^m_{i,j}\right]_{i,j=1}^n, &K^m_{i,j} & =
                     \begin{cases}
                        k^m(x^m_i,x^m_j), & \text{ if } i \ne j \\
                        0 & \text{ if } i=j
                     \end{cases}, \quad i,j\in [n],m\in [M], \label{eq:def_K^m_ij}
                \end{align}
and with this notation $\b K = \odot_{m\in [M]} \b K^m$ by the assumption that $k=\otimes_{m=1}^M k^m$.           
This means that by denoting the adjacency matrix of $G^m$ as $\b A^m=\left[A^m_{i,j}\right]_{i,j=1}^{n} \in \{0,1\}^{n \times n}$ with entries
$ A^m_{i,j}=1 \ \mathrm{if} \ (i,j) \in E^m, \ \mathrm{and} \  A^m_{i,j}=0 \ \mathrm{otherwise}$,  
\begin{itemize}
    \item (i) the upper-diagonal entries of $\b A^m$ follow a conditionally independent Bernoulli distribution given the latent positions $(x^m_i)_{i=1}^{n}$:
                \begin{align} \label{eq:A^m}
                    A^m_{i,j} &\sim \mathrm{Bernoulli}(k^m(x^m_{i},x^m_{j})) \quad \text{for} \ i < j\,\, |\,\, (X^m_i)_{i=1}^{n} = (x^m_i)_{i=1}^{n},
                \end{align}
     (ii) the lower-diagonal entries equal to the upper-diagonal ones: $A_{i,j}^m = A_{j,i}^m$ ($i>j$, $m\in [M]$), (iii) the diagonal entries are $A^m_{i,i}=0$ for all $i\in [n]$ and $m\in [M]$.
     \item In short, $\b A^m$-s have zero diagonal ($\diag\left(\b A^m\right) = \b 0_n$), and they are symmetric ($\left(\b A^m\right)^\T = \b A^m$) with the upper triangular entries specified according to \eqref{eq:A^m}.
\end{itemize}

\tb{Remarks}: 

\begin{itemize}
\item For instance, $(G^m)_{m=1}^M$ can correspond to functional brain networks under $M$ different cognitive tasks (an example we will be using for numerical illustration in Section~\ref{sec:experiments}). 

\item Specific cases of LPRGs: Multiple popular graph models arise as a specific case of LPRGs (Fig.~\ref{fig:LPRG_related}).
\begin{enumerate}
    \item Indeed, by using a constant kernel ($k^m(x^m,\tilde{x}^m)=p \in [0,1]$ for all $x^m,\tilde{x}^m \in \X^m$), the LPRG $G^m$ reduces to the Erd{\H{o}}s-R{\'e}nyi (ER) model which assumes a fixed connection probability.
    \item With a linear $k^m(x^m,\tilde{x}^m)=(x^m)^{\top}\tilde{x}^m$, one gets back the random dot product graph (RDPG; \citealt{young2007random}).
\end{enumerate}
\end{itemize}

We posit that the LPRGs $(G^m)_{m=1}^{M}$ are independent if and only if $(X^m)_{m=1}^{M}$-s are independent, in other words $\P = \otimes_{m=1}^M \P^m$, and quantify the dependence of $(G^m)_{m=1}^{M}$-s with
\begin{align}\label{eq:target}
     \mathrm{GHSIC}^2_k(\P) = \left\| \mu_k(\P) - \mu_k(\otimes_{m=1}^M \P^m) \right\|^2_{\mH_k}, \quad k=\otimes_{m=1}^M k^m,
\end{align}
which we will refer to as the squared \emph{GHSIC (a shorthand for graph HSIC)}. This definition in the specific case of $M=2$ reduces to the one provided by \citet{wen2026rkhs}.

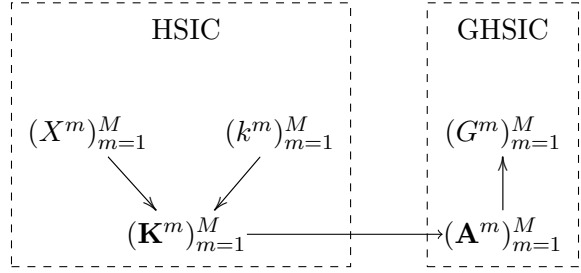
\begin{figure}
    \centering
\begin{tikzpicture}
    % Left dashed box
    \node[rectangle, draw, dashed, inner sep=3pt] (box1) {
        \xymatrix@C=-0.5cm@R=0.8cm{
            & \text{HSIC} \\
            (X^m)_{m=1}^{M} \ar[dr]  & &(k^m)_{m=1}^{M} \ar[dl] \\
            & \tikz[remember picture] \node[inner sep=0pt] (A1) {$(\mathbf{K}^m)_{m=1}^{M}$};
        }
    };
    
    % Right dashed box
    \node[rectangle, draw, dashed, inner sep=3pt, right=1cm of box1] (box2) {
        \xymatrix@C=0cm@R=0.8cm{
            \text{GHSIC} \\
            (G^m)_{m=1}^{M}  \\
            \tikz[remember picture] \node[inner sep=0pt] (B2) {$(\mathbf{A}^m)_{m=1}^{M}$}; \ar[u]
        }
    };
 
\end{tikzpicture}

\begin{tikzpicture}[overlay, remember picture]
        \draw[->] (A1.east) -- (B2.west);
\end{tikzpicture} 

\caption{Illustration of measuring the independence of latent position random graphs (r.h.s.; using GHSIC) vs.\ classical independence (l.h.s.; relying on HSIC). The traditional HSIC estimator relies on the Gram matrices $\left(\b{K}^m\right)_{m=1}^{M}$; the GHSIC estimator has to be based on the observable adjacency matrices $\left(\b{A}^m\right)_{m=1}^{M}$ determined by the hidden Gram matrices $\left(\b{K}^m\right)_{m=1}^{M}$.}
    \label{fig:HSIC-graphHSIC}
\end{figure}

It is important to note that while one has access to the adjacency matrices $\left(\b A^m\right)_{m=1}^{M}$ of the random graphs $\left(G^m\right)_{m=1}^{M}$, both the latent positions $\left(X^m\right)_{m=1}^{M}$ and the kernels $\left(k^m\right)_{m=1}^{M}$ are unobservable; this specifically means that one \emph{can not} evaluate the Gram matrices $\left(\b K^m\right)_{m=1}^{M}$ which the classical HSIC estimator \eqref{sample_HSIC1} is based on; see Fig.~\ref{fig:HSIC-graphHSIC} for an illustration. Our goal is to estimate \eqref{eq:target} based only on the indirect information present in $\left(\b A^m\right)_{m=1}^{M}$-s.  
 
\section{Proposed Estimator} \label{sec:graph-HSIC-estimation}

In this section we present our estimator for $\mathrm{GHSIC}_{k}$. Particularly, for our LPRGs $(G^m)_{m=1}^M$ with the adjacency matrices $\b A^m \in \R^{n \times n}$ ($m \in [M]$), let us consider the spectral decomposition of the positive semi-definite matrices $\left(\left(\b A^m\right)^{\top} \b A^m\right)^{1/2}$-s ($m\in [M]$; with non-negative eigenvalues ordered non-increasingly with the corresponding eigenvectors) and their $d^m$-truncated ($d^m \in [n]$) approximations:
 \begin{align}
 \left(\left(\b A^m\right)^{\top} \b A^m\right)^{1/2}  &= \b U^m \b S^m \left(\b U^m\right)^{\top}\approx  \b U^{m}_{d^m} \b S^{m}_{d^m} \left(\b U^{m}_{d^m}\right)^{\top} \nonumber \\
 &=: \hat{\b K}^m = \left[\hat{K}^m_{i,j}\right]_{i,j=1}^n =: \left(\hat{\bm{\Phi}}^m\right)^\T\hat{\bm{\Phi}}^m =: \left[\langle  \hat{\phi}_{k^m}(x_i^m),  \hat{\phi}_{k^m}(x_j^m)\rangle \right]_{i,j=1}^n, \label{eq:Km-hat}
 \end{align}
 where $m \in [M]$, $\b U^m\in \R^{n\times n}, \b S^m =\diag\left(\left(\hat{\lambda}_i^m\right)_{i=1}^n\right) \in \R^{n \times n}, \b U^m_{d^m} \in \R^{n\times d^m}, \left(\b S^{m}_{d^m}\right)^{1/2} := \diag \left (\sqrt{\hat{\lambda}^m_1},\dots,\sqrt{\hat{\lambda}^m_{d^m}}\right) \in \R^{d^m \times d^m}$, where $\left (\hat{\lambda}^m_i \right)_{i=1}^{d^m} $ denotes the top $d^m$ eigenvalues of $\left(\left(\b A^m\right)^{\top} \b A^m\right)^{1/2}$ and 
 \begin{align}
 \left(\hat{\bm{\Phi}}^m\right)^\T := \b U_{d^m}^m \left(\b S_{d^m}^m\right)^{1/2} =: \left(\left[\hat{\phi}_{k^m}\left(x_1^m\right),\ldots, \hat{\phi}_{k^m}\left(x_n^m\right)\right]_{i=1}^n\right)^\T \in \R^{n\times d^m}, 
 \label{eq:hat-Phi-m}
 \end{align}
 with $\hat{\phi}_{k^m}\left(x_i^m\right)\in \R^{d^m}$ denoting the feature associated to $x_i^m$, $i\in [n]$ and $m\in [M]$. 

 Our proposed GHSIC estimator consists of replacing $\b K^m$-s in \eqref{sample_HSIC1}  with their estimate $\hat{\b K}^m$ computed from $\b A^m$: 
 \begin{align}\label{est_HSIC2}
     \widehat{\mathrm{GHSIC}^2_{k}}&= \frac{1}{n^2} \bm{1}_n^{\top}\left ( \odot_{m\in[M]} \hat{\b K}^m \right )\bm{1}_n + \frac{1}{n^{2M}}\prod_{m=1}^{M} \bm{1}_n^{\top} \hat{\b K}^m \bm{1}_n -\frac{2}{n^{M+1}} \bm{1}_n^{\top} \left ( \odot_{m\in[M]} \hat{\b K}^m \bm{1}_n \right )\\
     & = \frac{1}{n^2} \bm{1}_n^{\top}\left ( \odot_{m\in[M]} \hat{\b K}^m \right )\bm{1}_n + \prod_{m=1}^{M} \left(\frac{1}{n^2}\bm{1}_n^{\top} \hat{\b K}^m \bm{1}_n\right) -\frac{2}{n} \bm{1}_n^{\top} \left ( \odot_{m\in[M]} \frac{1}{n}\hat{\b K}^m \bm{1}_n \right ),\nonumber
     \end{align} 
where the latter form is advantageous for large $n$ to avoid numerical instabilities.   
 
One can see (Lemma~\ref{lemma:HSIC2:Gram}) that \eqref{est_HSIC2} is non-negativity. Indeed, let $k = \otimes_{m=1}^M k^m$ and $X=(X^m)_{m=1}^M$ with distribution $\P$. 
Recall that $\hat{\P}_n = \left(\left(x_i^1,\dots,x_i^{M}\right) \right)_{i=1}^n$ denotes a sample of size $n$ from $\P$, and similarly $\hat{\P}^m_n = (x_i^m)_{i=1}^n$ is a sample of size $n$ from $\P^m$ $(m\in [M]$). 
Our following lemma shows that the estimator $\widehat{\mathrm{GHSIC}^2_{k}}$ in \eqref{est_HSIC2} is non-negative.
 \begin{lemma}[Non-negativity of $\widehat{\mathrm{GHSIC}^2_{k}}$, estimator $\widehat{\mathrm{GHSIC}_{k}}$] \label{lemma:HSIC2:Gram}
     The estimator in \eqref{est_HSIC2} can be expressed as
     \begin{align}\label{est_HSIC2:v0}
     \widehat{\mathrm{GHSIC}^2_{k}} = \left \| \hat{\mu}_k \left(\hat{\P}_n\right) - \otimes_{m=1}^M  \hat{\mu}_{k^m} \left(\hat{\P}^m_n\right) \right \|^2_2,
 \end{align}
 where 
 \begin{align}
     \hat{\mu}_k\left(\hat{\P}_n\right) &= \frac{1}{n} \sum_{i=1}^n \otimes_{m=1}^M  \hat{\phi}_{k^m}\left(x_i^m\right) \in \R^{d^1 \times \cdots \times d^M}, & \hspace{-0.45cm} \hat{\mu}_{k^m}\left(\hat{\P}^m_n\right) &= \frac{1}{n} \sum_{i=1}^n  \hat{\phi}_{k^m}\left(x_i^m\right) \in \R^{d^m}. \label{eq:def_hat_mu_k_hat_mu_k^m}
 \end{align}
 Specifically, this means that  the estimator $\widehat{\mathrm{GHSIC}_{k}}$ is also well-defined, as one can take the square root of the r.h.s.\ of \eqref{est_HSIC2:v0}.
 \end{lemma}

 Notice that the proposed estimator does not rely on prior knowledge of the kernel and latent positions, thereby eliminating the typical bottleneck of kernel choice. In the next section, we analyze our estimator $\widehat{\mathrm{GHSIC}_{k}}$ and discuss a selection criteria for the embedding dimensions $\left(d^m \right)_{m=1}^{M}$-s.

\subsection{Main Result}\label{main_result}
In this section we establish rates for the proposed estimator $\widehat{\mathrm{GHSIC}_{k}}$.

If $(\X_m)_{m=1}^M$-s are compact metric spaces, and $k^m: \X^m \times \X^m \to [0,1]$ are continuous kernels for $m\in [M]$ (hence $(k^m)_{m=1}^M$-s are Mercer kernels), then the operators 
$T_{k^m}=T_{k^m,\P^m}:\mathcal{L}_2\left(\X^m,\P^m\right) \to \mathcal{L}_2\left(\X^m,\P^m\right)$ ($m\in [M]$) are compact, positive, self-adjoint and they have countable set of non-negative eigenvalues $\left(\lambda_i^m \right)_{i\in I^m}$ ($m\in [M]$) converging to zero associated to an ONS $(\psi^m_i)_{i \in I^m}$; w.l.o.g.\ one can assume that the eigenvalues are ordered in a non-increasing fashion $\lambda_1^m \ge \lambda_2^m \ge \cdots$ ($m\in [M]$). If $(\X^m)_{m=1}^M$-s are compact metric, they are separable \citep[Proposition~1.9, Corollary~1.5]{howes95modern}, which together with continuous $(k^m)_{m=1}^M$-s imply the separability of $(\mH_{k^m})_{m=1}^M$-s.$^{\ref{footnote:sep-Hk}}$ Moreover, in this case $\int_{\X}k^m\left(x^m,x^m\right)\d \P^m\left(x^m\right)<\infty$ ($m\in[M]$) holds for bounded $(k^m)_{m=1}^M$-s implying the trace class property of the operators $T_{k^m}$ ($m\in [M]$).

Throughout the paper we assume that the number of nodes is $n\ge 2$. We also impose the following blanket assumptions which are summarized in Table~\ref{assumption_table} for the reader's convenience. 
\begin{itemize}
    \item[] 
    \begin{assumption}[density]\label{assum_Delta}
        Let $m \in [M]$ be arbitrary and fixed. There exist universal constants $C_0^m > 0$ and $C_1^m > 0$, and for any $\eta \in (0,1)$ there exist $n_{0,(i)}^{m}(\eta) \in \Zp$ and $ n_{0,(ii)}^{m}(\eta) \in \Zp$ with the following properties. (i) For all $n \ge n_{0,(i)}^m(\eta)$ and for any Gram matrix $\b K^m = \left[ K^m_{i,j} \right]_{i,j=1}^n =\left[ k^m(X_i,X_j) \right]_{i,j=1}^n \in \R^{n \times n}$ associated to some sequence $\left( X_i^m \right)_{i=1}^n \sim (\P^m)^{n}$, $\Delta^m > C_0^m \log(n)$ holds with $\left(\P^m\right)^{\otimes n}$-probability at least $1-\eta$, where $\Delta^m = \Delta^m (X_1^m,\dots, X_n^m;n):= \max_{i \in [n]} \sum_{j \in [n]} K^m_{i,j}$. (ii) For all $n \ge n_{0,(ii)}^m(\eta)$ and for any Gram matrix $\b K^m = \left[ K^m_{i,j} \right]_{i,j=1}^n =\left[ k^m(X_i,X_j) \right]_{i,j=1}^n \in \R^{n \times n}$ associated to some sequence $\left( X_i^m \right)_{i=1}^n \sim (\P^m)^{n}$, $\bar{\sigma}^m > C_1^m \sqrt{\frac{\log(n)}{n}}$ holds with $\left(\P^m\right)^{\otimes n}$-probability at least $1-\eta$, where $\bar{\sigma}^m := \frac{1}{n^2} \sum_{i,j\in [n]} K^m_{i,j}\left( 1- K^m_{i,j} \right)$.
        
    \end{assumption}
    \item[] \begin{assumption}[spectral control]\label{assum_lambda_delta}
        Let $m \in [M]$ be arbitrary and fixed. Let $d^m \ge 1$ and $a_{k^m}^2=\sup_{x^m\in \X^m}k^m(x^m,x^m) <\infty$. Let the eigenvalues (ordered in a non-increasing way) of the integral operator $T_{k^m}$ be $\left ( \lambda^m_i \right )_{i \in I^m} $ and the gap of $d^m$-th and $(d^m+1)$-th eigenvalues is defined as $\delta^m_{d^m} = \lambda^m_{d^m} - \lambda^m_{d^m+1} \ne 0$.  
        There exist universal constants $c_0^m > 0$ and $c_2^m > 0$, and for any $\eta \in (0,1)$ there exist $n_{1,(i)}^{m}(\eta)\in \Zp$ and $n_{1,(ii)}^{m}(\eta) \in \Zp$ with the following properties. (i) For all $n \ge n_{1,(i)}^{m}(\eta)$ and for any Gram matrix $\b K^m = \left[ K^m_{i,j} \right]_{i,j=1}^n =\left[ k^m_{i,j}(X_i,X_j) \right]_{i,j=1}^n \in \R^{n \times n}$ associated to some sequence $\left( X_i^m \right)_{i=1}^n \sim (\P^m)^{n}$, with eigenvalues $\left( \lambda^m_j(\b K^m) \right)_{j=1}^n$ ordered in non-increasing way, $\frac{\lambda^m_{d^m}(\b K^m)}{\Delta^m} \ge c^m_0$ holds with $\left(\P^m\right)^{\otimes n}$-probability at least $1-\eta$, with $\Delta^m$ defined in Assumption~\ref{assum_Delta}. (ii) For all $n \ge n_{1,(ii)}^{m}(\eta)$, $\delta^m_{d^m} \ge 16 c_1^m(n) \sqrt{\frac{\log \left( \frac{6n}{\eta} \right)}{n}}$ holds, where $c_1^m(n) := c_2^m \log\left( 2 + \log (2n) \right)$.

    \end{assumption}
    
    \item[] 
    \begin{assumption}[simple positive eigenvalues of $C_{k^m}$]\label{assum_eigenval_C_k} 
    Let the eigenvalues $\left ( \lambda^m_i \right )_{i \in I^m} $ of $T_{k^m}$ be simple and positive: $\lambda^m_1 > \lambda^m_2 > \cdots >0$, for all $m\in [M]$.\footnote{Assumption~\ref{assum_eigenval_C_k} implies that $\delta_{i}^m := \lambda_i^m - \lambda_{i+1}^m \ne 0$ for all $m\in [M]$ and for all $i$. \label{footnote_delta_ne_0}}
    \end{assumption}
    \item[] \begin{assumption}[simple positive eigenvalues of $\hat{C}_{k^m}$]
    \label{assum_eigenval_hat_C_k}
    Let the eigenvalues (ordered in a non-increasing way) of $\hat{C}_{k^m}$ be $\left ( \hat{\lambda}^m_i \right )_{i \in \hat{I}^m} $, which for any realization we assume to be simple ($|\hat{I}^m| =n$) and positive ($\hat{\lambda}^m_i>0$ for all $i$), for all $m \in [M]$.
    
    \end{assumption}
\end{itemize}
Assumption~\ref{assum_Delta}(i) means that the expected node degree $\Delta$ has to be larger than $\log(n)$ (up to constant multiplier), in other words, the random graph can not be too sparse. This is a common assumption used in the literature on the spectral analysis of random graphs \citep{oliveira2009concentration, tang2013universally, lyzinski2014perfect}, and allows to control the concentration of the adjacency matrix (see Lemma~\ref{concentration_dense1}). 
Assumption~\ref{assum_Delta}(ii) is the mean variance of $A_{i,j}^m$ controlling the density of $G^m$, thereby ensuring that the order of $\E \left\| \b A^m - \E \b A^m \right\|$ matches $\sqrt{\Delta}$ (as shown in Lemma~\ref{decomposition_LPRG}); matching of this order is common in the literature \citep{bandeira2016sharp, le2018concentration, benaych2020spectral}. 
While Assumption~\ref{assum_Delta}(i) is a lower bound on $\Delta$, the condition $\frac{\lambda_d(\b K)}{\Delta} \ge c_0$ in Assumption~\ref{assum_lambda_delta} is an upper bound on $\Delta$, hence the random graph is required to be not too dense. This assumption already shows up in the specific case of linear kernels \citep{athreya2018statistical} which we extend in the manuscript. The lower bound on $\delta_d$ in Assumption~\ref{assum_lambda_delta} is a technical condition helping to simplify the error bound of the adjacency spectral decomposition (in Lemma~\ref{decomposition_LPRG}). Assumptions~\ref{assum_eigenval_C_k}-\ref{assum_eigenval_hat_C_k} ensure the well-definedness of certain projection operators as detailed in Section~\ref{lemma_prove_bound1}; such assumptions are standard and widely adopted \citep{rudi2015less, sterge2020gain}.
Finally, we note that existing works \citep{oliveira2009concentration, athreya2018statistical} applying Assumptions~\ref{assum_Delta} and \ref{assum_lambda_delta} concern fixed sequences. In contrast, our Assumptions~\ref{assum_Delta} and \ref{assum_lambda_delta} are formulated in a probabilistic form. For completeness, we verify Assumptions~\ref{assum_Delta}-\ref{assum_eigenval_hat_C_k} for the Laplacian kernel and the uniform distribution; see Lemma~\ref{lemma:assumption_verify_laplacian}. 

Our next theorem describes our main result, the performance of the estimator $\widehat{\mathrm{GHSIC}_{k}}$.

\begin{theorem}[Error bound for $\widehat{\mathrm{GHSIC}_{k}} $]\label{errorbound_GHSIC1}
    Consider the LPRGs $(G^m)_{m =1}^{M}$ associated to the tuplet $\left(\X=\times_{m=1}^M\X^m, k=\otimes_{m=1}^M k^m, \P\right)$, with kernel $k^m:\X^m\times \X^m \rightarrow \R$ and $\P^m$ being the $m^{th}$ marginal of $\P$. 
    Let the embedding dimension be $d^m \in \Z^+$,
    the eigenvalues (ordered in a non-increasing way) of the integral operator $T_{k^m}=T_{k^m,\P^m}$ be $\left(\lambda^m_i\right)_{i\in I^m}$, let $\delta^m_{d^m}=\lambda^m_{d^m} - \lambda^m_{d^m+1}$ denote the gap between the $d^m$-th and $(d^m+1)$-th eigenvalue of $T_{k^m}$ with $m\in [M]$. 
    Let the canonical feature map of $k^m$ be $\phi_{k^m}(x^m)=k^m(\cdot,x^m)$, where $x^m \in \X^m$, and the (uncentered) covariance operator is $C_{k^m} = \E \left [ \phi_{k^m}(X^m) \otimes \phi_{k^m}(X^m) \right ]$ ($m\in [M]$). Define the effective dimension $\mathcal{N}_{C_{k^m}}(t) = \tr\left(C_{k^m}(C_{k^m}+tI)^{-1}\right)$, where $I$ denotes the identity operator, $t >0$, and $m\in [M]$.
    Suppose that Assumptions~\ref{assum_Delta}-\ref{assum_eigenval_hat_C_k} hold, with $n_{0,(i)}^{m}(\eta), n_{0,(ii)}^{m}(\eta), n_{1,(i)}^{m}(\eta), n_{1,(ii)}^{m}(\eta) \in \Zp$, $c^m_0>0$, $c_1^m(n) = c_2^m \log \left( 2 + \log(2n) \right)$, $c_2^m >0$, $C_0^m>0$, and $C_1^m>0$ specified therein.$^{\ref{footnote_delta_ne_0}}$ For $m \in [M]$, let $t_{k^m}(n, \eta) = \frac{9 a_{k^m}^2}{n} \log\left( \frac{3n}{\eta}\right)$ with $a_{k^m}:=\sup_{x^m\in \X^m} \sqrt{k^m\left(x^m,x^m\right)}$, and
   \begin{align}
        \tilde{n}^m_{0,A1(i)}(\eta) &:= n^m_{0,(i)}\left( \frac{\eta}{2304}\right), \quad \tilde{n}^m_{0,A1(ii)}(\eta) := n^m_{0,(ii)}\left( \frac{\eta}{1152} \right), \quad \tilde{n}^m_{0,A2(i)}(\eta) := n^m_{1,(i)}\left( \frac{\eta}{1152} \right), \\
        \tilde{n}^m_{0,A2(ii)}(\eta)
        &:= n^m_{1,(ii)}\left( \frac{\eta}{192} \right), \quad \quad \tilde{n}^m_{0,L5}(\eta) := \left( \frac{\eta}{2304} \right)^{-\frac{1}{c^m}}, \quad \quad \tilde{n}^m_{0,L6}(\eta) := \frac{4608}{\eta} ,\\
        \tilde{n}_{0,\text{else}-1}^m(\eta) & := 32 \left(c_1^m(n)\right)^2 (c_0^m)^{-2} \log \left( \frac{2304 n}{\eta} \right), \quad \quad \tilde{n}_{0,\text{else}-2}^m := e^{\frac{1}{\left( 2C_1^m - 4\sqrt{2} \right)^2}},
    \end{align}
    Assume for all $m\in [M]$ that (i) $k^m: \X^m \times \X^m \to [0,1]$ is universal (specifically, this means that $k^m$ is continuous and $\X^m$ is compact metric), and $\mathrm{supp}(\P^m)=\X^m$, (ii) $C_1^m > 2\sqrt{2}$ and $c_2^m \ge 2$, (iii) $\lambda^m_{d^m+1} \le 8 \left( \log 2 \right) \sqrt{\frac{d^m \log \left( \frac{64Mn}{\eta}  \right)}{n}}$ holds for $n \ge \left( \frac{\eta}{32M} \right)^{-\frac{1}{c^m}}$, (iv) $t_{k^m}\left(n, \frac{\eta}{8M} \right) \le \left \| C_{k^m} \right \|$, and (v) for all $\left(x_i^m \right)_{i=1}^n \subset \X^m$
    \begin{align}
        \left\| \hat{\phi}_{k^m}^{m}(x_i^m) \right\|_2 \le d_{k^m}, \label{assum:bounded_feature_map}
    \end{align}
    with $\hat{\phi}_{k^m}^{m}\left(x_i^m\right)$ ($i\in [n]$, $m\in [M]$) defined according to \eqref{eq:hat-Phi-m}; w.l.o.g.\ we assume that $d_{k^m} \ge 1$ for all $m \in [M]$.
    Then for any $\eta \in (0,1)$ with
    \begin{align}
        n &\ge \max_{m \in [M]} \Bigg\{ \tilde{n}^m_{0,A1(i)}\left( \frac{\eta}{4M} \right), \tilde{n}^m_{0,A1(ii)}\left( \frac{\eta}{4M} \right), \tilde{n}^m_{0,A2(i)}\left( \frac{\eta}{4M} \right), \tilde{n}^m_{0,A2(ii)}\left( \frac{\eta}{4M} \right), \tilde{n}^m_{0,L5}\left( \frac{\eta}{4M} \right), \\
        & \hspace{1.85cm} \tilde{n}^m_{0,L6}\left( \frac{\eta}{4M} \right), \tilde{n}_{0,\text{else-1}}^m\left( \frac{\eta}{4M} \right),\tilde{n}_{0,\text{else-2}}^m \Bigg\},
    \end{align}
    we have that
    \begin{align}
        \MoveEqLeft \left | \mathrm{GHSIC}_{k} (\P) - \widehat{\mathrm{GHSIC}_{k}} \right | \le 2 a_k \sqrt{\frac{2\log \left( \frac{16}{\eta} \right)}{n}} + 2 \sum_{m=1}^M a_{k^m} \sqrt{\frac{2\log \left( \frac{16M}{\eta} \right)}{n}}  \\
        & \hspace{-0.8cm} + 6 \sqrt{\frac{3}{2}} \sum_{m=1}^M b_{k^m} \sqrt{\mathcal{N}_{C_{k^m}}\left(t_{k^m} \left(n, \frac{\eta}{8M}\right) \right)} \left(\lambda^m_{d^m+1} + t_{k^m} \left(n, \frac{\eta}{8M}\right) \right) + 2 d_k \sum_{m=1}^M D_{k^m}\left( \frac{\eta}{4M} \right), \label{eq:GHSIC:error-bound}
    \end{align}
    holds with $\mathbb{H}$-probability at least $1-\eta$, where $\mathbb{H}$ denotes the joint distribution of the adjacency matrices $(\b A^m)_{m=1}^M$ and $(X_i)_{i=1}^n$, $d_k = \prod_{m=1}^M d_{k^m} $, $b_{k^m}= \left \| C^{-\frac{1}{2}}_{k^m} \right \|$, and
    \begin{align}
        D_{k^m}(\eta) & = 312 \sqrt{2} c_1^m(n) d^m \left( \delta^m_{d^m} \right)^{-\frac{1}{2}}  \sqrt{\frac{\log \left ( \frac{864 n}{\eta} \right ) }{n}} + 896 \frac{\left(d^m \right)^{\frac{1}{2}}\left(c_1^m(n) \right)^2}{\sqrt{c_0^m}} \left( \delta^m_{d^m} \right)^{-2} \frac{\log \left( \frac{432 n}{\eta} \right)}{n} \\
        & + 1024 \frac{\left(d^m \right)^{\frac{1}{2}}\left(c_1^m(n) \right)^3}{\sqrt{c_0^m}} \left( \delta^m_{d^m} \right)^{-2} \frac{\log^{\frac{3}{2}} \left( \frac{432 n}{\eta} \right)}{n^{\frac{3}{2}}} + 64 \frac{\left(c_1^m(n) \right)^2}{\sqrt{c_0^m}} \left( \delta^m_{d^m} \right)^{-1} \frac{\left(d^m \right)^{\frac{1}{2}} \log \left( \frac{432n}{\eta} \right)}{n} \\
        & + 4 \frac{(d^m)^{\frac{1}{2}} c_1^m(n)}{\sqrt{c_0^m}} \sqrt{\frac{\log \left( \frac{144n}{\eta}  \right)}{n}} .
    \end{align}
\end{theorem}
\tb{Remarks (assumptions)}: The usage of the already discussed Assumptions~\ref{assum_Delta}-\ref{assum_eigenval_hat_C_k} (in the derivation of our theorem) is summarized in Fig.~\ref{fig:lemm_relationship}. The remaining assumptions can be understood as follows.
\begin{itemize}
    \item Assumption (i): The universality of $(k^m)_{m=1}^M$-s guarantees that HSIC is a valid independence measure \citep[Theorem~5]{szabo2017characteristic}; the classical Mercer theorem \citep[Theorem~4.49]{steinwart08support} relies on the additional assumption $\t{supp}(\P^m) = \X^m$ ($m\in [M]$).    
    \item Assumption (ii): The mild assumptions $C_1^m > 2\sqrt{2}$ and $c_2^{m}\ge 2$ help to control (Lemma~\ref{lemma:bound_Cm}) the Lipschitz constant arising from Lemma \ref{bound_psd_trans}.
    \item Assumption (iii): This assumption means that the $(d^m+1)$-th eigenvalue of $T_{k^m}$ is small when $n$ is large enough, in the sense of  $\lambda_{d^m+1}^m = \O\left(\left( \frac{d^m \log n}{n} \right)^{1/2}\right)$. This property is satisfied in case of polynomial spectral decay; see the remark below on convergence rates. 
    \item Assumption (iv): This assumption is commonly adopted in the literature \citep{rudi2015less, sterge2020gain} and is used to control the deviation of the  covariance operator and its empirical counterpart. 
    \item Assumption (v):  This is a standard assumption in the literature \citep{xie2024entrywise}, implying for instance the widely-used assumption of bounded coherence of eigenvectors \citep{candes2011robust}.
\end{itemize}

\tb{Remarks (convergence rates)}: 
In the sequel, we assume that the embedding dimensions $\left(d^m \right)_{m=1}^M$-s can depend on the node size $n$: $d^m=d^m(n)$ for $m\in [M]$. We consider and discuss two examples: polynomial and exponential decay of the spectrum of $\left(T_{k^m}\right)_{m=1}^M$-s (or equivalently that of $\left(C_{k^m}\right)_{m=1}^M$-s). We summarize the implications (with proofs in Section~\ref{diss_opt_rate}) of the obtained error bound for two spectral decay conditions.
\begin{itemize}
[labelindent=0em,leftmargin=1.2em,topsep=0cm,partopsep=0cm,parsep=0cm,itemsep=2mm]

    \item \textbf{Polynomial spectral decay of $\left(T_{k^m}\right)_{m=1}^M$-s}: The trace class property and the non-negativity of $T_{k^m}$ mean that $\sum_{i\in I} \lambda^m_i < \infty$. Assume that (i) the  eigenvalues of $T_{k^m}$ decay polynomially, in other words 
    $\lambda^m_i = \Theta\left(i^{-\alpha_m}\right)$---which implies that $\alpha_m > 1$---and (ii) the embedding dimension is chosen as $d^m = n^{\frac{\theta_m}{\alpha_m}}$ for $0 < \theta_m \le \alpha_m$.\textsuperscript{\ref{footnote:reduction},\ref{footnote:d^m-integer}} \refstepcounter{footnote}\label{footnote:reduction}\footnotetext[\value{footnote}]{Note that to perform dimensionality reduction, the condition $\frac{\theta_m}{\alpha_m}\le 1$ is imposed for the exponent of $d^m$.}\refstepcounter{footnote}\label{footnote:d^m-integer}\footnotetext[\value{footnote}]{Notice that to keep the analysis simple, we do not apply additional rounding of $n^{\frac{\theta_m}{\alpha_m}}$ and $\frac{1}{\tau_m}\log \left(n^{\theta_m}\right) + c$.}
    In this case, the optimal choice of the parameters is $\theta_m=\frac{2\alpha_m +1}{6 \alpha_m + 5}$ ($m\in [M]$), hence the optimal embedding dimensions are $d^m = n^{\frac{2\alpha_m+1}{(6\alpha_m+5)\alpha_m}}$ ($m\in [M]$), and one can be arbitrarily close to a convergence rate of $\O \left(\log(\log(n))\frac{\log^{\frac{1}{2}}(n)}{n^{\frac{1}{3}}} \right) = \tilde{\O} \left(\frac{1}{n^{\frac{1}{3}}} \right)$ as $\alpha_m \rightarrow \infty$ for all $m\in [M]$. We also note (and prove) that $d^m=\Theta \left( \log^{\theta_m} (n) \right)$ is not enough for consistency.
    \item \textbf{Exponential spectral decay of $\left(T_{k^m}\right)_{m=1}^M$-s}: We assume now that the eigenvalues of $T_{k^m}$ decay exponentially, in other words  $\lambda^m_i = \Theta\left(e^{- \tau_m i}\right)$ for $\tau_m >0$. We also suppose that the  embedding dimension is chosen as $d^m =\frac{1}{\tau_m}\log \left(n^{\theta_m}\right) +c$ with $\theta_m >0$ and $c\ge 0$.\textsuperscript{\ref{footnote:d^m-integer}} In this case, the optimal choice is $\theta_m = \frac{1}{3}$ ($m\in [M]$), which leads to the embedding dimensions $d^m =\frac{1}{\tau_m} \log \left( n^{\frac{1}{3}} \right)+c$ ($m\in [M]$), and one obtains that the error of the GHSIC estimator is $\O\left(\frac{1}{\sqrt{\tau_m}}[\log(\log(n))]^2\frac{\log^{\frac{3}{2}}(n)}{n^{\frac{1}{3}}}\right)= \tilde{\O} \left(\frac{1}{\sqrt{\tau_m}} \frac{1}{n^{\frac{1}{3}}} \right)$. This rate is not effected by the parameters $(\tau_m)_{m=1}^M$-s. However, as the problem becomes easier (i.e.\ as $\tau_m$-s increase), the error bound becomes tighter and the embedding dimension can be chosen to be smaller by the $\left(\frac{1}{\tau_m}\right)_{m=1}^M$ factors in $\left(d^m \right)_{m=1}^M$-s.

        \item \tb{Comparison of the  polynomial and exponential spectrum}: We saw that as $\alpha_m$-s ($m \in [M]$) in the polynomial case go to $\infty$, the rate approaches that of the exponential case $\tilde{\O} \left(\frac{1}{n^{\frac{1}{3}}} \right)$. In addition, in the exponential case, the embedding dimensions are logarithmic instead of polynomial in the node size $n$ ($d^m= \frac{1}{\tau_m} \log \left(n^{\frac{1}{3}}\right) +c $ instead of $d^m= c n^{\frac{2\alpha_m+1}{(6\alpha_m+5)\alpha_m}} $, $m\in [M]$) which means an exponentially smaller embedding dimension.
        
        \item \tb{Comparison to HSIC estimators}: Classical HSIC estimators like \eqref{sample_HSIC1} achieve a rate $\O\left(\frac{1}{\sqrt{n}}\right)$, which in fact is also known to be optimal \citep{kalinke24minimax, cribeiro2026minimax}. The proposed GHSIC estimator---using only implicit information on the Gram matrices---can reach $\tilde{\O}\left(\frac{1}{n^{\frac{1}{3}}}\right)$. An important question is whether and how much this error can be improved, and whether one can reach the rate $\O\left(\frac{1}{\sqrt{n}}\right)$ (either by constructing an alternative GHSIC estimator or by a different analysis).        
\end{itemize}

\begin{table}[h!]
    \caption{Summary of our assumptions, with the constants appearing in them. The bounds in Assumption~\ref{assum_Delta}(i), Assumption~\ref{assum_Delta}(ii) and Assumption~\ref{assum_lambda_delta}(i) are probabilistic. Assumptions~\ref{assum_Delta}-\ref{assum_lambda_delta} hold for $n$ large enough. Assumptions~\ref{eq:poly-decay:lambda^m-decay}–\ref{eq:poly-decay:c(i)_assumption2} (Assumptions~\ref{eq:exp-decay:lambda^m-decay}–\ref{eq:exp-decay:c(i)_assumption1}) are used to derive the optimal embedding dimensions under polynomial and exponential eigenvalue decay, respectively, in Section~\ref{diss_opt_rate}; both sets of assumptions are assumed to hold for $n$ large enough.}
    \label{assumption_table}
    \begin{center}
    \begin{tabular}{lll}
    \toprule 
    Assumption & Constant & Content \\
    \midrule
    Assumption~\ref{assum_Delta}(i) & $C_0^m$ & $\Delta^m > C_0^m \log (n)$ \\
    Assumption~\ref{assum_Delta}(ii) & $C_1^m$ & $\bar{\sigma}^m > C_1^m \sqrt{\frac{\log (n)}{n}}$ \\
    Assumption~\ref{assum_lambda_delta}(i) & $c_0^m$ &  $\frac{\lambda^m_{d^m}(\b K^m)}{\Delta^m} \ge c^m_0$  \\
    Assumption~\ref{assum_lambda_delta}(ii) &  $c_2^m$ &  $\delta^m_{d^m} \ge 16 c_1^m(n) \sqrt{\frac{\log \left( \frac{6n}{\eta} \right)}{n}}$, $c_1^m(n) =c_2^m \log \left( 2 + \log (2n) \right) $ \\
    Assumption~\ref{assum_eigenval_C_k} &  & $C_{k^m}$ has simple and positive eigenvalues   \\
    Assumption~\ref{assum_eigenval_hat_C_k} &  & $\hat{C}_{k^m}$ has simple and positive eigenvalues   \\
    \midrule
    Assumption~\ref{eq:poly-decay:lambda^m-decay} & $c_l^m$, $c_u^m$ & $\lambda_i^m = c^m(i) i^{-\alpha_m}$, $c^m(i) \in \left[ c_l^m, c_u^m \right]$, $0<c_l^m < c_u^m$ \\
    Assumption~\ref{eq:poly-decay:c(i)_assumption1} & $\bar{c}_1^m$ & $\lim_{i \to \infty}c^m(i):= \bar{c}_1^m > c_l^m$ \\
    Assumption~\ref{eq:poly-decay:c(i)_assumption2} & $K^m$ & $\left| c^m(i) - c^m(i+1) \right| \le \frac{K^m}{i}$, $0\le K^m <\alpha_m c_l^m$ \\
    \midrule
    Assumption~\ref{eq:exp-decay:lambda^m-decay} & $\tilde{c}_l^m$, $\tilde{c}_u^m$ & $\lambda_i^m = \tilde{c}^m(i) e^{-\tau_m i}$, $\tilde{c}^m(i) \in \left[ \tilde{c}_l^m, \tilde{c}_u^m \right]$, $0< \tilde{c}_l^m < \tilde{c}_u^m$ \\
    Assumption~\ref{eq:exp-decay:c(i)_assumption1} & $\bar{c}_2^m$ & $\lim_{i \to \infty}\tilde{c}^m(i):= \bar{c}_2^m $ \\
    \bottomrule
    \end{tabular}
    \par\end{center}
\end{table}

\tb{Remarks (computational complexity)}: To analyze the computational complexity of \eqref{est_HSIC2}, we divide the calculation into three steps:
   \begin{enumerate}
   \item Firstly, we select a relatively large value $d_0 \in \Z^+$ and for each $m \in [M]$ (i) compute the $d_0$-truncated SVD of $\left(\left(\b A^m\right)^\top \b A^m\right)^{\frac{1}{2}}/n$ , (ii) estimate the rate of spectral decay from the resulting $d_0$ eigenvalues and (iii) from this rate estimate the embedding dimension $d^m$.
   \item Then, we estimate $\hat{\b K}^m$ using \eqref{eq:Km-hat} by relying on the first $\hat{d}^m$ eigenvalues and eigenvectors obtained from the previous step. If a sufficiently large $d_0$ has been selected $(d_0 \geq \hat{d}^m)$, there is no need to perform another truncated SVD in this step. Consequently, a trade-off exists between the choice of $d_0$ and the resulting running time which we explore further numerically in Section~\ref{sec:experiments}.
   
   \item Finally, we estimate $\widehat{\text{GHSIC}}_k^2$ using \eqref{est_HSIC2}. 
   \end{enumerate}
   The first step is the most computationally demanding, dominating the overall computational complexity.
   The computational cost of the $d_0$-truncated SVD of $(\mathbf{A}^m)^\top \mathbf{A}^m$ is $\mathcal{O}\left( (2q+2)d_0 n^2 \right)$ using the randomized SVD algorithm \citep[page~9]{halko2011finding}, where $q$ denotes the number of power iterations.

\section{Numerical Experiments} \label{sec:experiments}
In this section, we demonstrate the efficiency of our proposed GHSIC$_k$ estimator and independence test (relying on it) for LPRGs.\footnote{The source code replicating our experiments is available at the \href{https://github.com/wenxy123/Joint-Independence-Testing-for-Latent-Position-Random-Graphs}{GHSIC package}.} Our experiments were designed as follows.
\begin{itemize}[labelindent=0em,leftmargin=1em,topsep=0cm,partopsep=0cm,parsep=0cm,itemsep=0mm]
    \item \textbf{Experiment-1 (synthetic example, exponential spectral decay, accuracy and runtime)}: We illustrate the accuracy and the runtime of the proposed $\widehat{\t{GHSIC}_k^2}$ estimator for Gaussian latent position $X$ and Gaussian kernel $k$, an example when the eigenvalues of $T_{k^m}$-s are known \citep[Proposition~1]{shi2009data} to decay exponentially. We compared four strategies to obtain the eigenvalue decay rate, including (i) the known decay rate $\tau_m$ (highly idealized setting), (ii) the estimated decay rate $\hat{\tau}_m$ from $\b K^m/n$ (idealized setting), (iii) the estimated decay rate $\hat{\hat{\tau}}_m$ from SVD of $\left((\b A^m)^\top \b A^m\right)^{\frac{1}{2}}/n $ (slower approximation using observable quantities) and (iv) the estimated decay rate $\hat{\hat{\tau}}_{m,d_0}$ from $d_0$-truncated SVD of $\left((\b A^m)^\top \b A^m\right)^{\frac{1}{2}}/n$ (faster approximation using observable quantities). After comparing the performance of the aforementioned strategies, our investigation continues by exploring how the selection of $d_0$ influences the runtime. Additionally, we illustrate the impact on estimation accuracy of the correlation $\rho$ (in $X$), kernel parameter $\gamma$, and embedding constant $c$.
       
    \item \textbf{Experiment-2 (synthetic example, exponential spectral decay, test power)}: 
    Under the setting of Experiment-1, we investigated the test power of $\widehat{\t{GHSIC}_k^2}$ in the independence testing procedure proposed in Section~\ref{Testing_Procedure}. We investigated three eigenvalue decay approximation strategies, particularly (ii)-(iv) from above. Additionally, we compared the runtimes of (iii) and (iv), i.e., those of relying on non-idealized $\tau_m$ estimation.
    We note that Experiment-1 and Experiment-2 also serve as testbeds going (slightly) beyond the assumptions under which the $\mathrm{GHSIC}_{k}$ estimator has been analyzed (Theorem~\ref{errorbound_GHSIC1}); in this case only local compactness holds for $(\X_m)_{m=1}^M$-s.
    
    \item \textbf{Experiment-3 (fMRI data analysis, study of functional connectivity in human brain)}:  We investigated the role of different human brain functional connectivity subnetworks in natural language perception by evaluating the independence of distinct tasks within each subnetwork. This is an example when the spectral decay is polynomial; one can use the same strategy (iv) as in described in Experiment-1 to estimate the decay rates $(\alpha_m)_{m=1}^M$.
    
\end{itemize}

\subsection{Experiment-1} \label{sec:experiment-1}

We considered Gaussian $X^m$ with Gaussian kernel $k^m$, in which case it is known that the eigenvalues of $T_{k^m}$-s decay exponentially \citep[Proposition~1]{shi2009data}: $\lambda^m_i = \Theta\left(e^{- \tau_m i}\right)$, $i\in \Z^+$, $m\in [M]$. In this case the optimal embedding dimension (by Theorem~\ref{errorbound_GHSIC1}) is 
\begin{align}\label{exp_d_m}
    d^m &= \frac{1}{\tau_m} \log \left( n^{\theta_m} \right)+c, & \theta_m &=\frac{1}{3}, & c & >0.
\end{align}
The latent position variable was $X=(X^m)_{m=1}^M \sim \mathcal{N}(\b m, \bm{\Sigma})$, with $\b m = (\b m^m)_{m=1}^M \in \R^p$, and $\bm{\Sigma}=\left [ \bm{\Sigma}_{i,j} \right ]_{i,j=1}^M \in \R^{p \times p} $, where 
\begin{align}
    \R^{p_i \times p_j} \ni \bm{\Sigma}_{i,j} = \begin{cases}
 1 & \text{ if } i= j\\
 \rho & \text{ if } i \ne j
\end{cases} 
\end{align}
$p = \sum_{m=1}^M p_m$, and we chose $p_1 = \cdots = p_M = 1$. In our experiments we used $X^m\sim \mathcal{N}(0,1)$ and the Gaussian kernel $k^m(x, y) = e^{- \frac{\gamma}{2}\left (x -y \right )^2}$ ($\gamma>0$). 
Hence, the covariance $\rho$ equals to the correlation of $X_i$ and $X_j$. In this context, we refer to $\rho$ as the correlation.
It is worth noting that, according to the expression of $\lambda_i$ in Lemma~\ref{Tk_eigenvalue}, the spectral decay rate $\tau_m$ under our setting can be written as 
\begin{align}\label{exp_tau_m}
    \tau_m & =\log\left( \frac{\lambda^m_1}{\lambda^m_2}\right) = \log \left( \frac{1+ 2 \gamma + \sqrt{1+ 4 \gamma}}{2 \gamma}  \right) . 
\end{align} 
Therefore, an increase in the bandwidth $\gamma$ of the Gaussian kernel leads to an decrease in $\tau_m$, which results in a larger $d^m$ by \eqref{exp_d_m}.
In this Gaussian-Gaussian case, one can compute \citep[Lemma~1]{kalinke24minimax} the value of $\t{HSIC}_k^2$ (and thus in our case $\t{GHSIC}_k^2$), which we use to measure the performance of the estimators. In our experiments, the number of iterations in the truncated-SVD approximation was set as $q=5$ and the sample size $n$ changed from $100$ to $3000$, and for each selection of $n$ we ran $50$ simulations.

The embedding dimensions $\left(d^m\right)_{m=1}^M$ rely on the spectral decay rate, which was $\tau_m$ in the case of exponential decay; see \eqref{exp_d_m}. We investigated in total $4$ strategies to estimate $\left(d^m\right)_{m=1}^M$-s. For the first $3$ techniques, we started with the least realistic (but fastest) and moved progressively towards more and more realistic  (but slower) estimators. The fourth strategy is an approximation of the 3rd one, improving scalability. The strategies were as follows.
\begin{enumerate}
    \item For all $m\in[M]$, $\tau_m$ is assumed to be known, and hence $d^m$ can be calculated analytically.
    \item For all $m\in[M]$, $\b K^m$ is assumed to be known, and $\b K^m/n$ is used to estimate $\tau_m$. We use the first and second eigenvalue of $\b K^m/n$ to estimate $\lambda_1^m$ and $\lambda_2^m $ in \eqref{exp_tau_m}. The resulting  $\hat{\tau}_m$ is substituted to the expression of $d^m$ in \eqref{exp_d_m}, giving $\hat{d}^m$.
    \item For all $m\in[M]$, the full SVD of $\left(\left(\b A^m\right)^\top \b A^m\right)^{\frac{1}{2}}/n $ (i.e., $d_0=n$; computed from that of $\left(\b A^m\right)^\top \b A^m/n$) is used to estimate $\tau_m$. We use the first and second eigenvalue of $\left(\left(\b A^m\right)^\top \b A^m\right)^{\frac{1}{2}}/n$ to estimate $\lambda_1^m$ and $\lambda_2^m $ in \eqref{exp_tau_m}. The resulting $\hat{\hat{\tau}}_m$ is plugged into \eqref{exp_d_m} to get the embedding dimension estimate $\hat{\hat{d}}^m$.
    \item For all $m\in[M]$, the $d_0$-truncated SVD of $\left(\left(\b A^m\right)^\top \b A^m\right)^{\frac{1}{2}}/n $ ($d_0 \ll n$; computed from that of $\left(\b A^m\right)^\top \b A^m/n$) is used to estimate $\tau_m$. We use the first and second eigenvalue of $\left(\left(\b A^m\right)^\top \b A^m\right)^{\frac{1}{2}}/n$ to estimate $\lambda_1^m$ and $\lambda_2^m $ in \eqref{exp_tau_m}. The obtained $\hat{\hat{\tau}}_{m,d_0}$ is substituted to \eqref{exp_d_m} to arrive at the embedding dimension estimate $\hat{\hat{d}}^m_{d_0}$.  
    
\end{enumerate}

For the selection of $d_0$, as discussed in Section~\ref{main_result} regarding computational complexity, choosing a relatively large $d_0$ can avoid the need for another $d^m$-truncated SVD in the calculation of $\hat{\b K}^m$, albeit at the cost of increased computational burden in estimating $d^m$. Additionally, aside from $d_0$, several other parameters may influence the performance of $\widehat{\t{GHSIC}_k^2}$, including the embedding constant $c$, the correlation $\rho$, the kernel bandwidth $\gamma$ and the number of components $M$. Therefore, considering these factors alongside the four strategies for calculating $d^m$, we designed the following experiments to explore the influence of these parameters:
\begin{itemize}
    \item \textbf{Experiment-1A (changing $M$)}: We chose the correlation $\rho=0.9$, the kernel bandwidth $\gamma=2$, the $d_0=10$, the constant in the embedding dimension $c=0$, while varying the number of components $M\in \{2,3,5\}$.
    \item \textbf{Experiment-1B (changing $d_0$)}: We selected the number of components $M=2$, the correlation $\rho=0.9$, the kernel bandwidth $\gamma=2$, the constant in the embedding dimension $c=0$, while we varied $d_0$ in $\{ 2,10,20 \}$.
    \item \textbf{Experiment-1C (changing $\rho$)}: We set the number of components $M=2$, the kernel bandwidth $\gamma=2$, the $d_0=10$, the constant in the embedding dimension $c=0$, while changing $\rho \in \{ 0.1,0.5,0.9 \}$.
    \item \textbf{Experiment-1D (changing $\gamma$)}: We studied the case of number of components $M=2$, the correlation $\rho=0.9$, the $d_0=10$, the constant in the embedding dimension $c=0$, while working with kernel bandwidth $\gamma \in \{ 1,2,4 \}$.
    \item \textbf{Experiment-1E (changing $c$)}: We set the  number of components $M=2$, the correlation $\rho=0.9$, the kernel bandwidth $\gamma=4$, the $d_0=10$, while changing the constant in the embedding dimension $c \in \{ 0,2,4 \}$.
\end{itemize}

\graphicspath{{figs/}}
	\begin{figure}
		\scriptsize
		\begin{center}
			\begin{tabular}{c}			\includegraphics[width=1.0\textwidth]{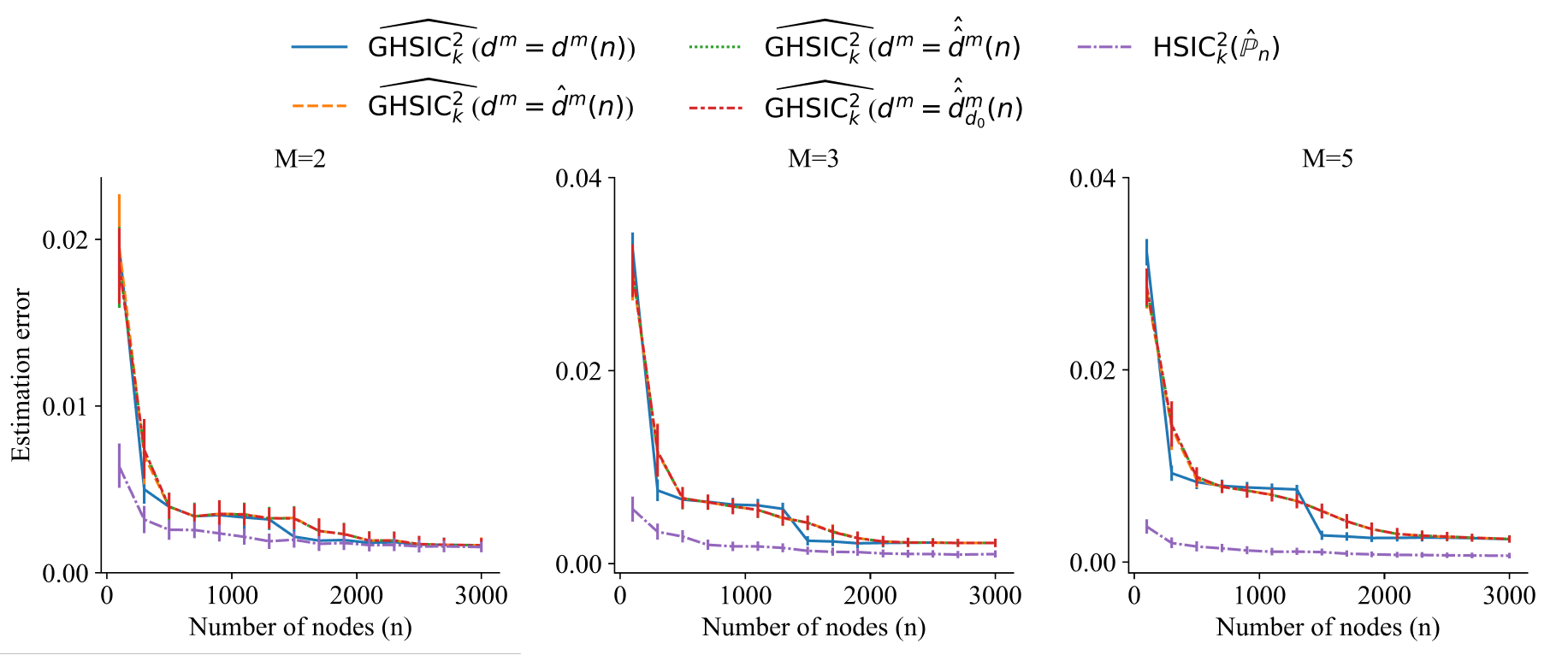}
				\\
			\end{tabular}
		\end{center}
		\caption{Estimation error for different number of components $M$ (Experiment-1A). Line: mean. Upper bar: $97.5\%$ quantile. Lower bar: $2.5\%$ quantile. }
		\label{fig_compare_M}
\end{figure}

\graphicspath{{figs/}}
	\begin{figure}
		\scriptsize
		\begin{center}
			\begin{tabular}{c}	\includegraphics[width=1.0\textwidth]{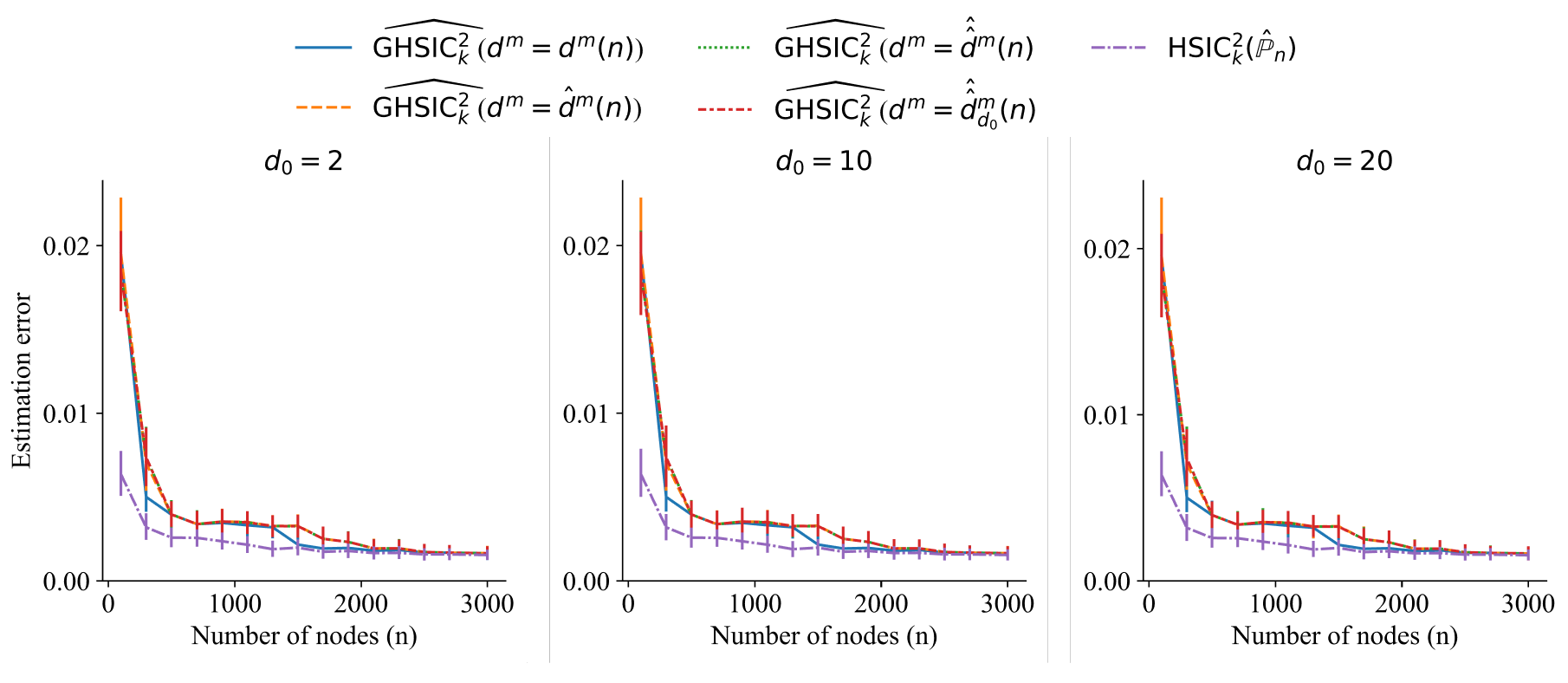} \\
            (a) Comparison of estimation error. \\
            \includegraphics[width=1.0\textwidth]{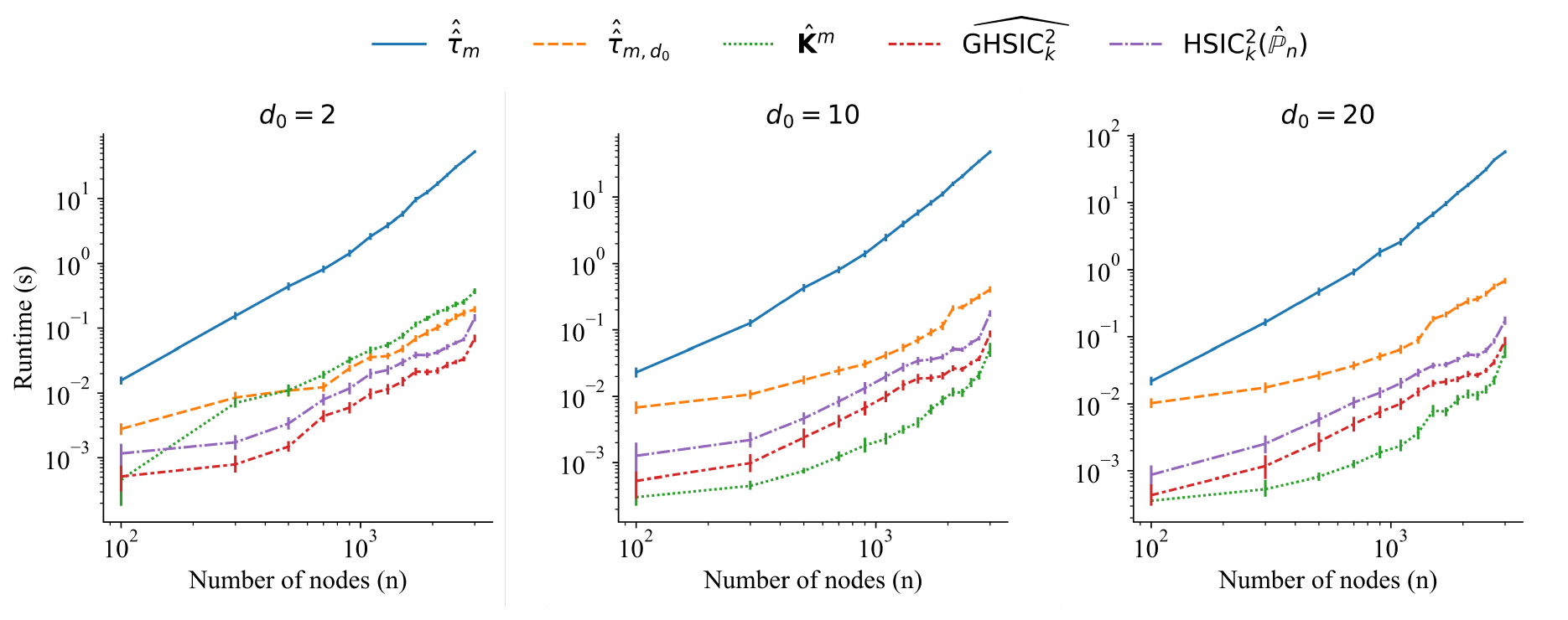}
				\\
            (b) Comparison of runtime.
			\end{tabular}
		\end{center}
		\caption{1st row: estimation error  for different selection of $d_0$ (Experiment-1B); 2nd row: associated runtime of $\t{HSIC}_k^2(\hat{\P}_n)$ and that of the $3$ estimation steps in $\widehat{\t{GHSIC}_k^2}$ under different selection of $d_0$ . The x and y axis are on logarithmic scale. The legend $\widehat{\t{GHSIC}_k^2}$ refers to the 3rd step in the discussion on computational after Theorem~\ref{errorbound_GHSIC1}.  Line: mean. Upper bar: $97.5\%$ quantile. Lower bar: $2.5\%$ quantile.}
		\label{fig_compare_time}
\end{figure}

\graphicspath{{figs/}}
	\begin{figure}
		\scriptsize
		\begin{center}
			\begin{tabular}{ccc}
				\psfig{figure=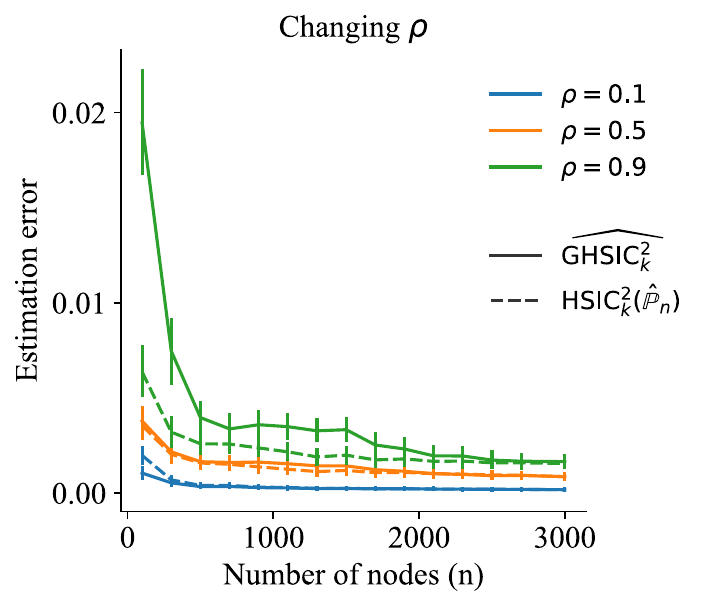,width=2in,angle=0} & \psfig{figure=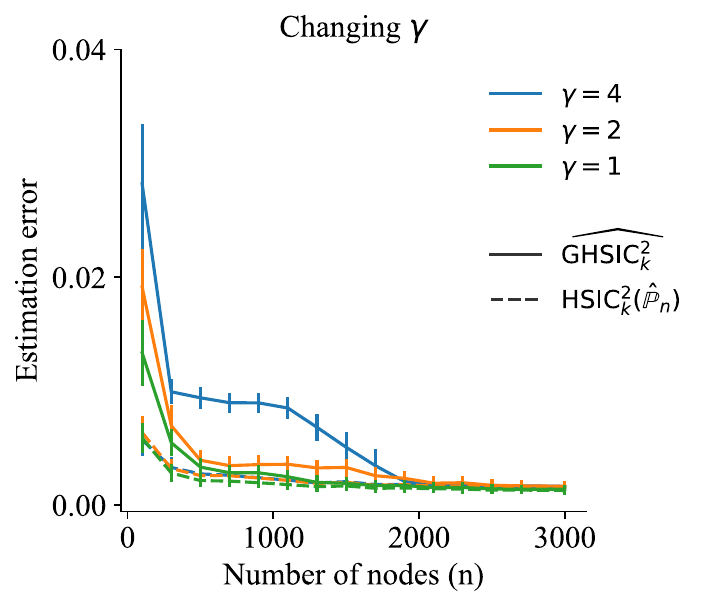,width=2in,angle=0}& \psfig{figure=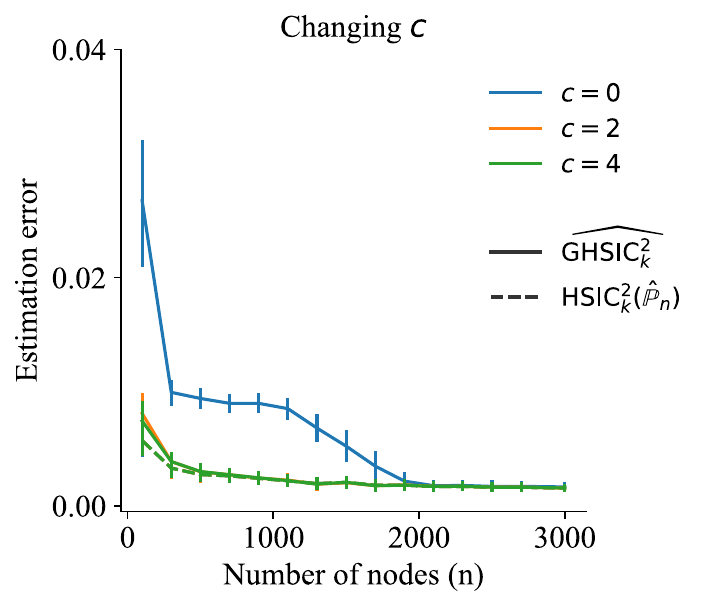,width=2in,angle=0}\\
				(a) & (b) & (c)\\
			
			\end{tabular}
		\end{center}
		\caption{Estimation error with various parameters changing. (a): changing $\rho$ (Experiment-1C). (b): varying $\gamma$ (Experiment-1D). (c): changing $c$ (Experiment-1E). Line: mean. Upper bar: $97.5\%$ quantile. Lower bar: $2.5\%$ quantile.
  }
		\label{fig_compare_para}
\end{figure}

The results of Experiment-1A are summarized in Fig.~\ref{fig_compare_M} and compared with those of the oracle estimator $\t{HSIC}_k^2(\hat{\P}_n)$. The figure demonstrates that all estimators approach the accuracy of $\t{HSIC}_k^2(\hat{\P}_n)$ as the number of nodes $n \to \infty$, showing that even the $d_0$-truncated SVD (with $\hat{\hat{\tau}}_{m,d_0}$) is an effective method for estimating $\tau_m$. However, with an increase in the number of components $M$, the estimation error of $\widehat{\t{GHSIC}_k^2}$ approaches the error of $\t{HSIC}_k^2(\hat{\P}_n)$ at a slower rate, showing  that larger $M$ leads to a larger estimation error. 

For the results of Experiment-1B, Fig.~\ref{fig_compare_time}(a) depicts the estimation error for various choices of $d_0$. The figure shows that there is no significant difference in estimation error across different values of $d_0$. The runtime of the estimators is summarized in Fig.~\ref{fig_compare_time}(b). As expected, the estimation of $\tau_m$ is the most time-demanding. However, by leveraging the $d_0$-truncated SVD approximation, we achieve a quadratic runtime in $n$, i.e.\ $\O(n^2)$ (assuming that $q$ and $d_0$ are fixed), which is identical to the computational complexity of $\t{HSIC}_k^2(\hat{\P}^n)$. The selection of $d_0$ affects both the runtime of $\hat{\hat{\tau}}_{m,d_0}$ and that of $\hat{\b K}^m$. Particularly, the estimation of $\hat{\hat{\tau}}_{m,d_0}$ is obtained through the $d_0$-truncated SVD of $(\b A^m)^\top \b A^m$, making $d_0=2$ the fastest option for estimating $\hat{\hat{\tau}}_{m,d_0}$. Regarding the calculation of $\hat{\b K}^m$, when $d^m \le d_0$, we can directly utilize the eigenvalues and eigenvectors obtained from the preceeding step, obviating the need for another $d^m$-truncated SVD. In this step, the runtime of $\hat{\b K}^m$ for $d_0=10$ and $d_0=20$ is significantly lower than for $d_0=2$. Therefore for the exponential decay rate condition, in the considered example $d_0=10$ is a practical choice.

Combining the conclusions drawn from  Fig.~\ref{fig_compare_M} and  Fig.~\ref{fig_compare_time}, in sequel we adopt the $d_0$-truncated SVD strategy in the $\widehat{\t{GHSIC}_k^2}$ estimation with $d^m=\hat{\hat{d}}^m_{d_0}(n)$ and $d_0=10$, and examine the influence of the other three parameters (Fig.~\ref{fig_compare_para}). Fig.~\ref{fig_compare_para}(a) illustrates the results of Experiment-1C, and shows that increase in the covariance $\rho$ results in larger estimation error. In Experiment-1D, as illustrated by Fig.~\ref{fig_compare_para}(b), the kernel bandwidth $\gamma$ not only affects the $d^m$  but also significantly influences the estimation error. Indeed, our method performs a low-rank approximation, and larger value of $\gamma$ leads to smaller $\tau_m$ by \eqref{exp_tau_m} which leads to
larger $d^m$ by \eqref{exp_d_m}, thereby increasing the estimation error under small sample size $n$. However, the impact of $\gamma$ can be mitigated by the constant $c$ in the embedding dimension, as explored in Experiment-1E. The results of Experiment-1E are shown in Fig.~\ref{fig_compare_para}(c). We compared the effect of the embedding constant $c$ by selecting a case where the performance was sub-optimal. We observe that the incorporation of $c$ can reduce the estimation error, and hence it is a good practical solution.
We note that from Experiment-1C to 1E, we always kept the most challenging parameter to the next experiment; particularly, in Experiment-1D we set $\rho=0.9$, followed by $\gamma=4$ in Experiment-1E, ultimately achieving good convergence.

\subsection{Experiment-2} \label{sec:experiment-2}
In this experiment, we performed independence testing of LPRGs (with permutation testing detailed in Section~\ref{Testing_Procedure} and LPRGs defined analogously to Section~\ref{sec:experiment-1}), using the $\t{GHSIC}_k^2$ approximations investigated in Section~\ref{sec:experiment-1} and used here as test statistics. We considered two cases: $\rho=0.3$ and $\rho=0.4$; in both cases the alternative ($H_1$) holds. In this experiment, we consider the three strategies [(ii)-(iv)] we mentioned in Experiment 1 to estimate $\left( d^m \right)_{m=1}^M$. We chose the kernel bandwidth $\gamma=2$, truncation dimension $d_0=10$ and embedding constant $c=0$, and $M \in \{ 2,3,5 \}$. The level of the test was $\alpha=0.05$, the number of random permutations chosen to estimate the quantile in the permutation test was
$B=500$ and we estimated the power based on $200$ repetitions.

\begin{figure}
		\scriptsize
		\begin{center}
			\begin{tabular}{c}			\includegraphics[width=0.9\textwidth]{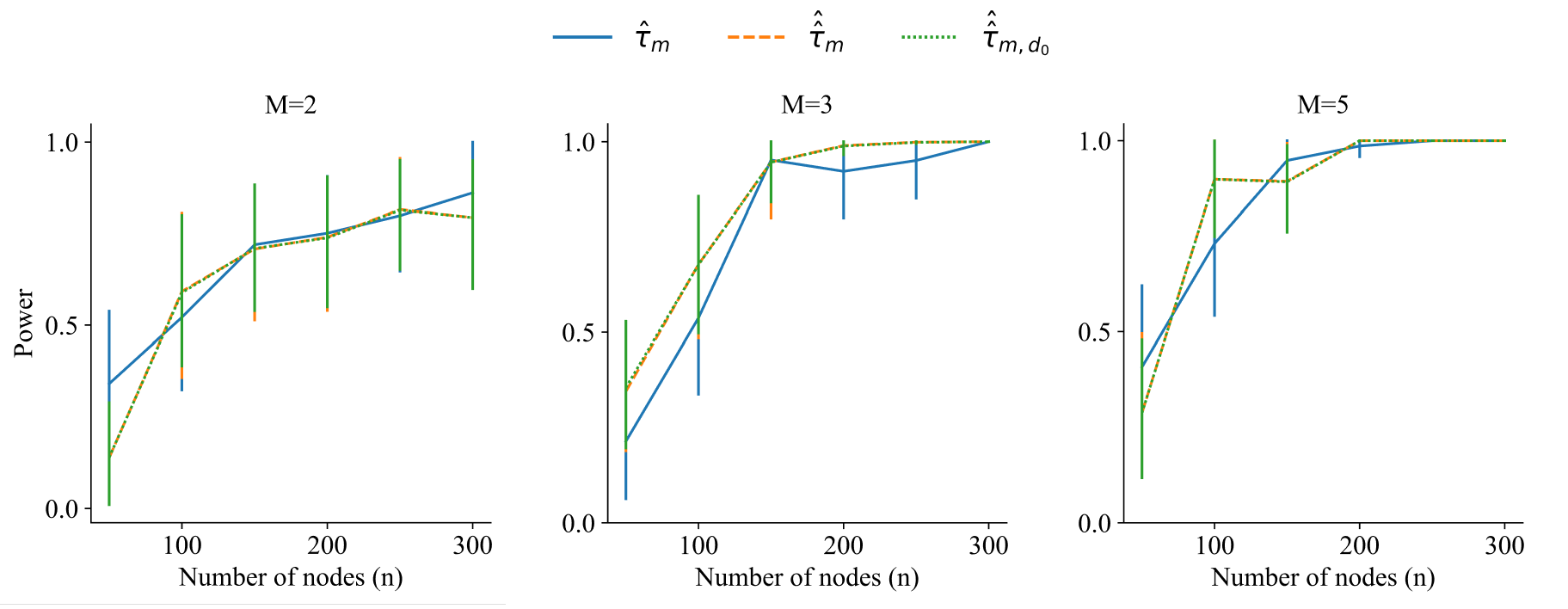}
				\\
            (a) Testing power for correlation $\rho=0.3$. \\
            \includegraphics[width=0.9\textwidth]{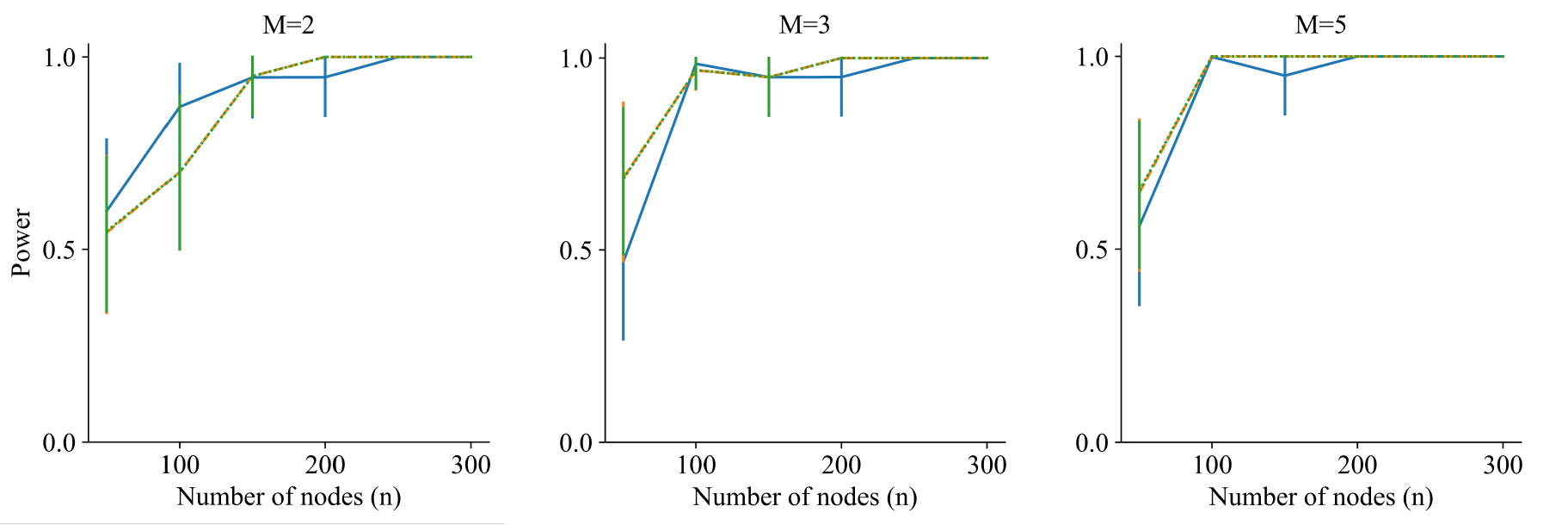}
				\\
            (b) Testing power for correlation $\rho=0.4$. \\
			\end{tabular}
		\end{center}
		\caption{Test power of $\widehat{\t{GHSIC}_k^2}$ with $\hat{d}^m=\frac{1}{\hat{\tau}_m} \log^{\frac{1}{3}}(n)$,  $\hat{\hat{d}}^m=\frac{1}{\hat{\hat{\tau}}_m} \log^{\frac{1}{3}}(n)$ and $\hat{\hat{d}}^m_{d_0}=\frac{1}{\hat{\hat{\tau}}_{m,d_0}} \log^{\frac{1}{3}}(n)$. (a): $\rho = 0.3$. (b): $\rho = 0.4$. Line: mean. Upper bar: $97.5\%$ quantile. Lower bar: $2.5\%$ quantile. }
		\label{fig_compare_power}
\end{figure}

\begin{figure}
		\scriptsize
		\begin{center}
			\begin{tabular}{c}			\includegraphics[width=0.9\textwidth]{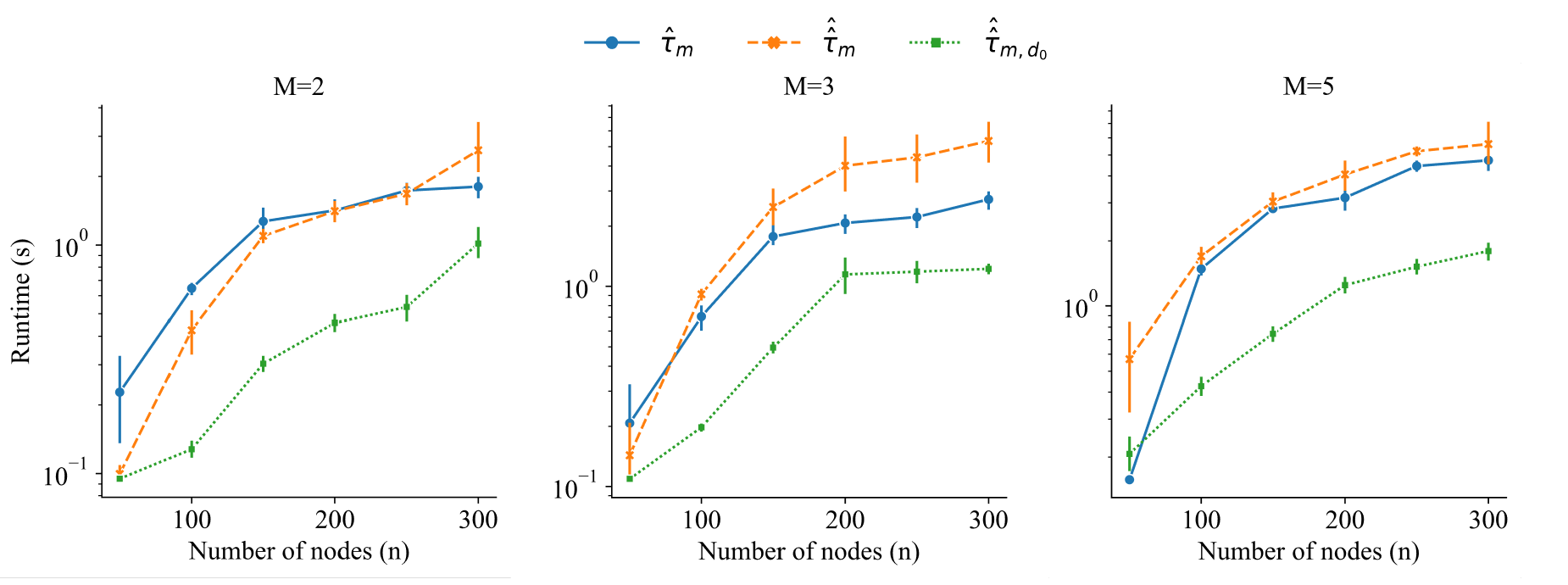}
				\\
			\end{tabular}
		\end{center}
		\caption{Runtime of calculating a quantile of $\widehat{\t{GHSIC}_k^2}$ $(\hat{d}^m=\frac{1}{\hat{\tau}_m} \log^{\frac{1}{3}}(n))$, $\widehat{\t{GHSIC}_k^2}$ $(\hat{\hat{d}}^m=\frac{1}{\hat{\hat{\tau}}_m} \log^{\frac{1}{3}}(n))$ and $\widehat{\t{GHSIC}_k^2}$ $(\hat{\hat{d}}^m_{d_0}=\frac{1}{\hat{\hat{\tau}}_{m,d_0}} \log^{\frac{1}{3}}(n))$, with logarithmic scale on the y axis.}
		\label{fig_compare_ptime}
\end{figure}

The resulting test powers are summarized in Fig.~\ref{fig_compare_power}. The figure shows that the $\widehat{\t{GHSIC}_k^2}$ calculated using $\hat{\hat{\tau}}_m$ and $\hat{\hat{\tau}}_{m,d_0}$ are almost identical (as it can be observed by their highly overlapping curves), and they perform similarly to $\hat{\tau}_m$; this is in line with the $\t{GHSIC}_k^2$ approximation results of Experiment-1A (Fig.~\ref{fig_compare_M}).
Moreover, the performance observed with $\rho=0.4$ surpassed that with $\rho=0.3$, illustrating that a smaller $\rho$ renders the testing problem more challenging.
Fig.~\ref{fig_compare_para}(a) showed that smaller value of $\rho$ results in smaller estimation error. However, it is important to note that smaller $\rho$ values also lead to numerically smaller values of $\widehat{\t{GHSIC}_k^2}$, and hence to expectedly harder testing. On the other hand, comparing the runtime of quantile calculation in the permutation test, Fig.~\ref{fig_compare_ptime} shows that the $d_0$-truncation brings a significant speedup; this figure is about $\rho=0.3$, we got similar results $\rho=0.4$.
\graphicspath{{figs/}}

\graphicspath{{figs/}}

\subsection{Experiment-3}

\begin{figure}
		\scriptsize
		\begin{center}
			\begin{tabular}{ccc}
				\psfig{figure=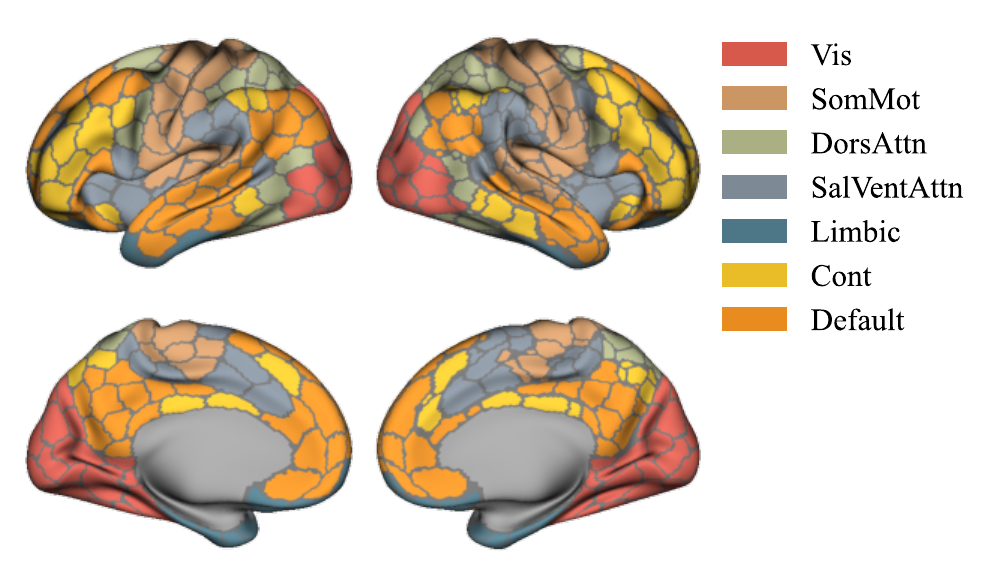,width=4in,angle=0} & \psfig{figure=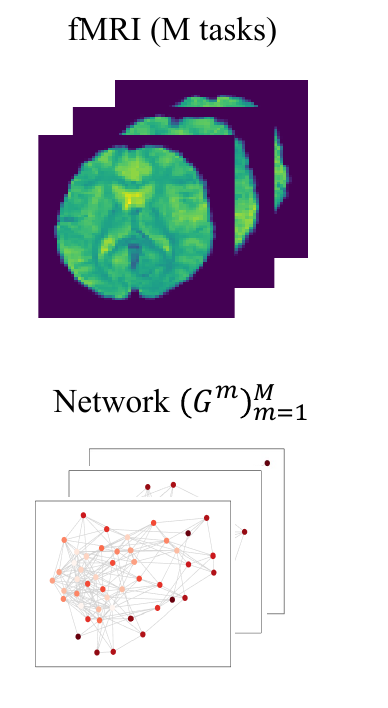,width=1.4in,angle=0}\\
				(a) Brain parcellation & (b) fMRI preprocessing\\
			
			\end{tabular}
		\end{center}
		\caption{The overview of the preprocessing of the fMRI data.}
		\label{fig_brain_area}
\end{figure}

The analysis of functional connectivity in human brain can provide insights into the neural mechanisms underlying natural language perception \citep{chai2016functional}. In this experiment, we consider the brain parcellation scheme proposed by \cite{schaefer2018local}, which subdivides the brain into $N=400$ distinct parcellations organized into $J=7$ subnetworks as defined by \cite{yeo2011organization} and illustrated in Fig.~\ref{fig_brain_area}(a). In Fig.~\ref{fig_brain_area}(a), for instance the term `Vis' refers to the visual cortex of the brain, which comprises of $61$ distinct parcellations. For each subnetwork one can construct a functional connectivity network  \citep{van2010exploring}. Specifically, for each subnetwork one can compute the correlation matrix of signals corresponding to these parcellations, and a connection between parcellations $i$ and $j$ ($i\ne j$) is established if the absolute value of their correlation exceeds a predefined threshold $t>0$. In this study, we utilized fMRI data obtained during a passive natural language listening task \citep{lebel2023natural}. The dataset comprises MRI data from $P=8$ participants across $S=5$ scanning sessions, with each session encompassing $M=6$ tasks involving passive listening to distinct narrative stories. For instance, the first session comprises of six stories titled ``Alternate Ithaca Tom", ``Avatar", ``Legacy", ``Ode to Stepfather", ``Souls" and ``Where there’s Smoke." 

\begin{figure}
		\scriptsize
		\begin{center}
			\begin{tabular}{ccc}				\psfig{figure=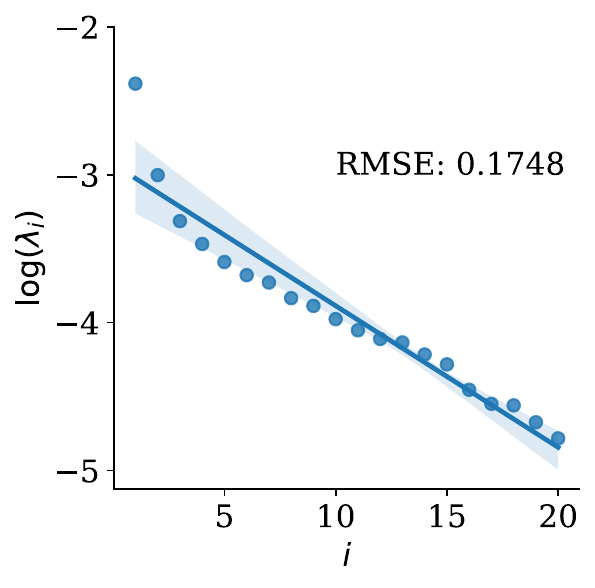,width=1.7in,angle=0} & \psfig{figure=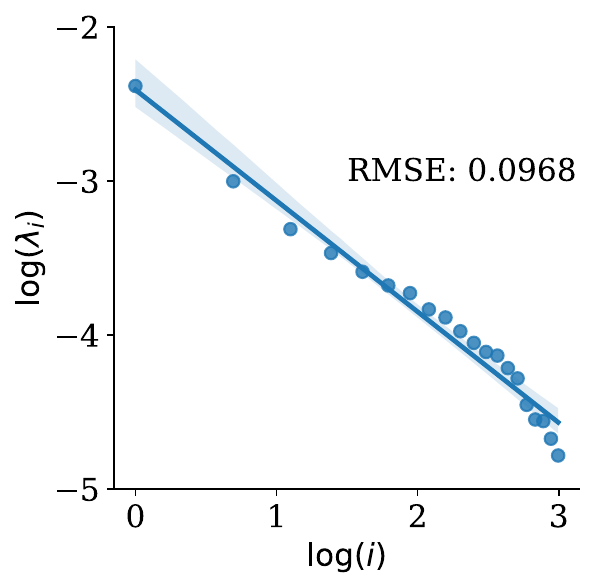,width=1.7in,angle=0}& \psfig{figure=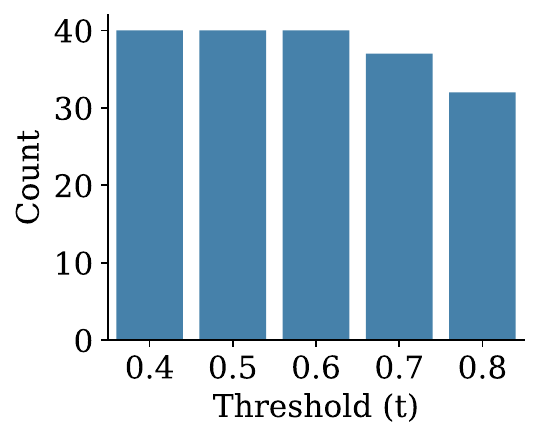,width=1.7in,angle=0}\\
				(a) Exponential decay fitting & (b) Polynomial decay fitting &
                (c) Count of acceptances of $H_1$
            \\
            \end{tabular}
		\end{center}
		\caption{Visual subnetwork. (a)-(b): checking the validity of exponential and polynomial spectral decay. (c): Number of times the alternative ($H_1$) was accepted under different threshold $t$.}
		\label{fig_Vis_result}
\end{figure}

We consider the visual subnetwork first. For fixed participant and scanning session, one can generate functional connectivity subnetworks $(G^m)_{m=1}^M$ ($M=6$); for illustration Fig.~\ref{fig_brain_area}(b). 
One can intuitively expect that the visual subnetwork is involved in multiple speech recognition task. Hence with the hypotheses
\begin{align}
    H_0&: (G^m)_{m=1}^M \t{ are independent}, &H_1&: (G^m)_{m=1}^M \t{ are dependent}
\end{align}
$H_1$ is expected to hold. Across all participants and scanning sessions, $40=P\times S= 8 \times 5$ tests were conducted, using the discussed permutation test where the number of permutations was $B=500$ with a significance level of $\alpha=0.01$. The count of acceptances of $H_1$ enables us to assess the level of activation in visual areas of the brain during language recognition tasks. These acceptance counts may be influenced by various factors, such as the selection of threshold $t$ during the construction of the functional connectivity subnetwork is a consideration. To assess this influence threshold was varied as $t \in \{0.4,0.5,0.6,0.7,0.8\}$. 
Prior to testing, we examined the spectral decay pattern of the visual subnetwork by fitting the first $20$ positive eigenvalues of $\left( (\b A^m)^\top \b A^m \right)^{\frac{1}{2}}/n$, where $\b A^m \in \R^{n\times n}$ ($m\in [M]$), $n=61$. The results of an exponential and polynomial decay fitting are illustrated (for a fixed $m \in [M]$; we got similar results for other values of $m$) in Fig.~\ref{fig_Vis_result}(a) and Fig.~\ref{fig_Vis_result}(b), respectively. 
Based on the root mean square error (RMSE) of the fitting lines, we determined that the spectral decay pattern of this dataset is better described by the polynomial decay condition. Consequently, the estimated embedding dimension is assumed to take the form $ n^{\frac{2\alpha_m+1}{(6\alpha_m+5)\alpha_m}}$, where $\alpha_m$ represents the decay rate. Similar to the case of exponential decay (see Experiment~1-2), we employed the $d_0$-truncated SVD of $\left( (\b A^m)^\top \b A^m \right)^{\frac{1}{2}}/n$ to estimate $\alpha_m$, with $d_0=10$ and $q=5$. The counts of acceptances of $H_1$ under different thresholds are illustrated in Fig.~\ref{fig_Vis_result}(c). These results show that increasing the threshold $t$ makes accepting $H_1$ more challenging.

\begin{figure}
		\scriptsize
		\begin{center}
			\begin{tabular}{c}				\psfig{figure=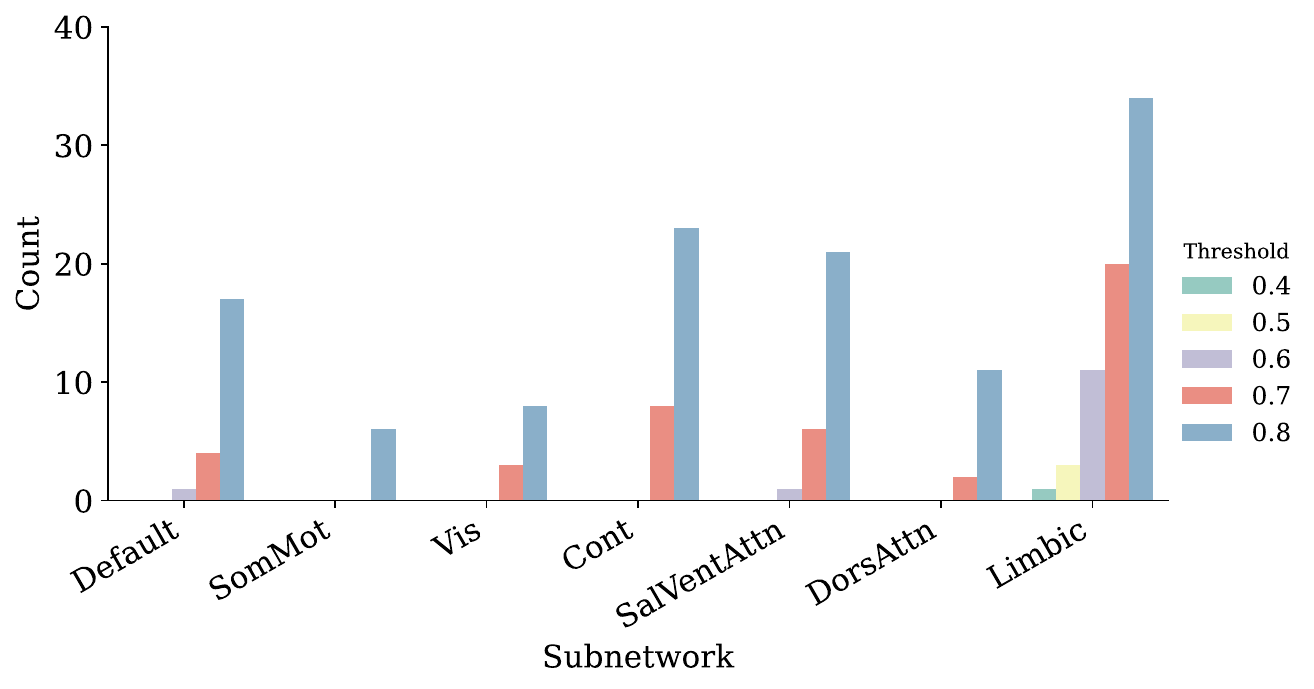,width=0.9\textwidth} 
            \end{tabular}
		\end{center}
		\caption{Count of rejection of $H_1$ for different subnetworks. The subnetworks are ordered decreasingly according to their size.}
        \label{fig_compare_count}
\end{figure}

In addition to the threshold selection, the number of nodes also influences the acceptance of $H_1$. Given that each brain subnetwork consists of a different number of parcellations, variations in the testing results indicate the impact of node count on the acceptance of $H_1$. 
In addition to the Vis subnetwork with $n=61$ nodes, we considered the six other subnetworks named SoMot, DorsAttn, SalVenAttn, Limbic, Cont, and Default, with node sizes of $n=77$, $46$, $47$, $26$, $52$, and $91$, respectively; they also followed a polynomial spectral decay. Under the same setting as with visual subnetwork, we conducted $40=P\times S=8\times 5$ tests on each other subnetwork and recorded the rejections of $H_1$. The results are illustrated in Fig.~\ref{fig_compare_count} with the subnetworks arranged according their size in a descending order. The results in the figure show that accepting $H_1$ was more feasible in subnetworks with larger node size. Notably, the Limbic subnetwork, which has the smallest node size, exhibited the lowest acceptance of $H_1$. Exceptions to this trend present opportunities for understanding this phenomenon from a neuroscience perspective. 
Indeed, recent studies have highlighted the significant activity of the SomMot subnetwork during speech perception processes \citep{schomers2016sensorimotor, trudeau2019auditory, abu2023effect}. Additionally, the participation of the visual cortex during speech perception has been confirmed \citep{besle2008visual}. Therefore, despite the Default subnetwork having the largest number of nodes, it appears that the SomMot and Vis subnetworks are the most active during natural language perception tasks; these neuroscience observations are in line with our test results obtained using GHSIC. 

These experiments demonstrate the applicability of the GHSIC measure and its proposed estimator.

\graphicspath{{figs/}}

\graphicspath{{figs/}}

\graphicspath{{figs/}}

% Acknowledgements and Disclosure of Funding should go at the end, before appendices and references

%\acks{}

\section{Conclusions}\label{sec:conclusions}

In this paper, we focused on the joint independence testing of $M\ge 2$ LPRGs, proposed an independence measure with an estimator based on adjacency spectral embeddings. We proved error bounds for the estimator, which when using the derived optimal embedding dimension  (under the spectrum of the corresponding integral operator decaying at a polynomial or exponential rate) lead to a convergence rate $\tilde{\mathcal{O}} \left( n^{-1/3}\right)$. We demonstrated the efficiency of the proposed approach both on synthetic examples and in the analysis of brain functional connectivity subnetworks arriving in case of the latter to conclusions compatible with recent results in neuroscience.

Several questions remain open including whether (i) the obtained rate $\tilde{\mathcal{O}} \left( n^{-1/3}\right)$ [slower than that of estimating HSIC at speed $\O(n^{-1/2}$)] is an artifact of the current analysis or reflects the more challenging nature of the problem due to the Gram matrices being only implicitly observable, (ii)  estimators with faster converge rate can be designed, (iii) minimax lower bounds can be derived, (iv) the proposed framework and analysis can be extended to other classes of random graph models.

%\newpage
\appendix

\begin{center}
{\large\bf Supplement}
\end{center}

The supplement is divided into four parts: Section~\ref{external_thm}  contains external lemmas that we use. Section~\ref{sec:internal-lemmas} contains all the own results of our paper, including the proof of statements in the main text and the auxiliary lemmas for the proof. Fig.~\ref{fig:lemm_relationship} summarizes the relations among the results in the supplement. Table~\ref{lemma_table} provides a summary of the external lemmas and own results, alongside with a brief description of the content of each statement.
In our independence testing experiments we used the permutation approach which we briefly summarize in Section~\ref{Testing_Procedure}. The detailed derivation of the optimal embedding dimensions is provided in Section~\ref{diss_opt_rate}.

In the supplement we will use a few \tb{additional notations}: $\Rnn$, $\P^{\otimes n}$, $S_n$, $\b P_\sigma$, $\b I_d$, $\b 0_{d_1 \times d_2}$, $[\b A_1;\b A_2]$, $U[a,b]$, $\Kernel(A)$, $\Image(A)$, $\eqd$, $A\backslash B$.

Let the set of nonnegative real numbers be denoted by $\Rnn$. The $n$-fold product of a probability measure $\P$ is denoted by $\P^{\otimes n}$. For two matrices $\b U \in \R^{n \times d}$ and $\b V \in \R^{n \times d}$ ($n\ge d$) with orthonormal columns, the $i$-th principal angle 
(\citealt[Theorem 1]{bjorck1973numerical}, \citealt[Exercise~2 on page 45]{stewart1990matrix}) (also called $i$-th canonical angle) between their column space is defined as $\theta_i := \cos^{-1} (\lambda_i) \in [0,\pi/2]$ for $i \in [d]$, where $\left( \lambda_i  \right)_{i \in [d]}$ are the singular value of $\b U^\top \b V \in \R^{d \times d}$ ordered in non-increasing fashion. Let $S_n$ be the set of permutations of $[n]$; the permutation matrix associated to $\sigma \in S_n$ is denoted by $\b P_\sigma \in \{0,1\}^{n\times n}$. The $d\times d$-sized identity matrix is $\b I_d$; the $d_1\times d_2$-sized zero matrix is written as $\b 0_{d_1\times d_2}$. The vertical concatenation of matrix $\b A_1 \in \R^{d_1 \times d}$ and $\b A_2 \in \R^{d_2 \times d}$ is $[\b A_1; \b A_2] \in \R^{(d_1+d_2) \times d}$. Let $U[a, b]$ denote the uniform distribution on the interval $[a,b]$. For a linear operator $A: X\to Y$, let its null space and image space be denoted by $\Kernel(A) = \{x\in X\,:\, Ax = 0\}$ and $\Image(A) = \{Ax\,:\,x\in X\} $, respectively. Equality in distributions is denoted by $\eqd$. We say that 2 normed spaces  $\left( X, \left\| \cdot \right\|_X \right)$ and  $\left( Y, \left\| \cdot \right\|_Y \right)$ are isometrically isomorphic if there exists a linear isomorphism $L: X \to Y$ satisfying
\begin{align}
    \left\| Lx \right\|_Y = \left\| x \right\|_X \quad \text{for all } x \in X. \label{eq:norm-preserving}
\end{align}
Such a mapping $L$ is called an isometric isomorphism \citep{li1998isometric} from $X$ to $Y$. When $\left(X,\langle \cdot, \cdot \rangle_X\right)$ and $\left(Y,\langle \cdot, \cdot \rangle_Y\right)$ are Hilbert spaces, the norm preserving property \eqref{eq:norm-preserving} of $L:X\to Y$ is equivalent to inner product preserving:
\begin{align}
\langle La, Lb \rangle_Y = \langle a,b \rangle_X\t{ for all }a,b\in X.
\end{align}
The set difference of $A$ and $B$ is written as $A\backslash B$.  

\begin{figure}%[htbp!]
    \scriptsize
		\begin{center}
			\begin{tabular}{c}			\includegraphics[width=1.0\textwidth]{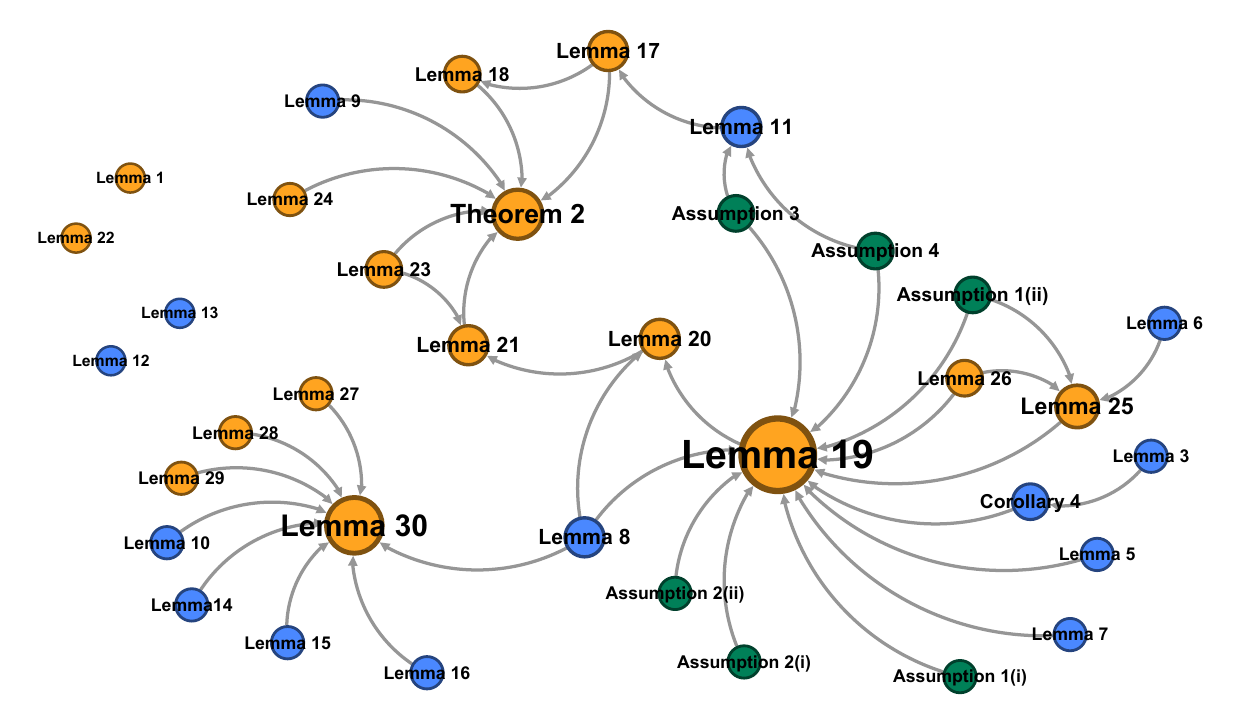}
				\\
			\end{tabular}
		\end{center}
\caption{Illustration of the relationships between results and assumptions. Blue nodes: external results. Orange nodes: own results. Green nodes: assumptions. An arrow $R_1 \rightarrow R_2$  from a result node $R_1$ to a result node $R_2$ means that result $R_1$ is used to prove result $R_2$; each result node is either blue or orange. An arrow $A\rightarrow R$ from an assumption node $A$ to a result node $R$ means that assumption A is imposed (for the first time) in result $R$.} 
    \label{fig:lemm_relationship}
\end{figure}

\begin{table}%\footnotesize
    %\captionsetup{font=footnotesize}
    \caption{Summary of  external and our own results (indicated in the 2nd column).}
    \label{lemma_table}
    \begin{center}
    \begin{tabular}{l l l}
    \toprule 
    Statement & Type & Content \\
    \midrule
    Lemma~\ref{lemma:HSIC2:Gram} & own & non-negativity of $\widehat{\mathrm{GHSIC}^2_{k}}$, estimator $\widehat{\mathrm{GHSIC}_{k}}$ \\
    Theorem~\ref{errorbound_GHSIC1} & own & error bound for $\widehat{\mathrm{GHSIC}_{k}}$ \\
    Lemma~\ref{lemm_sin_theta_theorem} & external & $\sin \left(\Theta \right)$ theorem \\
    Corollary~\ref{lemm_sin_theta_var1} & external & a variant of the $\sin \left(\Theta \right)$ theorem \\
    Lemma~\ref{concentration_dense1} & external & concentration inequality of adjacency matrices \\
    Lemma~\ref{lowerbound_A_diff_K} & external & concentration inequality of symmetric matrices \\
    Lemma~\ref{bound_psd_trans} & external &  Lipschitz continuity of $ A \mapsto \left|  A\right|$ \\
    Lemma~\ref{eigen_Tk} & external & concentration of eigenvalues of Gram matrices\\
    Lemma~\ref{concentration_ineq1} & external & concentration of bounded Hilbert space-valued random variables \\
    Lemma~\ref{mcdiarmid_ineq} & external & McDiarmid’s inequality\\
    Lemma~\ref{concentration_effective_dimension} & external & concentration of the empirical covariance operator\\
    Lemma~\ref{eigenvalue_decay} & external & effective dimension under various eigenvalue decays \\
    Lemma~\ref{Tk_eigenvalue} & external& analytical spectrum of the integral operator in the Gaussian case\\
    Lemma~\ref{lemm:totally_positive_matrix} & external & conditions for a Green's matrix to be totally positive \\
    Lemma~\ref{lemm:oscillation_matrix} & external & conditions for a matrix to be an oscillation matrix \\
    Lemma~\ref{lemm:oscillation_matrix_eigval} & external & eigenvalues of an oscillation matrix \\
    Lemma~\ref{truncated_ineq} & own & approximation of the empirical mean embedding with $d^m$-truncation \\
    Lemma~\ref{truncated_ineq2} & own & approximation of the joint empirical mean embedding with $\b d$-truncation \\
    Lemma~\ref{decomposition_LPRG} & own & approximation of the spectral features of $\b K^m$ from the adjacency matrix \\
    Lemma~\ref{concentration_adj} & own & improved approximation of the spectral features of $\b K^m$ from the  \\
    & & adjacency matrix \\    Lemma~\ref{concentration_joint_adj} & own & approximation of the joint spectral features from the adjacency matrix\\
    Lemma~\ref{lemma:f-properties} & own & properties of $f$ in \eqref{eq:f:rate} \\
    Lemma~\ref{lemma:properties-of-tensor-products} & own & properties of tensor product in a Hilbert space \\
    Lemma~\ref{lemma:isometric_embedding} & own & isometric property of the truncated feature maps\\
    Lemma~\ref{lemma:bound_Cm} & own & upper and lower bound of the Lipschitz constant in Lemma~\ref{bound_psd_trans} \\
    Lemma~\ref{lemm:combine_probs}  & own & combination of probability events \\
    Lemma~\ref{lemma:universality_verify_laplacian} & own & $c$-universality of the Laplacian kernel \\
    Lemma~\ref{lemma:assumption_verify_laplacian0} & own & Assumptions~\ref{eq:poly-decay:lambda^m-decay}-\ref{eq:poly-decay:c(i)_assumption2} for the Laplacian kernel and the uniform distribution\\
    Lemma~\ref{lemma:exp_triangular_dist} & own & expectations for the triangular distribution \\
    Lemma~\ref{lemma:assumption_verify_laplacian} & own & Assumptions~\ref{assum_Delta}-\ref{assum_eigenval_hat_C_k} for the Laplacian kernel and the uniform distribution \\
    \bottomrule
    \end{tabular}
    \par\end{center}
\end{table}

\section{External Results}\label{external_thm}
This section is about external results used in our proofs (Section~\ref{appendix_proof}).

The next lemma provides a bound of the $\sin$ of the principal angles spanned by the eigenvectors of two symmetric matrices.

\begin{lemma}[$\sin \left(\Theta \right)$ theorem; \citealt{davis1970rotation,yu2015useful}]\label{lemm_sin_theta_theorem}
Let $\b H, \hat{\b H}  \in \R^{n \times n}$ be two symmetric matrices with  $\lambda_1 \ge \lambda_2 \ge \cdots \ge \lambda_n$ and $\hat{\lambda}_1 \ge \hat{\lambda}_2 \ge \cdots \ge \hat{\lambda}_n$ eigenvalues, and $\left(\b v_i \right)_{i=1}^n$ and $\left(\hat{\b v}_i \right)_{i=1}^n$ associated orthonormal eigenvectors. 
For fixed $1 \le r \le s \le n$, let $d= s-r+1$ and let $\b V = (\b v_r, \b v_{r+1}, \dots, \b v_s) \in \R^{n \times d}$ and $\hat{\b V} = (\hat{\b v}_r, \hat{\b v}_{r+1}, \dots, \hat{\b v}_s) \in \R^{n \times d}$. Define the eigengap (see Fig.~\ref{fig:delta-demo} for a visual illustration) as
\begin{align}
    \delta &:= \inf \left \{ \left | \lambda-\hat{\lambda} \right |:\lambda \in \left [ \lambda_s,\lambda_r \right ],\hat{\lambda} \in \left( -\infty,\hat{\lambda}_{s+1}\right]\cup \left[\hat{\lambda}_{r-1},+\infty \right)   \right \} \label{eq:delta-def}\\
    &= \min \left\{  \left|  \lambda_s - \hat{\lambda}_{s+1} \right|, \left|  \lambda_r - \hat{\lambda}_{r-1} \right|  \right\},  \nonumber
\end{align}
and let $\hat{\lambda}_0 = -\infty$ and $\hat{\lambda}_{n+1} = + \infty$. If $\delta >0$ then we have
\begin{align}\label{sin_theta_theorem1}
    \left \| \sin \left(\Theta \left( \b V,\hat{\b V} \right)\right) \right \| \le \frac{\left \| \b H - \hat{\b H} \right \| }{\delta},
\end{align}
where $ \Theta \left( \b V,\hat{\b V} \right) \in \R^{n \times n}$ is a diagonal matrix whose $i$-th diagonal entry is the $i$-th principal angle between $\b V$ and $\hat{\b V}$, and the function $\sin$ is acting element-wise. 
\begin{figure}%[h]
\centering
\begin{tikzpicture}
    % Define coordinates
    \coordinate (L1) at (0,0);
    \coordinate (Ls) at (2,0);
    \coordinate (Lr) at (3,0);
    \coordinate (Ld) at (5,0);
    
    \coordinate (Lr_minus_1) at (3.5,-1);
    \coordinate (Ls_plus_1) at (1.5,-1);

    % Draw main horizontal line
    \draw[dashed] (L1) -- (Ld);
    \draw[dashed] ($(L1) - (0,1)$) -- ($(Ld) - (0,1)$);

    % Draw eigengap line
    \draw[ultra thick] (Lr) -- (Ls);

    % Draw vertical dotted lines
    \draw[dotted] (Lr) -- (Lr |- Lr_minus_1);
    \draw[dotted] (Ls) -- (Ls |- Ls_plus_1);

    % Draw bottom lines
    \draw[ultra thick] ($(L1) - (0,1)$) -- (Ls_plus_1);
    \draw[ultra thick] (Lr_minus_1) -- ($(Ld) - (0,1)$);

    \node at (Ls_plus_1) {$]$};
    \node at (Lr_minus_1) {$[$};

    % Draw intervals with notation
    \node at (Ls) {$[$};
    \node at (Lr) {$]$};
    \node at ($(Lr) + (0,0.15)$) [above] {$\lambda_r$};
    \node at ($(Ls) + (0,0.15)$) [above] {$\lambda_s$};

    \node at (Lr_minus_1) [below] {$\hat{\lambda}_{r-1}$};
    \node at (Ls_plus_1) [below] {$\hat{\lambda}_{s+1}$};
    
    \node at ($(L1) - (0,0.75)$) [left] {$\hat{\lambda}$};
    \node at ($(L1) + (0,0.25)$) [left] {$\lambda$};

\end{tikzpicture}
\caption{Illustration of the eigengap $\delta$ defined in \eqref{eq:delta-def}. \label{fig:delta-demo}}
\end{figure}
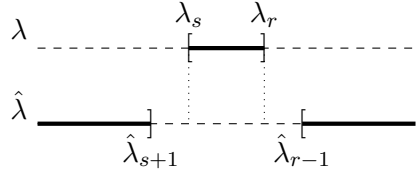
\end{lemma}
\tb{Remarks:} Let 
    \begin{align}
    \mathcal{P}_{\b V} &:= \b V \b V^\top \in \R^{n \times n}, &
    \mathcal{P}_{\hat{\b V}} &:=\hat{\b V} \hat{\b V}^\top \in \R^{n \times n} \label{eq:proj:V-Vhat}
    \end{align}
    be the projections onto the subspaces spanned by the columns of $\b V$ and $\hat{\b V}$, respectively. Then we have \cite[Theorem 4.5 on page~92 and Corollary 4.6 on page~93]{stewart1990matrix}
    \begin{align}\label{proj_sin_theta}
        \left \| \mathcal{P}_{\b V} - \mathcal{P}_{\hat{\b V}} \right \| = \left \| \sin \left(\Theta \left( \b V,\hat{\b V} \right) \right)\right \| .
    \end{align}
    Therefore, by Lemma~\ref{lemm_sin_theta_theorem} for the projections spanned by $\b V$ and $\hat{\b V}$ we have
    \begin{align}\label{sin_theta_theorem2}
        \left \| \mathcal{P}_{\b V} - \mathcal{P}_{\hat{\b V}} \right \| \le  \frac{\left \| \b H - \hat{\b H} \right \| }{\delta} .
    \end{align}

The following corollary is a consequence of Lemma~\ref{lemm_sin_theta_theorem}.

\begin{corollary}[A variant of the $\sin \left(\Theta \right)$ theorem]\label{lemm_sin_theta_var1}
Let $\b H$, $\hat{\b H}$, $(\lambda_i)_{i=1}^n$, 
$\b V$, $\hat{\b V}$, $r$ and $s$ be specified as in Lemma~\ref{lemm_sin_theta_theorem}, $\mathcal{P}_{\b V}$ and $\mathcal{P}_{\hat{\b V}}$ defined in \eqref{eq:proj:V-Vhat}. If $r=1$ and $\gamma_n := \left \| \b H - \hat{\b H} \right \| \le \frac{\left| \lambda_s - \lambda_{s+1} \right |}{2} \left( =\frac{\lambda_s - \lambda_{s+1}}{2} \right)$ then we have
\begin{align}\label{sin_theta_theorem3}
        \left \| \mathcal{P}_{\b V} - \mathcal{P}_{\hat{\b V}} \right \| \le  \frac{2 \left \| \b H - \hat{\b H} \right \| }{\lambda_s - \lambda_{s+1}}.
\end{align}
    
\end{corollary}

\begin{proof}(of Corollary~\ref{lemm_sin_theta_var1}) Let $(\hat{\lambda}_i)_{i=0}^n$ be defined as in Lemma~\ref{lemm_sin_theta_theorem}.
   When $r=1$, we have
    \begin{align}
        \delta = \min \left\{  \left | \lambda_s - \hat{\lambda}_{s+1} \right |,\left | \lambda_1 - \hat{\lambda}_0 \right |   \right\} \stackrel{(a)}{=} \left | \lambda_s - \hat{\lambda}_{s+1} \right | \stackrel{(b)}{\ge} \frac{\left | \lambda_s - \lambda_{s+1} \right |}{2} 
        \stackrel{(c)}{=} \frac{ \lambda_s - \lambda_{s+1}}{2},
        \label{eq:delta-lower-bound}
    \end{align}
    where (a) holds by using that $\hat{\lambda}_0=-\infty$. To obtain (b), one can use the Weyl's Perturbation Theorem \citep[Corollary~III.2.6 on page~63]{bhatia1997matrix}
    \begin{align}
    \left | \lambda_{s+1} - \hat{\lambda}_{s+1} \right | \le \left \| \b H - \hat{\b H} \right \|=\gamma_n.
    \end{align}
     Indeed, by using this property, one has
    \begin{align}
            & \quad \left | \lambda_s - \lambda_{s+1} \right | - \left | \lambda_s -\hat{\lambda}_{s+1} \right |  \le \left|  \left | \lambda_s - \lambda_{s+1} \right | - \left | \lambda_s -\hat{\lambda}_{s+1} \right |   \right| \\
            & \stackrel{(d)}{\le} \left| (\lambda_s - \lambda_{s+1}) - (\lambda_s -\hat{\lambda}_{s+1})   \right| = \left| \hat{\lambda}_{s+1} -\lambda_{s+1}   \right| \le \gamma_n,
        \end{align}
        where  $(d)$ was obtained by the reverse triangle inequality. Hence, one gets
        \begin{align}
            \left | \lambda_s -\hat{\lambda}_{s+1} \right | \ge \left | \lambda_s - \lambda_{s+1} \right | - \gamma_n,
        \end{align}
        which in case of $\gamma_n \le \frac{\left| \lambda_s - \lambda_{s+1} \right |}{2}$ implies that $\left| \lambda_s - \lambda_{s+1} \right | - \gamma_n \ge \frac{\left| \lambda_s - \lambda_{s+1} \right |}{2}$ and gives (b). (c) holds since $\lambda_i$-s are ordered non-increasingly. Combining the lower bound on $\delta$ obtained in \eqref{eq:delta-lower-bound} with \eqref{sin_theta_theorem2}, proves the claim in \eqref{sin_theta_theorem3}.
\end{proof}

The next lemma shows that the adjacency matrix of an LPRG can be used to approximate its hidden Gram matrix with high probability.

\begin{lemma}[Concentration inequality of adjacency matrices; Theorem 3.1 of \citealt{oliveira2009concentration}]\label{concentration_dense1}
Let $m \in [M]$ be arbitrary and fixed.
Let $\left( x^m_i \right)_{i=1}^n \subset \X^m$ be  fixed latent positions from a set $\X^m$ enriched with a continuous symmetric function $k^m:\X^m\times \X^m \to [0,1]$, giving rise to the Gram matrix $\b K^m = [K^m_{i,j}]_{i,j=1}^{n} = \left [ k^m(x^m_i,x^m_j) \right ]_{i,j=1}^{n}\in \R^{n \times n}$.
Let $\b A^m=[A^m_{i,j}]_{i,j=1}^n \in \R^{n \times n}$ be a symmetric matrix with upper-diagonal entries following a conditionally independent Bernoulli distribution given $\left( x^m_i \right)_{i=1}^n$-s and zero diagonal, in other words $A^m_{i,j} \sim \t{Bernoulli}(k^m(x^m_i,x^m_j))$ for $i < j$, $A^m_{i,j} = A^m_{j,i}$ for $i>j$, and $A^m_{i,i}=0$ for $i \in [n]$. Let $\Delta^m = \Delta^m (x^m_1,\ldots,x^m_n)= \max_{i \in [n]} \sum_{j=1}^n K^m_{i,j}$. Then for any constant $c^m > 0$, there exists a constant $C_0^m=C_0^m(c^m)>0$, independent of $n$ and $\b K^m$, such that the following holds.
If $\Delta^m > C_0^m \log (n)$,\footnote{This requirement is guaranteed by  Assumption~\ref{assum_Delta}(i) with high probability.} 
then for all $n^{-c^m} \le \eta \le \frac{1}{2}$, one has
\begin{align}
    \Q^m \left( \left \| \b A^m  - \b K^m \right \| \le 4 \sqrt{\Delta^m \log \left(  \frac{n}{\eta}\right)} \mid (X^m_i)_{i=1}^n=(x^m_i)_{i=1}^n \right) \ge 1-\eta,
\end{align}  
where $\Q^m$ denotes the distribution of the entries in $\b A^m$ given the latent positions $(x^m_i)_{i=1}^n$.

\end{lemma}

The next statement shows strong concentration around its expectation of the spectral norm of a random symmetric matrix with bounded elements and zero mean off-diagonal.
\begin{lemma}[Concentration inequality of symmetric matrices; Theorem~1.2 of \citealt{vu2007spectral}]\label{lowerbound_A_diff_K}
Let $\b A = \left[ A_{i,j} \right]_{i,j=1}^n \in \R^{n \times n}$ be a random symmetric matrix determined by the independent (not necessarily i.i.d.) entries $A_{i,j}$ ($1\le i \le j \le n$).  Assume that the following three properties hold: (i) the elements of $\b A$ are bounded, i.e.\ $\left| A_{i,j} \right| \le K$ for all $1 \le i \le j \le n$, and (ii) the off-diagonal elements of $\b A$ are zero mean, in other words $\E (A_{i,j}) =0$ for all $1\le i < j \le n$. 
Then for any $t>0$:
\begin{align}
    \Q \left[ \left|  \left\| \b A  \right\| - \E \left(\left\| \b A  \right\| \right)  \right|  \ge t \right] \le 4 e^{-\frac{t^2}{32}},
\end{align}
where $\Q$ denotes the distribution of the entries in $\b A$. 
\end{lemma}
\textbf{Remark:} 
\begin{itemize}
    \item The original statement of Theorem 1.2 in \citealt{vu2007spectral} is $ \left|  \left\| \b A  \right\| - \E \left(\left\| \b A  \right\| \right)  \right|  \ge c_1 t $ for some constant $c_1 > 0$. However, the detailed proofs by \cite{krivelevich2002approximating} show that Lemma~\ref{lowerbound_A_diff_K} holds with $c_1=1$. 
    \item According to Lemma~\ref{lowerbound_A_diff_K}, for any $\eta \in (0,1)$, we have $\left|  \left\| \b A  \right\| - \E \left(\left\| \b A  \right\| \right)  \right|  < \sqrt{32 \log \left( \frac{4}{\eta} \right)}$ holds with probability at least $1-\eta$ by letting $t=\sqrt{32 \log \left( \frac{4}{\eta} \right)}$. Then we have
    \begin{align}\label{eq:lowerbound_A_diff_K}
        - \sqrt{32 \log \left( \frac{4}{\eta} \right)} & < \left\| \b A  \right\| - \E \left(\left\| \b A  \right\| \right) < \sqrt{32 \log \left( \frac{4}{\eta} \right)} \Rightarrow \nonumber\\
        \left\| \b A  \right\| &> \E \left(\left\| \b A  \right\| \right) - \sqrt{32 \log \left( \frac{4}{\eta} \right)} \stackrel{(a)}{\ge} \E \left(\left\| \b A  \right\| \right) - \sqrt{32 \log(n)} ,
    \end{align}
    where (a) holds if $n\ge \frac{4}{\eta}$.
\end{itemize}

We use a positive semi-definite transformation of the adjacency matrices ($\b A^m \mapsto \left|\b A^m\right|:=\left(\left(\b A^m\right)^\T \b A^m\right)^{\frac{1}{2}}$, $m\in [M]$) in our estimation. The following lemma shows that this mapping is almost Lipschitz continuous, even slightly more generally with bounded linear operators.

\begin{lemma}[Lipschitz continuity of $ A \mapsto \left|  A\right|$; Theorem 6.7 in \citealt{aleksandrov2011estimates}\label{bound_psd_trans}]
Let $\mH$ be a Hilbert space. Then there exists a constant $C > 0$  such that 
\begin{align}
    \left \| \left | A \right | - \left | B \right |   \right \| \le C \log \left( 2 + \log\left ( \frac{\left \| A \right \| + \left \| B \right \| }{\left \| A-B \right \| }  \right )  \right) \left \| A-B \right \|,
\end{align}
for any two bounded linear operator $A,B:\mH \to \mH$, where  $\left | M \right | :=\left ( M^* M \right ) ^{\frac{1}{2} }$, $M \in \{A,B\}$.
\end{lemma}

The next lemma provides a bound on the difference between the eigenvalues of the Gram matrix and that of the integral operator associated to a kernel $k$.
\begin{lemma}[Concentration of eigenvalues of Gram matrices; Proposition 10 in \citealt{rosasco2010learning}]\label{eigen_Tk} Let $(X_i)_{i=1}^n$ be i.i.d.\ $\X$-valued random variables with distribution $\P$. Let $\X$ be endowed with a continuous bounded kernel $k: \X \times \X \to \R$, the associated (normalized) Gram matrix $\b K$ is given by $\b K = [K_{i,j}]_{i,j=1}^n=\left [ \frac{1}{n} k(X_i,X_j)\right ] _{i,j=1}^n \in \R^{n \times n}$. Let the eigenvalues (ordered in a non-increasing way) of the integral operator $T_k$ and the Gram matrix $\b K$ be $\left ( \lambda_i \right )_{i=1}^{\infty} $ and $\left ( \lambda_i (\b K) \right )_{i=1}^{n} $. Let $a_k^2=\sup_{x\in \X}k(x,x) <\infty$. Then for any $\tau>0$ we have
\begin{align}
    \P^{\otimes n}\left(\sum_{i=1}^\infty \left[ \lambda_i - \lambda_i(\b K) \right]^2 \le \frac{8 a_k^4 \tau}{n}\right) \ge 1-2e^{-\tau}.
\end{align}
In particular, this means that 
\begin{align}\label{eq:gap_between_eigenvals}
\P^{\otimes n}\left(\sup_{i \in \Z^+} \left | \lambda_i - \lambda_i(\b K) \right | \le 2\sqrt{2} a^2_k \sqrt{\frac{\tau}{n} }\right) \ge 1-2e^{-\tau}.
\end{align}
\end{lemma}

The following lemma shows the concentration of the empirical average around the mean for bounded random variables. 
\begin{lemma}[Concentration of bounded Hilbert space-valued random variables; Lemma E.1 in \citealt{chatalic22nystrom}]\label{concentration_ineq1} Let $(X_i)_{i=1}^n$ be i.i.d.\ random variables with zero mean in a separable Hilbert space $(\mH, \left\|\cdot\right\|_\mH)$ and assume that $\sup_{i \in [n]} \left \| X_i \right \|_{\mH}  \le A$ a.s.\ for some $A >0$. Then for any $\eta \in (0,1)$, it holds 
with probability at least $1-\eta$, that 
\begin{align}
    \left \| \frac{1}{n}\sum_{i=1}^n X_i \right \|_{\mH} \le A \sqrt{\frac{2 \log (2/\eta)}{n}}.
\end{align}

\end{lemma}

The next statement is about the concentration of independent random variables combined with a function satisfying the bounded difference property.

\begin{lemma}[McDiarmid’s inequality; \citealt{mcdiarmid1989method}]\label{mcdiarmid_ineq}
Let $(Z_i)_{i=1}^n\sim \otimes_{i=1}^n \P^i$ with $\P^i \in \mathcal{M}_{+}^1(\mathcal{Z}^i)$. Let $Y=f(Z_1,\ldots, Z_n)$, where the function $f: \times_{i=1}^n \mathcal{Z}^i \to \R$ satisfies the bounded difference property, in other words, there are constants $(c_i)_{i=1}^n\subset \R$ such that for all $i \in [n]$ one has 
\begin{align}
    \sup_{\substack{z_1\in \mathcal{Z}^1,\ldots,z_n\in \mathcal{Z}^n, \\z_i'\in \mathcal{Z}^i}} \hspace{-0.5cm} \left| f(z_1,\ldots,z_{i-1},z_i, z_{i+1},\ldots, z_n) - f(z_1,\ldots,z_{i-1},\tilde{z}_i, z_{i+1},\ldots, z_n) \right| \le c_i. \label{eq:bounded-diff-property}
\end{align}
Then for any $t>0$
\begin{align}
    \P^{\otimes n} \left ( \left | Y - \E(Y) \right | \ge t  \right ) \le 2 \exp \left ( -\frac{2 t^2}{\sum_{i=1}^n c_i^2}  \right ).
\end{align}

\end{lemma}

The following lemma provides useful bounds for the the difference of the  covariance operator and its empirical counterpart.
\begin{lemma}[Concentration of the empirical covariance operator; Lemma A.1 of \citealt{sterge2020gain}]\label{concentration_effective_dimension} Assume that $\X^m $ is a separable topological space and $k^m:\X^m \times \X^m \to \R$ is a continuous bounded kernel.
Define the uncentered covariance operator $C_{k^m}:= \E \left [ \phi_{k^m}(X^m) \otimes \phi_{k^m}(X^m) \right ]$ for $X^m \sim \P^m \in \mathcal{M}^1_{+}(\X^m)$ and its empirical estimator $\hat{C}_{k^m}:= \frac{1}{n} \sum_{i=1}^n \phi_{k^m}(X^m_i) \otimes \phi_{k^m}(X^m_i)$ for $X^m_1,\dots,X^m_n \stackrel{i.i.d}{\sim} \P^m$.
Suppose that  Assumption~\ref{assum_eigenval_C_k}-\ref{assum_eigenval_hat_C_k} hold for  $C_{k^m}$ and $\hat{C}_{k^m}$, respectively.
Let $\lambda^m_l(C_{k^m})$ be the $l^{th}$ largest eigenvalue of $C_{k^m}$ and $\lambda^m_l(\hat{C}_{k^m})$ be the $l^{th}$ largest eigenvalue of $\hat{C}_{k^m}$, where $l\le n$. Let $a_{k^m}^2=\sup_{x^m \in \X^m}k^m(x^m,x^m) <\infty$. Then for any $\delta > 0$ and $\frac{9 a_{k^m}^2}{n} \log \left( \frac{n}{\delta} \right) \le t_{k^m} \le \left \| C_{k^m} \right \| $, the followings hold. 
\begin{itemize}
    \item[(i)] $\left( \P^m \right)^{\otimes n} \left((X^m_i)_{i=1}^n\,:\,\sqrt{\frac{2}{3} } \le \left \| (C_{k^m} + t_{k^m} I)^{\frac{1}{2} } (\hat{C}_{k^m} + t_{k^m} I)^{-\frac{1}{2} } \right \| \le \sqrt{2}  \right) \ge 1-\delta$,
    \item[(ii)] $\left( \P^m \right)^{\otimes n} \left((X^m_i)_{i=1}^n\,:\,\left \| (C_{k^m} + t_{k^m} I)^{-\frac{1}{2} } (\hat{C}_{k^m} + t_{k^m} I)^{\frac{1}{2} } \right \| \le \sqrt{\frac{3}{2}}  \right) \ge 1-\delta$,
    \item[(iii)] $\left( \P^m \right)^{\otimes n} \left( (X^m_i)_{i=1}^n\,:\,\lambda^m_l(\hat{C}_{k^m})+t_{k^m} \le \frac{3}{2}(\lambda^m_l(C_{k^m}) + t_{k^m})  \right) \ge 1-\delta$,
    \item[(iv)] $\left( \P^m \right)^{\otimes n} \left( (X^m_i)_{i=1}^n\,:\,\lambda^m_l(C_{k^m})+t_{k^m} \le 2(\lambda^m_l(\hat{C}_{k^m}) + t_{k^m})  \right) \ge 1-\delta$.
\end{itemize}
\end{lemma}

The next lemma is about the order of the effective dimension  under two eigenvalue decay conditions (polynomially and exponentially decaying spectrum) on the integral operator.
\begin{lemma}[Effective dimension under various eigenvalue decays; Lemma C.9 of \citealt{sriperumbudur2022approximate}]\label{eigenvalue_decay}
For any trace class self-adjoint operator $C$ with its countable many eigenvalues $\left(\lambda_i \right)_{i \in I}  $, the following results hold.
\begin{itemize}
    \item[(i)] If $\lambda_i = \Theta (i^{-\alpha})$ for $\alpha >1$, then $\mathcal{N}_{C}(t) := \tr\left(C(C+tI)^{-1}\right) = \Theta \left(t^{-\frac{1}{\alpha}} \right)$, where $t>0$.
    \item[(ii)] If $\lambda_i= \Theta(e^{-\tau i})$ for some $\tau >0$, then $\mathcal{N}_{C}(t) = \Theta \left( \log \left( \frac{1}{t} \right)\right)$, where $t>0$.
\end{itemize}

\end{lemma}

The following result gives the analytic solution of the eigenvalues and the eigenfunctions of the integral operator associated to the  Gaussian kernel and the Gaussian distribution.
\begin{lemma}[Analytical spectrum of the integral operator in the Gaussian case; Proposition 1 of \citealt{shi2009data}]\label{Tk_eigenvalue} For $X \sim \mathcal{N}(\mu, \sigma^2)$ and a Gaussian kernel $k(x,y)=e^{-\frac{1}{2 \omega^2} (x-y)^2}$, let $\beta=\frac{2 \sigma^2}{\omega^2}$, and let $H_i$ be the $i^{th}$ order Hermite polynomial. Then the eigenvalues and eigenfunctions of $T_k$ for $i=1,2,\dots$ are given by
\begin{align}
    \lambda_i & = \sqrt{\frac{2}{1+\beta+\sqrt{1+2 \beta}}} \left( \frac{\beta}{1+\beta+\sqrt{1+2 \beta}}  \right)^{i-1}, \\
    \phi_i(x) & = \frac{(1+2 \beta)^{\frac{1}{8}}}{\sqrt{2^{i-1} (i-1)!}} e^{-\frac{(x-\mu)^2}{2 \sigma^2} \frac{\sqrt{1+2 \beta}-1}{2}} H_{i-1} \left( \left( \frac{1}{4} + \frac{\beta}{2} \right)^{\frac{1}{4}} \frac{x-\mu}{\sigma} \right).
\end{align}
    
\end{lemma}
\textbf{Remark:} Let us choose $X^m \sim \mathcal{N}(0,1)$, and the kernel function is defined as $k^m(x,y)=e^{-\frac{\gamma}{2}(x-y)^2}$. Thus, $\sigma=1$, $\gamma =\frac{1}{\omega^2}$,  and 
$\beta=\frac{2\sigma^2}{\omega^2} = 2\gamma$. According to Lemma~\ref{Tk_eigenvalue}, this means that 
\begin{align}
\lambda_1 &= cq^{1-1}= c, &
\lambda_2 &= cq^{2-1}= cq  \label{eq:lambda-1-2}
\end{align}
with $c=\sqrt{\frac{2}{1+\beta+\sqrt{1+2 \beta}}}$ and $q=\frac{\beta}{1+\beta+\sqrt{1+2 \beta}}$. Then, by \eqref{eq:lambda-1-2} and $\lambda_i = \Theta\left(e^{-i\tau_m}\right)$ we have
\begin{align}
    \frac{\lambda_1}{\lambda_2} &= \frac{1}{q} = \frac{e^{-\tau_m}}{e^{-2\tau_m}} = e^{-\tau_m + 2\tau_m}= e^{\tau_m} \Rightarrow\\
    \tau_m &= \log \left( \frac{1}{q} \right) = \log \left( \frac{1+\beta+\sqrt{1+2 \beta}}{\beta} \right) = \log \left( \frac{1+2 \gamma+\sqrt{1+4 \gamma}}{2 \gamma} \right).
\end{align}

The following statement provides a characterization of when the Green's matrix is totally positive matrix.

\begin{lemma}[Conditions for a Green's matrix to be totally positive; Theorem 4.2 of \citealt{pinkus2010totally}]\label{lemm:totally_positive_matrix}
    Assume two sequences $\left\{ c_i \right\}_{i=1}^n$ and $\left\{ d_i \right\}_{i=1}^n$ with nonzero values (in other words, $c_i \neq 0$ and  $d_i \neq 0$ for all $i\in [n]$), and let the associated Green's matrix be defined as $ \b A = \left[ A_{i,j} \right]_{i,j=1}^n \in \R^{n \times n}$ where $A_{i,j} = c_{\min \{i,j\}} d_{\max \{i,j\}}$. Then $\b A$ is totally positive\footnote{
    A matrix $\b A = \left[A_{i,j} \right]_{i,j=1}^n \in \R^{n \times n}$ is said to be totally positive if all its minors are nonnegative \citep[Definition~1.1]{pinkus2010totally}, i.e., for every $p \in [n]$ and all index sets $1 \le i_1 < \ldots < i_p \le n$ and $1 \le j_1 < \ldots < j_p \le n$, the submatrix $\b B = [A_{i_k,j_l}]_{k,l=1}^p$ satisfies $\det(\b B) \ge 0$. It is said to be strictly totally positive if $\det(\b B) > 0$ always holds.} if and only if the $\left\{ c_i \right\}_{i=1}^n$ and $\left\{ d_i \right\}_{i=1}^n$ are all of one strict sign (in other words, $\sign(c_1)=\ldots=\sign(c_n)=\sign(d_1)=\ldots=\sign(d_n)$) and $0 < \frac{c_1}{d_1} \le \ldots \le \frac{c_n}{d_n}$.
\end{lemma}

The next lemma yields a characterization of when a totally positive matrix is an oscillation matrix.

\begin{lemma}[Conditions for a matrix to be an oscillation matrix; Theorem 5.2 of \citealt{pinkus2010totally}]\label{lemm:oscillation_matrix}
    A matrix $\b A = \left[ A_{i,j} \right]_{i,j=1}^n \in \R^{n \times n}$ is an oscillation matrix\footnote{A matrix $\b A \in \R^{n \times n}$ is said to be an oscillation matrix if $\b A$ is totally positive and some (positive integer) power of $\b A$ is strictly totally positive \citep[Definition~5.1]{pinkus2010totally}.} if and only if $\b A$ is totally positive, nonsingular, and $A_{i,i+1}, A_{i+1,i} > 0$ for $i = 1, \ldots , n-1$.
\end{lemma}

According to the following lemma the eigenvalues of an oscillation matrix are positive and simple.

\begin{lemma}[Eigenvalues of an oscillation matrix; Theorem 5.3 of \citealt{pinkus2010totally}]\label{lemm:oscillation_matrix_eigval}
    Let $\b A\in \R^{n\times n}$ be an oscillation matrix. Then the $n$ eigenvalues of $\b A$ are positive and simple. 
\end{lemma}

\section{Own Results}\label{sec:internal-lemmas}

Section~\ref{lemma_prove_bound1} is about own auxiliary results used in our proofs (Section~\ref{appendix_proof}) and their discussions.

\subsection{Lemmas to the Proof of Theorem~\ref{errorbound_GHSIC1}}\label{lemma_prove_bound1}

Section~\ref{lemma_prove_bound1} can be divided into two parts. In the first part, we show an error bound for the $d^m$-truncated approximation of the feature map of $k^m$ (Lemma~\ref{truncated_ineq}); we then extend this result to the tensor product kernel $k=\otimes_{m=1}^M k^m$ (Lemma~\ref{truncated_ineq2}). In the second part, we bound the error between the $d^m$-truncated spectral decomposition of $\left( (\b A^m)^\top \b A^m \right)^{\frac{1}{2}}$ and $\b K^m$ (Lemma~\ref{concentration_adj}) by using the decomposition of the error bound provided in Lemma~\ref{decomposition_LPRG}. Subsequently, we extend these results to the multiple component case in Lemma~\ref{concentration_joint_adj}. The discussion of Theorem~\ref{errorbound_GHSIC1} relies on Lemma~\ref{lemma:f-properties}. Lemma~\ref{lemma:properties-of-tensor-products} is an auxiliary result on the norm of tensor products in Hilbert spaces; it is used in the proof of Lemma~\ref{concentration_joint_adj}. Lemma~\ref{lemma:isometric_embedding} is about the isometric property of the truncated feature maps. Lemma~\ref{lemma:bound_Cm} provides upper and lower bound for the Lipschitz constant in Lemma~\ref{bound_psd_trans}; it is used in the proof of Lemma~\ref{decomposition_LPRG}.

We start with a few definitions.
\begin{itemize}[labelindent=0em,leftmargin=1em,topsep=0cm,partopsep=0cm,parsep=0cm,itemsep=1mm]
    \item \textbf{Definition of the approximated feature map}: Let us fix an arbitrary $m\in [M]$. Recall that $k^m:\X^m \times \X^m \to [0,1]$ is a continuous kernel on the compact metric space $\X^m$ and $T_{k^m}$ is the integral operator associated to $k^m$. $T_{k^m}$ is a positive and trace-class operator with eigenvalues $\lambda^m_1 \ge \lambda^m_2 \ge \dots \ge 0$ and countable orthonormal eigenvectors $\left(\psi^m_i\right)_{i\in I^{m}}$. One can assume without loss of generality \citep[page~149]{steinwart08support} that $\left(\psi^m_i\right)_{i\in I^{m}}\subset \mH_{k^m}$, $S_{k^m}\psi^m_i = \lambda_i^m \psi^m_i$, and $\left(\sqrt{\lambda_i^m}\psi^m_i\right)_{i\in I^{m}}\subset \mH_{k^m}$ forms an ONS in  $\mH_{k^m}$.
    Then, provided that $\mathrm{supp}(\P^m)=\X^m$, by the Mercer's theorem \citep[Theorem~4.49]{steinwart08support} 
    \begin{align}
  k^m(x^m,\tilde{x}^m)=\sum_{i \in I^{m}} \lambda^m_i \psi^m_i (x^m) \psi^m_i (\tilde{x}^m), 
    \quad x^m, \tilde{x}^m \in \X^m.
    \end{align}
    Let 
    \begin{align}
    C_{k^m,\J} & = \E_{\tilde{X}^m \sim \J} \left [ k^m(\cdot,\tilde{X}^m) \otimes k^m(\cdot,\tilde{X}^m) \right ];
    \end{align}
    when $\J= \P^m$ and $\J = \hat{\P}^m_n$, we use the shorthands
    \begin{align}
    C_{k^m} & := C_{k^m,\P^m}= \E_{\tilde{X}^m \sim \P^m} \left [ k^m(\cdot,\tilde{X}^m) \otimes k^m(\cdot,\tilde{X}^m) \right ],\\ 
    \hat{C}_{k^m} &:= C_{k^m,\hat{\P}_n^m} = \frac{1}{n} \sum_{i=1}^n k^m(\cdot,\tilde{X}^m_i) \otimes k^m(\cdot,\tilde{X}^m_i). \label{eq:C-emp-m}
    \end{align} 
    According to the note above, $C_{k^m}$ has a spectral decomposition
    \begin{align}
    C_{k^m}=\sum_{i \in I^m} \lambda_i^m \underbrace{\left(\sqrt{\lambda_i^m}\psi^m_i\right)}_{=:f_i^m} \otimes \underbrace{\left(\sqrt{\lambda_i^m}\psi^m_i\right)}_{=f_i^m}.
    \end{align}
    Having introduced this notation, we define the projection operator associated to $C_{k^m}$ and to truncated level $d^m\in \Z^+$  as
    \begin{align}
   P^m_{d^m}\left(C_{k^m}\right)= \sum_{i=1}^{d^m} f^m_i \otimes f^m_i: \mH_{k^m} \rightarrow \Span\left(f^m_i \,:\, i\in \left[d^m\right]\right) \subset  \mH_{k^m},\label{eq:projection-op_Ck} 
    \end{align}
    where Assumption \ref{assum_eigenval_C_k} ensures that the $P^m_{d^m}\left(C_{k^m}\right)$ is well-defined for any $d^m \in \Z^+$. 
    By the spectral theorem, the empirical covariance operator can be written in the form
    \begin{align}\label{eq:C-emp_decompose}
     \hat{C}_{k^m} &= \sum_{i \in \hat{I}^m}\hat{\lambda}_i^m \hat{f}_i^m \otimes \hat{f}_i^m,
    \end{align}
    where $\hat{\lambda}_i^m$ and $\hat{f}_i^m$ are the (decreasingly ordered) non-negative eigenvalues and corresponding orthogonal eigenvectors of $\hat{C}_{k^m}$. Let us now define the projection operator associated to $\hat{C}_{k^m}$ and to truncation level $d^m \le |\hat{I}^m|=n$ (where the equality holds by Assumption~\ref{assum_eigenval_hat_C_k}) as:
    \begin{align}
     P^m_{d^m}\left(\hat{C}_{k^m}\right)= \sum_{i=1}^{d^m} \hat{f}^m_i \otimes \hat{f}^m_i: \mH_{k^m} \rightarrow \Span\left(\hat{f}^m_i \,:\, i\in \left[d^m\right]\right) \subset  \mH_{k^m}, \label{eq:projection-op} 
    \end{align}  
        and let the truncation of $k^m(\cdot,z)$ for all $z \in \X^m$ be similarly be defined as 
    \begin{align}
k^m_{d^m}\left(\cdot,z\right) &:= P^{m}_{d^m} \left(\hat{C}_{k^m}\right)k^m(\cdot,z) \in \mH_{k^m}.\label{eq:def:km}
    \end{align}
    With these notations at hand, and with $(X^m_i)_{i=1}^n \stackrel{\t{i.i.d.}}{\sim} \P^m$, let 
    \begin{align}
    \tilde{\mu}_{k^m} \left(\hat{\P}^m_n\right) & = \frac{1}{n} \sum_{i=1}^n k^m \left(\cdot, X^m_i\right)\in \mH_{k^m}, & \tilde{\mu}_{k^m,d^m} \left(\hat{\P}^m_n\right) &= \frac{1}{n} \sum_{i=1}^n k^m_{d^m} \left(\cdot, X^m_i\right) \in \mH_{k^m}.  \label{eq:mu-tilde-and-truncated}
    \end{align}
    \item \textbf{Definition of the reconstruction error}: Recall that for any orthogonal projection $P$, the following two properties hold:
    \begin{align}
    \left(P\right)^* &= P, \label{eq:P:self-adjoint}\\
    \left(P\right)^2 &= P. \label{eq:P:idempotent}
    \end{align}
    Let the reconstruction error of $k^m(\cdot,X^m)$ w.r.t.\ a  probability measure $\Q \in \mathcal{M}_{+}^1(\X^m)$ and an orthogonal projection $P$ be defined as
    \begin{align}
    R_{\Q}(P) &:= \E_{X^m \sim \Q} \left \| (I-P) k^m(\cdot,X^m) \right \|^2_{\mH_{k^m}}.
    \end{align}    

    Then we have\footnote{The relation \eqref{deduce_RE} has been derived in \cite{sterge2020gain}; here we included it for self-containedness with additional details, as we use also parts of the derivation in Lemma~\ref{truncated_ineq2}.}
    \begin{align}\label{deduce_RE}
     \lefteqn{R_{\Q}\left(P\right) = \E_{X^m \sim \Q} \left \| (I-P) k^m(\cdot,X^m) \right \|^2_{\mH_{k^m}}} \\
     &\stackrel{(a)}{=} \E_{X^m\sim \Q} \left \langle \left(I-P\right) k^m(\cdot,X^m),\left(I-P\right) k^m(\cdot,X^m) \right \rangle_{\mH_{k^m}}  \nonumber\\ 
    &\stackrel{(b)}{=} \E_{X^m\sim \Q}  \langle\left(I-P\right)^* \left(I-P\right)  k^m(\cdot,X^m), k^m(\cdot,X^m)\rangle_{\mH_{k^m}} \nonumber\\
    &\stackrel{(c)}{=} \E_{X^m\sim \Q}  \left \langle \left(I-P\right) k^m(\cdot,X^m), k^m(\cdot,X^m) \right \rangle_{\mH_{k^m}} \\
    & \stackrel{(d)}{=} \E_{X^m\sim \Q} \left \langle I-P, k^m(\cdot,X^m) \otimes k^m(\cdot,X^m) \right \rangle_{\t{HS}\left(\mH_{k^m}\right)}\\
    & \stackrel{(e)}{=} \left \langle (I-P), \E_{X^m\sim \Q} \left [ k^m(\cdot,X^m) \otimes k^m(\cdot,X^m) \right ]  \right \rangle_{\t{HS}\left(\mH_{k^m}\right)} \\
    &\stackrel{(f)}{=} \left \langle (I-P), C_{k^m,\Q}\right \rangle_{\t{HS}\left(\mH_{k^m}\right)} 
    \stackrel{(g)}{=} \tr \left [ (I-P) C_{k^m,\Q} \right ] \nonumber\\
    & \stackrel{(h)}{=} \tr \left [ C_{k^m,\Q}^{1/2}(I-P)(I-P) C_{k^m,\Q}^{1/2} \right ]
    \stackrel{(i)}{=} \left \| (I-P) C_{k^m,\Q}^{1/2} \right \|^2_{\t{HS}\left(\mH_{k^m}\right)}.\nonumber
    \end{align}

    In (a) we used that $\left\|h\right\|_{\mH_{k^m}} = \langle h,h\rangle_{\mH_{k^m}}$ ($  h\in \mH_{k^m}$), (b) follows from the definition of the adjoint operator, (c) comes from $\left( I-P\right)^* (I-P) \stackrel{(k)}{=} \left(I-P\right)^2 = I-2P+\left(P\right)^2 \stackrel{(l)}{=} I-2P+P = I-P$ where (k) is implied by the linearity of the adjoint and by \eqref{eq:P:self-adjoint} and (l) follows from \eqref{eq:P:idempotent},
    in (d) we used that $\left \langle A f,g \right \rangle_{\mH_{k^m}}=\left \langle A, f \otimes g \right \rangle_{\t{HS}\left(\mH_{k^m}\right)}$ ($f,g\in \mH_{k^m}$, $A\in \t{HS}\left(\mH_{k^m}\right)$), in (e) the expectation and the inner product were swapped \citep[(A.32)]{steinwart08support},
    in (f)  we applied the definition of $C_{k^m,\Q}$, (g) is implied by $\left \langle A,B \right \rangle_{\t{HS}\left(\mH_{k^m}\right)} = \tr (A^* B)$, the linearity of the adjoint, and \eqref{eq:P:self-adjoint},
    in (h) we used the property $\left(I-P\right)^2 = I-P$ already seen in (c), the uniquely existing square root of $C_{k^m,\Q}$, and the cyclic invariance of the trace, (i) follows from the self adjointness of $C_{k^m,\Q}$ and that of $I-P$, and from $\left \langle A,B \right \rangle_{\t{HS}\left(\mH_{k^m}\right)} = \tr (A^* B)$. Specifically, by choosing $P$ to be $P_{d^m}^m \left( C_{k^m} \right)$ and $P_{d^m}^m \left( \hat{C}_{k^m} \right) $,  we got that
    \begin{align}
        R_{\Q} \left( P_{d^m}^m \left( C_{k^m} \right)  \right) & = \E_{X^m \sim \Q} \left \| \left(I-P^{m}_{d^m}\left( C_{k^m}  \right) \right) k^m(\cdot,X^m) \right \|^2_{\mH_{k^m}}, \\
        R_{\Q} \left( P_{d^m}^m \left( \hat{C}_{k^m} \right)  \right) &= \E_{X^m \sim \Q} \left \| \left(I-P^{m}_{d^m}\left( \hat{C}_{k^m}  \right) \right) k^m(\cdot,X^m) \right \|^2_{\mH_{k^m}}.
    \end{align}
\end{itemize}
The following lemma gives an error bound on the truncated feature map based $\tilde{\mu}_{k^m,d^m}(\hat{\P}^m_n)$ approximation of the empirical mean embedding $\tilde{\mu}_{k^m} \left(\hat{\P}^m_n\right)$.

\begin{lemma}[Approximation of the empirical mean embedding with $d^m$-truncation]\label{truncated_ineq}
Let $m\in [M]$ be fixed, and assume that the kernel $k^m: \X^m \times \X^m \to [0,1]$ is continuous on the compact metric space $\X^m$ and $\mathrm{supp}(\P^m)=\X^m$. Let $\tilde{\mu}_{k^m} \left(\hat{\P}^m_n\right)  = \frac{1}{n} \sum_{i=1}^n k^m \left(\cdot, X^m_i\right)$, $\tilde{\mu}_{k^m,d^m} \left(\hat{\P}^m_n\right) = \frac{1}{n} \sum_{i=1}^n k^m_{d^m} \left(\cdot, X^m_i\right)$ be as defined in \eqref{eq:mu-tilde-and-truncated}. Let $C_{k^m}$, $\mathcal{N}_{C_{k^m}}$, $b_{k^m}$, $a_{k^m}$, and $\lambda^m_{d^m+1}$ ($m\in [M]$) as defined in Theorem~\ref{errorbound_GHSIC1}. Let
\begin{align}
    t_{k^m}(n, \eta)= \frac{9 a_{k^m}^2}{n} \log\left( \frac{3n}{\eta}\right).\label{eq:def_t_km}
\end{align}

Assume that the pair $(n, \eta)$ is such that $t_{k^m}(n, \eta) \le \left \| C_{k^m} \right \| $; suppose also that Assumption~\ref{assum_eigenval_C_k}-\ref{assum_eigenval_hat_C_k} hold.
Then for any $\eta \in (0,1)$ 
\begin{align}
    \left\| \tilde{\mu}_{k^m}\left(\hat{\P}^m_n\right) - \tilde{\mu}_{k^m,d^m}\left(\hat{\P}^m_n\right) \right\|_{\mH_{k^m}} \le 3 \sqrt{\frac{3}{2}} b_{k^m} \sqrt{\mathcal{N}_{C_{k^m}}\left(t_{k^m} (n, \eta) \right)} \left(\lambda^m_{d^m+1} + t_{k^m}(n, \eta) \right)
\end{align} 
holds with $\left(\P^m\right)^{\otimes n}$-probability at least $1-\eta$    .

\end{lemma}

\begin{proof}(of Lemma \ref{truncated_ineq}):
Below we will use the shorthand $P_{d^m}^m := P_{d^m}^m \left( \hat{C}_{k^m}  \right)$. We have
\begin{align}\label{trans_inner_sum}
    \lefteqn{\left\| \tilde{\mu}_{k^m}\left(\hat{\P}^m_n\right) - \tilde{\mu}_{k^m,d^m}\left(\hat{\P}^m_n\right) \right\|^2_{\mH_{k^m}}  \stackrel{(a)}{=} \left\| \frac{1}{n} \sum_{i=1}^n k^m (\cdot, X^m_i) - \frac{1}{n} \sum_{i=1}^n k^m_{d^m} (\cdot, X_i^m) \right\|^2_{\mH_{k^m}}}  \\
    & \stackrel{(b)}{=} \left\| \frac{1}{n} \sum_{i=1}^n \left(I-P^{m}_{d^m}\right)k^m (\cdot, X^m_i) \right\|^2_{\mH_{k^m}}
     \stackrel{(c)}{\le} \frac{1}{n^2}\left ( \sum_{i=1}^n \left \| \left(I-P^{m}_{d^m}\right)k^m (\cdot, X^m_i) \right \|_{\mH_{k^m}}  \right ) ^2.
\end{align}
In (a) the definitions of $\tilde{\mu}_{k^m}\left(\hat{\P}^m_n\right)$ and $\tilde{\mu}_{k^m,d^m}\left(\hat{\P}^m_n\right)$ [\eqref{eq:mu-tilde-and-truncated}] were invoked, (b) follows from \eqref{eq:def:km}, and (c) is implied by the homogeneity of norms and the triangle inequality.
Taking the square root of both sides, one arrives at
\begin{align}
    \left\| \tilde{\mu}_{k^m}\left(\hat{\P}^m_n\right) - \tilde{\mu}_{k^m,d^m}\left(\hat{\P}^m_n\right) \right\|_{\mH_{k^m}} \le \frac{1}{n} \sum_{i=1}^n \left \| \left(I-P^{m}_{d^m}\right)k^m (\cdot, X^m_i) \right \|_{\mH_{k^m}}. 
\end{align}

We continue the proof by upper bounding the (square of the) obtained r.h.s.; by the Jensen’s inequality---$f(\E(Z)) \le \E (f(Z))$ with $f(z)=z^2$---we get 
\begin{align}
    \left( \E_{X^m \sim \hat{\P}_n^m}   \left \| \left(I-P^{m}_{d^m}\right)k^m (\cdot, X^m) \right \|_{\mH_{k^m}}  \right)^2 & \le \E_{X^m \sim \hat{\P}_n^m} \left \| \left(I-P^{m}_{d^m}\right)k^m (\cdot, X^m) \right \|^2_{\mH_{k^m}}  \\
    & \stackrel{(a)}{=}  \left \| \left(I-P^{m}_{d^m}\right) \hat{C}_{k^m}^{1/2} \right \|^2_{\t{HS}\left(\mH_{k^m}\right)},
\end{align}
where (a) was obtained from \eqref{deduce_RE} with $\Q=\hat{\P}_n^m$ and from \eqref{eq:C-emp-m}. Taking the square root of both sides, we have
\begin{align}\label{trans_exp}
    \E_{X^m \sim \hat{\P}_n^m} \left \| \left(I-P^{m}_{d^m}\right)k^m (\cdot, X^m) \right \|_{\mH_{k^m}}  \le \left \| \left(I-P^{m}_{d^m}\right) \hat{C}_{k^m}^{1/2} \right \|_{\t{HS}\left(\mH_{k^m}\right)}.
\end{align}

We proceed by upper bounding further the r.h.s.\ in \eqref{trans_exp}; we get
\begin{align}
    & \quad \left \| \left(I-P^{m}_{d^m}\right) \hat{C}_{k^m}^{1/2} \right \|_{\t{HS}\left(\mH_{k^m}\right)}\\ &\stackrel{(a)}{=} \Big\| C_{k^m}^{-1/2}C_{k^m}^{1/2}(C_{k^m}+t_{k^m} I)^{-1/2}(C_{k^m}+t_{k^m} I)^{1/2}\left(I-P^{m}_{d^m}\right) \\
    & \quad \hspace{0.1cm} \times (C_{k^m}+t_{k^m} I)^{1/2} (C_{k^m}+t_{k^m} I)^{-1/2} \hat{C}_{k^m}^{1/2} \Big\|_{\t{HS}\left(\mH_{k^m}\right)} \\
    & \stackrel{(b)}{\le} \left \| C_{k^m}^{-1/2}C_{k^m}^{1/2}(C_{k^m}+t_{k^m} I)^{-1/2} \right \|_{\t{HS}\left(\mH_{k^m}\right)}\\
    & \quad \hspace{0.1cm} \times \left \| (C_{k^m}+t_{k^m} I)^{1/2}\left(I-P^{m}_{d^m}\right)(C_{k^m}+t_{k^m} I)^{1/2} (C_{k^m}+t_{k^m} I)^{-1/2} \hat{C}_{k^m}^{1/2} \right \|_{\t{HS}\left(\mH_{k^m}\right)} \\
    & \stackrel{(c)}{\le}  \underbrace{\left \| C_{k^m}^{-1/2} \right \|}_{=:A_1} \underbrace{\left \| C_{k^m}^{1/2}(C_{k^m}+t_{k^m} I)^{-1/2} \right \|_{\t{HS}\left(\mH_{k^m}\right)}}_{=:A_2}  \\
    & \quad \hspace{0.1cm}\times \underbrace{\left \| (C_{k^m}+t_{k^m} I)^{1/2}\left(I-P^{m}_{d^m}\right)(C_{k^m}+t_{k^m} I)^{1/2}  \right \|_{\t{HS}\left(\mH_{k^m}\right)}}_{=:A_3} \underbrace{\left \| (C_{k^m}+t_{k^m} I)^{-1/2} \hat{C}_{k^m}^{1/2} \right \|}_{=:A_4},
\end{align}
where at this point $t_{k^m}$ can be any positive real; we will choose the specific value of $t_{k^m}$ as a function of $n$ and $\eta$ later. (a) follows by adding the term $I=C_{k^m}^{-1/2}C_{k^m}^{1/2}(C_{k^m}+t_{k^m} I)^{-1/2}(C_{k^m}+t_{k^m} I)^{1/2}$ and $I=(C_{k^m}+t_{k^m} I)^{1/2} (C_{k^m}+t_{k^m} I)^{-1/2}$, (b) is implied by $\left \| AB \right \|_{\t{HS}\left(\mH_{k^m}\right)} \le \left \| A \right \|_{\t{HS}\left(\mH_{k^m}\right)} \left \| B \right \|_{\t{HS}\left(\mH_{k^m}\right)}$, (c) comes from the well-known fact \citep[page~267]{conway90course} that for any $A\in \L(\mH_{k^m})$ and $B \in HS(\mH_k)$, one has
\begin{align}
    \left \| AB \right \|_{\t{HS}\left(\mH_{k^m}\right)}  & \le  \left \| A \right \| \left \| B \right \|_{\t{HS}\left(\mH_{k^m}\right)}, &  
    \left \| B A\right \|_{\t{HS}\left(\mH_{k^m}\right)}  & \le   \left \| B \right \|_{\t{HS}\left(\mH_{k^m}\right)}\left \| A \right \|. \label{eq:HS-opnorm-relation}
\end{align}
\begin{itemize}[labelindent=0em,leftmargin=1em,topsep=0cm,partopsep=0cm,parsep=0cm,itemsep=1mm]
    \item \tb{Term} $A_1$: For $A_1$, according to our assumption we have $A_1 \le b_{k^m}$.
    \item \tb{Term} $A_2$: Considering $A_2$, we have
    \begin{align}
    A_2 &= \left \|C_{k^m}^{1/2} (C_{k^m}+t_{k^m}I)^{-1/2} \right \|_{\t{HS}\left(\mH_{k^m}\right)} \\
    &\stackrel{(a)}{=} \sqrt{\left\langle C_{k^m}^{1/2} (C_{k^m}+t_{k^m}I)^{-1/2} , C_{k^m}^{1/2} (C_{k^m}+t_{k^m}I)^{-1/2}  \right\rangle_{\t{HS}\left(\mH_{k^m}\right)}}\\
    &\stackrel{(b)}{=} \sqrt{\tr \left( C_{k^m}+t_{k^m}I)^{-1/2} C_{k^m}^{1/2} C_{k^m}^{1/2} \left(C_{k^m}+t_{k^m}I\right)^{-1/2}\right)}\\
    & \stackrel{(c)}{=}  \sqrt{\tr\left( C_{k^m} (C_{k^m}+t_{k^m}I)^{-1}\right)} \stackrel{(d)}{=} \sqrt{\mathcal{N}_{C_{k^m}}(t_{k^m})},
    \end{align}
    where (a) is implied by $\left\|h\right\|_{\t{HS}\left(\mH_{k^m}\right)} = \sqrt{\langle h,h\rangle_{\t{HS}\left(\mH_{k^m}\right)}}$ ($h\in \t{HS}\left(\mH_{k^m}\right)$), (b) follows from the self-adjointness of $C_{k^m}$ and $\left \langle A,B \right \rangle_{\t{HS}\left(\mH_{k^m}\right)} = \tr (A^* B)$, in (c) the cyclic invariance of the trace was used, followed by the definition of the effective dimension $\mathcal{N}_{C_{k^m}}$ in (d).
    \item \tb{Term} $A_3$:  
            Regarding $A_3$, we have
    \begin{align}
    A_3 & \stackrel{(a)}{\le} \underbrace{\left \| (C_{k^m}+t_{k^m} I)^{\frac{1}{2}}(\hat{C}_{k^m}+t_{k^m} I)^{-\frac{1}{2}}(\hat{C}_{k^m}+t_{k^m} I)^{\frac{1}{2}}\left(I-P^{m}_{d^m}\right)(\hat{C}_{k^m}+t_{k^m}I)^{\frac{1}{2}} \right \|_{\t{HS}\left(\mH_{k^m}\right)}}_{=:B_1} \\
    & \quad \times \underbrace{\left \|(\hat{C}_{k^m}+t_{k^m}I)^{-\frac{1}{2}} (C_{k^m}+t_{k^m}I)^{\frac{1}{2}}  \right \|}_{=:B_2} .
    \end{align}
    The (a) is obtained by adding the terms $I=(\hat{C}_{k^m}+t_{k^m} I)^{-\frac{1}{2}}(\hat{C}_{k^m}+t_{k^m} I)^{\frac{1}{2}}$ and $I = (\hat{C}_{k^m}+t_{k^m}I)^{\frac{1}{2}} (\hat{C}_{k^m}+t_{k^m}I)^{-\frac{1}{2}}$, followed by using \eqref{eq:HS-opnorm-relation}.
    \begin{itemize}[labelindent=0em,leftmargin=1em,topsep=0cm,partopsep=0cm,parsep=0cm,itemsep=1mm]
        \item \tb{Term} $B_2$: Using that $\left\|A\right\| = \left\|A^*\right\|$ for any $A \in \L(\mH_{k^{m}})$, and the self-adjointness of $C_{k^m}$ and $\hat{C}_{k^m}$, by Lemma~\ref{concentration_effective_dimension}(i), for any $\delta > 0$ and $\frac{9 a_{k^m}^2}{n} \log \left( \frac{n}{\delta} \right) \le t_{k^m} \le \left \| C_{k^m} \right \| $ we have $B_2 \le \sqrt{2}$ 
        with probability at least $1-\delta$. 
        \item \tb{Term} $B_1$: For $B_1$ we have
        \begin{align}
            B_1 &\stackrel{(a)}{\le} \underbrace{\left \| (C_{k^m}+t_{k^m}I)^{\frac{1}{2}}(\hat{C}_{k^m}+t_{k^m}I)^{-\frac{1}{2}}\right \|_{\t{HS}\left(\mH_{k^m}\right)}}_{=:C_1} \\
            &\quad \hspace{0.1cm} \times \underbrace{\left \| (\hat{C}_{k^m}+t_{k^m}I)^{\frac{1}{2}}\left(I-P^{m}_{d^m}\right)(\hat{C}_{k^m}+t_{k^m}I)^{\frac{1}{2}}  \right \|}_{=:C_2} \\
            & \stackrel{(b)}{\le} \sqrt{2} \left( \lambda_{d^m+1}^m(\hat{C}_{k^m})+t_{k^m}  \right) \stackrel{(c)}{\le}  \frac{3 \sqrt{2} }{2}(\lambda_{d^m+1}^m(C_{k^m})+t_{k^m}).
        \end{align}
        (a) is  implied by \eqref{eq:HS-opnorm-relation}. In (b) we used that $C_1 \le \sqrt{2}$ with probability at least 1-$\delta$ [by Lemma~\ref{concentration_effective_dimension}(i) as we saw it for term $B_2$] and $C_2 = \lambda_{d^m+1}^m(\hat{C}_{k^m})+t_{k^m}$ for $\frac{9 a_{k^m}^2}{n} \log \left( \frac{n}{\delta} \right) \le t_{k^m} \le \left \| C_{k^m} \right \| $, (c) follows from Lemma \ref{concentration_effective_dimension}(iii) with probability at least $1-\delta$.  
    \end{itemize}
    \item \tb{Term} $A_4$:  
            For $A_4$, we have
        \begin{align}
            A_4 & \stackrel{(a)}{=} \left \| (C_{k^m}+t_{k^m}I)^{-1/2} (\hat{C}_{k^m}+t_{k^m}I)^{1/2} (\hat{C}_{k^m}+t_{k^m}I)^{-1/2} \hat{C}_{k^m}^{1/2} \right \| \\
            & \stackrel{(b)}\le \underbrace{\left \| (C_{k^m}+t_{k^m}I)^{-1/2} (\hat{C}_{k^m}+t_{k^m}I)^{1/2} \right \|}_{=:D_1} \underbrace{\left \| (\hat{C}_{k^m}+t_{k^m}I)^{-1/2} \hat{C}_{k^m}^{1/2} \right \|}_{=:D_2} \stackrel{(c)}{\le } \sqrt{\frac{3}{2}},
        \end{align}
        (a) was obtained by adding the term $I = (\hat{C}_{k^m}+t_{k^m}I)^{1/2} (\hat{C}_{k^m}+t_{k^m}I)^{-1/2}$, in (b) we used $\left \| AB \right \| \le \left \| A \right \|  \left \| B \right \| $. In (c), we have $D_1 \le \sqrt{\frac{3}{2}}$ according to Lemma~\ref{concentration_effective_dimension}(ii) with probability at least $1-\delta$ for $\frac{9 a_{k^m}^2}{n} \log \left( \frac{n}{\delta} \right) \le t_{k^m} \le \left \| C_{k^m} \right \| $ and $D_2 \le 1$.
\end{itemize}

Using the bounds obtained above with $\delta := \frac{\eta}{3}$ and
\begin{align}
t_{k^m}(n, \eta)&:=\frac{9 a_{k^m}^2}{n} \log\left( \frac{n}{\eta/3}\right)= \frac{9 a_{k^m}^2}{n} \log\left( \frac{3n}{\eta}\right)
\end{align} 
above and requiring that $t_{k^m}(n, \eta) \le \left \| C_{k^m} \right \| $, and combining with a union bound the results obtained for $A_1$, $A_2$, $A_3$ and $A_4$, we got that
 \begin{align}
    \E_{X^m \sim \hat{\P}_n^m} \left [ \left \| (I-P^{m}_{d^m})k^m (\cdot, X^m) \right \|_{\mH_{k^m}} \right ]  
    &\le 3 \sqrt{\frac{3}{2}} b_{k^m} \sqrt{\mathcal{N}_{C_{k^m}}\left(t_{k^m} (n, \eta) \right)} \left(\lambda_{d^m+1}^m + t_{k^m}(n, \eta) \right),
\end{align}
and hence
\begin{align}\label{exp_truncate_error}
    \hspace{-0.5cm}\left\| \tilde{\mu}_{k^m}\left(\hat{\P}^m_n\right) - \tilde{\mu}_{k^m,d^m}\left(\hat{\P}^m_n\right) \right\|_{\mH_{k^m}}  \le 3 \sqrt{\frac{3}{2}} b_{k^m} \sqrt{\mathcal{N}_{C_{k^m}}\left(t_{k^m} (n, \eta) \right)} \left(\lambda_{d^m+1}^m + t_{k^m}(n, \eta) \right)
\end{align}
holds with probability at least $1-\eta$ for $\eta \in (0,1)$.
\end{proof}

Having obtained the finite-sample bound of the approximate kernel mean embedding of $k^m$ $(m \in [M])$, we now extend our result to the tensor product kernel $k = \otimes_{m=1}^M k^m$.

\begin{lemma}[Approximation of the joint empirical mean embedding with $\b d$-truncation]\label{truncated_ineq2}
Let 
\begin{align}
\tilde{\mu}_{k}(\hat{\P}_n) & =\frac{1}{n} \sum_{i=1}^n k(\cdot,X_i) \in \mH_k, &
\tilde{\mu}_{k,\b d}(\hat{\P}_n) &= \frac{1}{n} \sum_{i=1}^n \otimes_{m=1}^M  k^m_{d^m} (\cdot,X^m_i) \in \mH_k \label{eq:def_joint_tilde_mu}
\end{align}
with $k=\otimes_{m=1}^M k^m$ and $\b d:=\left(d^m \right)_{m=1}^M$. 
Let $C_{k^m}$, $\mathcal{N}_{C_{k^m}}$, $b_{k^m}$, $a_{k^m}$, and $\lambda^m_{d^m+1}$ ($m\in [M]$) as defined in Theorem~\ref{errorbound_GHSIC1}. 
Suppose that Assumption~\ref{assum_eigenval_C_k}-\ref{assum_eigenval_hat_C_k} hold.
Assume also that the kernel $k^m: \X^m \times \X^m \to [0,1]$ is continuous on the compact metric space $\X^m$ and $\mathrm{supp}(\P^m)=\X^m$ for $m\in [M]$.
Then for any $\eta \in (0,1)$, 
\begin{align}
    \left\| \tilde{\mu}_{k}(\hat{\P}_n) - \tilde{\mu}_{k,\b d}(\hat{\P}_n) \right\|_{\mH_{k}} \le  3 \sqrt{\frac{3}{2}} \sum_{m=1}^M b_{k^m} \sqrt{\mathcal{N}_{C_{k^m}}\left( t_{k^m}\left( n, \frac{\eta}{M} \right) \right)}\left(\lambda_{d^m+1}^m + t_{k^m} \left( n, \frac{\eta}{M} \right) \right),
\end{align}
holds with $\P^{\otimes n}$-probability at least $1-\eta$, while assuming that $t_{k^m}\left( n, \frac{\eta}{M} \right) \le \left \| C_{k^m} \right \| $ for all $m\in [M]$, with $t_{k^m}$ defined in \eqref{eq:def_t_km}.
\end{lemma}

\begin{proof}(Lemma \ref{truncated_ineq2})
Similarly to the proof of Lemma \ref{truncated_ineq}, we will use the shorthand $P_{d^m}^m := P_{d^m}^m \left( \hat{C}_{k^m}  \right)$, where $m\in [M]$. With this notation at hand, we have
\begin{align}
    & \left\| \tilde{\mu}_k \left(\hat{\P}_n \right) - \tilde{\mu}_{k,\b d}\left( \hat{\P}_n \right) \right\|^2_{\mH_k} 
    \stackrel{(a)}{=} \left\| \frac{1}{n} \sum_{i=1}^n k \left(\cdot,X_i \right) - \frac{1}{n} \sum_{i=1}^n \otimes_{m=1}^M   k^m_{d^m}\left( \cdot,X^m_i \right) \right\|^2_{\mH_k} \\
    & \stackrel{(b)}{=} \left\| \frac{1}{n} \sum_{i=1}^n \otimes_{m=1}^M k^m (\cdot,X^m_i) - \frac{1}{n} \sum_{i=1}^n \otimes_{m=1}^M P^{m}_{d^m} k^m(\cdot,X^m_i) \right\|^2_{\mH_k} \\
    &\stackrel{(c)}{\le} \frac{1}{n^2} \left( \sum_{i=1}^n \left\|  \otimes_{m=1}^M k^m (\cdot,X^m_i) -  \otimes_{m=1}^M P^m_{d^m}  k^m (\cdot,X^m_i) \right\|_{\mH_k}  \right)^2. 
\end{align}
In (a) we applied the definition of $\tilde{\mu}_k \left( \hat{\P}_n \right)$  and that of $\tilde{\mu}_{k,\b d} \left( \hat{\P}_n \right)$.
(b) was obtained by using that $k = \otimes_{m=1}^M k^m$ and that $ k^m_{d^m}(\cdot,X^m_i) = P^m_{d^m} k^m(\cdot,X^m_i)$ by \eqref{eq:def:km}. (c) follows from the homogeneity of the norms and the triangle inequality.

By taking the square root of both sides, we have
\begin{align}
    \left\| \tilde{\mu}_k \left(\hat{\P}_n \right) - \tilde{\mu}_{k,\b d}\left(\hat{\P}_n \right) \right\|_{\mH_k} & \le \frac{1}{n} \sum_{i=1}^n \left\|  \otimes_{m=1}^M k^m (\cdot,X^m_i) -  \otimes_{m=1}^M P^m_{d^m} k^m (\cdot,X^m_i) \right\|_{\mH_k}.
\end{align}

We continue the reasoning by bounding the (square of the) obtained r.h.s.; by the Jensen’s inequality---$f(\E(Z)) \le \E (f(Z))$ with $f(z)=z^2$---, one gets
\begin{align}
    & \left( \E_{X \sim \hat{\P}_n}\left[  \left\|  \otimes_{m=1}^M k^m (\cdot,X^m) -  \otimes_{m=1}^M P^m_{d^m} k^m (\cdot,X^m) \right\|_{\mH_k} \right] \right)^2 \\
    & \le \E_{X \sim \hat{\P}_n} \left[  \left\|  \otimes_{m=1}^M k^m (\cdot,X^m) -  \otimes_{m=1}^M P^m_{d^m} k^m (\cdot,X^m) \right\|^2_{\mH_k} \right].
\end{align}

Then we have
\begin{align}
\begin{split}
    &  \E_{X\sim \hat{\P}_n} \left[  \left\|  \otimes_{m=1}^M k^m(\cdot,X^m)  -  \otimes_{m=1}^M P^m_{d^m} k^m(\cdot,X^m) \right\|^2_{\mH_k} \right] \\
    & \stackrel{(a)}{=} \E_{X \sim \hat{\P}_n} \left[  \left \langle \otimes_{m=1}^M k^m(\cdot,X^m)  -  \otimes_{m=1}^M P^m_{d^m}k^m(\cdot,X^m),\otimes_{m=1}^M k^m(\cdot,X^m)  -  \otimes_{m=1}^M P^m_{d^m} k^m (\cdot,X^m)\right \rangle_{\mH_k}  \right] \\
    & \stackrel{(b)}{=} \E_{X\sim \hat{\P}_n} \left[  \left \langle \otimes_{m=1}^M k^m(\cdot,X^m) ,\otimes_{m=1}^M k^m(\cdot,X^m) \right \rangle_{\mH_k} \right.  - \left \langle \otimes_{m=1}^M k^m (\cdot,X^m),\otimes_{m=1}^M P^m_{d^m} k^m (\cdot,X^m)\right \rangle_{\mH_k}  \\
    & \quad \left. - \left \langle \otimes_{m=1}^M P^m_{d^m} k^m (\cdot,X^m),\otimes_{m=1}^M k^m(\cdot,X^m) \right \rangle_{\mH_k} + \left \langle \otimes_{m=1}^M P^m_{d^m} k^m (\cdot,X^m),\otimes_{m=1}^M P^m_{d^m} k^m (\cdot,X^m)\right \rangle_{\mH_k}  \right] \\
    & \stackrel{(c)}{=} \E_{X \sim \hat{\P}_n} \left[  \prod_{m=1}^M \left \langle  k^m (\cdot,X^m), k^m (\cdot,X^m)\right \rangle_{\mH_{k^m}}  - \prod_{m=1}^M \left \langle k^m (\cdot,X^m), P^m_{d^m} k^m(\cdot,X^m) \right \rangle_{\mH_{k^m}} \right. \\
    & \quad \left. - \prod_{m=1}^M \left \langle P^m_{d^m} k^m (\cdot,X^m), k^m (\cdot,X^m)\right \rangle_{\mH_{k^m}} + \prod_{m=1}^M \left \langle P^m_{d^m} k^m (\cdot,X^m), P^m_{d^m} k^m (\cdot,X^m)\right \rangle_{\mH_{k^m}}  \right] \\
    & \stackrel{(d)}{=} \E_{X \sim \hat{\P}_n} \left[  \prod_{m=1}^M \left \langle  k^m (\cdot,X^m), k^m (\cdot,X^m)\right \rangle_{\mH_{k^m}}  - \prod_{m=1}^M \left \langle k^m (\cdot,X^m), P^m_{d^m} k^m(\cdot,X^m) \right \rangle_{\mH_{k^m}} \right]\\
    & \stackrel{(e)}{\le} \E_{X \sim \hat{\P}_n} \left[ \sum_{m=1}^M \left | \left \langle  k^m (\cdot,X^m) , k^m (\cdot,X^m)\right \rangle_{\mH_{k^m}} - \left \langle k^m (\cdot,X^m), P^m_{d^m} k^m (\cdot,X^m) \right \rangle_{\mH_{k^m}} \right |   \right] \\
    & \stackrel{(f)}{=} \sum_{m=1}^M \E_{X^m \sim \hat{\P}_n^m}  \left \langle  k^m (\cdot,X^m), \left(I-P^m_{d^m}\right)k^m(\cdot,X^m) \right \rangle_{\mH_{k^m}} \\
    & \stackrel{(g)}{=} \sum_{m=1}^M \E_{X^m \sim \hat{\P}_n^m} \left \| (I-P_{d^m}^m)k^m(\cdot,X^m) \right \|^2_{\mH_{k^m}} \stackrel{(h)}{=} \sum_{m=1}^M \left \| \left(I-P^m_{d^m}\right) \hat{C}_{k^m}^{1/2} \right \|^2_{\t{HS}\left(\mH_{k^m}\right)}.
\end{split}    
\end{align}
In (a) we used that $\left\|h\right\|_{\mH_{k}}^2 = \langle h,h\rangle_{\mH_{k}}$ ($  h\in \mH_{k}$); (b) was obtained by expanding the inner product. (c) follows from the definition of the tensor product [\eqref{eq:def:tensor-product}]. (d) is implied by 
\begin{align}
    \left \langle P^m_{d^m} k^m (\cdot,X^m), P^m_{d^m} k^m (\cdot,X^m)\right \rangle_{\mH_{k^m}} & \stackrel{(i)}{=} \left \langle (P^m_{d^m})^*P^m_{d^m} k^m (\cdot,X^m), k^m (\cdot,X^m)\right \rangle_{\mH_{k^m}} \\
    & \stackrel{(j)}{=} \left \langle P^m_{d^m} k^m (\cdot,X^m), k^m (\cdot,X^m)\right \rangle_{\mH_{k^m}},
\end{align}
where (i) holds by the definition of the adjoint operator and (j) comes from \eqref{eq:P:self-adjoint}
and \eqref{eq:P:idempotent}. (e) is implied by the inequality \citep[page~190]{steele2004cauchy}: 
\begin{equation}\label{prod_sum_ineq}
    \prod_{m=1}^M w_m - \prod_{m=1}^M v_m \le \left | \prod_{m=1}^M w_m - \prod_{m=1}^M v_m  \right |  \le \sum_{m=1}^M \left | w_m - v_m \right |, \quad \left | w_m \right |,\left | v_m \right | \le 1 \ \t{for all }  m \in [M]. 
\end{equation}
Indeed, according to our assumption $k^m:\X^m \times \X^m \to [0,1]$; therefore with 
\begin{align}
w_m &:= \left \langle  k^m (\cdot,X^m) , k^m (\cdot,X^m)\right \rangle_{\mH_{k^m}},&
v_m &:= \left \langle k^m (\cdot,X^m), P^m_{d^m} k^m (\cdot,X^m) \right \rangle_{\mH_{k^m}},
\end{align}
one gets
\begin{align}
v_m &\le |v_m| = \left|\left \langle k^m (\cdot,X^m), P^m_{d^m} k^m (\cdot,X^m) \right \rangle_{\mH_{k^m}}\right| 
\stackrel{(k)}{\le} \left\| k^m (\cdot,X^m) \right\|_{\mH_{k^m}}  \left\| P^m_{d^m} k^m (\cdot,X^m)  \right\|_{\mH_{k^m}}\\
&\stackrel{(l)}{\le} \left\|k^m (\cdot,X^m)\right\|_{\mH_{k^m}}^2 \stackrel{(m)}{=}
\left \langle  k^m (\cdot,X^m) , k^m (\cdot,X^m)\right \rangle_{\mH_{k^m}} \stackrel{(n)}{=} k^m\left(X^m,X^m\right) =: w_m  = |w_m| \le 1,
\end{align}
where (k) holds by the CBS inequality, (l) follows as $P^m_{d^m}$ is a projection (hence by the Pythagorean theorem $\left\|h\right\|_{\mH_{k^m}}^2=\left\|P^m_{d^m}(h)\right\|_{\mH_{k^m}}^2+\left\|h-P^m_{d^m}(h)\right\|_{\mH_{k^m}}^2$ holds for any $h \in \mH_{k^m}$), (m) holds as the norm is inner product induced in an RKHS, (n) comes from the reproducing property. (f) follows from the linearity of the expectation, marginalization, and by the property
\begin{align}
    \left \langle k^m(\cdot,X^m), (I-P^{m}_{d^m})k^m(\cdot,X^m) \right\rangle_{\mH_{k^m}} =\left \|(I-P^{m}_{d^m})k^m(\cdot,X^m)  \right \|_{\mH_{k^m}}^2 \ge 0, 
\end{align}
where the equality was seen in the derivation of \eqref{deduce_RE}; hence the absolute value was discarded. (g) is implied by the derivation of (d)-(i) in \eqref{deduce_RE} with $\Q:=\hat{\P}^m_n$.

Thus, we got that 
\begin{align}
    & \E_{X \sim \hat{\P}_n} \left[  \left\|  \otimes_{m=1}^M k^m (\cdot,X^m) -  \otimes_{m=1}^M P^m_{d^m} k^m (\cdot,X^m) \right\|_{\mH_k} \right] \\
    & \le \sqrt{\sum_{m=1}^M \left \| \left(I-P^m_{d^m}\right) \hat{C}_{k^m}^{1/2} \right \|^2_{\t{HS}\left(\mH_{k^m}\right)}} \stackrel{(a)}{\le} \sum_{m=1}^M \left( \left \| \left(I-P^m_{d^m} \right) \hat{C}_{k^m}^{1/2} \right \|^2_{\t{HS}\left(\mH_{k^m}\right)}\right)^{\frac{1}{2}} \\
    & = \sum_{m=1}^M \left \| \left(I-P^m_{d^m}\right) \hat{C}_{k^m}^{1/2} \right \|_{\t{HS}\left(\mH_{k^m}\right)} \stackrel{(b)}{\le} \sum_{m=1}^M 3 \sqrt{\frac{3}{2}} b_{k^m} \sqrt{\mathcal{N}_{C_{k^m}}\left(t_{k^m}(n, \eta) \right)}\left(\lambda_{d^m+1}^m + t_{k^m}(n, \eta) \right),
\end{align}
with probability at least $1-M \eta$ for $\eta\in (0,1)$ if $t_{k^m}\left( n, \eta \right) \le \left \| C_{k^m} \right \| $ ($m\in [M]$), with $t_{k^m}$ defined in \eqref{eq:def_t_km}. (a) was obtained by $\sqrt{\sum_{m=1}^M a_m} \le \sum_{m=1}^M \sqrt{a_m}$ holding for $a_m\ge 0$ ($m\in [M]$), and (b) comes from \eqref{exp_truncate_error} with a union bound.

Applying a substitution $\eta \leftarrow \frac{\eta}{M}$, we have that 
\begin{align}
    \left\| \tilde{\mu}_k(\hat{\P}_n) - \tilde{\mu}_{k,\b d}(\hat{\P}_n) \right\|_{\mH_k}  \le 3 \sqrt{\frac{3}{2}} \sum_{m=1}^M b_{k^m} \sqrt{\mathcal{N}_{C_{k^m}} \left(t_{k^m} \left( n, \frac{\eta}{M} \right) \right)} \left(\lambda_{d^m+1}^m + t_{k^m} \left( n, \frac{\eta}{M} \right) \right),
\end{align}
holds with $\P^{\otimes n}$-probability at least $1-\eta$ for $\eta \in (0,1)$ where $t_{k^m} \left( n, \frac{\eta}{M} \right) \le \left\| C_{k^m} \right\|$ for all $m \in [M]$.
\end{proof}

We are aiming for the estimation of the features associated to the $d^m$-truncated approximation of the Gram matrix $\b K^m$ (see $\bm{\Phi}_d^{m}$ below). In the following lemma we show how well the $d^m$-truncated approximation of $\left | \b A^m \right |:=\left( (\b A^m)^\top \b A^m \right)^{\frac{1}{2}}$ (via $\hat{\bm{\Phi}}_d^{m}$ below)---which can be computed using the observations--- can achieve this goal with high probability, in Frobenius norm sense.

\begin{lemma}[Approximation of the spectral features of $\b K^m$ from the adjacency matrix]\label{decomposition_LPRG}
Let $\eta \in \left(0,1 \right)$, $m\in [M]$, where $M \in \Zp$ denotes the number of LPRGs, and $c^m > 0$ be fixed.\footnote{The value of $c^m$ determines that of $C_0^m$ in Lemma~\ref{concentration_dense1}, which affects the value of $\Delta^m$ in Assumption~\ref{assum_Delta}. \label{footnote:cm-meaning}} Let $n$ denote the number of nodes of each LPRG. Suppose that Assumptions~\ref{assum_Delta}-\ref{assum_eigenval_hat_C_k} hold, 
with $n_{0,(i)}^{m}(\eta), n_{0,(ii)}^{m}(\eta), n_{1,(i)}^{m}(\eta), n_{1,(ii)}^{m}(\eta) \in \Zp$, $c^m_0>0$, $c_1^m(n)=c_2^m \log \left( 2 + \log(2n) \right)$, $c_2^m > 0$, $C_0^m>0$, and $C_1^m>0$  specified therein. Assume that $C_1^m > 2\sqrt{2}$ and $c_2^m \ge 2$. Let 
\begin{align}
    \tilde{n}^m_{0,A1(i)}(\eta) &:= n^m_{0,(i)}\left( \frac{\eta}{1152}\right), \quad \tilde{n}^m_{0,A1(ii)}(\eta) := n^m_{0,(ii)}\left( \frac{\eta}{576}\right), \quad \tilde{n}^m_{0,A2(i)}(\eta) := n^m_{1,(i)}\left( \frac{\eta}{576} \right), \\
    \tilde{n}^m_{0,A2(ii)}(\eta) &:= n_{1,(ii)}^m \left( \frac{\eta}{96} \right), \quad \quad \tilde{n}^m_{0,L5}(\eta) := \left( \frac{\eta}{1152} \right)^{-\frac{1}{c^m}}, \quad \quad \tilde{n}^m_{0,L6}(\eta) :=  \frac{2304}{\eta}, \\
    \tilde{n}_{0,\text{else}-1}^m(\eta) & := 32 \left(c_1^m(n)\right)^2 (c_0^m)^{-2} \log \left( \frac{1152 n}{\eta} \right),
    \quad \quad \tilde{n}_{0,\text{else}-2}^m := e^{\frac{1}{\left( 2C_1^m - 4\sqrt{2} \right)^2}},
\end{align}
and 
\begin{align}
    n \ge \max & \Big\{ \tilde{n}^m_{0,A1(i)}(\eta), \tilde{n}^m_{0,A1(ii)}(\eta), \tilde{n}^m_{0,A2(i)}(\eta), \tilde{n}^m_{0,A2(ii)}(\eta), \tilde{n}^m_{0,L5}(\eta), \tilde{n}^m_{0,L6}(\eta), \\
    & \quad \tilde{n}_{0,\text{else-1}}^m(\eta), \tilde{n}_{0,\text{else-2}}^m \Big\}. \label{eq:L19:n-constraint}
\end{align}
Let $(X^m_i)_{i=1}^n$ be i.i.d.\ $\X^m$-valued random variables, where $\X^m$ is assumed to be a compact metric space. Let the Gram matrix $\b K^m = [K^m_{i,j}]_{i,j=1}^{n} = \left [ k^m(X^m_i,X^m_j) \right ]_{i,j=1}^{n}\in \R^{n \times n}$ be associated to a continuous kernel $k^m:\X^m \times \X^m \to [0,1]$ defined on  $\X^m$, with corresponding adjacency matrix $\b A^m \in \R^{n \times n}$ as specified in \eqref{eq:A^m}. Let $d^m \in \Zp$, the eigenvalues (ordered in a non-increasing way) of the integral operator $T_{k^m}$ and $\b K^m$ be $\left ( \lambda^m_i \right )_{i \in I^m} $ and $\left ( \lambda^m_i (\b K^m)\right )_{i \in [n]}$, respectively. Let the gap of the $d^m$-th and the $(d^m+1)$-th eigenvalues of $T_{k^m}$ be defined as $\delta^m_{d^m} = \lambda^m_{d^m} - \lambda^m_{d^m+1} \ne 0$ (by Assumption~\ref{assum_eigenval_C_k}).$^{\ref{footnote_delta_ne_0}}$ Let the spectral decomposition of $\left | \b A^m \right |:=\left( (\b A^m)^\top \b A^m \right)^{\frac{1}{2}} \in \R^{n\times n}$ and its $d^m$-truncated approximation  be denoted by 
\begin{align}
\left | \b A^m \right | &= \b U^m \b S^m  \left(\b U^m\right)^{\top} \approx \b U^m_{d^m} \b S^m_{d^m} \left(\b U^m_{d^m}\right)^{\top}, \label{Am:SVD-SVDm}
\end{align}
where $\b U^m$, $\b S^m \in \R^{n\times n}$, and $\b S^m_{d^m} \in \R^{d^m \times d^m}$ is the diagonal matrix composed of the $d^m$ largest eigenvalues of $\left | \b A^m \right | $ and $\b U^m_{d^m} \in \R^{n \times d^m}$ consists of the corresponding orthogonal eigenvectors. Similarly, for $\b K^m$ let its spectral decomposition and $d^m$-truncated approximation be 
\begin{align}
\b K^m &= \b U^m_{\b K^m} \b S^m_{\b K^m} \left(\b U^m_{\b K^m}\right)^{\top} \approx \b U^m_{\b K^m,d^m} \b S^m_{\b K^m,d^m} \left(\b U^m_{\b K^m,d^m}\right)^{\top}, \label{Km:SVD-SVDm}
\end{align}
where $\b U^m_{\b K^m}$, $\b S^m_{\b K^m}\in \R^{n\times n}$, $\b S^m_{\b K^m,d^m} \in \R^{d^m \times d^m}$ and $\b U^m_{\b K^m, d^m} \in \R^{n \times d^m}$. Let 
\begin{align}
\hat{\bm{\Phi}}^m_{d^m} := \b U^m_{d^m} \left(\b S^m_{d^m}\right)^{\frac{1}{2}} \in \R^{n \times d^m}. \label{eq:hat-Phi-m-def}
\end{align}
Then, there exists an orthogonal matrix $\b W^m \in \R^{d^m \times d^m}$ such that with 
\begin{align}
\bm{\Phi}^m_{d^m} := \b U^m_{\b K^m,d^m} \left( \b S^m_{\b K^m,d^m} \right)^{\frac{1}{2}} \b W^m \in \R^{n \times d^m},
\end{align}
one has 
\begin{align}
    \J^m & \Big(\left \| \hat{\b \Phi}^m_{d^m} - \b \Phi^m_{d^m}  \right \|_F \le \left \| \left(\b U^m_{d^m} \left( \b U^m_{d^m} \right)^{\top} \left | \b A^m \right | - \b U^m_{\b K^m,d^m} \left( \b U^m_{\b K^m,d^m} \right)^\top \b K^m\right)\b U^m_{d^m} \left( \b S^m_{d^m} \right)^{-\frac{1}{2}} \right \|_F\\
    & \hspace{3.7cm} + \xi^m_n \Big) \ge 1-\eta,
\end{align}
where $\J^m$ denotes the joint distribution of the entries of $\b A^m$ and $\left(X^m_i\right)_{i=1}^n$, and 
\begin{align}
    \xi^m_n &:= 264 \sqrt{2} c_1^m(n) d^m \left( \delta^m_{d^m} \right)^{-\frac{1}{2}}  \sqrt{\log \left ( \frac{432 n}{\eta} \right ) } + 896 \frac{\left(d^m \right)^{\frac{1}{2}} \left(c_1^m(n)\right)^2} {\sqrt{c_0^m}} \left( \delta^m_{d^m} \right)^{-2} \frac{\log \left( \frac{216 n}{\eta} \right)}{\sqrt{n}}\\
     & \quad + 1024 \frac{\left(d^m \right)^{\frac{1}{2}} \left(c_1^m(n)\right)^3}{\sqrt{c_0^m}} \left( \delta^m_{d^m} \right)^{-2} \frac{\log^{\frac{3}{2}} \left( \frac{216 n}{\eta} \right)}{n} + 64 \frac{ \left(c_1^m(n)\right)^2}{\sqrt{c_0^m}} \left( \delta^m_{d^m} \right)^{-1} \frac{\left(d^m \right)^{\frac{1}{2}} \log \left( \frac{216 n}{\eta} \right)}{\sqrt{n}} \\
     & \quad + 4 \frac{(d^m)^{\frac{1}{2}} c_1^m(n)}{\sqrt{c_0^m}} \sqrt{\log \left( \frac{72n}{\eta}  \right)}. \label{eq:def_xi^m} 
\end{align}

\end{lemma}

\noindent\tb{Remark}: 
\begin{itemize}
    \item \tb{Interpretation}: The relevance of the probabilistic bound in Lemma~\ref{decomposition_LPRG} is that by controlling the error of the $d^m$-truncated approximation ($\left \| \hat{\b \Phi}^m_{d^m} - \b \Phi^m_{d^m}  \right \|_F$) via the expression on the right-hand side, one can leverage (see later Lemma~\ref{concentration_adj} and Lemma~\ref{concentration_joint_adj}) the concentration between $\left|\b A^m \right|$ and $\b K^m$ to establish an upper bound on the former quantity. 
    \item \tb{$\J^m$ events, propagation of $n$ constraints}: For the reader's convenience, we summarize all the $\J^m$ events appearing throughout the proof in Fig.~\ref{fig:Jm_events}. We note that Assumption~\ref{assum_Delta}(i), Assumption~\ref{assum_Delta}(ii), Assumption~\ref{assum_lambda_delta}(i), Assumption~\ref{assum_lambda_delta}(ii), Lemma~\ref{concentration_dense1} and Lemma~\ref{lowerbound_A_diff_K} (these 2 lemmas are used in the proof) all give rise `$n$ is large enough' constraints---indicated by color \colAonei{CadetBlue}, \colAoneii{NavyBlue}, \colAi{Teal}, \colAii{Magenta}, \colLfive{Violet}, \colLsix{Orange} throughout the proof. The $n\ge \tilde{n}_{0,\text{else-1}}^m(\eta)$ constraint guarantees that $\left\| \left( \b S^m_{d^m} \right)^{-\frac{1}{2}} \right\| \le 2\sqrt{2}(n \delta_{d^m}^m)^{-\frac{1}{2}}$ (see \eqref{eq:bound_eig_A} and \eqref{eq:source_else1}); the $n\ge \tilde{n}_{0,\text{else-2}}^m$ requirement ensures that $\left\| \b D^m\right\|\ge 1$ (with $\b D^m:= \b A^m- \b K^m$) in \eqref{eq:D^m-lower-bound->n-constraint}; these constraints are indicated by color \coleone{BrickRed} and \coletwo{LimeGreen}, respectively.  
\end{itemize}

\begin{proof}(Lemma \ref{decomposition_LPRG})

%-------Illustration of events
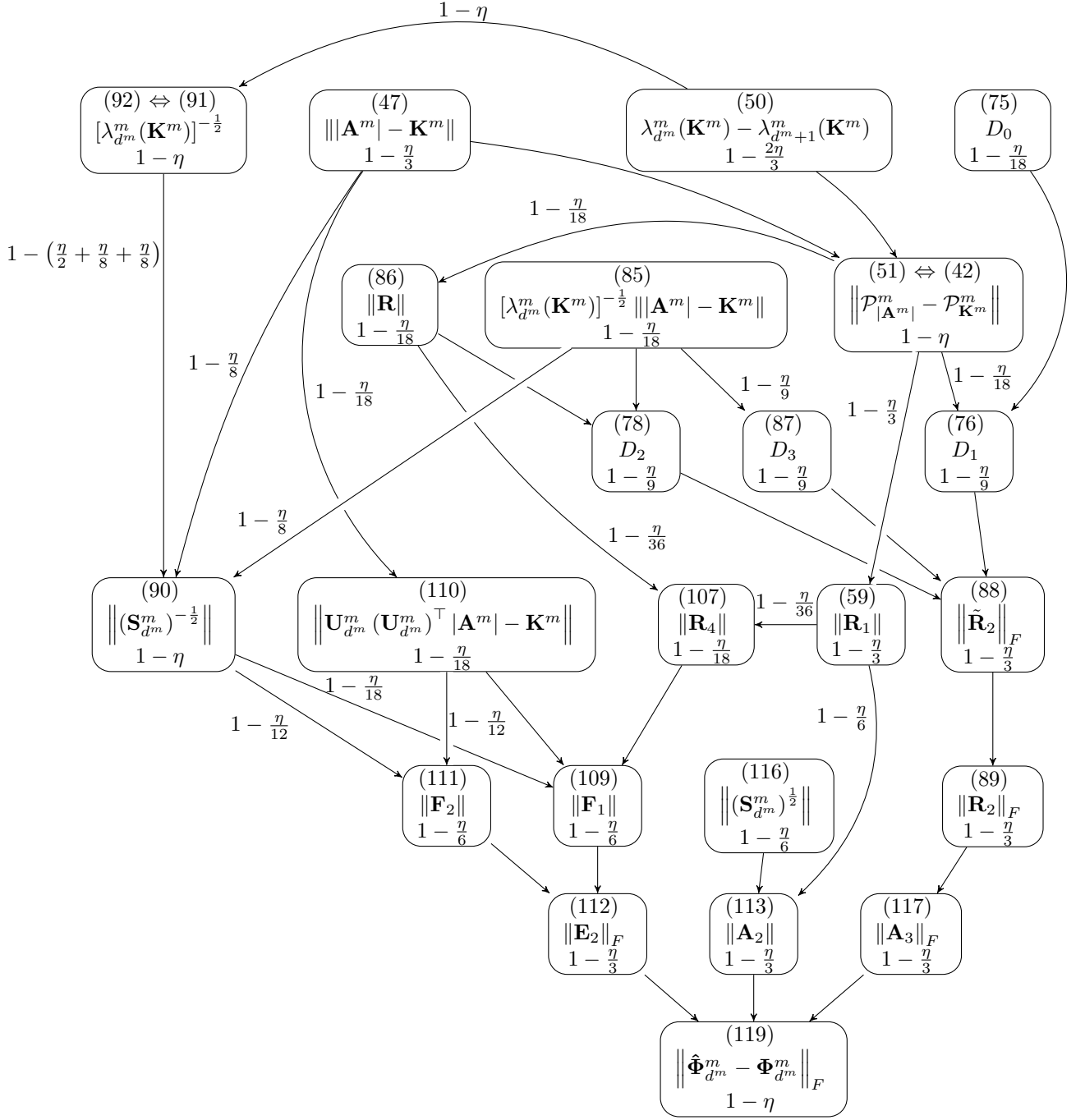
\begin{figure}
    \centering

\begin{tikzpicture}[
    ->,                % Directed edges
    >=stealth',        % Stylish arrow tips
    node distance=1cm,
    every node/.style={
    draw,
    rectangle,
    rounded corners=3mm,
    align=center,
    font=\small,
    minimum height=1mm,
    inner sep=0.5pt
    }
]

%----------------------
% First layer
%----------------------
\node (E1) {
\begin{tabular}{c}
\eqref{eig_third_bound} $\Leftrightarrow$ \eqref{eig_third_bound0}\\
$\left[ \lambda_{d^m}^m(\b K^m) \right]^{-\frac{1}{2}}$\\
$1-\eta$
\end{tabular}
};
\node (E2) [right=1cm of E1] {\begin{tabular}{c}
\eqref{bound_diff_psd}\\
$\left\| \left| \b A^m \right| - \b K^m \right\|$\\
$1-\frac{\eta}{3}$
\end{tabular}};
\node (E3) [right=2.5cm of E2] {
\begin{tabular}{c}
\eqref{eig_second_bound} \\
$\lambda_{d^m}^m(\b K^m)-\lambda_{d^m+1}^m(\b K^m)$ \\
$1-\frac{2\eta}{3}$
\end{tabular}
};
\node (E4) [right=1cm of E3] {
\begin{tabular}{c}
\eqref{eq:bound_D0}\\
$D_0$\\
$1-\frac{\eta}{18}$
\end{tabular}
};

% Edges

\draw (E3) to [out=150,in=30] node[pos=0.5, above, draw=none, fill=none]{$1-\eta$} (E1);

%----------------------
% Second layer
%----------------------

\node (E5) [below=1.5cm of E2] {
\begin{tabular}{c}
\eqref{bound_R} \\
$\left\| \b R \right\|$ \\
$1-\frac{\eta}{18}$
\end{tabular}
};
\node (E6) [right=0.75cm of E5] {
\begin{tabular}{c}
\eqref{bound_D2_sub} \\
$\left[\lambda^m_{d^m}(\b K^m) \right]^{-\frac{1}{2}} \left \| \left | \b A^m \right | - \b K^m \right \|$ \\
$1-\frac{\eta}{18}$
\end{tabular}
};
\node (E7) [right=0.75cm of E6] {
\begin{tabular}{c}
\eqref{bound_proj} $\Leftrightarrow$ \eqref{bound_proj0}\\
$\left \| \mathcal{P}^m_{\left| \b A^m \right|} - \mathcal{P}^m_{\b K^m} \right \|$\\
$1-\eta$
\end{tabular}
};

%--edges------

\draw[bend left=10] (E2) to (E7);
\draw[bend left=10] (E3) to (E7);
\draw (E7) to [out=155,in=25] node[pos=0.7, above=0.2cm, draw=none, fill=none]{$1-\frac{\eta}{18}$} (E5);

%--Layer 3

\node (E8) [below=1cm of E6] {
\begin{tabular}{c}
\eqref{bound_D2} \\
$D_2$ \\
$1-\frac{\eta}{9}$
\end{tabular}
};
\node (E9) [right=1cm of E8] {
\begin{tabular}{c}
\eqref{bound_D3} \\
$D_3$ \\
$1-\frac{\eta}{9}$
\end{tabular}
};
\node (E10) [right=1.5cm of E9] {
\begin{tabular}{c}
\eqref{decompose_D1} \\
$D_1$ \\
$1-\frac{\eta}{9}$
\end{tabular}
};

%--edges----------

\draw (E5) to (E8);
\draw (E6) to (E8);
\draw (E6) -- node[pos=0.4, below, xshift=2.5em, draw=none, fill=none] {$1-\frac{\eta}{9}$} (E9);
\draw (E7) -- node[pos=0.2, below, xshift=1.6em, draw=none, fill=none] {$1-\frac{\eta}{18}$} (E10);
\draw[bend left=40] (E4) to (E10);

%----layer 4
\node (E11) [below=6.5cm of E1] {
\begin{tabular}{c}
\eqref{eq:bound_eig_A} \\
$\left\| \left(\b S_{d^m}^m \right)^{-\frac{1}{2}} \right\|$ \\
$1-\eta$
\end{tabular}
};
\node (E12) [right=1cm of E11] {
\begin{tabular}{c}
\eqref{bound_diff_psd2} \\
 $\left \| \b U_{d^m}^m \left( \b U_{d^m}^m \right)^\T \left | \b A^m \right | - \b K^m \right \|$ \\
 $1-\frac{\eta}{18}$
\end{tabular}
};
\node (E13) [right=1cm of E12] {
\begin{tabular}{c}
\eqref{bound_R4} \\
$\left\| \b R_4 \right\|$ \\
$1-\frac{\eta}{18}$
\end{tabular}
};
\node (E17) [right=1cm of E13] {
\begin{tabular}{c}
\eqref{bound_Jm_R1} \\
$\left\| \b R_1 \right\|$ \\
$1-\frac{\eta}{3}$ 
\end{tabular}
};
\node (E14) [right=3cm of E13] {
\begin{tabular}{c}
\eqref{bound_tilde_R2} \\
$\left\| \tilde{\b R}_2 \right\|_F$ \\
$1-\frac{\eta}{3}$ 
\end{tabular}
};

%-edges-----------
\draw (E8) to (E14);
\draw (E9) to (E14);
\draw (E10) to (E14);
\draw [bend right=10] (E5) to node[pos=0.7, below, xshift=2.75em, draw=none, fill=none] {$1-\frac{\eta}{36}$} (E13);
\draw [bend right=40] (E2) to node[pos=0.5, below, xshift=1.75em, draw=none, fill=none] {$1-\frac{\eta}{18}$} (E12);
\draw [bend right=10] (E2) to node[pos=0.5, left, xshift=-0.25em, draw=none, fill=none] {$1-\frac{\eta}{8}$} (E11);
\draw (E1) to node[pos=0.2, left, draw=none, fill=none] {$1-\left(\frac{\eta}{2}+\frac{\eta}{8}+\frac{\eta}{8}\right)$} (E11);
\draw[->, preaction={draw=white, -, line width=6pt}] (E6) to node[pos=0.7, below, xshift=-3em, draw=none, fill=none] {$1-\frac{\eta}{8}$} (E11);
\draw[->, preaction={draw=white, -, line width=6pt}] (E7) -- node[pos=0.2, below, xshift=-1.5em, draw=none, fill=none] {$1-\frac{\eta}{3}$} (E17);

%--Layer 5
\node (E15) [below=1.5cm of E12] {
\begin{tabular}{c}
\eqref{bound_d2} \\
$\left\| \b F_2 \right\|$ \\
$1-\frac{\eta}{6}$
\end{tabular}
};
\node (E16) [right=1cm of E15] {
\begin{tabular}{c}
\eqref{bound_d1} \\
$\left\| \b F_1 \right\|$ \\
$1-\frac{\eta}{6}$
\end{tabular}
};
\node (E23) [right=1cm of E16] {
\begin{tabular}{c}
\eqref{eq:bound_adj_eig} \\
$\left\| \left( \b S_{d^m}^m \right)^{\frac{1}{2}} \right\|$ \\
$1-\frac{\eta}{6}$
\end{tabular}
};
\node (E18) [below=1.5cm of E14] {
\begin{tabular}{c}
\eqref{bound_R2} \\
$\left\| \b R_2 \right\|_F$ \\
$1-\frac{\eta}{3}$
\end{tabular}
};

%--edges
\draw (E11) to node[pos=0.4, below, xshift=-1.75em, draw=none, fill=none] {$1-\frac{\eta}{12}$} (E15);
\draw (E11) to node[pos=0.15, below, xshift=3em, draw=none, fill=none] {$1-\frac{\eta}{18}$} (E16);
\draw (E12) -- (E16);
\draw[->, preaction={draw=white, -, line width=6pt}] (E12) to node[pos=0.4, below, xshift=1.3em, draw=none, fill=none] {$1-\frac{\eta}{12}$} (E15);
\draw (E13) to (E16);
\draw (E17) to node[above, yshift=0.3em, draw=none, fill=none] {$1-\frac{\eta}{36}$} (E13);
\draw (E14) to (E18);

%---Layer 6

\node (E19) [below=0.75cm of E16] {
\begin{tabular}{c}
\eqref{bound_E2} \\
$\left\| \b E_2 \right\|_F$ \\
$1-\frac{\eta}{3}$
\end{tabular}
};
\node (E20) [right=1cm of E19] {
\begin{tabular}{c}
\eqref{bound_A2} \\
$\left\| \b A_2 \right\|$ \\
$1-\frac{\eta}{3}$
\end{tabular}
};
\node (E21) [right=1cm of E20] {
\begin{tabular}{c}
\eqref{bound_A3} \\
$\left\| \b A_3 \right\|_F$ \\
$1-\frac{\eta}{3}$
\end{tabular}
};

% Edges 

\draw (E15) -- (E19);
\draw (E16) -- (E19);
\draw (E23) -- (E20);
\draw [bend left=30] (E17) to node[pos=0.2, left, xshift=-0.25em, draw=none, fill=none] {$1-\frac{\eta}{6}$} (E20);
\draw (E18) -- (E21);

%Final layer
\node (E22) [below=0.75cm of E20] {
\begin{tabular}{c}
\eqref{eq:final_bound} \\
$\left \| \bm{\hat{\Phi}}^m_{d^m} - \bm{\Phi}^m_{d^m}  \right \|_F$ \\
$1-\eta$
\end{tabular}
};

\draw (E19) -- (E22);
\draw (E20) -- (E22);
\draw (E21) -- (E22);

\end{tikzpicture}

\caption{Illustration of $\J^m$ events (boxes) leading to our probabilistic bound on $\left \| \bm{\hat{\Phi}}^m_{d^m} - \bm{\Phi}^m_{d^m}  \right \|_F$ (bottom of the figure). In each box, the associated equation and the estimated quantity are included with the probability of the event. The notation $B \rightarrow B'$ indicates that the event $B'$ can be derived from the event $B$; $B_1, B_2, \ldots, B_N \rightarrow B'$ ($N\ge 2$) means that the events $(B_i)_{i=1}^N$ are combined with a union bound to control the event $B'$. When an $\eta$ reparameterization is applied, the corresponding probability shows up on the associated arrow. } 
    \label{fig:Jm_events}
\end{figure}

%%%%%%%%%%%%%%%%%

Let us define the projections $\mathcal{P}^m_{\left| \b A^m \right|}:= \b U^m_{d^m} \left(\b U^m_{d^m}\right)^\top$ and $\mathcal{P}^m_{\b K^m}:= \b U^m_{\b K^m,d^m} \left(\b U^m_{\b K^m,d^m}\right)^\top$. Let
\begin{align}\label{eq:def_C^m}
    C^m:= c_2^m \log \left( 2 + \log\left ( \frac{\left \| \b A^m \right \| + \left \| \b K^m \right \| }{\left \| \b A^m - \b K^m \right \| }  \right )  \right)
\end{align}
for the constant $c_2^m > 0$. Under the imposed assumption of $C_1^m > 2 \sqrt{2}$ and $c_2^m \ge 2$, one can show (Lemma~\ref{lemma:bound_Cm}) that
\begin{align}
    2 \log 2 \le C^m \le c_2^m \log \left( 2 + \log(2n) \right) = c_1^m(n),
\end{align}
where the lower bound always holds and the upper bound holds for any $\eta \in (0,1)$ and $n \ge \max \left\{ \colAoneii{n^m_{0,(ii)}\left( \frac{\eta}{2}\right)}, \colLsix{\frac{8}{\eta}}, \coletwo{e^{\frac{1}{\left( 2C_1^m - 4\sqrt{2} \right)^2}} } \right\}$ with $\J^m$-probability at least $1-\eta$. Then, for any $\eta \in \left( 0,1\right)$ and $n \ge \max \left\{ \colAonei{ n^m_{0,(i)}\left( \frac{\eta}{12}\right)}, \colAoneii{n^m_{0,(ii)}\left( \frac{\eta}{12}\right)}, \colAii{ n_{1,(ii)}^m(\eta)}, \colLfive{\left( \frac{\eta}{12} \right)^{-\frac{1}{c^m}}} , \colLsix{\frac{48}{\eta}}, \coletwo{e^{\frac{1}{\left( 2C_1^m - 4\sqrt{2} \right)^2}} } \right\}$
\begin{align}
    \left \| \mathcal{P}^m_{\left| \b A^m \right|} - \mathcal{P}^m_{\b K^m} \right \| & \stackrel{(a)}{\le} \frac{2 \left \| \left | \b A^m \right | - \b K^m \right \| }{\lambda^m_{d^m}(\b K^m) - \lambda^m_{d^m+1}(\b K^m)} \stackrel{(b)}{\le} \frac{8 c_1^m(n) \sqrt{n \log \left( \frac{12n}{\eta}\right)}}{n \delta^m_{d^m}/2}\\
    & = 16 c_1^m(n) \left( \delta^m_{d^m} \right)^{-1} \sqrt{\frac{\log \left(  \frac{12n}{\eta}\right)}{n}}, \label{bound_proj0}
\end{align}
holds with $\J^m$-probability at least $1-\eta$. Indeed, the inequalities above hold for the following reasons.
\begin{itemize}
    \item (a) in \eqref{bound_proj0} follows from \eqref{sin_theta_theorem3} in Corollary~\ref{lemm_sin_theta_var1} and setting $s=d^m$. In order to use \eqref{sin_theta_theorem3} in Corollary~\ref{lemm_sin_theta_var1} we need to verify that 
    \begin{align}
     \left \| \left | \b A^m \right | -  \b K^m    \right \| \le \frac{\lambda^m_{d^m}(\b K^m) - \lambda^m_{d^m+1}(\b K^m)}{2}.\label{eq:diff-target}
    \end{align}
    
    \begin{itemize}
        \item Concerning $\left \| \left | \b A^m \right | -  \b K^m  \right \|$: 
        \begin{itemize}
            \item \textbf{Fixed $(X^m_i)_{i=1}^n=(x^m_i)_{i=1}^n$, randomness in $\b A^m$:}  We have for any $\eta \in  \left(0, 6 \right] $ and $n \ge \colLfive{ \left( \frac{\eta}{12} \right)^{-\frac{1}{c^m}} }$ that
        \begin{align}
            & \quad \left \| \left | \b A^m \right | -  \b K^m \right \| \stackrel{(a)}{=} \left \| \left | \b A^m \right | -  \left| \b K^m  \right|   \right \|  \\
            &\stackrel{(b)}{\le}\underbrace{ c_2^m \log \left( 2 + \log\left ( \frac{\left \| \b A^m \right \| + \left \| \b K^m \right \| }{\left \| \b A^m - \b K^m \right \| }  \right )  \right)}_{=:\tilde{C}^m\left(x_1^m,\ldots,x_n^m\right)=: \tilde{C}^m} \left \| \b A^m - \b K^m \right \| \stackrel{(c)}{\le} 4 \tilde{C}^m \sqrt{\Delta^m \log \left(  \frac{n}{\eta /12}\right)}  \label{eq:fix_X_spectral_diff}
        \end{align}
        holds with $\Q^m$-probability at least $1-\frac{\eta}{12}$, where $\Q^m$ denotes the distribution of the entries in $\b A^m$ given the latent positions $\left(x^m_i\right)_{i=1}^n$ and $\tilde{C}^m$ denotes the Lipschitz constant for random $\b A^m$ and fixed $\left( X_i^m \right)_{i=1}^n$. Indeed, (a) holds since $\left | \b K^m \right |=((\b K^m)^\top \b K^m)^{\frac{1}{2}}=\b K^m$ by the positive semi-definiteness of  $\b K^m$. (b) is implied by Lemma~\ref{bound_psd_trans}, where $c_2^m > 0$ is a constant. 
        (c) follows with $\Q^m$-probability at least $1-\frac{\eta}{12}$ for any $\eta \in \left(0, \frac{12}{2} \right] =(0,6]$ and $n \ge \colLfive{ \left( \frac{\eta}{12} \right)^{-\frac{1}{c^m}} }$ from Lemma~\ref{concentration_dense1} by a substitution $\eta \leftarrow \frac{\eta}{12}$.
        Therefore, we got that for any $\eta \in  \left(0, 6 \right]$ and $n \ge \colLfive{ \left( \frac{\eta}{12} \right)^{-\frac{1}{c^m}} }$
        \begin{align}\label{event_C}
            \hspace{-0.6cm} \Q^m \left( \left \| |\b A^m|  - \b K^m \right \| \le 4 \tilde{C}^m \sqrt{\Delta^m \log \left(  \frac{12 n}{\eta}\right)} \mid (X^m_i)_{i=1}^n=(x^m_i)_{i=1}^n \right) & \ge 1-\frac{\eta}{12},
        \end{align}  
        where $\Delta^m:= \max_{i \in [n]} \sum_{j \in [n]} K^m_{i,j}$ with $K^m_{i,j} = k^m\left(x^m_i,x^m_j\right)$ ($i,j\in [n]$).
        \item \textbf{Randomness in $(X^m_i)_{i=1}^n$}: Let now $\Delta^m =\Delta^m\left(X^m_1,\ldots,X^m_n;n\right)$ be random, where $\Delta^m:= \max_{i \in [n]} \sum_{j \in [n]} K^m_{i,j}$ with $K^m_{i,j} = k^m\left(X^m_i,X^m_j\right)$ ($i,j\in [n]$). According to Assumption~\ref{assum_Delta}(i), we have that $\Delta^m > C_0^m \log(n)$ holds with $\left(\P^m\right)^{\otimes n}$-probability at least $1-\frac{\eta}{12}$ for any $\eta \in (0,12)$ and $n \ge \colAonei{ n^m_{0,(i)}\left( \frac{\eta}{12}\right)}$ by a substitution $\eta \leftarrow \frac{\eta}{12}$. Besides, one has $\left(\P^m \right)^{\otimes n} \left( \Delta^m \le n  \right) =1$ since $K^m_{i,j} \le 1$ according to the boundedness of the kernel $k^m$ (i.e., $k^m(x^m,\tilde{x}^m) \le 1$ for any $x^m, \tilde{x}^m \in \X^m$). Therefore, for any $\eta \in (0,12)$ and $n \ge \colAonei{ n^m_{0,(i)}\left( \frac{\eta}{12}\right)}$, one has
        \begin{align}
        \left(\P^m\right)^{\otimes n} \left(C_0^m \log(n) < \Delta^m \le n\right) \ge 1-\frac{\eta}{12}. \label{eq:event-B}
        \end{align}
        
        \item \textbf{Combination}: Now we combine the above mentioned results. Define the following events:
        \begin{align}
            A & = \left \{ \left( (x^m_i)_{i=1}^n, \b A^m \right): \left \| \left | \b A^m \right | -  \b K^m  \right \| \le 4 \tilde C^m \sqrt{n \log \left(  \frac{12 n}{\eta}\right)} \right \} \\
            & = \left \{ \left( (x^m_i)_{i=1}^n, \b A^m \right): \underbrace{\left \| \left | \b A^m \right | -  \b K^m  \right \|}_{=:f(x,y)} \le \underbrace{4 \tilde C^m \sqrt{\log \left(  \frac{12 n}{\eta}\right)}}_{=:g(x)} \underbrace{\sqrt{n}}_{=:b} \right \} ,\\
            B & = \left \{ (x^m_i)_{i=1}^n : C_0^m \log(n) < \Delta^m \le n  \right \} \\
            & = \left \{ (x^m_i)_{i=1}^n : \underbrace{\sqrt{C_0^m \log(n)}}_{=:a} < \underbrace{\sqrt{\Delta^m}}_{=: S_0(x)} \le \underbrace{\sqrt{n}}_{=:b}  \right \} \subseteq \left( \X^m \right)^n, \\
            C & = \left \{ \left( (x^m_i)_{i=1}^n, \b A^m \right): \left \| \left | \b A^m \right | -  \b K^m  \right \| \le 4 \tilde C^m \sqrt{\Delta^m \log \left(  \frac{12n}{\eta}\right)}, (x^m_i)_{i=1}^n \in B   \right \} \\
            &= \left \{ \left( (x^m_i)_{i=1}^n, \b A^m \right): \underbrace{\left \| \left | \b A^m \right | -  \b K^m  \right \|}_{=f(x,y)} \le \underbrace{4 \tilde C^m \sqrt{\log \left(  \frac{12n}{\eta}\right)}}_{=g(x)} \underbrace{\sqrt{\Delta^m}}_{=S_0(x)}, (x^m_i)_{i=1}^n \in B   \right \},
        \end{align}
        and by definition $C \subseteq A$ holds. With these notations, we have (i) by \eqref{event_C} for any $\eta \in (0,6]$ and  $n \ge \colLfive{ \left( \frac{\eta}{12} \right)^{-\frac{1}{c^m}} }$ that $\Q^m\left( C \mid (x^m_i)_{i=1}^n \right) \ge 1-\frac{\eta}{12}$, (ii) by \eqref{eq:event-B} for any $\eta \in (0,12)$ and $n \ge \colAonei{ n^m_{0,(i)}\left( \frac{\eta}{12}\right)}$ that $\left(\P^m\right)^{\otimes n}(B) \ge 1-\frac{\eta}{12}$. Applying  Lemma~\ref{lemm:combine_probs}\ref{group:combine_probs1} with the notations $f$, $g$, $S_0$, $a$ and $b$ introduced above, and with $\delta_1 = \delta_2 = \frac{\eta}{12}$, one gets for any $\eta \in (0,6] \cap (0,12) = (0,6]$ and $n \ge \max\left\{ \colAonei{ n^m_{0,(i)}\left( \frac{\eta}{12}\right)}, \colLfive{ \left( \frac{\eta}{12} \right)^{-\frac{1}{c^m}}} \right\}$
        \begin{align}\label{eq:prob_event1}
            (\J^m)(A) \ge  1-\left(\frac{\eta}{12} + \frac{\eta}{12} \right) = 1-\frac{\eta}{6},
        \end{align}
        where $\J^m$ denotes the joint distribution of $(X^m_i)_{i=1}^n$ and the entries in $\b A^m$. In the event $A$ appearing in $(\J^m)(A)$, one is considering $(X_i^m)_{i=1}^n$ to be random, thus $\tilde{C}^m$ is also replaced by $C^m$.
        \end{itemize}

        Therefore, we have for any $\eta \in (0,6]$ and $n \ge \max\left\{ \colAonei{ n^m_{0,(i)}\left( \frac{\eta}{12}\right)}, \colLfive{ \left( \frac{\eta}{12} \right)^{-\frac{1}{c^m}}} \right\}$ that
        \begin{align}%\label{bound_diff_psd}
            \J^m\left(\left \| |\b A^m|  - \b K^m \right \| \le 4 C^m \sqrt{n \log \left(  \frac{12n}{\eta}\right)} \right)  \ge 1-\frac{\eta}{6}.
        \end{align}
        According to Lemma~\ref{lemma:bound_Cm}, by a substitution of $\eta \leftarrow \frac{\eta}{6}$, we have for any $\eta \in (0,12)$ and $n \ge \max \left\{ \colAoneii{n^m_{0,(ii)}\left( \frac{\eta}{12}\right)}, \colLsix{\frac{48}{\eta}}, \coletwo{e^{\frac{1}{\left( 2C_1^m - 4\sqrt{2} \right)^2}} } \right\}$
        \begin{align}
            \J^m \left( C^m \le c_1^m(n) \right) \ge 1 - \frac{\eta}{6}.
        \end{align}
        By combining the above two events, we have for any $\eta \in (0, 6] \cap (0,12) \cap (0,3) = (0, 3)$ and for $n \ge \max \left\{ \colAonei{ n^m_{0,(i)}\left( \frac{\eta}{12}\right)}, \colAoneii{n^m_{0,(ii)}\left( \frac{\eta}{12}\right)}, \colLfive{ \left( \frac{\eta}{12} \right)^{-\frac{1}{c^m}}}, \colLsix{\frac{48}{\eta}}, \coletwo{e^{\frac{1}{\left( 2C_1^m - 4\sqrt{2} \right)^2}} } \right\}$ that    \begin{align}\label{bound_diff_psd}
            \J^m\left(\left \| |\b A^m|  - \b K^m \right \| \le 4 c_1^m(n) \sqrt{n \log \left(  \frac{12n}{\eta}\right)}  \right) \ge 1-\frac{\eta}{3},
        \end{align}
        where restricting $\eta$ to $(0,3)$ guarantees that the r.h.s.\ of \eqref{bound_diff_psd} is positive, hence the bound is non-vacuous. 
        \item Concerning $\lambda^m_{d^m}(\b K^m) - \lambda^m_{d^m+1}(\b K^m)$: We have that for any $\eta\in \left(0,6 \right)$ 
        \begin{align}\label{eigen_K_delta_d}
        %\MoveEqLeft
             & \left( \lambda^m_{d^m} - \lambda^m_{d^m+1} \right) - \left[ \frac{1}{n}\lambda^m_{d^m}(\b K^m) - \frac{1}{n}\lambda^m_{d^m+1}(\b K^m) \right] \nonumber \\
             & \stackrel{(a)}{=}\left[ \lambda^m_{d^m} - \frac{1}{n}\lambda^m_{d^m}(\b K^m) \right] + \left[ \frac{1}{n}\lambda^m_{d^m+1}(\b K^m) - \lambda^m_{d^m+1} \right] 
             \\
             &\stackrel{(b)}{\le} \left | \lambda^m_{d^m} - \frac{1}{n}\lambda^m_{d^m}(\b K^m) \right | + \left | \frac{1}{n}\lambda^m_{d^m+1}(\b K^m) - \lambda^m_{d^m+1} \right | \stackrel{(c)}{\le} 4 \sqrt{2} a^2_{k^m} \sqrt{\frac{\log \left ( \frac{6}{\eta}  \right ) }{n} }  \nonumber
        \end{align}
        holds with $(\P^m)^{\otimes n}$-probability at least $1-\frac{2\eta}{3}$. 
        In (a) the terms were rearranged, (b) follows from $a\le |a|$ ($a\in \R$) and the triangle inequality, (c)  comes from \eqref{eq:gap_between_eigenvals} in Lemma~\ref{eigen_Tk} by letting $2e^{-\tau}=\frac{\eta}{3}$, i.e.\ $\tau = \log \left( \frac{6}{\eta} \right)$ and using the definition of $a_{k^m}:=\sup_{x^m \in \X^m} \sqrt{k^m(x^m,x^m)}$.\footnote{The eigenvalues $\lambda^m_{d^m}(\b K^m)$ and  $\lambda^m_{d^m+1}(\b K^m)$ in Lemma~\ref{eigen_Tk} are replaced by $\frac{1}{n}\lambda^m_{d^m}(\b K^m)$ and $\frac{1}{n}\lambda^m_{d^m+1}(\b K^m)$ in our setting, respectively, to adapt the normalized setting of the lemma to our case.} 

        Rearranging the terms in \eqref{eigen_K_delta_d} and applying the definition $\delta^m_{d^m} = \lambda^m_{d^m} - \lambda^m_{d^m+1}$, we get that 
        \begin{align}
            \frac{1}{n} \lambda^m_{d^m}(\b K^m) - \frac{1}{n} \lambda^m_{d^m+1}(\b K^m)  &\ge \delta^m_{d^m} - 4 \sqrt{2} a^2_{k^m} \sqrt{\frac{\log \left ( \frac{6}{\eta}  \right ) }{n} } \stackrel{\cdot n}{\Rightarrow}\\
            \lambda^m_{d^m}(\b K^m) - \lambda^m_{d^m+1}(\b K^m) &\ge n \delta^m_{d^m} - 4 \sqrt{2} a^2_{k^m} \sqrt{n \log \left ( \frac{6}{\eta}  \right ) } \stackrel{(a)}{\ge} \frac{n \delta^m_{d^m}}{2} \label{eigen_K_delta_d1}
        \end{align}
        holds with $(\P^m)^{\otimes n}$-probability at least $1-\frac{2 \eta}{3}$ for any $\eta \in \left( 0,1\right) \cap (0,6)=\left( 0,1\right) $ and $n\ge \colAii{ n_{1,(ii)}^m(\eta) }$. (a) is implied by the Assumption~\ref{assum_lambda_delta}(ii); indeed
        \begin{align}
        (a) & \Leftrightarrow \frac{n \delta^m_{d^m}}{2} \ge  4 \sqrt{2} a^2_{k^m} \sqrt{n \log \left ( \frac{6}{\eta}  \right ) } \stackrel{\cdot\frac{2}{n}}{\Leftrightarrow} \delta^m_{d^m} \ge 8 \sqrt{2} a^2_{k^m} \sqrt{\frac{\log \left ( \frac{6}{\eta}  \right ) }{n}},\\
        \delta^m_{d^m} & \stackrel{(b)}{\ge} 16 \left[2 \log \left( 2 + \log(2n) \right)\right]\sqrt{\frac{\log\left(\frac{6n}{\eta}\right)}{n} } \stackrel{(c)}{\ge}  32 \log \left( 2 + \log(4) \right)   \sqrt{\frac{\log\left(\frac{6}{\eta}\right)}{n} } \\
        & \stackrel{(d)}{\ge}8 \sqrt{2} a^2_{k^m} \sqrt{\frac{\log \left ( \frac{6}{\eta}  \right ) }{n}},
        \end{align}
        where (b) follows from Assumption~\ref{assum_lambda_delta}(ii) for any $\eta \in \left( 0,1\right)$ with $n\ge \colAii{ n_{1,(ii)}^m(\eta)}$ and the assumption $c_2^m \ge 2$, (c) is implied by $n\ge 2$ and the monotonicity of the $\log$ and the square root function, (d) holds since $32 \log \left( 2 + \log(4) \right) \ge 8\sqrt{2}$ and $1 \ge a_{k^m}^2$.  
        Therefore, we got that for any $\eta \in \left( 0,1\right)$ and $n \ge \colAii{n_{1,(ii)}^m(\eta)}$
        \begin{align}
            (\P^m)^{\otimes n} \left( \lambda^m_{d^m}(\b K^m) - \lambda^m_{d^m+1}(\b K^m) \ge \frac{n \delta^m_{d^m}}{2}  \right) \ge 1-\frac{2\eta}{3}. \label{eq:prob_event_A}
        \end{align}         
        
        \end{itemize}
        Define the following event:
        \begin{align}
            A = \left\{ (x^m_i)_{i=1}^n \,:\, \lambda^m_{d^m}(\b K^m) - \lambda^m_{d^m+1}(\b K^m) \ge \frac{n \delta^m_{d^m}}{2}   \right\} \subseteq \left( \X^m  \right)^n,
        \end{align}
        let $\Y^m:=\{ 0,1\}^{\binom{n}{2}}$ be the sample space of r.v.s from $\Q^m$. Then we have
        \begin{align}
            \J^m(A \times \Y^m) & =(\P^m)^{\otimes n} (A) \stackrel{(a)}{\ge} 1-\frac{2\eta}{3},
        \end{align} 
        where (a) is implied by \eqref{eq:prob_event_A}. Hence we got that for any $\eta \in \left( 0,1\right)$ and $n\ge \colAii{ n_{1,(ii)}^m(\eta)}$
        \begin{align}\label{eig_second_bound}
            \J^m \left( \lambda^m_{d^m}(\b K^m) - \lambda^m_{d^m+1}(\b K^m) \ge \frac{n \delta^m_{d^m}}{2}  \right) \ge 1-\frac{2\eta}{3}.
        \end{align}
        \eqref{eq:diff-target} is established if the upper bound of the $\left \| \left | \b A^m \right | -  \b K^m    \right \|$ is smaller or equal than the lower bound of $\frac{\lambda^m_{d^m}(\b K^m) - \lambda^m_{d^m+1}(\b K^m)}{2}$; therefore we want that 
     \begin{align}\label{delta_constriant2}
            4 c_1^m(n) \sqrt{n \log \left(  \frac{12 n}{\eta}\right)}  \le \frac{n \delta^m_{d^m}}{4} \Leftrightarrow \delta^m_{d^m}  \ge  16 c_1^m(n) \sqrt{\frac{\log \left(  \frac{12 n}{\eta}\right)}{n}};
        \end{align}
        this inequality holds for $n\ge n^m_{1,(ii)}\left(\frac{\eta}{2}\right)$ using Assumption~\ref{assum_lambda_delta}(ii).
        Hence (a) in \eqref{bound_proj0} holds with $\J^m$-probability at least $1-\eta$ for any $\eta \in \left( 0,1 \right) \cap (0, 3) =\left( 0,1\right)$ and $n \ge \max \left\{ \colAonei{ n^m_{0,(i)}\left( \frac{\eta}{12}\right)}, \colAoneii{n^m_{0,(ii)}\left( \frac{\eta}{12}\right)}, \colAii{ n_{1,(ii)}^m(\eta)}, \colAii{n^m_{1,(ii)}\left(\frac{\eta}{2}\right)}, \colLfive{\left( \frac{\eta}{12} \right)^{-\frac{1}{c^m}}} , \colLsix{\frac{48}{\eta}}, \coletwo{e^{\frac{1}{\left( 2C_1^m - 4\sqrt{2} \right)^2}} } \right\}$ by 
        combining the events
        \begin{align}
        A & = \left \{ \left( (x^m_i)_{i=1}^n, \b A^m \right): \left \| \left | \b A^m \right | -  \b K^m  \right \| \le 4 c_1^m(n) \sqrt{n \log \left(  \frac{12n}{\eta}\right)}  \right \} ,\\
        B & = \left \{ \left( (x^m_i)_{i=1}^n, \b A^m \right):  \lambda^m_{d^m}(\b K^m) - \lambda^m_{d^m+1}(\b K^m) \ge \frac{n \delta^m_{d^m}}{2}  \right \},
    \end{align}
       i.e., union bounding \eqref{bound_diff_psd} and \eqref{eig_second_bound}.
    \item For (b) in \eqref{bound_proj0}, the same combination of the events $A$ and $B$ as used at the proof of (a) in \eqref{bound_proj0} gives that
    \begin{align}
        \J^m & \left( \frac{2 \left \| \left | \b A^m \right | - \b K^m \right \| }{\lambda^m_{d^m}(\b K^m) - \lambda^m_{d^m+1}(\b K^m)} \le \frac{8 c_1^m(n) \sqrt{n \log \left(  \frac{12n}{\eta}\right)}}{n \delta^m_{d^m}/2}  \right) \ge 1-\left(\frac{\eta}{3} + \frac{2 \eta}{3}\right) = 1-\eta.
    \end{align}
    Thus, for any $\eta \in \left( 0,1\right)$ and 
    \begin{align}
        n \ge \max \left\{ \colAonei{ n^m_{0,(i)}\left( \frac{\eta}{12}\right)}, \colAoneii{n^m_{0,(ii)}\left( \frac{\eta}{12}\right)}, \colAii{ n_{1,(ii)}^m(\eta)}, \colAii{n^m_{1,(ii)}\left(\frac{\eta}{2}\right)}, \colLfive{\left( \frac{\eta}{12} \right)^{-\frac{1}{c^m}}} , \colLsix{\frac{48}{\eta}}, \coletwo{e^{\frac{1}{\left( 2C_1^m - 4\sqrt{2} \right)^2}} } \right\},
    \end{align}
    we have that 
    \begin{align}\label{bound_proj}
    \J^m \left( \left \| \mathcal{P}^m_{\left| \b A^m \right|} - \mathcal{P}^m_{\b K^m} \right \| \le 16 c_1^m(n) \left( \delta^m_{d^m} \right)^{-1} \sqrt{\frac{\log \left(  \frac{12n}{\eta}\right)}{n}} \right) \ge 1-\eta.
    \end{align}
\end{itemize}
Having established \eqref{bound_proj}, we now move on to bounding the difference between $\bm{\hat{\Phi}}^m_{d^m}$ and $\bm{\Phi}^m_{d^m}$. Based on the SVD of $\left(\b U^m_{\b K^m,d^m}\right)^\top \b U^m_{d^m} \in \R^{d^m\times d^m}$
\begin{align}
\left(\b U^m_{\b K^m,d^m}\right)^\top \b U^m_{d^m} &=\b W_1 \b \Sigma \b W_2^\top, \quad \left(\b W_1 \coloneq \b W_1^m, \bm{\Sigma} \coloneq \bm{\Sigma}^m, \b W_2 \coloneq \b W_2^m \in \R^{d^m \times d^m}\right),
\end{align}
let us define the matrices $\b W^m$, $\b R_1$, and $\b R_2$ as follows:
\begin{align}
\b W^m &:= \b W_1 \b W_2^\top \in \R^{d^m \times d^m}, \label{eq:Wm-def}\\
\b R_1 \coloneq \b R_1^m &:= \b U^m_{\b K^m,d^m} \left(\b U^m_{\b K^m,d^m}\right)^\top \b U^m_{d^m} - \b U^m_{\b K^m,d^m} \b W^m \in \R^{n \times d^m},\label{eq:R1-def}\\
\b R_2 \coloneq \b R_2^m &:= \b W^m \left(\b S^m_{d^m}\right)^{\frac{1}{2}} - \left(\b S^m_{\b K^m,d^m}\right)^{\frac{1}{2}} \b W^m \in \R^{d^m \times d^m}.\label{eq:R2-def}
\end{align}
Then, $\b W^m$ is orthogonal by the orthogonality of $\b W_1$ and $\b W_2$, and we have
\begin{align}
    & \quad \bm{\hat{\Phi}}^m_{d^m} - \bm{\Phi}^m_{d^m} = \hat{\b \Phi}^m_{d^m} - \b U^m_{\b K^m,d^m} \left(\b S^m_{\b K^m,d^m} \right)^{\frac{1}{2}} \b W^m  = \label{eq:initial_decompose-line1}\\
    & \stackrel{(a)}{=} \b U^m_{d^m} \left( \b S^m_{d^m} \right)^{\frac{1}{2}} - \b U^m_{\b K^m,d^m} \b W^m \left( \b S^m_{d^m} \right)^{\frac{1}{2}} + \b U^m_{\b K^m,d^m} \b W^m \left( \b S^m_{d^m} \right)^{\frac{1}{2}} - \b U^m_{\b K^m,d^m} \left( \b S^m_{\b K^m,d^m} \right)^{\frac{1}{2}} \b W^m \\
    & \stackrel{(b)}{=} \left(\b U^m_{d^m} - \b U^m_{\b K^m,d^m} \b W^m \right) \left( \b S^m_{d^m} \right)^{\frac{1}{2}} + \b U^m_{\b K^m,d^m} \underbrace{\left[ \b W^m \left( \b S^m_{d^m} \right)^{\frac{1}{2}}-\left( \b S^m_{\b K^m,d^m} \right)^{\frac{1}{2}} \b W^m \right]}_{= \b R_2 \in \R^{d^m\times d^m}} \\
    & \stackrel{(c)}{=} \left[ \b U^m_{d^m} - \b U^m_{\b K^m,d^m} \left( \b U^m_{\b K^m,d^m} \right)^\top \b U^m_{d^m} + \underbrace{\b U^m_{\b K^m,d^m} \left( \b U^m_{\b K^m,d^m} \right)^\top \b U^m_{d^m} - \b U^m_{\b K^m,d^m} \b W^m}_{=\b R_1 \in \R^{n\times d^m}} \right] \left( \b S^m_{d^m} \right)^{\frac{1}{2}} \\
    & \quad + \b U^m_{\b K^m,d^m} \b R_2\\
    &= \underbrace{\left[\b U^m_{d^m} - \b U^m_{\b K^m,d^m} \left( \b U^m_{\b K^m,d^m} \right)^\top \b U^m_{d^m} \right] \left( \b S^m_{d^m} \right)^{\frac{1}{2}}}_{=: \b A_1^m =:\b A_1 \in \R^{n \times d^m}} + \underbrace{\b R_1 \left( \b S^m_{d^m} \right)^{\frac{1}{2}}}_{=:\b A_2^m =:\b A_2 \in \R^{n \times d^m}} + \underbrace{\b U^m_{\b K^m,d^m} \b R_2}_{=:\b A_3^m =: \b A_3 \in \R^{n \times d^m}}\label{eq:initial_decompose}.
\end{align}
In (a), we used the definition of $\hat{\b \Phi}^m_{d^m}$ [see \eqref{eq:hat-Phi-m-def}] and added $\pm \b U^m_{\b K^m,d^m} \b W^m \left( \b S^m_{d^m} \right)^{\frac{1}{2}} \in \R^{n\times d^m}$; (b) follows by collecting terms. In (c), we added $\pm \b U^m_{\b K^m,d^m} \left( \b U^m_{\b K^m,d^m} \right)^\top \b U^m_{d^m} \in \R^{n\times d^m}$. To bound $\b A_2$ and $\b A_3$, we first study the upper bounds of $\left\|\b R_1\right\|$ and $\left\|\b R_2\right\|_F$; afterwards we present our bound controlling $\b A_1$, $\b A_2$ and $\b A_3$. 
\begin{itemize}
    \item \textbf{Term} $\left\|\b R_1\right\|$: For $\left\|\b R_1\right\|$, for any $\eta \in \left( 0,1 \right)$ and 
    \begin{align}
        n \ge \max \left\{ \colAonei{ n^m_{0,(i)}\left( \frac{\eta}{12}\right)}, \colAoneii{n^m_{0,(ii)}\left( \frac{\eta}{12}\right)}, \colAii{ n_{1,(ii)}^m(\eta)}, \colAii{n^m_{1,(ii)}\left(\frac{\eta}{2}\right)}, \colLfive{\left( \frac{\eta}{12} \right)^{-\frac{1}{c^m}}} , \colLsix{\frac{48}{\eta}}, \coletwo{e^{\frac{1}{\left( 2C_1^m - 4\sqrt{2} \right)^2}} } \right\}
    \end{align}
    we have
    \begin{align}
        & \quad \left \| \b R_1 \right \|  \stackrel{\eqref{eq:R1-def}}{=} \left \| \b U^m_{\b K^m,d^m} \left( \b U^m_{\b K^m,d^m} \right)^\top \b U^m_{d^m} - \b U^m_{\b K^m,d^m} \b W^m \right \| \\
        & \stackrel{(a)}{=} \left \| \b U^m_{\b K^m,d^m} \left[ \left( \b U^m_{\b K^m,d^m} \right)^\top \b U^m_{d^m} - \b W_1 \b W_2^\top \right] \right \| \nonumber \\
        & \stackrel{(b)}{\le} \left \| \b U^m_{\b K^m,d^m} \right \| \left \| \left( \b U^m_{\b K^m,d^m} \right)^\top \b U^m_{d^m} - \b W_1 \b W_2^\top \right \| \stackrel{(c)}{\le } \left \| \left( \b U^m_{\b K^m,d^m} \right)^\top \b U^m_{d^m} - \b W_1 \b W_2^\top \right \|  \nonumber \\
        & \stackrel{(d)}{=} \left \| \b W_1 \left( \b \Sigma - \b I \right) \b W_2^\top \right \| \stackrel{(e)}{=} \left \| \b \Sigma - \b I \right \|, 
        \label{bound_R1}
    \end{align}    
    with $\J^m$-probability at least $1-\eta$. 
    (a) follows from the definition of $\b W^m$ [see \eqref{eq:Wm-def}]. (b) is implied by $\left \| \b A \b B \right \| \le \left \| \b A \right \| \left \| \b B \right \| $; (c) is obtained by $\left \| \b U^m_{\b K^m,d^m} \right \| \le 1$ following from $\left \| \mathcal{P}^m_{\b K^m} \right \| = \left \| \b U^m_{\b K^m,d^m} (\b U^m_{\b K^m,d^m})^{\top} \right \| = \left \| \b U^m_{\b K^m,d^m} \right \|^2 \le 1 $. 
    In (d), we used that $(\b U^m_{\b K^m,d^m})^{\top} \b U^m_{d^m} = \b W_1 \b \Sigma \b W_2$; $(e)$ follows from the orthogonality of $\b W_1$ and $\b W_2$. Further bounding the r.h.s. of the \eqref{bound_R1}, let $\sigma_1, \dots, \sigma_{d^m}$ denote the singular values (ordered in a non-increasing fashion) of $\left( \b U^m_{\b K^m,d^m} \right)^\top \b U^m_{d^m} \in \R^{d^m \times d^m}$; in this case $0\le \sigma_i = \cos (\theta_i) \le 1$ for $i \in [d^m]$ where $\theta_i$ is the $i$-th principal angle between the column space of $\b U^m_{d^m}$ and $\b U^m_{\b K^m,d^m}$. 
    Then we have 
    \begin{align}
        \left \| \b \Sigma - \b I \right \| &= \max_{i \in [d^m]} \left | 1-\sigma_i \right |  \stackrel{(f)}{\le} \max_{i \in [d^m]} (1-\sigma_i^2) \stackrel{(g)}{=} \max_{i \in [d^m]} \sin^2(\theta_i) \\
        & \stackrel{(h)}{=} \left \| \sin \Theta \left(\b U^m_{d^m} ,\b U^m_{\b K^m,d^m} \right) \right \|^2  \stackrel{(i)}{=} \left \| \mathcal{P}^m_{\left| \b A^m \right|} - \mathcal{P}^m_{\b K^m} \right \|^2 \\
        & \stackrel{(j)}{\le} 256 \left( c_1^m(n) \right)^2 \left( \delta^m_{d^m} \right)^{-2} \frac{\log \left(  \frac{12n}{\eta}\right)}{n}, \label{eq:bound_signma_diff_I}
    \end{align}
    where (f) is implied by $\sigma^2_i \le \sigma_i$ where this relation on $\sigma_i$ holds by $0 \le \sigma_i \le 1$, (g) follows from $\sigma_i=\cos (\theta_i)$ and the identity $\cos^2(\theta_i)+\sin^2(\theta_i)=1$. Let $\Theta \left(\b U^m_{d^m} ,\b U^m_{\b K^m,d^m} \right):= \diag \left(\left( \theta_i \right)_{i=1}^{d^m} \right) \in \R^{d^m \times d^m}$.  Using this notation,
    (h) holds by $\left\|  \diag(\b v) \right\| = \max_{i \in [d^m]} \left| v_i \right|$
    with $\b v := \left( \sin (\theta_i)\right)_{i=1}^{d^m}$, which gives $\left\|  \diag(\b v) \right\|^2 = \left(\max_{i \in [d^m]} \left| v_i \right|\right)^2 = \max_{i \in [d^m]} \left| v_i \right|^2$; the function $\sin$ acts coordinate-wise.      
    (i) comes from \eqref{proj_sin_theta}. (j) follows from \eqref{bound_proj} with $\J^m$-probability at least $1-\eta$ for any $\eta \in \left( 0,1\right)$ and 
    \begin{align}
        n \ge \max \left\{ \colAonei{ n^m_{0,(i)}\left( \frac{\eta}{12}\right)}, \colAoneii{n^m_{0,(ii)}\left( \frac{\eta}{12}\right)}, \colAii{ n_{1,(ii)}^m(\eta)}, \colAii{n^m_{1,(ii)}\left(\frac{\eta}{2}\right)}, \colLfive{\left( \frac{\eta}{12} \right)^{-\frac{1}{c^m}}} , \colLsix{\frac{48}{\eta}}, \coletwo{e^{\frac{1}{\left( 2C_1^m - 4\sqrt{2} \right)^2}} } \right\}.
    \end{align}
    Combining \eqref{bound_R1} and \eqref{eq:bound_signma_diff_I}, and choosing $\eta/3$ we get that for any $\eta \in \left( 0,3\right)$ and 
    \begin{align}
        n \ge \max \left\{ \colAonei{ n^m_{0,(i)}\left( \frac{\eta}{36}\right)}, \colAoneii{n^m_{0,(ii)}\left( \frac{\eta}{36}\right)}, \colAii{ n_{1,(ii)}^m\left( \frac{\eta}{3} \right)}, \colAii{n^m_{1,(ii)}\left(\frac{\eta}{6}\right)}, \colLfive{\left( \frac{\eta}{36} \right)^{-\frac{1}{c^m}}} , \colLsix{\frac{144}{\eta}}, \coletwo{e^{\frac{1}{\left( 2C_1^m - 4\sqrt{2} \right)^2}} } \right\},
    \end{align}
    we have 
    \begin{align}\label{bound_Jm_R1}
        \J^m \left( \left \| \b R_1 \right \| \le 256 \left( c_1^m(n) \right)^2 \left( \delta^m_{d^m} \right)^{-2} \frac{\log \left(  \frac{36 n}{\eta}\right)}{n} \right) \ge 1-\frac{\eta}{3}.
    \end{align}
    
    \item \textbf{Term} $\left\|\b R_2\right\|_F$: The bound on $\left\|\b R_2\right\|_F$ will be derived by that of the auxiliary quantity  $\left\|\tilde{\b R}_2\right\|_{F}$, where    
    \begin{align}
        \tilde{\b R}_2 &:= \tilde{\b R}_2^m :=\left( \b S^m_{\b K^m,d^m} \right)^{-\frac{1}{2}} \left(\b W^m \b S^m_{d^m} - \b S^m_{\b K^m,d^m} \b W^m \right) \in \R^{d^m \times d^m}. \label{eq:R2-tilde-def}
    \end{align}
    Notice that $\left( \b S^m_{\b K^m,d^m} \right)^{-\frac{1}{2}} = \diag \left( \left( \left( \lambda^m_{j}(\b K^m) \right)^{-\frac{1}{2}} \right)_{j=1}^{d^m} \right)$ and $\lambda^m_{j}(\b K^m) > 0$.\footnote{The inequality $\lambda_j^m(\b K^m)>0$ ($j\in [n]$) holds by Assumption~\ref{assum_eigenval_hat_C_k} and the relation $\lambda^m_j(\b K^m) = n \hat{\lambda}^m_j$ \citep[page~3]{sterge2020gain}; hence $(\lambda_j^m( \left|\b A^m\right| ))^{\frac{1}{2}} + (\lambda_j^m(\b K^m))^{\frac{1}{2}}>0$ and dividing by it is valid.\label{footnote:Km-has-positive-eigenvalues}}
    Particularly,
    \begin{align}\label{eq:decompose_tilde_R2}
        \left \| \tilde{\b R}_2 \right \|_F = \left \| \left( \b S^m_{\b K^m,d^m} \right)^{-\frac{1}{2}} \underbrace{\left(\b W^m \b S^m_{d^m} - \b S^m_{\b K^m,d^m} \b W^m \right)}_{=:\b B_1^m =: \b B_1 \in \R^{d^m \times d^m}} \right \|_F \stackrel{(a)}{\le} \left \| \left( \b S^m_{\b K^m,d^m} \right)^{-\frac{1}{2}} \right \|_F \left \| \b B_1 \right \|, 
    \end{align}
    where (a) follows from $\left \| \b A \b B \right \|_F \le \left \| \b A \right \|_F \left \| \b B \right \|  $. We proceed by upper bounding $\left\|\tilde{\b R}_2\right\|_F$, via upper bounding $\left\|\b B_1\right\|$.
   
    \textbf{Term} $\left\|\b B_1\right\|$: For the term $\b W^m \b S^m_{d^m}$ in $\b B_1$, we have the following decomposition:
        \begin{align}
        \b W^m \b S^m_{d^m} & \stackrel{(a)}{=} \left[ \b W^m - \left( \b U^m_{\b K^m,d^m} \right)^\top \b U^m_{d^m} \right] \b S^m_{d^m} + \left( \b U^m_{\b K^m,d^m} \right)^\top \b U^m_{d^m} \b S^m_{d^m}\\
        & \stackrel{(b)}{=} \left[\b W^m - \left( \b U^m_{\b K^m,d^m} \right)^\top \b U^m_{d^m} \right] \b S^m_{d^m} \\
        & \quad + \underbrace{\left( \b U^m_{\b K^m,d^m} \right)^\top \left[\b U^m_{d^m} \left( \b U^m_{d^m} \right)^\top \left | \b A^m \right |  - \b K^m \right]}_{=: \b R_3^m =: \b R_3 \in \R^{d^m\times n}} \b U^m_{d^m} + \left( \b U^m_{\b K^m,d^m} \right)^\top \b K^m \b U^m_{d^m}  \label{eq:R3-def}\\
        & \stackrel{(c)}{=} \left[\b W^m - \left( \b U^m_{\b K^m,d^m} \right)^\top \b U^m_{d^m} \right] \b S^m_{d^m} + \b S^m_{\b K^m,d^m} \left( \b U^m_{\b K^m,d^m} \right)^\top \b U^m_{d^m}  \\
        & \quad + \b R_3 \underbrace{ \left[\b U^m_{d^m}-\b U^m_{\b K^m,d^m} \left( \b U^m_{\b K^m,d^m} \right)^\top \b U^m_{d^m} \right]}_{=:\b R^m =:\b R \in \R^{n\times d^m}} + \b R_3 \b U^m_{\b K^m,d^m} \left( \b U^m_{\b K^m,d^m} \right)^\top \b U^m_{d^m} \label{eq:R-def}\\
        & \stackrel{(d)}{=} \left[ \b W^m - \left( \b U^m_{\b K^m,d^m} \right)^\top \b U^m_{d^m} \right] \b S^m_{d^m} + \b R_3 \b R + \b R_3 \b U^m_{\b K^m,d^m} \left( \b U^m_{\b K^m,d^m} \right)^\top \b U^m_{d^m} \\
        & \quad + \b S^m_{\b K^m,d^m} \left[ \left( \b U^m_{\b K^m,d^m} \right)^\top \b U^m_{d^m} -\b W^m \right] + \b S^m_{\b K^m,d^m} \b W^m \\
        & \stackrel{(e)}{=}  \b S^m_{\b K^m,d^m} \left[ \left( \b U^m_{\b K^m,d^m} \right)^\top \b U^m_{d^m} -\b W^m \right] - \left[\left( \b U^m_{\b K^m,d^m} \right)^\top \b U^m_{d^m} -\b W^m \right] \b S^m_{d^m}  \\
        & \quad + \b R_3 \b R + \b R_3 \b U^m_{\b K^m,d^m} \left( \b U^m_{\b K^m,d^m} \right)^\top \b U^m_{d^m}  + \b S^m_{\b K^m,d^m} \b W^m, \label{eq:W-S-decomposition}
        \end{align}
        where (a) follows from adding $\pm \left( \b U^m_{\b K^m,d^m} \right)^\top \b U^m_{d^m} \b S^m_{d^m} \in \R^{d^m\times d^m}$. (b) is obtained by (i) noticing that $\b U^m_{d^m} \b S^m_{d^m}=\b U^m_{d^m} \left( \b U^m_{d^m} \right)^\top \left | \b A^m \right | \b U^m_{d^m}$---this holds by 
        \begin{align}\label{eq:projection0}
            \b U^m_{d^m} \left( \b U^m_{d^m} \right)^\top \left | \b A^m \right | & = \b U^m_{d^m} \left( \b U^m_{d^m} \right)^\top \b U^m \b S^m \left(\b U^m\right)^\T 
            = \b U^m_{d^m} \b S^m_{d^m} \left( \b U^m_{d^m} \right)^\top, 
        \end{align}
        implied by \eqref{Am:SVD-SVDm} and multiplying both sides of \eqref{eq:projection0} with $\b U^m_{d^m}\in \R^{n\times d^m}$ from the right---, and (ii)  adding the term $\pm \left( \b U^m_{\b K^m,d^m} \right)^\top \b K^m \b U^m_{d^m} \in \R^{d^m \times d^m}$.            
        (c) follows by (i) adding the term $\pm \b R_3 \b U^m_{\b K^m,d^m} \left( \b U^m_{\b K^m,d^m} \right)^\top \b U^m_{d^m}$ and (ii) noticing that $\left( \b U^m_{\b K^m,d^m} \right)^\top \b K^m \b U_{d^m}^m= \b S^m_{\b K^m,d^m} \left( \b U^m_{\b K^m,d^m} \right)^\top \b U_{d^m}^m$ holds; the latter equality is implied by 
        \begin{align}
        \b U^m_{\b K^m,d^m} \left( \b U^m_{\b K^m,d^m} \right)^\top \b K^m & = \b U^m_{\b K^m,d^m} \left( \b U^m_{\b K^m,d^m} \right)^\top \b U^m_{\b K^m} \b S^m_{\b K^m} \left( \b U^m_{\b K^m} \right)^\T  \\
        &=\b U^m_{\b K^m,d^m} \b S^m_{\b K^m,d^m} \left( \b U^m_{\b K^m,d^m} \right)^\top \label{eq:projection}
        \end{align}
        coming from \eqref{Km:SVD-SVDm}, after multiplying \eqref{eq:projection} with $\left(\b U^m_{\b K^m,d^m}\right)^\T \in \R^{d^m \times n}$ from the left and with $\b U^m_{d^m}\in \R^{n\times d^m}$ from the right.      
        (d) comes from the equality
        \begin{align}
            \b S^m_{\b K^m,d^m} \left( \b U^m_{\b K^m,d^m} \right)^\top \b U^m_{d^m} = \b S^m_{\b K^m,d^m} \left[\left( \b U^m_{\b K^m,d^m} \right)^\top \b U^m_{d^m} - \b W^m + \b W^m  \right].
        \end{align}
        (e) is obtained by extracting the negative sign in $\b W^m  - \left( \b U^m_{\b K^m,d^m} \right)^\top \b U^m_{d^m} $. 
        
        Then by using the decomposition of $\b W^m \b S^m_{d^m}$ obtained in \eqref{eq:W-S-decomposition} with a rearrangement to $\b B_1=\b W^m \b S^m_{d^m} - \b S^m_{\b K^m,d^m}\b W^m$, the relation $\left\|\b{AB}\right\|\le \left\|\b A \right\|  \left\|\b B \right\|$ and the triangular inequality, one gets
        \begin{align}
            \left \| \b B_1 \right \| & \le \left \| \left( \b U^m_{\b K^m,d^m} \right)^\top \b U^m_{d^m} -\b W^m \right \|  \left( \left \| \b S^m_{\b K^m,d^m} \right\| + \left\| \b S^m_{d^m} \right \| \right) + \left \| \b R_3 \b R \right \| \\
            & \quad + \left \| \b R_3 \b U^m_{\b K^m,d^m} \left( \b U^m_{\b K^m,d^m} \right)^\top \b U^m_{d^m} \right \|. \label{eq:bound_B1}
        \end{align}
        Using this result for $\left \| \b B_1 \right \|$, we have for the bound on $\left\|\tilde{\b R}_2\right\|_F$ that
        \begin{align}
            & \left \| \tilde{\b R}_2 \right \|_F
            \stackrel{\eqref{eq:decompose_tilde_R2}}{\le} \left\|  \left( \b S_{\b K^m,d^m}^m \right)^{-\frac{1}{2}}  \right\|_F \left\| \b B_1  \right\| \\
            & \stackrel{(a)}{\le} \underbrace{\left(d^m \right)^{\frac{1}{2}} \left \| \left( \b U^m_{\b K^m,d^m} \right)^\top \b U^m_{d^m} -\b W^m \right \| \overbrace{ \left( \left \| \b S^m_{\b K^m,d^m} \right\| + \left\| \b S^m_{d^m} \right \| \right) \left[\lambda^m_{d^m}(\b K^m) \right]^{-\frac{1}{2}} }^{=:D_0^m=:D_0} }_{=:D_1^m=:D_1} \label{eq:D0-D1-def} \\
            &\quad + \underbrace{\left(d^m \right)^{\frac{1}{2}} \left[\lambda^m_{d^m}(\b K^m) \right]^{-\frac{1}{2}} \left \| \b R_3 \b R \right \| }_{=:D_2^m=:D_2} \label{eq:D2-def}\\
            &\quad + \underbrace{\left(d^m \right)^{\frac{1}{2}} \left[\lambda^m_{d^m}(\b K^m) \right]^{-\frac{1}{2}} \left \| \b R_3 \b U^m_{\b K^m,d^m} \left( \b U^m_{\b K^m,d^m} \right)^\top \b U^m_{d^m} \right \| }_{=:D_3^m=:D_3},
            \label{eq:D3-def}
        \end{align}
        where (a) comes from $\left \| (\b S^m_{\b K^m,d^m})^{-\frac{1}{2} } \right \|_F \stackrel{(b)}{\le} \sqrt{d^m}\left \| (\b S^m_{\b K^m,d^m})^{-\frac{1}{2} } \right \| \stackrel{(c)}{=}\left(d^m \right)^{\frac{1}{2}} \left[\lambda^m_{d^m}(\b K^m) \right]^{-\frac{1}{2}}$ and \eqref{eq:bound_B1}; (b) is implied by 
        \begin{align}
            \left\| (\b S^m_{\b K^m,d^m})^{-\frac{1}{2} } \right\|_F \le \sqrt{\rank \left( (\b S^m_{\b K^m,d^m} )^{-\frac{1}{2} } \right)} \left\|(\b S^m_{\b K^m,d^m})^{-\frac{1}{2} }\right\| = \sqrt{d^m} \left\| (\b S^m_{\b K^m,d^m})^{-\frac{1}{2} } \right\| 
        \end{align}
        since $\rank \left( (\b S^m_{\b K^m,d^m} )^{-\frac{1}{2} } \right) = d^m$, (c) follows from the identity $\left\| \diag(\b v) \right\| = \max_{i \in [d^m]} \left| v_i \right|$ with $\b v:=\left(\left[\lambda^m_i\left(\b K^m\right)\right]^{-\frac{1}{2}}\right)_{i=1}^{d^m}$, and using that the eigenvalues of $\b K^m$ are non-negative and they are ordered in a non-increasing fashion. We proceed by bounding the term $D_1$ (via controlling $D_0$), $D_2$ and $D_3$.
        \begin{itemize}
            \item \textbf{Term} $D_0$: Below we bound $D_0$.
            \begin{itemize}
                \item \textbf{Fixed $(X_i^m)_{i=1}^n=(x_i^m)_{i=1}^n$, randomness in $\b A^m$:} We have
                \begin{align}
                    D_0 & = \left[\lambda^m_{d^m}(\b K^m) \right]^{-\frac{1}{2}} \left( \left \| \b S^m_{\b K^m,d^m} \right\| + \left\| \b S^m_{d^m} \right \| \right) \\
                    & \stackrel{(a)}{\le} \left[\lambda^m_{d^m}(\b K^m) \right]^{-\frac{1}{2}} \left( 2 \Delta^m + 4 \tilde{C}^m \sqrt{\Delta^m \log \left( \frac{n}{\eta /144} \right)} \right) 
                \end{align}
                holds with $\Q^m$-probability at least $1-\frac{\eta}{144}$ for any $\eta \in \left( 0, 72 \right]$ and $n \ge \colLfive{\left( \frac{\eta}{144} \right)^{-\frac{1}{c^m}}}$, where (a) is established as follows. 
                \begin{itemize}
                    \item Term $\left\| \b S^m_{\b K^m,d^m} \right\|$: We have
                    \begin{align}\label{eig_delta_bound}
                        \left\| \b S^m_{\b K^m,d^m} \right\| \stackrel{(b)}{=} \lambda_1^m (\b K^m) \stackrel{(c)}{\le} \max_{\i \in [n]} \sum_{j=1}^n K^m_{i,j} =\Delta^m,
                \end{align}
                where (b) holds by $\left\| \diag(\b v) \right\| = \max_{i \in [d^m]} \left| v_i \right|$ with $\b v = \left( \lambda_i^m(\b K^m) \right)_{i=1}^{d^m}$ and
                (c)
                follows from \citep[Theorem 5.6.9, Corollary 6.1.5]{horn2012matrix}. 

                \item Term $\left\| \b S_{d^m}^m \right\|$: We have for any $\eta \in \left(0,72 \right] $ and $n\ge \colLfive{\left( \frac{\eta}{144} \right)^{-\frac{1}{c^m}}} $ that
                \begin{align}
                    & \quad \max_{i \in [n]} \left|  \lambda_i^m \left( \left| \b A^m \right| \right) - \lambda_i^m (\b K^m) \right| \stackrel{(d)}{\le} \left\| \left| \b A^m \right| - \b K^m \right\| \stackrel{(e)}{\le} 4 \tilde{C}^m \sqrt{\Delta^m \log \left( \frac{n}{\eta /144} \right)} \\
                    & \Rightarrow \left\|  \b S_{d^m}^m \right\| = \lambda_1^m \left( \left| \b A^m \right| \right) \le \lambda_1^m \left( \b K^m \right) + 4 \tilde{C}^m \sqrt{\Delta^m \log \left( \frac{n}{\eta /144} \right)} \\
                    & \quad \stackrel{ \eqref{eig_delta_bound}:(d)}{\le} \Delta^m + 4 \tilde{C}^m \sqrt{\Delta^m \log \left( \frac{n}{\eta /144} \right)}, \label{eig_adj_bound0}
                \end{align}
                with $\Q^m$-probability at least $1-\frac{\eta}{144}$. (d) is implied by the Weyl's Perturbation Theorem \citep[Corollary~III.2.6 on page~63]{bhatia1997matrix}, (e) follows with $\Q^m$-probability at least $1-\frac{\eta}{144}$ for any $\eta \in \left(0, 72 \right]$ and $ n\ge \colLfive{ \left( \frac{\eta}{144} \right)^{-\frac{1}{c^m}}} $  from \eqref{event_C} by a substitution $\frac{\eta}{12} \leftarrow \frac{\eta}{144}$. 
                
                \end{itemize}
                                   
                \item \textbf{Randomness in $(X_i^m)_{i=1}^n$}: For the random variables $(X_i^m)_{i=1}^n$, to guarantee that (f) in \eqref{eig_adj_bound0} holds, we invoke Assumption~\ref{assum_Delta}(i). By the substitution $\eta \leftarrow \frac{\eta}{144}$, we obtain that $\Delta^m > C_0^m \log(n)$ holds with $\left(\P^m\right)^{\otimes n}$-probability at least $1-\frac{\eta}{144}$ for any $\eta \in (0,144)$ and $n \ge \colAonei{n^m_{0,(i)}\left( \frac{\eta}{144}\right)}$. Define the following events:
                \begin{align}\label{eq:combine_prob_events}
                    A & = \left \{ \left( (x^m_i)_{i=1}^n, \b A^m \right): \left\| \left| \b A^m \right| - \b K^m \right\| \le 4 \tilde{C}^m \sqrt{\Delta^m \log \left( \frac{144n}{\eta} \right)}  \right \} \\
                    & = \left \{ \left( (x^m_i)_{i=1}^n, \b A^m \right): \underbrace{\left\| \left| \b A^m \right| - \b K^m \right\|}_{=:f(x,y)} \le \underbrace{4 \tilde{C}^m \sqrt{ \log \left( \frac{144n}{\eta} \right)}}_{=:g(x)} \underbrace{\sqrt{\Delta^m}}_{=:S_0(x)} \right \},\\
                    B & = \left \{ (x^m_i)_{i=1}^n : C_0^m \log(n) < \Delta^m  \right \}= \left \{ (x^m_i)_{i=1}^n, : \underbrace{\sqrt{C_0^m \log(n)}}_{=:a} < \underbrace{\sqrt{\Delta^m}}_{=S_0(x)}  \right \} \subseteq \left( \X^m \right)^n, \\
                    C & = \left \{ \left( (x^m_i)_{i=1}^n, \b A^m \right): \left\| \left| \b A^m \right| - \b K^m \right\| \le 4 \tilde{C}^m \sqrt{\Delta^m \log \left( \frac{144n}{\eta} \right)}, (x^m_i)_{i=1}^n \in B \right\} \\
                    & = \left \{ \left( (x^m_i)_{i=1}^n, \b A^m \right): \underbrace{\left\| \left| \b A^m \right| - \b K^m \right\|}_{=f(x,y)} \le \underbrace{4 \tilde{C}^m \sqrt{\log \left( \frac{144n}{\eta} \right)}}_{=g(x)} \underbrace{\sqrt{\Delta^m}}_{=S_0(x)}, (x^m_i)_{i=1}^n \in B \right\}
                \end{align}
                and $C \subseteq A$. Then we have (i) for any $\eta \in (0,72]$ and $n\ge \colLfive{ \left( \frac{\eta}{144} \right)^{-\frac{1}{c^m}}}$ that $\Q^m\left( C \mid (x^m_i)_{i=1}^n \right) \ge 1-\frac{\eta}{144}$, and (ii) for any $\eta \in (0,144)$ and $n \ge \colAonei{n^m_{0,(i)}\left( \frac{\eta}{144}\right)}$ that $\left(\P^m\right)^{\otimes n}(B) \ge 1-\frac{\eta}{144}$. Applying  Lemma~\ref{lemm:combine_probs}\ref{group:combine_probs2} with the notations $f$, $g$, $S_0$ and $a$ introduced above, and with $\delta_1 = \delta_2 = \frac{\eta}{144}$, we have 
                for all $\eta \in (0,144) \cap \left(0, 72 \right] = \left(0, 72 \right]$ and $n \ge \max \left\{ \colAonei{n^m_{0,(i)}\left( \frac{\eta}{144}\right)}, \colLfive{ \left( \frac{\eta}{144} \right)^{-\frac{1}{c^m}}} \right\}$ that
                \begin{align}
                    (\J^m)(A) \ge  1-\left(\frac{\eta}{144} + \frac{\eta}{144} \right) = 1-\frac{\eta}{72}. \label{eq:event_combined1}
                \end{align} 
                Thus, for any $\eta \in \left(0, 72 \right]$ and $n \ge \max \left\{ \colAonei{n^m_{0,(i)}\left( \frac{\eta}{144}\right)}, \colLfive{ \left( \frac{\eta}{144} \right)^{-\frac{1}{c^m}}} \right\}$ we have
                \begin{align}
                    \J^m \left( \left\|  \b S_{d^m}^m \right\| \le \Delta^m + 4 C^m \sqrt{\Delta^m \log \left( \frac{n}{\eta /144} \right)} \right) \ge 1-\frac{\eta}{72}; \label{eq:S_d^m_J^m_prob}
                \end{align}
                where due to the randomness of $(X_i^m)_{i=1}^n$, $\tilde{C}^m$ was replaced by $C^m$.
                Therefore we have for any $\eta \in \left(0, 72 \right]$ and $n \ge \max \left\{ \colAonei{n^m_{0,(i)}\left( \frac{\eta}{144}\right)}, \colLfive{ \left( \frac{\eta}{144} \right)^{-\frac{1}{c^m}}} \right\}$ that
                \begin{align}
                    \J^m \left( D_0 \stackrel{(a)}{\le} 
                    \left[\lambda_{d^m}(\b K^m) \right]^{-\frac{1}{2}} \left( 2 \Delta^m + 4 C^m \sqrt{\Delta^m \log \left( \frac{n}{\eta /144} \right)} \right) \right) \ge  1-\frac{\eta}{72}. \label{eq:prob_event2}
                \end{align}
                Next, by Assumption~\ref{assum_lambda_delta}(i) with the substitution $\eta \leftarrow \frac{\eta}{72}$, for any $\eta \in (0,72)$ and $n \ge \colAi{n^m_{1,(i)}\left( \frac{\eta}{72}\right)}$ we have that $\frac{\lambda^m_{d^m}(\b K^m)}{\Delta^m} \ge c^m_0$ holds with $\left(\P^m\right)^{\otimes n}$-probability at least $1-\frac{\eta}{72}$. Substituting this lower bound on $\lambda^m_{d^m}(\b K^m)$ into the previous inequality yields for any $\eta \in (0,72] \cap(0,72) \cap(0,36) =(0,36)$ and $n \ge \max \left\{ \colAonei{n^m_{0,(i)}\left( \frac{\eta}{144}\right)}, \colAi{n^m_{1,(i)}\left( \frac{\eta}{72}\right)}, \colLfive{ \left( \frac{\eta}{144} \right)^{-\frac{1}{c^m}}} \right\} $
                \begin{align}
                    \MoveEqLeft D_0 \le 
                    \left(c_0^m \right)^{-\frac{1}{2}} (\Delta^m)^{-\frac{1}{2}} \left( 2 \Delta^m + 4 C^m \sqrt{\Delta^m \log \left( \frac{n}{\eta /144} \right)} \right) \\
                    & = 2 \left(c_0^m \right)^{-\frac{1}{2}} (\Delta^m)^{\frac{1}{2}} + 4 \left(c_0^m \right)^{-\frac{1}{2}} C^m \sqrt{\log \left( \frac{n}{\eta /144} \right)}
                \end{align}
                holds with $\J^m$-probability at least $1-\frac{\eta}{36}$, where restricting $\eta$ to $(0,36)$ guarantees that the probability is positive.
                By the definition of $\Delta^m$ we have $\Delta^m \le n$ for any realization $(X_i^m)_{i=1}^n=(x_i^m)_{i=1}^n$. Therefore, for any $\eta \in \left( 0, 36 \right)$ and $n \ge \max \left\{ \colAonei{n^m_{0,(i)}\left( \frac{\eta}{144}\right)}, \colAi{n^m_{1,(i)}\left( \frac{\eta}{72}\right)}, \colLfive{ \left( \frac{\eta}{144} \right)^{-\frac{1}{c^m}}} \right\}$ one has 
                \begin{align}%\label{eq:bound_D0}
                    & \quad \J^m \left( D_0 \le 2 \left(c_0^m \right)^{-\frac{1}{2}} n^{\frac{1}{2}} + 4C^m \left(c_0^m \right)^{-\frac{1}{2}} \sqrt{\log \left( \frac{n}{\eta /144} \right)} \right) \ge 1-\frac{\eta}{36},
                \end{align}
                According to Lemma~\ref{lemma:bound_Cm}, by a substitution of $\eta \leftarrow \frac{\eta}{36}$, we have $C^m \le c_1^m(n)$ holds with $\J^m$-probability at least $1-\frac{\eta}{36}$ for any $\eta \in (0,36)$ and for any $n \ge \max \left\{ \colAoneii{n^m_{0,(ii)}\left( \frac{\eta}{72}\right)}, \colLsix{\frac{288}{\eta}}, \coletwo{e^{\frac{1}{\left( 2C_1^m - 4\sqrt{2} \right)^2}} } \right\}$. Therefore, for any $\eta \in (0,36) \cap (0,18) = (0,18)$ and 
                \begin{align}
                    n \ge \max\left\{ \colAonei{n^m_{0,(i)}\left( \frac{\eta}{144}\right)}, \colAoneii{n^m_{0,(ii)}\left( \frac{\eta}{72}\right)}, \colAi{n_{1,(i)}^m\left(\frac{\eta}{72}\right)}, \colLfive{\left( \frac{\eta}{144} \right)^{-\frac{1}{c^m}}}, \colLsix{\frac{288}{\eta}}, \coletwo{e^{\frac{1}{\left( 2C_1^m - 4\sqrt{2} \right)^2}} } \right\},
                \end{align}
                one has 
                \begin{align}\label{eq:bound_D0}
                    & \quad \J^m \left( D_0 \le 2 \left(c_0^m \right)^{-\frac{1}{2}} n^{\frac{1}{2}} + 4c_1^m(n) \left(c_0^m \right)^{-\frac{1}{2}} \sqrt{\log \left( \frac{144 n}{\eta} \right)} \right) \ge 1-\frac{\eta}{18},
                \end{align}
                where restricting $\eta$ to $(0,18)$ guarantees that the r.h.s.\ of \eqref{eq:bound_D0} is positive. Having established a bound on $D_0$, we now move on to controlling $D_1$.
            \end{itemize}
            \item \textbf{Term} $D_1$: We have that 
            \begin{align}
                D_1 &\stackrel{(a)}{=} \left(d^m \right)^{\frac{1}{2}} \left \| \left( \b U^m_{\b K^m,d^m} \right)^\top \b U^m_{d^m} - \b W_1 \b W_2^\top \right \| D_0 \\
                & \stackrel{(b)}{\le} \left(d^m \right)^{\frac{1}{2}} \left \| \mathcal{P}^m_{\left| \b A^m \right|} - \mathcal{P}^m_{\b K^m} \right \|^2 \left[ 2 \left(c_0^m \right)^{-\frac{1}{2}} n^{\frac{1}{2}} + 4 c_1^m(n) \left(c_0^m \right)^{-\frac{1}{2}} \sqrt{\log \left( \frac{144n}{\eta } \right)} \right] \\
                & \stackrel{(c)}{\le} \left(d^m \right)^{\frac{1}{2}} 256 \left( c_1^m(n) \right)^2 \left( \delta^m_{d^m} \right)^{-2} \frac{\log \left(  \frac{12n}{\eta / 18}\right)}{n}\\
                & \quad \hspace{0.1cm} \times \left[ 2 \left(c_0^m \right)^{-\frac{1}{2}} n^{\frac{1}{2}} + 4 c_1^m(n) \left(c_0^m \right)^{-\frac{1}{2}} \sqrt{\log \left( \frac{144 n}{\eta} \right)} \right] \\
                & \stackrel{(d)}{\le} 512 \frac{\left(d^m \right)^{\frac{1}{2}} \left( c_1^m(n) \right)^2}{\sqrt{c_0^m}} \left( \delta^m_{d^m} \right)^{-2} \frac{\log \left( \frac{12n}{\eta/18} \right)}{\sqrt{n}}\\& \quad \hspace{0.1cm} + 1024 \frac{\left(d^m \right)^{\frac{1}{2}} \left( c_1^m(n) \right)^3}{\sqrt{c_0^m}} \left( \delta^m_{d^m} \right)^{-2} \frac{\log^{\frac{3}{2}} \left( \frac{12n}{\eta/18} \right)}{n} \label{decompose_D1}
            \end{align}
            holds with $\J^m$-probability at least $1-\frac{\eta}{9}$ for any $\eta \in \left( 0,18 \right) \cap (0,9) = (0,9)$ and
            \begin{align}
                n & \ge \max \Bigg\{ \colAonei{ n^m_{0,(i)}\left( \frac{\eta}{144}\right)}, \colAonei{ n^m_{0,(i)}\left( \frac{\eta}{216}\right)}, \colAoneii{n^m_{0,(ii)}\left( \frac{\eta}{72}\right)}, \colAoneii{n^m_{0,(ii)}\left( \frac{\eta}{216}\right)}, \colAi{n_{1,(i)}^m\left(\frac{\eta}{72}\right)},  \\
                & \hspace{1.7 cm} \colAii{n_{1,(ii)}^m \left(\frac{\eta}{18}\right)}, \colAii{n^m_{1,(ii)}\left(\frac{\eta}{36}\right)}, \colLfive{\left( \frac{\eta}{216} \right)^{-\frac{1}{c^m}}}, \colLfive{\left( \frac{\eta}{144} \right)^{-\frac{1}{c^m}}}, \colLsix{\frac{288}{\eta}}, \colLsix{\frac{864}{\eta}}, \\
                & \hspace{1.7 cm} \coletwo{e^{\frac{1}{\left( 2C_1^m - 4\sqrt{2} \right)^2}} } \Bigg\} \\
                & \stackrel{(e)}{=} \max \Bigg\{ \colAonei{ n^m_{0,(i)}\left( \frac{\eta}{144}\right)}, \colAonei{ n^m_{0,(i)}\left( \frac{\eta}{216}\right)}, \colAoneii{n^m_{0,(ii)}\left( \frac{\eta}{72}\right)}, \colAoneii{n^m_{0,(ii)}\left( \frac{\eta}{216}\right)}, \colAi{n_{1,(i)}^m\left(\frac{\eta}{72}\right)}, \\
                & \hspace{1.7 cm} \colAii{n_{1,(ii)}^m \left(\frac{\eta}{18}\right)},\colAii{n^m_{1,(ii)}\left(\frac{\eta}{36}\right)}, \colLfive{\left( \frac{\eta}{216} \right)^{-\frac{1}{c^m}}}, \colLsix{\frac{864}{\eta}}, \coletwo{e^{\frac{1}{\left( 2C_1^m - 4\sqrt{2} \right)^2}} } \Bigg\},
            \end{align}
            where (e) follows from the decreasing property of the function 
            \begin{align}
            x\mapsto x^a,\quad a=-\frac{1}{c^m}<0\quad (\t{as } c^m>0).\label{eq:x^a-monotonicity}
            \end{align}
            Indeed, in (a) in \eqref{decompose_D1}, the definition of $D_1$ and $\b W^m$ [see \eqref{eq:D0-D1-def} and \eqref{eq:Wm-def}] were used. (b) is implied by (d)-(e) in \eqref{bound_R1} combined with (f)-(i) in \eqref{eq:bound_signma_diff_I}, and by \eqref{eq:bound_D0} with $\J^m$-probability at least $1-\frac{\eta}{18}$ for any $\eta \in  \left( 0,18\right)$ and 
            \begin{align}
                n \ge \max\left\{ \colAonei{ n^m_{0,(i)}\left( \frac{\eta}{144}\right)}, \colAoneii{n^m_{0,(ii)}\left( \frac{\eta}{72}\right)}, \colAi{n_{1,(i)}^m\left(\frac{\eta}{72}\right)}, \colLfive{\left( \frac{\eta}{144} \right)^{-\frac{1}{c^m}}}, \colLsix{\frac{288}{\eta}}, \coletwo{e^{\frac{1}{\left( 2C_1^m - 4\sqrt{2} \right)^2}} }  \right\}.
            \end{align}
            (c) follows with $\J^m$-probability at least $1-\frac{\eta}{18}$ for any $\eta \in \left( 0,18 \right)$ and 
            \begin{align}
                n & \ge \max \Big\{ \colAonei{ n^m_{0,(i)}\left( \frac{\eta}{216}\right)}, \colAoneii{n^m_{0,(ii)}\left( \frac{\eta}{216}\right)}, \colAii{ n_{1,(ii)}^m\left(\frac{\eta}{18}\right)}, \colAii{n^m_{1,(ii)}\left(\frac{\eta}{36}\right)}, \colLfive{\left( \frac{\eta}{216} \right)^{-\frac{1}{c^m}}} , \colLsix{\frac{864}{\eta}}, \\
                & \hspace{1.7cm} \coletwo{e^{\frac{1}{\left( 2C_1^m - 4\sqrt{2} \right)^2}} } \Big\}
            \end{align}
            from \eqref{bound_proj}  by a substitution $\eta \leftarrow \frac{\eta}{18}$. (d) holds by $\log \left( \frac{n}{\eta/144} \right) \le \log \left( \frac{12n}{\eta/18} \right)$ and combining terms. A union bound gives the claimed result in \eqref{decompose_D1}, with restricting $\eta$ to $(0,9)$ to guaranteeing that the probability is positive.
            
            \item \textbf{Term} $D_2$: We have  
            \begin{align}
                & D_2  \stackrel{(a)}{\le} \left(d^m \right)^{\frac{1}{2}}\left[\lambda^m_{d^m}(\b K^m) \right]^{-\frac{1}{2}} \left \| \b R_3 \right \| \left \| \b R \right \| 
                \\
                & \stackrel{(b)}{\le} \left(d^m \right)^{\frac{1}{2}} \left[\lambda^m_{d^m}(\b K^m) \right]^{-\frac{1}{2}} \left \| \b U^m_{d^m} \left( \b U^m_{d^m} \right)^\top \left | \b A^m \right |  - \b K^m  \right \|  \left \| \b R \right \|  \\
                & \stackrel{(c)}{\le} \left(d^m \right)^{\frac{1}{2}} \left[\lambda^m_{d^m}(\b K^m) \right]^{-\frac{1}{2}} \left \| \left | \b A^m \right |  - \b K^m  \right \|  \left \| \b R \right \| \stackrel{(d)}{\le} \left(d^m \right)^{\frac{1}{2}} 4 \frac{ c_1^m(n)}{\sqrt{c_0^m}} \sqrt{\log \left( \frac{144n}{\eta}  \right)} \left\| \b R \right\| \\
                & \stackrel{(e)}{\le} \left(d^m \right)^{\frac{1}{2}} 4 \frac{c_1^m(n)}{\sqrt{c_0^m}} \sqrt{\log \left( \frac{144 n}{\eta}  \right)} 16 c_1^m(n) \left( \delta^m_{d^m} \right)^{-1}\sqrt{\frac{ \log \left( \frac{216n}{\eta} \right)}{n}} \\
                & \stackrel{(f)}{\le} 64 \frac{\left( c_1^m(n) \right)^2}{\sqrt{c_0^m}} \left( \delta^m_{d^m} \right)^{-1} \frac{\left(d^m \right)^{\frac{1}{2}} \log \left( \frac{216n}{\eta} \right)}{\sqrt{n}} \label{bound_D2}
            \end{align}
            with $\J^m$-probability at least $1-\frac{\eta}{9}$ for any $\eta \in \left( 0,9 \right) \cap \left(0,18 \right) = \left( 0,9 \right) $ and 
            \begin{align}
                n & \ge \max \Bigg\{ \colAonei{ n^m_{0,(i)}\left( \frac{\eta}{144}\right)}, \colAonei{ n^m_{0,(i)}\left( \frac{\eta}{216}\right)}, \colAoneii{n^m_{0,(ii)}\left( \frac{\eta}{216}\right)}, \colAoneii{n^m_{0,(ii)}\left( \frac{\eta}{72}\right)}, \colAi{n^m_{1,(i)}\left( \frac{\eta}{72} \right)}, \\
                & \hspace{1.7cm} \colAii{ n_{1,(ii)}^m\left(\frac{\eta}{18}\right)}, \colAii{n^m_{1,(ii)}\left(\frac{\eta}{36}\right)}, \colLfive{\left( \frac{\eta}{144} \right)^{-\frac{1}{c^m}}}, \colLfive{\left( \frac{\eta}{216} \right)^{-\frac{1}{c^m}}}, \colLsix{\frac{288}{\eta}}, \colLsix{\frac{864}{\eta}}, \coletwo{e^{\frac{1}{\left( 2C_1^m - 4\sqrt{2} \right)^2}} } \Bigg\} \\
                & \stackrel{(g)}{=} \max \Bigg\{ \colAonei{ n^m_{0,(i)}\left( \frac{\eta}{144}\right)}, \colAonei{ n^m_{0,(i)}\left( \frac{\eta}{216}\right)}, \colAoneii{n^m_{0,(ii)}\left( \frac{\eta}{216}\right)}, \colAoneii{n^m_{0,(ii)}\left( \frac{\eta}{72}\right)}, \colAi{n^m_{1,(i)}\left( \frac{\eta}{72} \right)},  \\
                & \hspace{1.7cm} \colAii{ n_{1,(ii)}^m\left(\frac{\eta}{18}\right)}, \colAii{n^m_{1,(ii)}\left(\frac{\eta}{36}\right)}, \colLfive{\left( \frac{\eta}{216} \right)^{-\frac{1}{c^m}}}, \colLsix{\frac{864}{\eta}}, \coletwo{e^{\frac{1}{\left( 2C_1^m - 4\sqrt{2} \right)^2}} } \Bigg\} ,
            \end{align} 
            where (g) follows from \eqref{eq:x^a-monotonicity}. (a) in \eqref{bound_D2} is implied by the definition of $D_2$ [see \eqref{eq:D2-def}] and the inequality $\left\| \b A \b B\right\| \le \left\| \b A \right\| \left\| \b B \right\|$. (b) in \eqref{bound_D2} follows from 
            \begin{align}
                \left\| \b R_3 \right\| \stackrel{(h)}{\le} \left\|  \b U_{\b K^m,d^m}^m \right\| \left\|  \b U^m_{d^m} \left( \b U^m_{d^m} \right)^\top \left | \b A^m \right |  - \b K^m \right\| \stackrel{(i)}{\le} \left\|  \b U^m_{d^m} \left( \b U^m_{d^m} \right)^\top \left | \b A^m \right |  - \b K^m \right\|,
            \end{align}
            where (h) is obtained by the definition of $\b R_3$ [see \eqref{eq:R3-def}] and the relations $\left\| \b A \b B\right\| \le \left\| \b A \right\| \left\| \b B \right\|$ and $\left\|\b A^\T \right\| = \left\|\b A\right\|$, (i) is implied by $\left\| \b U_{\b K^m,d^m}^m \right\| \le 1$. 
            
            For (c) in \eqref{bound_D2}, we have
            \begin{align}\label{eq:matrix_ordering}
                    \b U^m_{d^m} \left( \b U^m_{d^m} \right)^\top \left | \b A^m \right | \stackrel{(i)}{\preceq} \left | \b A^m \right | \stackrel{(j)}{\Rightarrow } \b U^m_{d^m} \left( \b U^m_{d^m} \right)^\top \left | \b A^m \right | - \b K^m \preceq \left | \b A^m \right | - \b K^m,
                \end{align}
                where $\preceq$ refers to the positive semi-definite ordering of matrices. (i) holds since for the positive semi-definite $\left | \b A^m \right |$, 
                \begin{align}
                \b M:=\left[ \b I - \b U^m_{d^m} \left( \b U^m_{d^m} \right)^\top \right] \left | \b A^m \right | \label{eq:M-mtx}
                \end{align}
               is still positive semi-definite. Indeed, $\b M$ has $n$ eigenvalues, 
               \begin{align}
               \lambda^m_1(\b M)&=\dots=\lambda^m_{d^m}(\b M)=0, \label{eq:M:eigs-1}\\
               \lambda^m_{d^m+1}( \b M)&= \lambda^m_{d^m+1}(\left | \b A^m \right |) \ge \ldots \ge \lambda^m_{n}(\b M )= \lambda^m_{n}(\left | \b A^m \right |) \ge 0 \label{eq:M:eigs-2}
               \end{align}
               and nonnegative eigenvalues imply positive semi-definiteness for a real symmetric matrix \citep[page~1]{bhatia2007positive}; $\b M$ is symmetric since
                \begin{align}
                    \b M & =\left[ \b I - \b U^m_{d^m} \left( \b U^m_{d^m} \right)^\top \right] \left | \b A^m \right | = \left | \b A^m \right | -  \b U^m_{d^m} \left( \b U^m_{d^m} \right)^\top \left | \b A^m \right | \\&\stackrel{\eqref{eq:projection0}}{=} \left | \b A^m \right | - \b U^m_{d^m} \b S^m_{d^m} \left( \b U^m_{d^m} \right)^\top,\label{eq:M-mtx:symmetric}
                \end{align} 
                and both $\left | \b A^m \right |$ and $\b U^m_{d^m} \b S^m_{d^m} \left( \b U^m_{d^m} \right)^\top$ are symmetric; hence there difference is also so.
                (j) is implied by
                \begin{align}
                    \left( \left | \b A^m \right | - \b K^m \right) - \left[ \b U^m_{d^m} \left( \b U^m_{d^m} \right)^\top \left | \b A^m \right | - \b K^m \right] = \left | \b A^m \right | - \b U^m_{d^m} \left( \b U^m_{d^m} \right)^\top \left | \b A^m \right | \stackrel{(k)}{\succeq} 0,
                \end{align}
                where (k) follows from (i) in \eqref{eq:matrix_ordering}. Therefore, we have that
                \begin{align}\label{bound_diff_psd1}
                    \left \| \b U^m_{d^m} \left( \b U^m_{d^m} \right)^\top \left | \b A^m \right |  - \b K^m \right \|  \le \left \| \left | \b A^m \right |  - \b K^m  \right \|
                \end{align}
                Indeed, \eqref{bound_diff_psd1} follows from \eqref{eq:matrix_ordering} as $\b A \preceq \b B$ implies $\left\| \b A \right\| \le \left\| \b B \right\|$ \citep[2.2-10 on page~57]{ciarlet1989introduction}.
            For (d) in \eqref{bound_D2}, according to \eqref{eq:event_combined1}, we have for any $\eta \in (0,72]$ and $n \ge \max \left\{ \colAonei{n^m_{0,(i)}\left( \frac{\eta}{144}\right)}, \colLfive{ \left( \frac{\eta}{144} \right)^{-\frac{1}{c^m}}} \right\}$ that
            \begin{align}
                \J^m\left( \left \| \left | \b A^m \right | - \b K^m \right \| \le 4 C^m \sqrt{\Delta^m \log \left( \frac{144n}{\eta} \right)}\right) \ge 1-\frac{\eta}{72}.
            \end{align}
            Therefore, we have for any $\eta \in (0,72]$ and $n \ge \max \left\{ \colAonei{n^m_{0,(i)}\left( \frac{\eta}{144}\right)}, \colLfive{ \left( \frac{\eta}{144} \right)^{-\frac{1}{c^m}}} \right\}$ that
            \begin{align}
                \J^m\left( \left[\lambda^m_{d^m}(\b K^m) \right]^{-\frac{1}{2}} \left \| \left | \b A^m \right | - \b K^m \right \| \le \left[\lambda^m_{d^m}(\b K^m) \right]^{-\frac{1}{2}} 4 C^m \sqrt{\Delta^m \log \left( \frac{144n}{\eta} \right)}\right) \ge 1-\frac{\eta}{72}. \label{eq:prob_event3}
            \end{align}
            Next, by Assumption~\ref{assum_lambda_delta}(i) with the substitution $\eta \leftarrow \frac{\eta}{72}$, we have $\frac{\lambda^m_{d^m}(\b K^m)}{\Delta^m} \ge c^m_0$ holds with $\left(\P^m\right)^{\otimes n}$-probability at least $1-\frac{\eta}{72}$ for any $\eta \in (0,72)$ and $n \ge \colAi{n^m_{1,(i)}\left( \frac{\eta}{72}\right)}$. Substituting this lower bound on $\lambda^m_{d^m}(\b K^m)$ into the previous inequality yields for any $\eta \in (0,72) \cap(0,72] \cap (0,36) =(0,36)$ and $n \ge \max \left\{ \colAonei{n^m_{0,(i)}\left( \frac{\eta}{144}\right)}, \colAi{n^m_{1,(i)}\left( \frac{\eta}{72}\right)}, \colLfive{ \left( \frac{\eta}{144} \right)^{-\frac{1}{c^m}}} \right\} $
            \begin{align}
              \left[\lambda^m_{d^m}(\b K^m) \right]^{-\frac{1}{2}} \left \| \left | \b A^m \right | - \b K^m \right \| & \le \left( c_0^m\right)^{-\frac{1}{2}} \left(\Delta^m \right)^{-\frac{1}{2}} 4 C^m \sqrt{\Delta^m \log \left( \frac{144n}{\eta} \right)} \\
              & = 4 \frac{C^m}{\sqrt{c_0^m}} \sqrt{\log \left( \frac{144n}{\eta} \right)}
            \end{align}
            holds with $\J^m$-probability at least $1-\frac{\eta}{36}$, with restricting $\eta$ to $(0,36)$ guaranteeing the positivity of the probability. According to Lemma~\ref{lemma:bound_Cm}, by a substitution of $\eta \leftarrow \frac{\eta}{36}$, for any $\eta \in (0,36)$ and $n \ge \max \left\{ \colAoneii{n^m_{0,(ii)}\left( \frac{\eta}{72}\right)}, \colLsix{\frac{288}{\eta}}, \coletwo{e^{\frac{1}{\left( 2C_1^m - 4\sqrt{2} \right)^2}} } \right\}$, we have $C^m \le c_1^m(n)$ holds with $\J^m$-probability at least $1-\frac{\eta}{36}$. Therefore, for any $\eta \in \left( 0,18 \right) \cap (0,36) = (0,18)$ and $n \ge \max\left\{ \colAonei{n^m_{0,(i)}\left( \frac{\eta}{144}\right)}, \colAoneii{n^m_{0,(ii)}\left( \frac{\eta}{72}\right)}, \colAi{n^m_{1,(i)}\left( \frac{\eta}{72} \right)}, \colLfive{\left( \frac{\eta}{144} \right)^{-\frac{1}{c^m}}}, \colLsix{\frac{288}{\eta}}, \coletwo{e^{\frac{1}{\left( 2C_1^m - 4\sqrt{2} \right)^2}} } \right\}$ one has 
            \begin{align}\label{bound_D2_sub}    \J^m\left(\left[\lambda^m_{d^m}(\b K^m) \right]^{-\frac{1}{2}} \left \| \left | \b A^m \right | - \b K^m \right \| \le 4 \frac{c_1^m(n)}{\sqrt{c_0^m}} \sqrt{\log \left( \frac{144n}{\eta}  \right)}\right)\ge 1-\frac{\eta}{18},
            \end{align}
            where restricting $\eta$ to $(0,18)$ guarantees that the r.h.s.\ of \eqref{bound_D2_sub} is positive.
                     
            For (e) in \eqref{bound_D2}, recall that $\b R = \b U^m_{d^m} - \b U^m_{\b K^m,d^m} \left( \b U^m_{\b K^m,d^m} \right)^\top \b U^m_{d^m} \in \R^{n \times d^m}$. Then, we have for any $\eta \in \left(0,18 \right)$ and 
            \begin{align}
                n & \ge \max \Big\{ \colAonei{ n^m_{0,(i)}\left( \frac{\eta}{216}\right)}, \colAoneii{n^m_{0,(ii)}\left( \frac{\eta}{216}\right)}, \colAii{ n_{1,(ii)}^m\left(\frac{\eta}{18}\right)}, \colAoneii{n^m_{1,(ii)}\left(\frac{\eta}{36}\right)}, \colLfive{\left( \frac{\eta}{216} \right)^{-\frac{1}{c^m}}} , \colLsix{\frac{864}{\eta}}, \\
                & \hspace{1.7cm} \coletwo{e^{\frac{1}{\left( 2C_1^m - 4\sqrt{2} \right)^2}} } \Big\}
            \end{align}
            that                     \begin{align}\label{bound_R}
                \left \| \b R \right \| & = \left \| \left[ \b I - \b U^m_{\b K^m,d^m} \left( \b U^m_{\b K^m,d^m} \right)^\top \right] \b U^m_{d^m} \right \| \stackrel{(l)}{\le} \left \| \sin \Theta \left( \b U^m_{d^m}, \b U^m_{\b K^m,d^m} \right) \right \| \nonumber \\
                & \stackrel{(m)}{=} \left \| \mathcal{P}^m_{\left| \b A^m \right|} - \mathcal{P}^m_{\b K^m} \right \| \stackrel{(n)}{\le} 16 c_1^m(n) \left( \delta^m_{d^m} \right)^{-1}\sqrt{\frac{ \log \left( \frac{216n}{\eta} \right)}{n}},
                \end{align}
                holds with $\J^m$-probability at least $1-\frac{\eta}{18}$. (l) follows from the proof of Theorem 4.5 on page 92 in \citep{stewart1990matrix}\footnote{In Theorem 4.5 of \citep{stewart1990matrix}, $\X, \Y \in \b C_l^n$ denote the sets of all $l$-dimensional subspaces of the space of $n \times n$ matrices (see page~90 in the referred work). The notation $\left\| \cdot \right\|_2$ refers to the spectral norm, as specified in (69) by \citet{stewart1990matrix}. Consequently, $\b U^m_{\b K^m,d^m} \left( \b U^m_{\b K^m,d^m} \right)^\top$ corresponds to the projection $P_{\Y}$, while $\b U_{d^m}^m$ corresponds to $x \in \X$.}. 
                (m) is implied by \eqref{proj_sin_theta} in Lemma~\ref{lemm_sin_theta_theorem} and (n) holds for any $\eta \in \left(0,18 \right)$ and 
                \begin{align}
                    n & \ge \max \Big\{ \colAonei{ n^m_{0,(i)}\left( \frac{\eta}{216}\right)}, \colAoneii{n^m_{0,(ii)}\left( \frac{\eta}{216}\right)}, \colAii{ n_{1,(ii)}^m\left(\frac{\eta}{18}\right)}, \colAoneii{n^m_{1,(ii)}\left(\frac{\eta}{36}\right)}, \colLfive{\left( \frac{\eta}{216} \right)^{-\frac{1}{c^m}}} , \colLsix{\frac{864}{\eta}}, \\
                    & \hspace{1.7cm} \coletwo{e^{\frac{1}{\left( 2C_1^m - 4\sqrt{2} \right)^2}} } \Big\}
                \end{align}
                by \eqref{bound_proj} with a substitution $\eta \leftarrow \frac{\eta}{18}$. 
                Finally, (f) in \eqref{bound_D2} follows from $\log \left( \frac{144n}{\eta}\right) \le \log \left( \frac{216n}{\eta}\right)$. Combining \eqref{bound_D2_sub} and \eqref{bound_R} with a union bound gives the claimed result in \eqref{bound_D2}, where restricting $\eta$ to $(0,9)$ guarantees that the probability is positive.
            \item \textbf{Term} $D_3$: For any $\eta \in \left( 0, 9 \right)$ and 
            \begin{align}
                n \ge \max\left\{ \colAonei{ n^m_{0,(i)}\left( \frac{\eta}{72}\right)}, \colAoneii{n^m_{0,(ii)}\left( \frac{\eta}{36}\right)}, \colAi{n^m_{1,(i)}\left( \frac{\eta}{36} \right)}, \colLfive{\left( \frac{\eta}{72} \right)^{-\frac{1}{c^m}}}, \colLsix{\frac{144}{\eta}}, \coletwo{e^{\frac{1}{\left( 2C_1^m - 4\sqrt{2} \right)^2}} } \right\}
            \end{align}
            we have that
            \begin{align}
                D_3  & \stackrel{(a)}{=} \left(d^m \right)^{\frac{1}{2}} \left[\lambda^m_{d^m}(\b K^m) \right]^{-\frac{1}{2}} \left \| \b R_3 \b U^m_{\b K^m,d^m} \left( \b U^m_{\b K^m,d^m} \right)^\top \b U^m_{d^m} \right \| \\
                &\stackrel{(b)}{\le } \left(d^m \right)^{\frac{1}{2}} \left[\lambda^m_{d^m}(\b K^m) \right]^{-\frac{1}{2}} \left \| \b R_3 \right \| \\
                & \stackrel{(c)}{=} \left(d^m \right)^{\frac{1}{2}} \left[\lambda^m_{d^m}(\b K^m) \right]^{-\frac{1}{2}} \left \| \left( \b U^m_{\b K^m,d^m} \right)^\top \left[\b U^m_{d^m} \left( \b U^m_{d^m} \right)^\top \left | \b A^m \right |  - \b K^m \right] \right \|\\
                & \stackrel{(d)}{\le} \left(d^m \right)^{\frac{1}{2}} \left[\lambda^m_{d^m}(\b K^m) \right]^{-\frac{1}{2}} \left \| \b U^m_{d^m} \left( \b U^m_{d^m} \right)^\top \left | \b A^m \right | - \b K^m \right \| \\
                & \stackrel{(e)}{\le} \left(d^m \right)^{\frac{1}{2}} \left[\lambda^m_{d^m}(\b K^m) \right]^{-\frac{1}{2}} \left \| \left | \b A^m \right | - \b K^m \right \| \stackrel{(f)}{\le} 4 \frac{(d^m)^{\frac{1}{2}} c_1^m(n)}{\sqrt{c_0^m}} \sqrt{\log \left( \frac{72n}{\eta}  \right)} ,\label{bound_D3}
            \end{align}
            holds with $\J^m$-probability at least $1-\frac{\eta}{9}$. (a) is by the definition of $D_3$ [see \eqref{eq:D3-def}]. (b) follows from using $\left\| \b A \b B \right\| \le \left\| \b A \right\| \left\| \b B \right\|$, $\left\|\b A^\T\right\| = \left\|\b A\right\|$, $\left \| \b U^m_{\b K^m,d^m} \right \|\le 1$, 
              $\left \| \b U^m_{d^m} \right \| \le 1$. In (c), the definition of $\b R_3$ [see \eqref{eq:R3-def}] was applied. (d) comes from $\left\| \b A \b B \right\| \le \left\| \b A \right\| \left\| \b B \right\|$, $\left\|\b A^\T\right\| = \left\|\b A\right\|$, $\left \| \b U^m_{\b K^m,d^m} \right \|\le 1$.
            (e) follows from \eqref{bound_diff_psd1}, (f) holds with $\J^m$-probability at least $\frac{\eta}{9}$ as implied by \eqref{bound_D2_sub} with a substitution $\frac{\eta}{18} \leftarrow \frac{\eta}{9}$ for any $\eta \in \left( 0, 9 \right)$ and $n \ge \max\left\{ \colAonei{ n^m_{0,(i)}\left( \frac{\eta}{72}\right)}, \colAoneii{n^m_{0,(ii)}\left( \frac{\eta}{36}\right)}, \colAi{n^m_{1,(i)}\left( \frac{\eta}{36} \right)}, \colLfive{\left( \frac{\eta}{72} \right)^{-\frac{1}{c^m}}}, \colLsix{\frac{144}{\eta}}, \coletwo{e^{\frac{1}{\left( 2C_1^m - 4\sqrt{2} \right)^2}} } \right\}$.
            
        \end{itemize}
    
    Combining the obtained results for $D_1$, $D_2$ and $D_3$ with a union bound, one gets   
    \begin{align}
    & \left \| \tilde{\b R}_2 \right \|_F \le D_1+D_2+D_3 \le \\
    & \le 512 \frac{\left(d^m \right)^{\frac{1}{2}}\left( c_1^m(n) \right)^2}{\sqrt{c_0^m}} \left( \delta^m_{d^m} \right)^{-2} \frac{\log \left( \frac{216n}{\eta} \right)}{\sqrt{n}} + 1024 \frac{\left(d^m \right)^{\frac{1}{2}}\left( c_1^m(n) \right)^3}{\sqrt{c_0^m}} \left( \delta^m_{d^m} \right)^{-2} \frac{\log^{\frac{3}{2}} \left( \frac{216n}{\eta} \right)}{n}\\
    & \quad + 64 \frac{\left( c_1^m(n) \right)^2}{\sqrt{c_0^m}} \left( \delta^m_{d^m} \right)^{-1} \frac{\left(d^m \right)^{\frac{1}{2}} \log \left( \frac{216n}{\eta} \right)}{\sqrt{n}} + 4 \frac{(d^m)^{\frac{1}{2}} c_1^m(n)}{\sqrt{c_0^m}} \sqrt{\log \left( \frac{72n}{\eta}  \right)} \label{bound_tilde_R2} 
    \end{align}
    holds with $\J^m$-probability at least $1-\frac{\eta}{3}$ for any $\eta \in \left( 0, 9 \right) \cap \left( 0, 3 \right) = \left( 0, 3 \right)$
    \begin{align}
        n & \ge \max \Bigg\{ \colAonei{ n^m_{0,(i)}\left( \frac{\eta}{72}\right)}, \colAonei{ n^m_{0,(i)}\left( \frac{\eta}{144}\right)}, \colAonei{ n^m_{0,(i)}\left( \frac{\eta}{216}\right)}, \colAoneii{n^m_{0,(ii)}\left( \frac{\eta}{72}\right)}, \colAoneii{n^m_{0,(ii)}\left( \frac{\eta}{216}\right)}, \colAoneii{n^m_{0,(ii)}\left( \frac{\eta}{36}\right)},\\
        & \hspace{1.5 cm} \colAi{n_{1,(i)}^m\left(\frac{\eta}{72}\right)}, \colAi{n^m_{1,(i)}\left( \frac{\eta}{36} \right)}, \colAii{n_{1,(ii)}^m \left(\frac{\eta}{18}\right)}, \colAii{n^m_{1,(ii)}\left(\frac{\eta}{36}\right)}, \colLfive{\left( \frac{\eta}{216} \right)^{-\frac{1}{c^m}}}, \colLfive{\left( \frac{\eta}{72} \right)^{-\frac{1}{c^m}}}, \colLsix{\frac{864}{\eta}}, \\
        & \hspace{1.5 cm} \colLsix{\frac{144}{\eta}}, \coletwo{e^{\frac{1}{\left( 2C_1^m - 4\sqrt{2} \right)^2}} } \Bigg\} \\
        & \stackrel{(a)}{=} \max \Bigg\{ \colAonei{ n^m_{0,(i)}\left( \frac{\eta}{72}\right)}, \colAonei{ n^m_{0,(i)}\left( \frac{\eta}{144}\right)}, \colAonei{ n^m_{0,(i)}\left( \frac{\eta}{216}\right)}, \colAoneii{n^m_{0,(ii)}\left( \frac{\eta}{72}\right)}, \colAoneii{n^m_{0,(ii)}\left( \frac{\eta}{216}\right)}, \colAoneii{n^m_{0,(ii)}\left( \frac{\eta}{36}\right)},\\
        & \hspace{1.5 cm} \colAi{n_{1,(i)}^m\left(\frac{\eta}{72}\right)}, \colAi{n^m_{1,(i)}\left( \frac{\eta}{36} \right)}, \colAii{n_{1,(ii)}^m \left(\frac{\eta}{18}\right)}, \colAii{n^m_{1,(ii)}\left(\frac{\eta}{36}\right)}, \colLfive{\left( \frac{\eta}{216} \right)^{-\frac{1}{c^m}}}, \colLsix{\frac{864}{\eta}}, \coletwo{e^{\frac{1}{\left( 2C_1^m - 4\sqrt{2} \right)^2}} } \Bigg\},
    \end{align}
    where (a) is implied by \eqref{eq:x^a-monotonicity} and restricting $\eta$ to $(0,3)$ guarantees that the probability is positive.

   Considering $\b R_2 \in \R^{d^m\times d^m}$, and specifically its $(i,j) \in [d^m] \times [d^m]$-th entry:
    \begin{align}
        \left| (\b R_2)_{i,j} \right|& \stackrel{(a)}{=} \left| W^m_{i,j} \left(\lambda_j^m\left(\left|\b A^m\right| \right))^{\frac{1}{2}} - (\lambda_i^m(\b K^m)\right)^{\frac{1}{2}} W^m_{i,j} \right| \stackrel{(b)}{=} \left| W^m_{i,j} \frac{\lambda_j^m\left( \left|\b A^m\right| \right)-\lambda_i^m(\b K^m)}{(\lambda_j^m( \left|\b A^m\right| ))^{\frac{1}{2}} + (\lambda_i^m(\b K^m))^{\frac{1}{2}}} \right| \\
        & \stackrel{(c)}{=} 
        \left| W^m_{i,j} \right|  \frac{ \left| \lambda_j^m\left( \left|\b A^m\right| \right)-\lambda_i^m(\b K^m) \right| }{(\lambda_j^m( \left|\b A^m\right| ))^{\frac{1}{2}} + (\lambda_i^m(\b K^m))^{\frac{1}{2}}} 
        \stackrel{(d)}{\le} \left| W^m_{i,j} \right| \frac{\left| \lambda_j^m\left( \left|\b A^m\right| \right)-\lambda_i^m(\b K^m) \right|}{ (\lambda_i^m(\b K^m))^{\frac{1}{2}}} \\
        & \stackrel{(e)}{=} \left| \left( \tilde{\b R}_2 \right)_{i,j} \right|,
    \end{align}
    where in (a) we used the definition of $\b R_2$ [see \eqref{eq:R2-def}], (b) follows from the identity $(a-b)(a+b)=a^2-b^2$, using that $\lambda_j^m(\left| \b A^m \right|) \ge 0$ and $\lambda_j^m(\b K^m)>0$,\textsuperscript{\ref{footnote:Km-has-positive-eigenvalues}}
    (c) follows from the positivity of $(\lambda_j^m( \left|\b A^m\right| ))^{\frac{1}{2}} + (\lambda_j^m(\b K^m))^{\frac{1}{2}}$, (d) holds by not increasing the positive denominator and in (e) we applied the definition of $\tilde{\b R}_2$ [see \eqref{eq:R2-tilde-def}]. Therefore, we have $\left \| \b R_2 \right \|_F \le  \left \| \tilde{\b R}_2 \right \|_F$ and for any $\eta \in \left( 0, 3 \right)$ and 
    \begin{align}
        n & \ge \max \Bigg\{ \colAonei{ n^m_{0,(i)}\left( \frac{\eta}{72}\right)}, \colAonei{ n^m_{0,(i)}\left( \frac{\eta}{144}\right)}, \colAonei{ n^m_{0,(i)}\left( \frac{\eta}{216}\right)}, \colAoneii{n^m_{0,(ii)}\left( \frac{\eta}{72}\right)}, \colAoneii{n^m_{0,(ii)}\left( \frac{\eta}{216}\right)}, \colAoneii{n^m_{0,(ii)}\left( \frac{\eta}{36}\right)},\\
        & \hspace{1.5 cm} \colAi{n_{1,(i)}^m\left(\frac{\eta}{72}\right)}, \colAi{n^m_{1,(i)}\left( \frac{\eta}{36} \right)}, \colAii{n_{1,(ii)}^m \left(\frac{\eta}{18}\right)}, \colAii{n^m_{1,(ii)}\left(\frac{\eta}{36}\right)}, \colLfive{\left( \frac{\eta}{216} \right)^{-\frac{1}{c^m}}}, \colLsix{\frac{864}{\eta}}, \coletwo{e^{\frac{1}{\left( 2C_1^m - 4\sqrt{2} \right)^2}} } \Bigg\},
    \end{align}
    we have
    \begin{align}
        \MoveEqLeft \left \| \b R_2 \right \|_F  \le 512 \frac{\left(d^m \right)^{\frac{1}{2}} \left(c_1^m(n) \right)^2}{\sqrt{c_0^m}} \left( \delta^m_{d^m} \right)^{-2} \frac{\log \left( \frac{216n}{\eta} \right)}{\sqrt{n}}\\
        & \quad \hspace{0.1cm} + 1024 \frac{\left(d^m \right)^{\frac{1}{2}}\left(c_1^m(n) \right)^3}{\sqrt{c_0^m}} \left( \delta^m_{d^m} \right)^{-2} \frac{\log^{\frac{3}{2}} \left( \frac{216n}{\eta} \right)}{n} \nonumber \\
        & \quad + 64 \frac{\left(c_1^m(n) \right)^2}{\sqrt{c_0^m}} \left( \delta^m_{d^m} \right)^{-1} \frac{\left(d^m \right)^{\frac{1}{2}} \log \left( \frac{216n}{\eta} \right)}{\sqrt{n}} + 4 \frac{(d^m)^{\frac{1}{2}} c_1^m(n)}{\sqrt{c_0^m}} \sqrt{\log \left( \frac{72n}{\eta}  \right)} \label{bound_R2}
    \end{align}
    holds with $\J^m$-probability at least $1-\frac{\eta}{3}$.
\end{itemize}
Having established a bound on $\left\|\b R_1\right\|$ and $\left\|\b R_2\right\|_F$, we proceed by controlling $\b A_1$, $\b A_2$, and $\b A_3$ in \eqref{eq:initial_decompose} 
after proving that the following inequality
\begin{align}\label{eq:bound_eig_A}
    \left\| \left( \b S^m_{d^m} \right)^{-\frac{1}{2}} \right\|= \underbrace{\left[ \lambda^m_{d^m}(\left| \b A^m \right| ) \right]^{-\frac{1}{2}} \le 2\sqrt{2}(n \delta_{d^m}^m)^{-\frac{1}{2}}}_{=:E}
\end{align}
holds with $\J^m$-probability at least $1-\eta$ for any $\eta \in \left( 0,1 \right)$ and 
\begin{align}
    n \ge \max &\Bigg\{ \colAonei{ n^m_{0,(i)}\left( \frac{\eta}{64}\right)}, \colAonei{ n^m_{0,(i)}\left( \frac{\eta}{32}\right)}, \colAoneii{n^m_{0,(ii)}\left( \frac{\eta}{32}\right)}, \colAi{n^m_{1,(i)}\left( \frac{\eta}{32} \right)}, \colAii{ n_{1,(ii)}^m \left( \frac{3\eta}{4} \right)}, \colAii{ n_{1,(ii)}^m \left( \frac{3\eta}{16} \right)}, \\
    & \hspace{0.3cm} \colLfive{\left( \frac{\eta}{64} \right)^{-\frac{1}{c^m}}},  \colLsix{\frac{128}{\eta}}, \coleone{32 \left(c_1^m(n)\right)^2 (c_0^m)^{-2} \log \left( \frac{64n}{\eta} \right)},\coletwo{e^{\frac{1}{\left( 2C_1^m - 4\sqrt{2} \right)^2}}} \Bigg\}.
\end{align}

Considering $\lambda_{d^m}^m(\b K^m)$, applying \eqref{eig_second_bound} with a substitution $\frac{2\eta}{3}\leftarrow \eta$, for any $\eta \in \left( 0,\frac{2}{3} \right)$ and $n\ge \colAii{ n_{1,(ii)}^m\left(\frac{3}{2}\eta\right) }$, we have
                \begin{align} \label{eig_third_bound0}
                    \J^m \left( \lambda^m_{d^m}(\b K^m) \ge \lambda^m_{d^m+1}(\b K^m) + \frac{n \delta^m_{d^m}}{2} \stackrel{(a)}{\ge} \frac{n \delta^m_{d^m}}{2}  \right) \ge 1- \eta,
                \end{align}
                (a) holds by the  positive semi-definiteness of $\b K^m$ which implies that $\lambda^m_{d^m+1}(\b K^m) \ge 0$. Thus, we have that for any $\eta \in \left( 0,\frac{2}{3} \right)$ and $n\ge \colAii{n_{1,(ii)}^m\left(\frac{3}{2}\eta\right) }$
                \begin{align}\label{eig_third_bound}
                \J^m \Big(\left[\lambda^m_{d^m}(\b K^m) \right]^{-\frac{1}{2}} \le \sqrt{2} (n \delta^m_{d^m})^{-\frac{1}{2}} \Big)  \ge 1- \eta.
                \end{align}
                Next we discuss the range of $\lambda^m_{d^m}(\left| \b A^m \right| )$. Getting an upper bound on $\left[\lambda^m_{d^m}(\left| \b A^m \right|) \right]^{-\frac{1}{2}}$ in \eqref{eq:bound_eig_A} is equivalent to obtaining a lower bound on $\lambda^m_{d^m}(\left| \b A^m \right|) $. Having the lower bound of $\lambda^m_{d^m}(\b K^m)$ as shown in \eqref{eig_third_bound0}, we next use the relationship between $\lambda^m_{d^m}(\left| \b A^m \right|) $ and $\lambda^m_{d^m}(\b K^m) $ to discuss the lower bound of $\lambda^m_{d^m}(\left| \b A^m \right|) $. 
                Let $F$ denote the event $\lambda^m_{d^m}(\left| \b A^m \right| )  \ge \lambda^m_{d^m}( \b K^m )$, and assume that $1-a=\J^m(F)$; then $a = 1- (1-a)= \J^m(\bar{F})$ where $\bar{F}$ is the complement of $F$. 
                
\begin{itemize}
    	\item \textbf{Event $E \cap F$}: In this case, we have $\lambda_{d^m}^m\left(|\b A^m|\right) \stackrel{F}{\ge} \lambda^m_{d^m}\left(\b K^m\right)>0$ (hence taking $[\lambda_{d^m}^m\left(|\b A^m|\right)]^{-\frac{1}{2}}$ is valid) and
    	\begin{align}
    	     \left[ \lambda^m_{d^m}(\left| \b A^m \right| ) \right]^{-\frac{1}{2}} \stackrel{F}{\le} \left[ \lambda^m_{d^m}(\b K^m) \right]^{-\frac{1}{2}} \stackrel{(a)}{\le} \sqrt{2}(n \delta_{d^m}^m)^{-\frac{1}{2}} \le 2\sqrt{2}(n \delta_{d^m}^m)^{-\frac{1}{2}},
    	\end{align}
    	holds with $\J^m$-probability at least $1- \left( a+\frac{\eta}{2} \right)$ for any $\eta \in \left( 0,\frac{4}{3} \right)$ and $n \ge \colAii{n_{1,(ii)}^m \left( \frac{3 \eta}{4} \right)}$ by using a union bound, where (a) follows \eqref{eig_third_bound} by a substitution $\eta \leftarrow \frac{\eta}{2}$ for any $\eta \in \left( 0,\frac{4}{3} \right)$ and $n \ge \colAii{ n_{1,(ii)}^m \left( \frac{3 \eta}{4} \right) }$.
    	\item  \textbf{Event $E \cap \bar{F}$}: 
        First we show that $\lambda^m_{d^m}(\left| \b A^m \right| )>0$ with high probability to ensure that we can divide with it. For $\lambda^m_{d^m}(\left| \b A^m \right| )$ we have that for any $\eta \in \left(0, 4 \right)$ and for $n \ge \max \left\{\colAonei{ n^m_{0,(i)}\left( \frac{\eta}{32}\right)}, \colAoneii{n^m_{0,(ii)}\left( \frac{\eta}{32}\right)}, \colAii{ n_{1,(ii)}^m \left( \frac{3\eta}{16} \right)}, \colLfive{\left( \frac{\eta}{32} \right)^{-\frac{1}{c^m}}}, \colLsix{\frac{128}{\eta}}, \coletwo{e^{\frac{1}{\left( 2C_1^m - 4\sqrt{2} \right)^2}} } \right\}$
        \begin{align}
            & \hspace{-0.7cm} \lambda^m_{d^m}(\left| \b A^m \right| )  \stackrel{(a)}{\ge} \lambda_{d^m}^m(\b K^m) - 4 c_1^m(n) \sqrt{n  \log \left( \frac{32n}{\eta} \right)} \stackrel{(b)}{\ge} \frac{n \delta^m_{d^m}}{2} - 4 c_1^m(n) \sqrt{n  \log \left( \frac{32n}{\eta} \right)} \label{eq:eig_adj_sub0}\\
            & \hspace{-0.1cm}  \stackrel{(c)}{\ge}  \frac{n}{2} 16 c_1^m(n) \sqrt{\frac{\log \left( \frac{32n}{\eta} \right)}{n}} - 4 c_1^m(n) \sqrt{n  \log \left( \frac{32n}{\eta} \right)} \label{eq:eig_adj_sub0-line2}\\
            & \hspace{-0.1cm}  = 8 c_1^m(n) \sqrt{n \log \left( \frac{32n}{\eta} \right)} - 4 c_1^m(n) \sqrt{n  \log \left( \frac{32n}{\eta} \right)} = 4 c_1^m(n) \sqrt{n  \log \left( \frac{32n}{\eta} \right)} >0 \label{eq:eig_adj_sub0-final}
        \end{align}
        holds with $\J^m$-probability at least $1-\frac{\eta}{4}$. (a) is implied by 
        \begin{align}
    		& \left|  \lambda^m_{d^m}(\left| \b A^m \right| )  -   \lambda_{d^m}^m(\b K^m)   \right| \stackrel{(d)}{\le} \left\|  \left| \b A^m \right| -\b K^m  \right\| \stackrel{(e)}{\le} 4 c_1^m(n) \sqrt{n  \log \left( \frac{4n}{\eta/8} \right)}  \Rightarrow  \label{eq:eig_adj_sub1-pre}\\
    		& \quad \lambda^m_{d^m}(\left| \b A^m \right| )  \ge \lambda_{d^m}^m(\b K^m) - 4 c_1^m(n) \sqrt{n  \log \left( \frac{32n}{\eta} \right)} , \label{eq:eig_adj_sub1}
    	\end{align}
    	holding with $\J^m$-probability at least $1-\frac{\eta}{8}$ for any 
        \begin{align}
            n \ge \max \left\{\colAonei{ n^m_{0,(i)}\left( \frac{\eta}{32}\right)}, \colAoneii{n^m_{0,(ii)}\left( \frac{\eta}{32}\right)}, \colLfive{ \left( \frac{\eta}{32} \right)^{-\frac{1}{c^m}}}, \colLsix{\frac{128}{\eta}}, \coletwo{e^{\frac{1}{\left( 2C_1^m - 4\sqrt{2} \right)^2}} }\right\},
        \end{align}
        where (d) holds by the Weyl's Perturbation Theorem \citep[Corollary~III.2.6 on page~63]{bhatia1997matrix}, (e) follows with $\J^m$-probability at least $1-\frac{\eta}{8}$ for any $\eta \in (0,8)$  and $n \ge \max \left\{\colAonei{ n^m_{0,(i)}\left( \frac{\eta}{32}\right)}, \colAoneii{n^m_{0,(ii)}\left( \frac{\eta}{32}\right)}, \colLfive{ \left( \frac{\eta}{32} \right)^{-\frac{1}{c^m}}}, \colLsix{\frac{128}{\eta}}, \coletwo{e^{\frac{1}{\left( 2C_1^m - 4\sqrt{2} \right)^2}} }\right\}$ from \eqref{bound_diff_psd} by a substitution $\frac{\eta}{3} \leftarrow \frac{\eta}{8}$. (b) in \eqref{eq:eig_adj_sub0} follows from $\lambda_{d^m}^m(\b K^m) \ge \frac{n \delta^m_{d^m}}{2}$ with $\J^m$-probability at least $1-\frac{\eta}{8}$ for any $\eta \in \left( 0,\frac{16}{3} \right)$ and $n \ge \colAii{ n_{1,(ii)}^m \left(\frac{3}{2}\frac{\eta}{8}\right)= n_{1,(ii)}^m \left( \frac{3\eta}{16} \right) }$  by using \eqref{eig_third_bound0} with a substitution $\eta\leftarrow \frac{\eta}{8}$. (c) in \eqref{eq:eig_adj_sub0-line2} is implied by Assumption~\ref{assum_lambda_delta}(ii) with a substitution $\frac{\eta}{3}  \leftarrow \frac{\eta}{16}$ for any $\eta \in \left( 0,\frac{16}{3} \right)$ and $n \ge \colAii{n_{1,(ii)}^m \left(  \frac{3\eta}{16} \right)}$. A union bound gives the claimed result in \eqref{eq:eig_adj_sub0-final} for any $\eta \in \left( 0,\frac{16}{3} \right) \cap \left(0, 8 \right) \cap (0,4) = \left( 0,4 \right)$, where restricting $\eta$ to $(0,4)$ guarantees that the probability is positive.
        
        For any $\eta \in (0,4)$ and 
        \begin{align}
            n \ge \max \left\{\colAonei{ n^m_{0,(i)}\left( \frac{\eta}{32}\right)}, \colAoneii{n^m_{0,(ii)}\left( \frac{\eta}{32}\right)}, \colAii{ n_{1,(ii)}^m \left( \frac{3\eta}{16} \right)}, \colLfive{\left( \frac{\eta}{32} \right)^{-\frac{1}{c^m}}}, \colLsix{\frac{128}{\eta}}, \coletwo{e^{\frac{1}{\left( 2C_1^m - 4\sqrt{2} \right)^2}} } \right\},
        \end{align}
        after verifying that $\lambda^m_{d^m}(\left| \b A^m \right| ) > 0$ holds with $\J^m$-probability at least $1-\frac{\eta}{4}$, we can discuss on this highly probable event further the upper bound of $\left[ \lambda^m_{d^m}(\left| \b A^m \right| ) \right]^{-\frac{1}{2}}$. We have  
    	  \begin{align}
    		& \quad \left[ \lambda^m_{d^m}(\left| \b A^m \right| ) \right]^{-\frac{1}{2}} - \left[ \lambda^m_{d^m}(\b K^m) \right]^{-\frac{1}{2}} \\
            & \stackrel{(a)}{=} \frac{ \lambda^m_{d^m}(\b K^m) - \lambda^m_{d^m}(\left| \b A^m \right| ) }{ \left[ \lambda^m_{d^m}(\b K^m) \right]^{\frac{1}{2}} \left( \left[ \lambda^m_{d^m}(\left| \b A^m \right| )  \lambda^m_{d^m}(\b K^m) \right]^{\frac{1}{2}} +  \lambda^m_{d^m}(\left| \b A^m \right| ) \right)}  \label{eq:bound_adj_sub1}\\
    		& \stackrel{(b)}{\le} \frac{ 4c_1^m(n) (c_0^m)^{-\frac{1}{2}} \sqrt{\log \left( \frac{64n}{\eta} \right)} }{ \left[ \lambda^m_{d^m}(\left| \b A^m \right| )  \lambda^m_{d^m}(\b K^m) \right]^{\frac{1}{2}} +  \lambda^m_{d^m}(\left| \b A^m \right| )  } \\
            &\stackrel{(c)}{\le} 4c_1^m(n) (c_0^m)^{-\frac{1}{2}} \sqrt{\log \left( \frac{64n}{\eta} \right)} \left[ \lambda^m_{d^m}(\left| \b A^m \right| )  \lambda^m_{d^m}(\b K^m) \right]^{-\frac{1}{2}} \label{eq:bound_adj_sub1-final}
    	\end{align} 
        holds with $\J^m$-probability at least $1-\left( 1-a+ \frac{\eta}{4}+ \frac{\eta}{8} \right)=1-\left( 1-a+ \frac{3\eta}{8} \right)$ for any $\eta \in \left( 0, 4 \right) \cap \left(0,8 \right) = \left(0,4 \right)$ and 
        \begin{align}
            n \ge \max & \Bigg\{ \colAonei{n^m_{0,(i)}\left( \frac{\eta}{64}\right)}, \colAonei{ n^m_{0,(i)}\left( \frac{\eta}{32}\right)}, \colAoneii{n^m_{0,(ii)}\left( \frac{\eta}{32}\right)}, \colAi{n^m_{1,(i)}\left( \frac{\eta}{32} \right)}, \colAii{ n_{1,(ii)}^m \left( \frac{3\eta}{16} \right)}, \colLfive{\left( \frac{\eta}{64} \right)^{-\frac{1}{c^m}}}, \colLsix{\frac{128}{\eta}}, \\
            & \coletwo{e^{\frac{1}{\left( 2C_1^m - 4\sqrt{2} \right)^2}} }  \Bigg\}.
        \end{align}
        (a) in \eqref{eq:bound_adj_sub1-final} follows from the identity $c^{-\frac{1}{2}} - b^{-\frac{1}{2}} = \frac{b-c}{b^{\frac{1}{2}} \left[(cb)^{\frac{1}{2}} + c\right]}$ and $\lambda^m_{d^m}(\left| \b A^m \right| ) > 0$ holds with $\J^m$-probability at least $1-\frac{\eta}{4}$ for $\eta \in (0,4)$ and 
        \begin{align}
            n \ge \max \left\{\colAonei{ n^m_{0,(i)}\left( \frac{\eta}{32}\right)}, \colAoneii{n^m_{0,(ii)}\left( \frac{\eta}{32}\right)}, \colAii{ n_{1,(ii)}^m \left( \frac{3\eta}{16} \right)}, \colLfive{\left( \frac{\eta}{32} \right)^{-\frac{1}{c^m}}}, \colLsix{\frac{128}{\eta}}, \coletwo{e^{\frac{1}{\left( 2C_1^m - 4\sqrt{2} \right)^2}} } \right\}.
        \end{align}
        (b) in \eqref{eq:bound_adj_sub1-final} holds since considering the numerator and the 1st term in the denominator on the r.h.s.\ of \eqref{eq:bound_adj_sub1}, one gets for $\eta \in \left( 0, 8 \right)$ and
        \begin{align}
            n \ge \max \left\{ \colAonei{n^m_{0,(i)}\left( \frac{\eta}{64}\right)}, \colAoneii{n^m_{0,(ii)}\left( \frac{\eta}{32}\right)}, \colAi{n^m_{1,(i)}\left( \frac{\eta}{32} \right)}, \colLfive{\left( \frac{\eta}{64} \right)^{-\frac{1}{c^m}}}, \colLsix{\frac{128}{\eta}}, \coletwo{e^{\frac{1}{\left( 2C_1^m - 4\sqrt{2} \right)^2}} } \right\}
        \end{align}
        that
    	\begin{align}
    			 & \quad \left[ \lambda^m_{d^m}(\b K^m) \right]^{-\frac{1}{2}} \left[ \lambda^m_{d^m}(\b K^m) - \lambda^m_{d^m}(\left| \b A^m \right| ) \right] \le \left[ \lambda^m_{d^m}(\b K^m) \right]^{-\frac{1}{2}} \max_{i \in [n]} \left|    \lambda^m_{i}(\b K^m) - \lambda^m_{i}(\left| \b A^m \right| )  \right| \\
                 & \stackrel{(d)}{\le} \left[ \lambda^m_{d^m}(\b K^m) \right]^{-\frac{1}{2}} \left\|  \b K^m - \left| \b A^m \right|   \right\|   \stackrel{(e)}{\le} 4 c_1^m(n) (c_0^m)^{-\frac{1}{2}} \sqrt{\log \left( \frac{64n}{\eta} \right)}
    		\end{align}
    		holds with $\J^m$-probability at least $1-\frac{\eta}{8}$. Indeed, (d) is implied by Weyl's Perturbation Theorem \citep[Corollary~III.2.6 on page~63]{bhatia1997matrix} and (e) follows with $\J^m$-probability at least $1-\frac{\eta}{8}$ for any $\eta \in \left( 0, 8 \right)$ and 
            \begin{align}
                n \ge \max\left\{ \colAonei{n^m_{0,(i)}\left( \frac{\eta}{64}\right)}, \colAoneii{n^m_{0,(ii)}\left( \frac{\eta}{32}\right)}, \colAi{n^m_{1,(i)}\left( \frac{\eta}{32} \right)}, \colLfive{\left( \frac{\eta}{64} \right)^{-\frac{1}{c^m}}}, \colLsix{\frac{128}{\eta}}, \coletwo{e^{\frac{1}{\left( 2C_1^m - 4\sqrt{2} \right)^2}} } \right\}
            \end{align}
            from \eqref{bound_D2_sub} with a substitution $\frac{\eta}{18} \leftarrow \frac{\eta}{8}$. Finally, (c) in \eqref{eq:bound_adj_sub1-final} follows from not increasing the denominator. 
            To sum up, until this point, we have for any $\eta \in \left( 0,4 \right) \cap (0,8) = (0,4)$  and 
            \begin{align}
                n & \ge \max \Bigg\{ \colAonei{n^m_{0,(i)}\left( \frac{\eta}{64}\right)}, \colAonei{ n^m_{0,(i)}\left( \frac{\eta}{32}\right)}, \colAoneii{n^m_{0,(ii)}\left( \frac{\eta}{32}\right)}, \colAi{n^m_{1,(i)}\left( \frac{\eta}{32} \right)}, \colAii{ n_{1,(ii)}^m \left( \frac{3\eta}{16} \right)}, \colLfive{\left( \frac{\eta}{64} \right)^{-\frac{1}{c^m}}}, \\
                & \hspace{1.5cm} \colLfive{\left( \frac{\eta}{32} \right)^{-\frac{1}{c^m}}}, \colLsix{\frac{128}{\eta}}, \coletwo{e^{\frac{1}{\left( 2C_1^m - 4\sqrt{2} \right)^2}} }  \Bigg\} \\
                & \stackrel{\eqref{eq:x^a-monotonicity}}{\ge} \max \Bigg\{ \colAonei{n^m_{0,(i)}\left( \frac{\eta}{64}\right)}, \colAonei{ n^m_{0,(i)}\left( \frac{\eta}{32}\right)}, \colAoneii{n^m_{0,(ii)}\left( \frac{\eta}{32}\right)}, \colAi{n^m_{1,(i)}\left( \frac{\eta}{32} \right)}, \colAii{ n_{1,(ii)}^m \left( \frac{3\eta}{16} \right)}, \colLfive{\left( \frac{\eta}{64} \right)^{-\frac{1}{c^m}}},  \\
                & \hspace{1.5cm} \colLsix{\frac{128}{\eta}}, \coletwo{e^{\frac{1}{\left( 2C_1^m - 4\sqrt{2} \right)^2}} }  \Bigg\},
            \end{align}
            that the following inequality holds
            \begin{align}
                & \J^m \Big( \left[ \lambda^m_{d^m}(\left| \b A^m \right| ) \right]^{-\frac{1}{2}} - \left[ \lambda^m_{d^m}(\b K^m) \right]^{-\frac{1}{2}}\\
                & \quad \hspace{0.2cm}\le \underbrace{4 c_1^m(n) (c_0^m)^{-\frac{1}{2}} \sqrt{\log \left( \frac{64n}{\eta} \right)} \left[ \lambda^m_{d^m}(\left| \b A^m \right| )  \lambda^m_{d^m}(\b K^m) \right]^{-\frac{1}{2}}}_{=:G^m=:G} \Big) \\
                & \ge 1- \left( 1-a + \frac{\eta}{4} + \frac{\eta}{8} \right) = 1- \left( 1-a + \frac{3\eta}{8} \right). \label{eq:eig_diff_Jm}          
            \end{align}            
            Now, we will show that $G$ in \eqref{eq:eig_diff_Jm} is smaller or equal than $\left[ \lambda^m_{d^m}(\b K^m) \right]^{-\frac{1}{2}}$ when $n$ is large enough. Indeed, we have for any $\eta \in \left( 0,4 \right)$  and 
            \begin{align}
                n &\ge \max \Bigg\{ \colAonei{n^m_{0,(i)}\left( \frac{\eta}{64}\right)}, \colAonei{ n^m_{0,(i)}\left( \frac{\eta}{32}\right)}, \colAoneii{n^m_{0,(ii)}\left( \frac{\eta}{32}\right)}, \colAi{n^m_{1,(i)}\left( \frac{\eta}{32} \right)}, \colAii{ n_{1,(ii)}^m \left( \frac{3\eta}{16} \right)}, \colLfive{\left( \frac{\eta}{64} \right)^{-\frac{1}{c^m}}}, \colLsix{\frac{128}{\eta}}, \\
                & \hspace{1.5cm} \coletwo{e^{\frac{1}{\left( 2C_1^m - 4\sqrt{2} \right)^2}}} \Bigg\}
            \end{align}
            the following inequality 
            \begin{align}
                G &= 4 c_1^m(n) (c_0^m)^{-\frac{1}{2}} \sqrt{\log \left( \frac{64n}{\eta} \right)} \left[ \lambda^m_{d^m}(\left| \b A^m \right| )  \lambda^m_{d^m}(\b K^m) \right]^{-\frac{1}{2}}\\
                &\le \left[ \lambda^m_{d^m}(\b K^m) \right]^{-\frac{1}{2}} \Leftrightarrow \\
                1 &\ge 4 c_1^m(n) (c_0^m)^{-\frac{1}{2}} \sqrt{\log \left( \frac{64n}{\eta} \right)} \left[ \lambda^m_{d^m}(\left| \b A^m \right| ) \right]^{-\frac{1}{2}} \Leftrightarrow \\
                 \lambda^m_{d^m}(\left| \b A^m \right| )  &\ge 16 \left( c_1^m(n) \right)^2  (c_0^m)^{-1} \log \left( \frac{64n}{\eta} \right) \stackrel{(f)}{\Leftarrow}\\
                 4 c_1^m(n) \sqrt{n  \log \left( \frac{32n}{\eta} \right)} &\ge 16 \left( c_1^m(n) \right)^2 (c_0^m)^{-1} \log \left( \frac{64n}{\eta} \right) \stackrel{(g)}{\Leftrightarrow}\\
                 n &\ge 16 \left(c_1^m(n)\right)^2 (c_0^m)^{-2} \frac{\log^2 \left( \frac{64n}{\eta} \right)}{\log \left( \frac{32n}{\eta} \right)} \stackrel{(h)}{\Leftarrow} \\
                 n & \ge 32 \left(c_1^m(n)\right)^2 (c_0^m)^{-2} \log \left( \frac{64n}{\eta} \right) \label{eq:source_else1}
            \end{align}
            holds with $\J^m$-probability at least $1-(1-a+\frac{\eta}{4} + \frac{\eta}{8}) = 1-\left(1-a+\frac{3\eta}{8} \right)$; in (f) we used the lower bound of $\lambda^m_{d^m}(\left| \b A^m \right| )$ provided by \eqref{eq:eig_adj_sub0-final}, in (g) we expressed $n$, (h) holds as
            \begin{align}
                \log \left( \frac{64n}{\eta} \right) = \log \left( 2 \frac{32n}{\eta} \right) = \log(2) + \log \left( \frac{32n}{\eta} \right) \le 2 \log \left( \frac{32n}{\eta} \right),
            \end{align}
            where the last inequality follows from $8 < \frac{32n}{\eta}$ [implied by $n\ge 1$ and $\eta \in (0,4)$] combined with the monotonicity of the function $\log$ [$\log(2)  \le \log(8) \le \log\left( \frac{32n}{\eta} \right)$].
            
            Therefore, for any $\eta \in \left( 0,4 \right)$  and 
            \begin{align}
                n \ge \max &\Bigg\{ \colAonei{ n^m_{0,(i)}\left( \frac{\eta}{64}\right)}, \colAonei{ n^m_{0,(i)}\left( \frac{\eta}{32}\right)}, \colAoneii{n^m_{0,(ii)}\left( \frac{\eta}{32}\right)}, \colAi{n^m_{1,(i)}\left( \frac{\eta}{32} \right)}, \colAii{ n_{1,(ii)}^m \left( \frac{3\eta}{16} \right)}, \colLfive{\left( \frac{\eta}{64} \right)^{-\frac{1}{c^m}}}, \colLsix{\frac{128}{\eta}}, \\
                & \hspace{0.3cm} \coleone{32 \left(c_1^m(n)\right)^2 (c_0^m)^{-2} \log \left( \frac{64n}{\eta} \right)},\coletwo{e^{\frac{1}{\left( 2C_1^m - 4\sqrt{2} \right)^2}}} \Bigg\}
            \end{align}
            we have
            \begin{align}
                \J^m \left( \left[ \lambda^m_{d^m}(\left| \b A^m \right| ) \right]^{-\frac{1}{2}} - \left[ \lambda^m_{d^m}(\b K^m) \right]^{-\frac{1}{2}} \le \left[ \lambda^m_{d^m}(\b K^m) \right]^{-\frac{1}{2}} \right) &\ge 1-\left(1-a+\frac{3\eta}{8} \right),\text{i.e.}\\
                \J^m \left( \left[ \lambda^m_{d^m}(\left| \b A^m \right| ) \right]^{-\frac{1}{2}} \le 2 \left[ \lambda^m_{d^m}(\b K^m) \right]^{-\frac{1}{2}} \right) &\ge 1-\left(1-a+\frac{3\eta}{8} \right).
            \end{align}
            Therefore, we got that for any $\eta \in \left( 0, 4 \right) \cap \left( 0,\frac{16}{3} \right) = (0,4)$ and
            \begin{align}
                n \ge \max &\Bigg\{ \colAonei{ n^m_{0,(i)}\left( \frac{\eta}{64}\right)}, \colAonei{ n^m_{0,(i)}\left( \frac{\eta}{32}\right)}, \colAoneii{n^m_{0,(ii)}\left( \frac{\eta}{32}\right)}, \colAi{n^m_{1,(i)}\left( \frac{\eta}{32} \right)}, \colAii{ n_{1,(ii)}^m \left( \frac{3\eta}{16} \right)}, \colLfive{\left( \frac{\eta}{64} \right)^{-\frac{1}{c^m}}}, \colLsix{\frac{128}{\eta}}, \\
                & \hspace{0.3cm} \coleone{32 \left(c_1^m(n)\right)^2 (c_0^m)^{-2} \log \left( \frac{64n}{\eta} \right)},\coletwo{e^{\frac{1}{\left( 2C_1^m - 4\sqrt{2} \right)^2}}} \Bigg\},
            \end{align}
            the inequality
            \begin{align}
            \J^m \left( \left[ \lambda^m_{d^m}(\left| \b A^m \right| ) \right]^{-\frac{1}{2}} \le 2\sqrt{2}(n \delta_{d^m}^m)^{-\frac{1}{2}} \right) &\ge 1-\left(1-a+\frac{3\eta}{8} + \frac{\eta}{8} \right) = 1-\left(1-a+\frac{\eta}{2}\right)
            \end{align}
            holds, as implied by \eqref{eig_third_bound} with substitution $\eta \leftarrow \frac{\eta}{8}$ for any $\eta \in \left( 0,\frac{16}{3} \right)$ and $n\ge n_{1,(ii)}^m \left( \frac{3}{2} \frac{\eta}{8} \right) = \colAii{n_{1,(ii)}^m \left(\frac{3\eta}{16} \right)}$.	
\end{itemize}
The event $E$ defined in \eqref{eq:bound_eig_A} can be obtained by the disjoint union of the events $E \cap F$ and $E \cap \bar{F}$; therefore we have that for any $\eta \in \left( 0,\frac{4}{3} \right) \cap \left( 0,4 \right) \cap (0,1) = \left( 0,1 \right)$ and
\begin{align}
    \J^m (E) = \J^m (E \cap F) + \J^m (E \cap \bar{F}) = 1-\left( a + \frac{\eta}{2} \right) + 1- \left( 1-a + \frac{\eta}{2} \right) = 1-\eta
\end{align}
when 
\begin{align}
    n \ge \max &\Bigg\{ \colAonei{ n^m_{0,(i)}\left( \frac{\eta}{64}\right)}, \colAonei{ n^m_{0,(i)}\left( \frac{\eta}{32}\right)}, \colAoneii{n^m_{0,(ii)}\left( \frac{\eta}{32}\right)}, \colAi{n^m_{1,(i)}\left( \frac{\eta}{32} \right)}, \colAii{ n_{1,(ii)}^m \left( \frac{3\eta}{4} \right)}, \colAii{ n_{1,(ii)}^m \left( \frac{3\eta}{16} \right)},  \\
    & \hspace{0.3cm} \colLfive{\left( \frac{\eta}{64} \right)^{-\frac{1}{c^m}}}, \colLsix{\frac{128}{\eta}},\coleone{32 \left(c_1^m(n)\right)^2 (c_0^m)^{-2} \log \left( \frac{64n}{\eta} \right)},\coletwo{e^{\frac{1}{\left( 2C_1^m - 4\sqrt{2} \right)^2}}} \Bigg\},
\end{align}
where restricting $\eta$ to $(0, 1)$ guarantees that the probability is positive.

Having established the probabilistic bound on $\left\| \left( \b S^m_{d^m} \right)^{-\frac{1}{2}} \right\|$ in \eqref{eq:bound_eig_A}, we continue the proof with the control of $\b A_1$, $\b A_2$ and $\b A_3$ in \eqref{eq:initial_decompose}. The details are as follows.

\begin{itemize}
    \item \textbf{Term} $\b A_1$: Considering $\b A_1$, we have
    \begin{align}
        \b A_1 & \stackrel{(a)}{=} \b U^m_{d^m} \left( \b S^m_{d^m} \right)^{\frac{1}{2}} - \b U^m_{\b K^m,d^m} \left( \b U^m_{\b K^m,d^m} \right)^\top \b U^m_{d^m} \left( \b S^m_{d^m} \right)^{\frac{1}{2}} \\
        & \stackrel{(b)}{=} \b U^m_{d^m} \left( \b U^m_{d^m} \right)^{\top} \left | \b A^m \right |  \b U^m_{d^m} \left( \b S^m_{d^m} \right)^{-\frac{1}{2}} \\
        & \quad - \b U^m_{\b K^m,d^m} \left( \b U^m_{\b K^m,d^m} \right)^\top \b U^m_{d^m} \left( \b U^m_{d^m} \right)^{\top} \left | \b A^m \right |  \b U^m_{d^m} \left( \b S^m_{d^m} \right)^{-\frac{1}{2}}\\
        & \stackrel{(c)}{=} \underbrace{\left[\b U^m_{d^m} \left( \b U^m_{d^m} \right)^{\top} \left | \b A^m \right |  - \b U^m_{\b K^m,d^m} \left( \b U^m_{\b K^m,d^m} \right)^\top \b K^m \right]\b U^m_{d^m} \left( \b S^m_{d^m} \right)^{-\frac{1}{2}}}_{=:\b E_1^m =: \b E_1 \in \R^{n \times d^m}} \label{eq:E1-def}\\
        & \quad - \underbrace{\b U^m_{\b K^m,d^m} \left(\b U^m_{\b K^m,d^m}\right)^\top \left[ \b U^m_{d^m} \left( \b U^m_{d^m} \right)^{\top} \left | \b A^m \right |  -  \b K^m \right] \b U^m_{d^m} \left( \b S^m_{d^m} \right)^{-\frac{1}{2}}}_{=:\b E_2^m =: \b E_2 \in \R^{n \times d^m}}. \label{eq:A_1-decomposition}
    \end{align}
    (a) is implied by the definition of $\b A_1$ [see \eqref{eq:initial_decompose}]. (b) holds by using the fact that the columns of $\b U^m_{d^m}$ are orthogonal:
    \begin{align}
        \b U^m_{d^m} \left( \b U^m_{d^m} \right)^\T \left | \b A^m \right | \b U^m_{d^m} \left( \b S^m_{d^m} \right)^{-\frac{1}{2}} &\stackrel{\eqref{eq:projection0} }{=} \b U^m_{d^m} \b S^m_{d^m} \overbrace{\left( \b U^m_{d^m} \right)^\T \b U^m_{d^m}}^{=\b I_{d^m}} \left( \b S^m_{d^m} \right)^{-\frac{1}{2}} = \b U^m_{d^m} \left( \b S^m_{d^m} \right)^{\frac{1}{2}}.
    \end{align}    
    (c) is obtained by adding $\pm \b U^m_{\b K^m,d^m} \left( \b U^m_{\b K^m,d^m} \right)^\top \b K^m \b U^m_{d^m} \left( \b S^m_{d^m} \right)^{-\frac{1}{2}} \in \R^{n \times d^m}$ and combining similar terms. Making the term $\b E_1$ of the final error bound appear, we focus on controlling $\left\|\b E_2\right\|_F$.
    
    \textbf{Term} $\left\|\b E_2\right\|_F$: To bound $\left\|\b E_2\right\|_F$, we define 
        \begin{align}
        \b R_4 &:=\b U^m_{d^m} - \b U^m_{\b K^m,d^m} \b W^m \in \R^{n \times d^m}, \label{eq:R4-def}
        \end{align}
        replace $\b U^m_{d^m}$ in $\b E_2$ (in the penultimate term) with $\b R_4 + \b U^m_{\b K^m,d^m} \b W^m$, and get that 
        \begin{align}
            \b E_2 & = \underbrace{\b U^m_{\b K^m,d^m} \left( \b U^m_{\b K^m,d^m} \right)^\top \left[\b U^m_{d^m} \left( \b U^m_{d^m} \right)^{\top} \left | \b A^m \right |  -  \b K^m \right]\b R_4 \left( \b S^m_{d^m} \right)^{-\frac{1}{2}}}_{=: \b F_1^m =: \b F_1 \in \R^{n \times d^m}}  \label{eq:F1-def} \\
            & \quad + \underbrace{\b U^m_{\b K^m,d^m} \left( \b U^m_{\b K^m,d^m} \right)^\top \left[\b U^m_{d^m} \left( \b U^m_{d^m} \right)^{\top} \left | \b A^m \right |  -  \b K^m \right] \b U^m_{\b K^m,d^m} \b W^m \left( \b S^m_{d^m} \right)^{-\frac{1}{2}}}_{=:\b F_2^m =: \b F_2 \in \R^{n \times d^m }}, \label{decomp_b2}
        \end{align}
        and bound $\left\|\b F_1\right\|$ (via $\left\|\b R_4\right\|$) and $\left\|\b F_2\right\|$.
        \begin{itemize}
            \item \textbf{Term} $\left\|\b R_4\right\|$: For any $\eta \in \left( 0,36\right) \cap (0,72) \cap (0,18) = (0,18)$ and 
            \begin{align}
                n \ge &\max \Bigg\{ \colAonei{ n^m_{0,(i)}\left( \frac{\eta}{432}\right)}, \colAoneii{n^m_{0,(ii)}\left( \frac{\eta}{432}\right)}, \colAii{ n_{1,(ii)}^m\left(\frac{\eta}{36}\right)}, \colAii{n^m_{1,(ii)}\left(\frac{\eta}{72}\right)}, \colLfive{\left( \frac{\eta}{432} \right)^{-\frac{1}{c^m}}} , \colLsix{\frac{1728}{\eta}}, \\
                & \hspace{1cm} \coletwo{e^{\frac{1}{\left( 2C_1^m - 4\sqrt{2} \right)^2}} } \Bigg\}
            \end{align}
            we have
            \begin{align}
                \left \| \b R_4 \right \|  &\stackrel{(a)}{=} \left \| \b U^m_{d^m} - \b U^m_{\b K^m,d^m} \b W^m \right \| \stackrel{(b)}{=} \left \| \b U^m_{d^m} - \b U^m_{\b K^m,d^m} \left( \b U^m_{\b K^m,d^m} \right)^\top \b U^m_{d^m} + \b R_1 \right \| \nonumber \\
                &\stackrel{(c)}{\le} \left \| \b U^m_{d^m} - \b U^m_{\b K^m,d^m} \left( \b U^m_{\b K^m,d^m} \right)^\top \b U^m_{d^m} \right \| + \left \| \b R_1 \right \| \nonumber \\
                &\stackrel{(d)}{\le} 16 c_1^m(n) \left( \delta^m_{d^m} \right)^{-1}\sqrt{\frac{ \log \left( \frac{432n}{\eta} \right)}{n}} + 256 \left( c_1^m(n) \right)^2 \left( \delta^m_{d^m} \right)^{-2} \frac{\log \left( \frac{432n}{\eta}\right)}{n} \label{bound_R4a}\\ 
                & \stackrel{(e)}{\le} 32 c_1^m(n) \left( \delta^m_{d^m} \right)^{-1}\sqrt{\frac{ \log \left( \frac{432n}{\eta} \right)}{n}},\label{bound_R4}
            \end{align}
            with $\J^m$-probability at least $1-\frac{\eta}{18}$, where restricting $\eta$ to $(0,18)$ guarantees that the probability is positive. (a) comes from the definition of $\b R_4$ [see \eqref{eq:R4-def}]. (b) follows from the definition of $\b R_1$ [see \eqref{eq:R1-def}]; (c) is implied by the triangle inequality.    
            The bound on the first term in (d) holds with $\J^m$-probability at least $1-\frac{\eta}{36}$ for any $\eta \in \left(0,36 \right)$ and 
            \begin{align}
                n &\ge \max \Bigg\{ \colAonei{ n^m_{0,(i)}\left( \frac{\eta}{432}\right)}, \colAoneii{n^m_{0,(ii)}\left( \frac{\eta}{432}\right)}, \colAii{ n_{1,(ii)}^m\left(\frac{\eta}{36}\right)}, \colAii{n^m_{1,(ii)}\left(\frac{\eta}{72}\right)},\colLfive{\left( \frac{\eta}{432} \right)^{-\frac{1}{c^m}}}, \colLsix{\frac{1728}{\eta}}, \\
                & \hspace{1.5 cm} \coletwo{e^{\frac{1}{\left( 2C_1^m - 4\sqrt{2} \right)^2}} } \Bigg\}
            \end{align}
            by using \eqref{bound_R} with a substitution $\frac{\eta}{18} \leftarrow \frac{\eta}{36}$; the bound on the 2nd term in (d) is satisfied for any $\eta \in \left( 0,36 \right)$ and 
            \begin{align}
                n & \ge \max \Bigg\{ \colAonei{ n^m_{0,(i)}\left( \frac{\eta}{432}\right)}, \colAoneii{n^m_{0,(ii)}\left( \frac{\eta}{432}\right)}, \colAii{ n_{1,(ii)}^m\left( \frac{\eta}{36} \right)}, \colAii{n^m_{1,(ii)}\left(\frac{\eta}{72}\right)}, \colLfive{\left( \frac{\eta}{432} \right)^{-\frac{1}{c^m}}} , \colLsix{\frac{1728}{\eta}}, \\ 
                & \hspace{1.5 cm} \coletwo{e^{\frac{1}{\left( 2C_1^m - 4\sqrt{2} \right)^2}} } \Bigg\},
            \end{align}
            with $\J^m$-probability at least $1-\frac{\eta}{36}$ by invoking \eqref{bound_Jm_R1} with a substitution $\frac{\eta}{3}\leftarrow \frac{\eta}{36}$.
            To obtain (e), we ensure that the first term is the dominant in \eqref{bound_R4a} by the following inequality
            \begin{align}
                \MoveEqLeft 16 c_1^m(n) \left( \delta^m_{d^m} \right)^{-1} \sqrt{\frac{ \log \left( \frac{432n}{\eta} \right)}{n}} \ge 256 \left( c_1^m(n) \right)^2 \left( \delta^m_{d^m} \right)^{-2} \frac{\log \left( \frac{432n}{\eta}\right)}{n} \Leftrightarrow \\
                & \delta^m_{d^m} \ge 16 c_1^m(n) \sqrt{\frac{ \log \left( \frac{432n}{\eta} \right)}{n}}, \label{eq:1st-term-dominates-condition}
            \end{align}
            The sufficient dominance condition \eqref{eq:1st-term-dominates-condition} is guaranteed by Assumption~\ref{assum_lambda_delta}(ii) applied with a substitution $\eta \leftarrow \frac{\eta}{72}$ for any $\eta \in \left( 0,72 \right)$ and $n\ge \colAii{n_{1,(ii)}^m\left(\frac{\eta}{72}\right)}$. The claimed statement holds by applying a union bound.
            \item \textbf{Term} $\left \| \b F_1 \right \|$: For $\left \| \b F_1 \right \|$, we have 
            \begin{align}%\label{bound_d1}
                \left \| \b F_1 \right \| & \stackrel{(a)}{=} \left\|\b U^m_{\b K^m,d^m} \left( \b U^m_{\b K^m,d^m} \right)^\top \left[\b U^m_{d^m} \left( \b U^m_{d^m} \right)^{\top} \left | \b A^m \right |  -  \b K^m \right]\b R_4 \left( \b S^m_{d^m} \right)^{-\frac{1}{2}}\right\|\\
                & \stackrel{(b)}{\le} \left \| \b U^m_{d^m} \left( \b U^m_{d^m} \right)^{\top} \left | \b A^m \right |  -  \b K^m \right \| \left \| \b R_4 \right \| \left \| \left( \b S^m_{d^m} \right)^{-\frac{1}{2}} \right \| \nonumber \\
                & \stackrel{(c)}{\le} \left[ 4 c_1^m(n) \sqrt{n \log \left( \frac{72 n}{\eta}\right)}\right] \left \| \b R_4 \right \| \left \| \left( \b S^m_{d^m} \right)^{-\frac{1}{2}} \right \| \\
                & \stackrel{(d)}{\le} \left[ 4 c_1^m(n) \sqrt{n \log \left( \frac{72 n}{\eta}\right)}\right] \left[ 32 c_1^m(n) \left( \delta^m_{d^m} \right)^{-1}\sqrt{\frac{ \log \left( \frac{432n}{\eta} \right)}{n}}\right] \left[ 2 \sqrt{2} (n \delta^m_{d^m})^{-\frac{1}{2}}\right] \nonumber\\
                & \stackrel{(e)}{\le} 256 \sqrt{2} \left( c_1^m(n) \right)^2 \left( \delta^m_{d^m} \right)^{-\frac{3}{2}}\frac{ \log \left( \frac{432n}{\eta} \right)}{\sqrt{n}} \label{bound_d1}
            \end{align}
            holds with $\J^m$-probability at least $1-\frac{\eta}{6}$ for any $\eta \in \left( 0,18\right) \cap \left(0,6 \right)= \left(0,6 \right)$, (restricting $\eta$ to $(0,6)$ guarantees that the probability is positive) and
            \begin{align}
                n \ge \max & \Bigg\{ \colAonei{ n^m_{0,(i)}\left( \frac{\eta}{72}\right)}, \colAonei{ n^m_{0,(i)}\left( \frac{\eta}{432}\right)}, \colAonei{ n^m_{0,(i)}\left( \frac{\eta}{576}\right)}, \colAonei{ n^m_{0,(i)}\left( \frac{\eta}{1152}\right)}, \colAoneii{n^m_{0,(ii)}\left( \frac{\eta}{72}\right)}, \colAoneii{n^m_{0,(ii)}\left( \frac{\eta}{432}\right)},  \\
                & \hspace{0.3cm} \colAoneii{ n^m_{0,(ii)}\left( \frac{\eta}{576}\right)},\colAi{n^m_{1,(i)}\left( \frac{\eta}{576} \right)}, \colAii{ n_{1,(ii)}^m\left(\frac{\eta}{36}\right)}, \colAii{n^m_{1,(ii)}\left(\frac{\eta}{72}\right)}, \colAii{ n_{1,(ii)}^m \left( \frac{\eta}{24} \right)}, \colAii{ n_{1,(ii)}^m \left( \frac{\eta}{96} \right)},  \\
                & \hspace{0.3cm} \colLfive{ \left( \frac{\eta}{72} \right)^{-\frac{1}{c^m}}}, \colLfive{\left( \frac{\eta}{432} \right)^{-\frac{1}{c^m}}}, \colLfive{\left( \frac{\eta}{1152} \right)^{-\frac{1}{c^m}}}, \colLsix{\frac{288}{\eta}},\colLsix{\frac{1728}{\eta}}, \colLsix{\frac{2304}{\eta}}, \\
                & \hspace{0.3cm} \coleone{32 \left(c_1^m(n) \right)^2 (c_0^m)^{-2} \log \left( \frac{1152 n}{\eta} \right)}, \coletwo{e^{\frac{1}{\left( 2C_1^m - 4\sqrt{2} \right)^2}} } \Biggl\} \\
                \stackrel{(f)}{=} \max & \Bigg\{ \colAonei{ n^m_{0,(i)}\left( \frac{\eta}{72}\right)}, \colAonei{ n^m_{0,(i)}\left( \frac{\eta}{432}\right)}, \colAonei{ n^m_{0,(i)}\left( \frac{\eta}{576}\right)},\colAonei{ n^m_{0,(i)}\left( \frac{\eta}{1152}\right)}, \colAoneii{n^m_{0,(ii)}\left( \frac{\eta}{72}\right)}, \colAoneii{n^m_{0,(ii)}\left( \frac{\eta}{432}\right)}, \\
                & \hspace{0.3cm} \colAoneii{ n^m_{0,(ii)}\left( \frac{\eta}{576}\right)},\colAi{n^m_{1,(i)}\left( \frac{\eta}{576} \right)}, \colAii{ n_{1,(ii)}^m\left(\frac{\eta}{36}\right)}, \colAii{n^m_{1,(ii)}\left(\frac{\eta}{72}\right)}, \colAii{ n_{1,(ii)}^m \left( \frac{\eta}{24} \right)}, \colAii{ n_{1,(ii)}^m \left( \frac{\eta}{96} \right)}, \\
                & \hspace{0.3cm} \colLfive{\left( \frac{\eta}{1152} \right)^{-\frac{1}{c^m}}},  \colLsix{\frac{2304}{\eta}}, \coleone{32 \left(c_1^m(n) \right)^2 (c_0^m)^{-2} \log \left( \frac{1152 n}{\eta} \right)}, \coletwo{e^{\frac{1}{\left( 2C_1^m - 4\sqrt{2} \right)^2}} } \Biggl\},
            \end{align}
            where (f) follows from \eqref{eq:x^a-monotonicity}. (a) in \eqref{bound_d1} was obtained by  the definition of $\b F_1$ [see \eqref{eq:F1-def}]. (b) holds by using that $\left\|\b A \b B\right\| \le \left\| \b A\right\| \left\| \b B\right\|$, $\left\| \b A \right\| = \left\| \b A^\T\right\| $ and $\left\|\b U^m_{\b K^m,d^m}\right\|\le 1$. (c) follows from          \begin{align}\label{bound_diff_psd2}
                \left \| \b U^m_{d^m} \left( \b U^m_{d^m} \right)^{\top} \left | \b A^m \right |  -  \b K^m \right \| & \stackrel{\eqref{bound_diff_psd1}}{\le} \left \| \left | \b A^m \right |  -  \b K^m \right \| \stackrel{(f)}{\le} 4 c_1^m(n) \sqrt{n \log \left( \frac{72 n}{\eta}\right)}
            \end{align}
            holding for any $\eta \in (0,18)$ with $\J^m$-probability at least $1-\frac{\eta}{18}$. Indeed, (f) holds with $\J^m$-probability at least $1-\frac{\eta}{18}$ by \eqref{bound_diff_psd} with a substitution $\frac{\eta}{3} \leftarrow \frac{\eta}{18}$ for any $\eta \in \left(0, 18 \right)$ and $n \ge \max \left\{ \colAonei{ n^m_{0,(i)}\left( \frac{\eta}{72}\right)}, \colAoneii{n^m_{0,(ii)}\left( \frac{\eta}{72}\right)}, \colLfive{ \left( \frac{\eta}{72} \right)^{-\frac{1}{c^m}}}, \colLsix{\frac{288}{\eta}}, \coletwo{e^{\frac{1}{\left( 2C_1^m - 4\sqrt{2} \right)^2}} } \right\}$.         
            In (d) the bound on $\left\|\b R_4\right\|$  holds with $\J^m$-probability at least $1-\frac{\eta}{18}$ for any $\eta \in \left( 0, 18 \right)$ and 
            \begin{align}
                n \ge &\max \Bigg\{ \colAonei{ n^m_{0,(i)}\left( \frac{\eta}{432}\right)}, \colAoneii{n^m_{0,(ii)}\left( \frac{\eta}{432}\right)}, \colAii{ n_{1,(ii)}^m\left(\frac{\eta}{36}\right)}, \colAii{n^m_{1,(ii)}\left(\frac{\eta}{72}\right)}, \colLfive{\left( \frac{\eta}{432} \right)^{-\frac{1}{c^m}}} , \colLsix{\frac{1728}{\eta}}, \\
                & \hspace{1cm} \coletwo{e^{\frac{1}{\left( 2C_1^m - 4\sqrt{2} \right)^2}} } \Bigg\}
            \end{align}
            by \eqref{bound_R4}. 
            In (d) the bound on $\left \| \left( \b S^m_{d^m} \right)^{-\frac{1}{2}} \right \|$ is satisfied with $\J^m$-probability at least $1-\frac{\eta}{18}$ for any $\eta \in \left(0, 18 \right)$ and  
            \begin{align}
                n \ge \max &\Bigg\{ \colAonei{ n^m_{0,(i)}\left( \frac{\eta}{1152}\right)}, \colAonei{ n^m_{0,(i)}\left( \frac{\eta}{576}\right)}, \colAoneii{n^m_{0,(ii)}\left( \frac{\eta}{576}\right)}, \colAi{n^m_{1,(i)}\left( \frac{\eta}{576} \right)}, \colAii{ n_{1,(ii)}^m \left( \frac{\eta}{24} \right)},  \\
                & \hspace{0.3cm} \colAii{ n_{1,(ii)}^m \left( \frac{\eta}{96} \right)}, \colLfive{\left( \frac{\eta}{1152} \right)^{-\frac{1}{c^m}}}, \colLsix{\frac{2304}{\eta}}, \coleone{32 \left(c_1^m(n)\right)^2 (c_0^m)^{-2} \log \left( \frac{1152n}{\eta} \right)},\\
                & \hspace{0.3cm} \coletwo{e^{\frac{1}{\left( 2C_1^m - 4\sqrt{2} \right)^2}}} \Bigg\}.
            \end{align}
            by using \eqref{eq:bound_eig_A} with a substitution $\eta \leftarrow \frac{\eta}{18}$. (e) was obtained by collecting terms and using that $\log \left( \frac{72n}{\eta} \right) \le \log \left( \frac{432n}{\eta} \right)$ by the monotonicity of the $\log$ function. Applying a union bound implies the claimed statement in \eqref{bound_d1}.
            \item \textbf{Term} $\left \| \b F_2 \right \|$: Considering $\left \| \b F_2 \right \|$ we have
            \begin{align}
                \left \| \b F_2 \right \| & \stackrel{(a)}{=}\left\| \b U^m_{\b K^m,d^m} \left( \b U^m_{\b K^m,d^m} \right)^\top \left[\b U^m_{d^m} \left( \b U^m_{d^m} \right)^{\top} \left | \b A^m \right |  -  \b K^m \right] \b U^m_{\b K^m,d^m} \b W^m \left( \b S^m_{d^m} \right)^{-\frac{1}{2}} \right\|\\
                &\stackrel{(b)}{\le}  \left \| \b U^m_{d^m} \left( \b U^m_{d^m} \right)^{\top} \left | \b A^m \right |  -  \b K^m \right \| \left \| \b W^m \right\| \left\|\left( \b S^m_{d^m} \right)^{-\frac{1}{2}} \right \|\\
                & \stackrel{(c)}{\le} \left[4 c_1^m(n) \sqrt{n \log \left ( \frac{48n}{\eta} \right)} \right] 1 \left[ 2 \sqrt{2} (n \delta^m_{d^m})^{-\frac{1}{2}}\right]\\
                & = 8 \sqrt{2} c_1^m(n)  \left( \delta^m_{d^m} \right)^{-\frac{1}{2}} \sqrt{\log \left ( \frac{48n}{\eta} \right ) }\label{bound_d2}
            \end{align}
            holds with $\J^m$-probability at least $1-\frac{\eta}{6}$ for any $\eta \in \left( 0,12 \right) \cap (0,6) = \left( 0,6 \right)$ (restricting $\eta$ to $(0,6)$ guarantees that the probability is positive) and 
            \begin{align}
                n \ge \max &\Bigg\{ \colAonei{ n^m_{0,(i)}\left( \frac{\eta}{48}\right)}, \colAonei{ n^m_{0,(i)}\left( \frac{\eta}{768}\right)}, \colAonei{ n^m_{0,(i)}\left( \frac{\eta}{384}\right)}, \colAoneii{n^m_{0,(ii)}\left( \frac{\eta}{48}\right)}, \colAoneii{n^m_{0,(ii)}\left( \frac{\eta}{384}\right)}, \\
                & \hspace{0.3cm} \colAi{n^m_{1,(i)}\left( \frac{\eta}{384} \right)}, \colAii{ n_{1,(ii)}^m \left( \frac{\eta}{16} \right)}, \colAii{ n_{1,(ii)}^m \left( \frac{\eta}{64} \right)}, \colLfive{ \left( \frac{\eta}{48} \right)^{-\frac{1}{c^m}}}, \colLfive{ \left( \frac{\eta}{768} \right)^{-\frac{1}{c^m}}}, \colLsix{\frac{192}{\eta}}, \colLsix{\frac{1536}{\eta}},  \\
                & \hspace{0.3cm} \coleone{32 \left(c_1^m(n)\right)^2 (c_0^m)^{-2} \log \left( \frac{768 n}{\eta} \right)},\coletwo{e^{\frac{1}{\left( 2C_1^m - 4\sqrt{2} \right)^2}} } \Bigg\} \\
                \stackrel{(d)}{=} \max &\Bigg\{ \colAonei{ n^m_{0,(i)}\left( \frac{\eta}{48}\right)}, \colAonei{ n^m_{0,(i)}\left( \frac{\eta}{768}\right)}, \colAonei{ n^m_{0,(i)}\left( \frac{\eta}{384}\right)}, \colAoneii{n^m_{0,(ii)}\left( \frac{\eta}{48}\right)}, \colAoneii{n^m_{0,(ii)}\left( \frac{\eta}{384}\right)}, \\
                & \hspace{0.3cm} \colAi{n^m_{1,(i)}\left( \frac{\eta}{384} \right)}, \colAii{ n_{1,(ii)}^m \left( \frac{\eta}{16} \right)}, \colAii{ n_{1,(ii)}^m \left( \frac{\eta}{64} \right)}, \colLfive{ \left( \frac{\eta}{768} \right)^{-\frac{1}{c^m}}}, \colLsix{\frac{1536}{\eta}},  \\
                & \hspace{0.3cm} \coleone{32 \left(c_1^m(n)\right)^2 (c_0^m)^{-2} \log \left( \frac{768 n}{\eta} \right)},\coletwo{e^{\frac{1}{\left( 2C_1^m - 4\sqrt{2} \right)^2}} } \Bigg\},
            \end{align}
            where (d) comes from \eqref{eq:x^a-monotonicity}.
            For (a) in \eqref{bound_d2}, the definition of $\b F_2$ [see \eqref{decomp_b2}] was used. (b) follows from $\left\| \b A \b B \right\| \le \left\| \b A \right\| \left\| \b B \right\|$, $\left\| \b A^\T \right\| = \left\| \b A \right\|$ and  $\left \| \b U^m_{\b K^m,d^m} \right \|\le 1$. In (c), the bound  on the 1st term holds with $\J^m$-probability at least $1-\frac{\eta}{12}$ for any $\eta \in \left( 0,12\right)$ and $n \ge \max \left\{ \colAonei{ n^m_{0,(i)}\left( \frac{\eta}{48}\right)}, \colAoneii{n^m_{0,(ii)}\left( \frac{\eta}{48}\right)}, \colLfive{ \left( \frac{\eta}{48} \right)^{-\frac{1}{c^m}}}, \colLsix{\frac{192}{\eta}}, \coletwo{e^{\frac{1}{\left( 2C_1^m - 4\sqrt{2} \right)^2}} } \right\}$ by using \eqref{bound_diff_psd2} with a substitution $\frac{\eta}{18}\leftarrow \frac{\eta}{12}$; considering the 2nd term, $\left \| \b W^m \right \| = 1$ since $\b W^m$ is orthogonal; the bound on the 3rd term holds with $\J^m$-probability at least $1-\frac{\eta}{12}$ for any $\eta \in \left( 0, 12 \right) $ and
            \begin{align}
                n \ge \max &\Bigg\{ \colAonei{ n^m_{0,(i)}\left( \frac{\eta}{768}\right)}, \colAonei{ n^m_{0,(i)}\left( \frac{\eta}{384}\right)}, \colAoneii{n^m_{0,(ii)}\left( \frac{\eta}{384}\right)}, \colAi{n^m_{1,(i)}\left( \frac{\eta}{384} \right)}, \colAii{ n_{1,(ii)}^m \left( \frac{\eta}{16} \right)},   \\
                & \hspace{0.3cm} \colAii{ n_{1,(ii)}^m \left( \frac{\eta}{64} \right)}, \colLfive{\left( \frac{\eta}{768} \right)^{-\frac{1}{c^m}}},\colLsix{\frac{1536}{\eta}}, \coleone{32 \left(c_1^m(n)\right)^2 (c_0^m)^{-2} \log \left( \frac{768n}{\eta} \right)},\coletwo{e^{\frac{1}{\left( 2C_1^m - 4\sqrt{2} \right)^2}}} \Bigg\}.
            \end{align}
             by applying \eqref{eq:bound_eig_A} with a substitution $\eta \leftarrow \frac{\eta}{12}$.
             Taking a union bound gives the statement in \eqref{bound_d2}.
        \end{itemize}
        Using the obtained results on $\left\|\b F_1\right\|$ and $\left\|\b F_2\right\|$, one can bound $\left\|\b E_2\right\|_F$: for any $\eta \in \left( 0,3 \right) $ and 
        \begin{align}
            n \ge \max & \Bigg\{ \colAonei{ n^m_{0,(i)}\left( \frac{\eta}{72}\right)}, \colAonei{ n^m_{0,(i)}\left( \frac{\eta}{432}\right)}, \colAonei{ n^m_{0,(i)}\left( \frac{\eta}{576}\right)}, \colAonei{ n^m_{0,(i)}\left( \frac{\eta}{1152}\right)}, \colAonei{ n^m_{0,(i)}\left( \frac{\eta}{48}\right)}, \colAonei{ n^m_{0,(i)}\left( \frac{\eta}{768}\right)},  \\
            & \hspace{0.3cm} \colAonei{ n^m_{0,(i)}\left( \frac{\eta}{384}\right)}, \colAoneii{n^m_{0,(ii)}\left( \frac{\eta}{72}\right)},\colAoneii{n^m_{0,(ii)}\left( \frac{\eta}{432}\right)}, \colAoneii{ n^m_{0,(ii)}\left( \frac{\eta}{576}\right)}, \colAoneii{n^m_{0,(ii)}\left( \frac{\eta}{48}\right)}, \colAoneii{n^m_{0,(ii)}\left( \frac{\eta}{384}\right)}, \\
            & \hspace{0.3cm} \colAi{n^m_{1,(i)}\left( \frac{\eta}{576} \right)}, \colAi{n^m_{1,(i)}\left( \frac{\eta}{384} \right)},  \colAii{ n_{1,(ii)}^m\left(\frac{\eta}{36}\right)}, \colAii{n^m_{1,(ii)}\left(\frac{\eta}{72}\right)}, \colAii{ n_{1,(ii)}^m \left( \frac{\eta}{24} \right)}, \colAii{ n_{1,(ii)}^m \left( \frac{\eta}{96} \right)}, \\
            & \hspace{0.3cm} \colAii{ n_{1,(ii)}^m \left( \frac{\eta}{16} \right)}, \colAii{ n_{1,(ii)}^m \left( \frac{\eta}{64} \right)}, \colLfive{\left( \frac{\eta}{1152} \right)^{-\frac{1}{c^m}}}, \colLsix{\frac{2304}{\eta}}, \coleone{32 \left(c_1^m(n) \right)^2 (c_0^m)^{-2} \log \left( \frac{1152 n}{\eta} \right)}, \\
            & \hspace{0.3cm} \coletwo{e^{\frac{1}{\left( 2C_1^m - 4\sqrt{2} \right)^2}} } \Biggl\},
        \end{align} 
        one gets that 
        \begin{align}
            \left \| \b E_2 \right \|_F&  \stackrel{(a)}{\le} \sqrt{d^m} \left \| \b E_2 \right \|  \stackrel{\eqref{decomp_b2}}{=} \sqrt{d^m} \left \| \b F_1 + \b F_2 \right \| \stackrel{(b)}{\le} \sqrt{d^m} \left(\left \| \b F_1 \right\| + \left\|\b F_2 \right \|\right)\\
            & \stackrel{(c)}{\le} 256 \sqrt{2} \left(c_1^m(n) \right)^2 \left(d^m \right)^{\frac{1}{2}} \left( \delta^m_{d^m} \right)^{-\frac{3}{2}}\frac{ \log \left( \frac{432n}{\eta} \right)}{\sqrt{n}}\\
            &\quad \hspace{0.1cm} + 8 \sqrt{2} c_1^m(n) \left(d^m \right)^{\frac{1}{2}} \left( \delta^m_{d^m} \right)^{-\frac{1}{2}} \sqrt{\log \left ( \frac{48n}{\eta} \right ) } \nonumber\\
            & \stackrel{(d)}{\le} 256 \sqrt{2} \left(c_1^m(n) \right)^2 \left(d^m \right)^{\frac{1}{2}} \left( \delta^m_{d^m} \right)^{-\frac{3}{2}}\frac{ \log \left( \frac{432n}{\eta} \right)}{\sqrt{n}}\\
            & \quad \hspace{0.1cm} + 8 \sqrt{2} c_1^m(n) \left(d^m \right)^{\frac{1}{2}} \left( \delta^m_{d^m} \right)^{-\frac{1}{2}} \sqrt{\log \left ( \frac{432n}{\eta} \right ) } \nonumber \\
            & \stackrel{(e)}{\le} 264 \sqrt{2}c_1^m(n) \left(d^m \right)^{\frac{1}{2}} \left( \delta^m_{d^m} \right)^{-\frac{1}{2}}  \sqrt{\log \left ( \frac{432 n}{\eta} \right ) } ,\label{bound_E2}
        \end{align}
        holds with $\J^m$-probability at least $1-\frac{\eta}{3}$. Indeed, (a) comes from $\left\|\b E_2\right\|_F \le \sqrt{\rank(\b E_2)} \left\|\b E_2\right\| \le \sqrt{d^m} \left\|\b E_2\right\| $ since $\rank(\b E_2)\le d^m$ by $\b E_2 \in \R^{n\times d^m}$. The triangle inequality was applied in (b). (c) holds with $\J^m$-probability at least $1-\frac{\eta}{3}$ any $\eta \in \left( 0,6 \right) \cap(0,3) = \left( 0,3 \right)$ (restricting $\eta$ to $(0,3)$ guarantees that the probability is positive) and
        \begin{align}
            n \stackrel{(f)}{\ge} \max & \Bigg\{ \colAonei{ n^m_{0,(i)}\left( \frac{\eta}{72}\right)}, \colAonei{ n^m_{0,(i)}\left( \frac{\eta}{432}\right)}, \colAonei{ n^m_{0,(i)}\left( \frac{\eta}{576}\right)}, \colAonei{ n^m_{0,(i)}\left( \frac{\eta}{1152}\right)}, \colAonei{ n^m_{0,(i)}\left( \frac{\eta}{48}\right)}, \colAonei{ n^m_{0,(i)}\left( \frac{\eta}{768}\right)},  \\
            & \hspace{0.3cm} \colAonei{ n^m_{0,(i)}\left( \frac{\eta}{384}\right)}, \colAoneii{n^m_{0,(ii)}\left( \frac{\eta}{72}\right)},\colAoneii{n^m_{0,(ii)}\left( \frac{\eta}{432}\right)}, \colAoneii{ n^m_{0,(ii)}\left( \frac{\eta}{576}\right)}, \colAoneii{n^m_{0,(ii)}\left( \frac{\eta}{48}\right)}, \colAoneii{n^m_{0,(ii)}\left( \frac{\eta}{384}\right)}, \\
            & \hspace{0.3cm} \colAi{n^m_{1,(i)}\left( \frac{\eta}{576} \right)}, \colAi{n^m_{1,(i)}\left( \frac{\eta}{384} \right)},  \colAii{ n_{1,(ii)}^m\left(\frac{\eta}{36}\right)}, \colAii{n^m_{1,(ii)}\left(\frac{\eta}{72}\right)}, \colAii{ n_{1,(ii)}^m \left( \frac{\eta}{24} \right)}, \colAii{ n_{1,(ii)}^m \left( \frac{\eta}{96} \right)}, \\
            & \hspace{0.3cm} \colAii{ n_{1,(ii)}^m \left( \frac{\eta}{16} \right)}, \colAii{ n_{1,(ii)}^m \left( \frac{\eta}{64} \right)}, \colLfive{\left( \frac{\eta}{1152} \right)^{-\frac{1}{c^m}}}, \colLfive{\left( \frac{\eta}{768} \right)^{-\frac{1}{c^m}}}, \colLsix{\frac{2304}{\eta}}, \colLsix{\frac{1536}{\eta}},  \\
            & \hspace{0.3cm} \coleone{32 \left(c_1^m(n) \right)^2 (c_0^m)^{-2} \log \left( \frac{1152 n}{\eta} \right)}, \coleone{32 \left(c_1^m(n) \right)^2 (c_0^m)^{-2} \log \left( \frac{768 n}{\eta} \right)}, \coletwo{e^{\frac{1}{\left( 2C_1^m - 4\sqrt{2} \right)^2}} } \Biggl\}
            \\
            \stackrel{(g)}{=} \max & \Bigg\{ \colAonei{ n^m_{0,(i)}\left( \frac{\eta}{72}\right)}, \colAonei{ n^m_{0,(i)}\left( \frac{\eta}{432}\right)}, \colAonei{ n^m_{0,(i)}\left( \frac{\eta}{576}\right)}, \colAonei{ n^m_{0,(i)}\left( \frac{\eta}{1152}\right)}, \colAonei{ n^m_{0,(i)}\left( \frac{\eta}{48}\right)}, \colAonei{ n^m_{0,(i)}\left( \frac{\eta}{768}\right)},  \\
            & \hspace{0.3cm} \colAonei{ n^m_{0,(i)}\left( \frac{\eta}{384}\right)}, \colAoneii{n^m_{0,(ii)}\left( \frac{\eta}{72}\right)},\colAoneii{n^m_{0,(ii)}\left( \frac{\eta}{432}\right)}, \colAoneii{ n^m_{0,(ii)}\left( \frac{\eta}{576}\right)}, \colAoneii{n^m_{0,(ii)}\left( \frac{\eta}{48}\right)}, \colAoneii{n^m_{0,(ii)}\left( \frac{\eta}{384}\right)}, \\
            & \hspace{0.3cm} \colAi{n^m_{1,(i)}\left( \frac{\eta}{576} \right)}, \colAi{n^m_{1,(i)}\left( \frac{\eta}{384} \right)},  \colAii{ n_{1,(ii)}^m\left(\frac{\eta}{36}\right)}, \colAii{n^m_{1,(ii)}\left(\frac{\eta}{72}\right)}, \colAii{ n_{1,(ii)}^m \left( \frac{\eta}{24} \right)}, \colAii{ n_{1,(ii)}^m \left( \frac{\eta}{96} \right)}, \\
            & \hspace{0.3cm} \colAii{ n_{1,(ii)}^m \left( \frac{\eta}{16} \right)}, \colAii{n_{1,(ii)}^m \left( \frac{\eta}{64} \right)}, \colLfive{\left( \frac{\eta}{1152} \right)^{-\frac{1}{c^m}}}, \colLsix{\frac{2304}{\eta}}, \coleone{32 \left(c_1^m(n) \right)^2 (c_0^m)^{-2} \log \left( \frac{1152 n}{\eta} \right)}, \\
            & \hspace{0.3cm} \coletwo{e^{\frac{1}{\left( 2C_1^m - 4\sqrt{2} \right)^2}} } \Biggl\},
        \end{align}
        (f) is implied by using \eqref{bound_d1} and \eqref{bound_d2},  (g) follows from \eqref{eq:x^a-monotonicity} and the fact that $\log \left( \frac{1152 n}{\eta} \right) > \log \left( \frac{768 n}{\eta} \right)$. (d) in \eqref{bound_E2} follows from the fact that $\log \left( \frac{48n}{\eta} \right) < \log \left( \frac{432n}{\eta} \right) $. Considering (e) in \eqref{bound_E2}, one can notice that 
        \begin{align}
            c_1^m(n) \left( \delta^m_{d^m} \right)^{-\frac{1}{2}} \sqrt{\log \left ( \frac{432 n}{\eta} \right )} \ge \left(c_1^m(n) \right)^2 \left( \delta^m_{d^m} \right)^{-\frac{3}{2}} \frac{\log \left ( \frac{432 n}{\eta} \right )}{\sqrt{n}} \Leftrightarrow \delta^m_{d^m} \stackrel{(h)}{\ge} c_1^m(n) \sqrt{\frac{\log \left ( \frac{432 n}{\eta} \right )}{n}},
        \end{align}
        where (h) can be guaranteed by ensuring (i) in 
        \begin{align}
        c_1^m(n) \sqrt{\frac{\log \left ( \frac{432 n}{\eta} \right )}{n}} \stackrel{(i)}{\le} 16 c_1^m(n) \sqrt{\frac{\log \left ( \frac{6 n}{\eta} \right )}{n}} \stackrel{(j)}{\le} \delta^m_{d^m} 
        \end{align}
        as (j) holds for any $\eta \in \left( 0,1 \right)$ and $n \ge \colAii{n_{1,(ii)}^m (\eta)}$ by Assumption~\ref{assum_lambda_delta}(ii). To ensure (i) it is sufficient to have
        \begin{align}
        \log\Bigg(\frac{\overbrace{432}^{72\times 6}n}{\eta}\Bigg) \le 256 \log\left(\frac{6n}{\eta}\right) &\Leftrightarrow \log(72) +\log\left(\frac{6n}{\eta}\right) \le 256 \log\left(\frac{6n}{\eta}\right)\\
        &\Leftrightarrow \frac{\log(72)}{255} \le \log\left(\frac{6n}{\eta}\right).
        \end{align}
        This condition is satisfied as $\log \left( \frac{6n}{\eta} \right) \stackrel{(k)}{>} \log (6n) \stackrel{(l)}{\ge} \log (6) \ge \frac{\log (72)}{255}$, where (k) holds as $\eta \in (0,1)$ and the $\log$ function is strictly increasing; (l) holds by $n \ge 1$ combined with the increasing property of the $\log$ function.
    \item \textbf{Term} $\left \| \b A_2 \right \|$: For $\left \| \b A_2 \right \|$, we have 
    \begin{align}
        \left \| \b A_2 \right \| & \stackrel{(a)}{=} \left \| \b R_1 \left( \b S^m_{d^m} \right)^{\frac{1}{2}}  \right \|
        \stackrel{(b)}{\le} \left \| \b R_1 \right \| \left \| \left( \b S^m_{d^m} \right)^{\frac{1}{2}} \right \| \stackrel{(c)}{\le} 256 \left(c_1^m(n) \right)^2 \left( \delta^m_{d^m} \right)^{-2} \frac{\log \left(  \frac{72n}{\eta}\right)}{n} \sqrt{2n} \\
        & \stackrel{(d)}{=}   256 \sqrt{2} \left(c_1^m(n) \right)^2 \left( \delta^m_{d^m} \right)^{-2} \frac{\log \left( \frac{72n}{\eta} \right)}{\sqrt{n}}, \label{bound_A2}
    \end{align}
    holds with $\J^m$-probability at least $1-\frac{\eta}{3}$ for any $\eta \in \left(0, 6 \right) \cap(0,3)=(0,3)$ (restricting $\eta$ to $(0,3)$ guarantees that the probability is positive) and 
    \begin{align}
        n & \ge \max \Bigg\{ \colAonei{ n^m_{0,(i)}\left( \frac{\eta}{72}\right)}, \colAonei{ n^m_{0,(i)}\left( \frac{\eta}{24}\right)}, \colAoneii{n^m_{0,(ii)}\left( \frac{\eta}{72}\right)}, \colAoneii{n^m_{0,(ii)}\left( \frac{\eta}{24}\right)}, \colAii{ n_{1,(ii)}^m\left( \frac{\eta}{6} \right)}, \colAii{n^m_{1,(ii)}\left(\frac{\eta}{12}\right)},  \\
        & \quad \hspace{1.3cm} \colLfive{\left( \frac{\eta}{72} \right)^{-\frac{1}{c^m}}} , \colLfive{\left( \frac{\eta}{24} \right)^{-\frac{1}{c^m}}},\colLsix{\frac{288}{\eta}}, \colLsix{\frac{96}{\eta}}, \coletwo{e^{\frac{1}{\left( 2C_1^m - 4\sqrt{2} \right)^2}} }, 16 \left(c_1^m(n) \right)^2 \log \left( \frac{24 n}{\eta} \right) \Bigg\} \\
        & \stackrel{(e)}{=} \max \Bigg\{ \colAonei{ n^m_{0,(i)}\left( \frac{\eta}{72}\right)}, \colAonei{ n^m_{0,(i)}\left( \frac{\eta}{24}\right)},\colAoneii{n^m_{0,(ii)}\left( \frac{\eta}{72}\right)}, \colAoneii{n^m_{0,(ii)}\left( \frac{\eta}{24}\right)}, \colAii{ n_{1,(ii)}^m\left( \frac{\eta}{6} \right)}, \colAii{n^m_{1,(ii)}\left(\frac{\eta}{12}\right)}, \\
        & \quad \hspace{1.3cm} \colLfive{\left( \frac{\eta}{72} \right)^{-\frac{1}{c^m}}} , \colLsix{\frac{288}{\eta}}, \coletwo{e^{\frac{1}{\left( 2C_1^m - 4\sqrt{2} \right)^2}} }, 16 \left(c_1^m(n) \right)^2 \log \left( \frac{24 n}{\eta} \right) \Bigg\},
    \end{align}
    where (e) follows from \eqref{eq:x^a-monotonicity}.
    Indeed, (a) in \eqref{bound_A2} follows from the definition of $\b A_2$ [see \eqref{eq:initial_decompose}].
    (b) in \eqref{bound_A2} comes from $\left\| \b A \b B \right\| \le \left\| \b A \right\| \left\| \b B \right\|$. (c) in \eqref{bound_A2} holds with $\J^m$-probability at least $1-\frac{\eta}{6}$ for any $\eta \in \left(0, 6 \right)$ and 
    \begin{align}
         n \ge \max \left\{ \colAonei{ n^m_{0,(i)}\left( \frac{\eta}{72}\right)}, \colAoneii{n^m_{0,(ii)}\left( \frac{\eta}{72}\right)}, \colAii{ n_{1,(ii)}^m\left( \frac{\eta}{6} \right)}, \colAii{n^m_{1,(ii)}\left(\frac{\eta}{12}\right)}, \colLfive{\left( \frac{\eta}{72} \right)^{-\frac{1}{c^m}}} , \colLsix{\frac{288}{\eta}}, \coletwo{e^{\frac{1}{\left( 2C_1^m - 4\sqrt{2} \right)^2}} } \right\} 
    \end{align}
    by using \eqref{bound_Jm_R1} with a substitution of $\frac{\eta}{3} \leftarrow \frac{\eta}{6}$ and that $\left \| \left( \b S^m_{d^m} \right)^{\frac{1}{2}} \right \|  \le \sqrt{2n}$ holds with $\J^m$-probability at least $1-\frac{\eta}{6}$ for any $\eta \in \left(0, 6 \right)$ and 
    \begin{align}
        n \ge \max\left\{\colAonei{n^m_{0,(i)}\left( \frac{\eta}{24}\right)}, \colAoneii{n^m_{0,(ii)}\left( \frac{\eta}{24}\right)}, \colLfive{\left( \frac{\eta}{24} \right)^{-\frac{1}{c^m}}}, \colLsix{\frac{96}{\eta}}, \coletwo{e^{\frac{1}{\left( 2C_1^m - 4\sqrt{2} \right)^2}} }, 16 \left( c_1^m(n) \right)^2 \log \left( \frac{24n}{\eta} \right) \right\}.
    \end{align}
    Indeed, the 2nd property can be proved as follows. With a substitution of $\eta \leftarrow 6\eta$ in \eqref{eq:S_d^m_J^m_prob} we have for any $\eta \in (0,12]$ and $n \ge \max \left\{ \colAonei{n^m_{0,(i)}\left( \frac{\eta}{24}\right)}, \colLfive{ \left( \frac{\eta}{24} \right)^{-\frac{1}{c^m}}} \right\}$ that 
    \begin{align}
        \left \| \b S^m_{d^m} \right \| \stackrel{(e)}{\le} \Delta^m + 4 C^m \sqrt{\Delta^m \log \left( \frac{n}{\eta / 24} \right)}\label{eq:prob_event4}
    \end{align}
    holds with $\J^m$-probability at least $1-\frac{\eta}{12}$.
    By the definition of $\Delta^m$ we have $\Delta^m \le n$ for any realization of $\left( X_i^m \right)_{i=1}^n = \left( x_i^m \right)_{i=1}^n$. Therefore, we have that for any $\eta \in \left(0, 12 \right]$ and $n \ge \max \left\{ \colAonei{n^m_{0,(i)}\left( \frac{\eta}{24}\right)}, \colLfive{\left( \frac{\eta}{24} \right)^{-\frac{1}{c^m}}} \right\}$    
        \begin{align}
            \left \| \b S^m_{d^m} \right \| \le n + 4C^m \sqrt{n \log \left( \frac{24n}{\eta } \right)} 
        \end{align}
        holds with $\J^m$-probability at least $1-\frac{\eta}{12}$. According to Lemma~\ref{lemma:bound_Cm}, by a substitution of $\eta \leftarrow \frac{\eta}{12}$, we have that $C^m \le c_1^m(n)$ holds with $\J^m$-probability at least $1-\frac{\eta}{12}$ for any $\eta \in (0,12)$ and $n \ge \max \left\{ \colAoneii{n^m_{0,(ii)}\left( \frac{\eta}{24}\right)}, \colLsix{\frac{96}{\eta}}, \coletwo{e^{\frac{1}{\left( 2C_1^m - 4\sqrt{2} \right)^2}} } \right\}$. Therefore, one has that
        \begin{align}\label{eq:bound_adj_eig0}
            \left \| \b S^m_{d^m} \right \| \le n + 4c_1^m(n) \sqrt{n \log \left( \frac{24n}{\eta } \right)} \stackrel{(a)}{\le} 2n 
        \end{align}
        holds with $\J^m$-probability at least $1-\frac{\eta}{6}$ for any $\eta \in \left( 0,12 \right] \cap (0,12) \cap (0,6) = \left( 0,6 \right)$ and $n \ge \max\left\{\colAonei{n^m_{0,(i)}\left( \frac{\eta}{24}\right)}, \colAoneii{n^m_{0,(ii)}\left( \frac{\eta}{24}\right)}, \colLfive{\left( \frac{\eta}{24} \right)^{-\frac{1}{c^m}}}, \colLsix{\frac{96}{\eta}}, \coletwo{e^{\frac{1}{\left( 2C_1^m - 4\sqrt{2} \right)^2}} } \right\}$, where the restriction of $\eta$ to $(0,6)$ guarantees that the probability is positive; (a) holds when $4c_1^m(n) \sqrt{n \log \left( \frac{24n}{\eta } \right)} \le n$, in other words when $n \ge 16 \left( c_1^m(n) \right)^2 \log \left( \frac{24n}{\eta} \right)$.

    Thus, for any $\eta \in (0,6)$ and 
    \begin{align}
        n\ge \max\left\{\colAonei{n^m_{0,(i)}\left( \frac{\eta}{24}\right)}, \colAoneii{n^m_{0,(ii)}\left( \frac{\eta}{24}\right)}, \colLfive{\left( \frac{\eta}{24} \right)^{-\frac{1}{c^m}}}, \colLsix{\frac{96}{\eta}}, \coletwo{e^{\frac{1}{\left( 2C_1^m - 4\sqrt{2} \right)^2}} }, 16 \left( c_1^m(n) \right)^2 \log \left( \frac{24n}{\eta} \right)\right\}
    \end{align}
    we have that
    \begin{align}\label{eq:bound_adj_eig}
        \left \| \left( \b S^m_{d^m} \right)^{\frac{1}{2}} \right \| \stackrel{(g)}{=} \left[ \lambda_1^m \left( \left| \b A^m \right| \right) \right]^{\frac{1}{2}} \stackrel{(h)}{=} \left\| \b S^m_{d^m} \right\|^{\frac{1}{2}} \stackrel{(i)}{\le} \sqrt{2n}
    \end{align}
    holds with $\J^m$-probability at least $1-\frac{\eta}{6}$.
    (g) is implied by $\left\| \diag (\b v) \right\| = \max_{i \in [d^m]} \left| v_i \right|$ where $\b v = \left( \left[ \lambda_i^m \left( \left| \b A^m \right| \right) \right]^{\frac{1}{2}} \right)_{i=1}^{d^m}$. (h) follows from $\left\| \b S^m_{d^m} \right\| = \lambda_1^m \left( \left| \b A^m \right| \right)$ and (i) comes from \eqref{eq:bound_adj_eig0}. 
      
    \item \textbf{Term} $\left \| \b A_3 \right \|_F$: For $\left \| \b A_3 \right \|_F$, we have that for any $\eta \in \left( 0,3 \right)$ and 
    \begin{align}
        n & \ge \max \Bigg\{ \colAonei{ n^m_{0,(i)}\left( \frac{\eta}{72}\right)}, \colAonei{ n^m_{0,(i)}\left( \frac{\eta}{144}\right)}, \colAonei{ n^m_{0,(i)}\left( \frac{\eta}{216}\right)}, \colAoneii{n^m_{0,(ii)}\left( \frac{\eta}{72}\right)}, \colAoneii{n^m_{0,(ii)}\left( \frac{\eta}{216}\right)}, \colAoneii{n^m_{0,(ii)}\left( \frac{\eta}{36}\right)},\\
        & \hspace{1.5 cm} \colAi{n_{1,(i)}^m\left(\frac{\eta}{72}\right)}, \colAi{n^m_{1,(i)}\left( \frac{\eta}{36} \right)}, \colAii{n_{1,(ii)}^m \left(\frac{\eta}{18}\right)}, \colAii{n^m_{1,(ii)}\left(\frac{\eta}{36}\right)}, \colLfive{\left( \frac{\eta}{216} \right)^{-\frac{1}{c^m}}}, \colLsix{\frac{864}{\eta}}, \coletwo{e^{\frac{1}{\left( 2C_1^m - 4\sqrt{2} \right)^2}} } \Bigg\},
    \end{align}
    the following inequality
    \begin{align}
         \MoveEqLeft \left \| \b A_3 \right \|_F \stackrel{(a)}{=} \left \| \b U^m_{\b K^m,d^m} \b R_2 \right \|_F \stackrel{(b)}\le \left \| \b U^m_{\b K^m,d^m} \right \|  \left \| \b R_2 \right \|_F \stackrel{(c)}{\le} \left \| \b R_2 \right \|_F  \\
        & \hspace{-0.5cm}\stackrel{(d)}{\le} 512 \frac{\left(d^m \right)^{\frac{1}{2}} \left(c_1^m(n)\right)^2}{\sqrt{c_0^m}} \left( \delta^m_{d^m} \right)^{-2} \frac{\log \left( \frac{216n}{\eta} \right)}{\sqrt{n}} + 1024 \frac{\left(d^m \right)^{\frac{1}{2}} \left(c_1^m(n)\right)^3}{\sqrt{c_0^m}} \left( \delta^m_{d^m} \right)^{-2} \frac{\log^{\frac{3}{2}} \left( \frac{216n}{\eta} \right)}{n} \\
        & \hspace{-0.5cm} \quad + 64 \frac{\left(c_1^m(n)\right)^2}{\sqrt{c_0^m}} \left( \delta^m_{d^m} \right)^{-1} \frac{\left(d^m \right)^{\frac{1}{2}} \log \left( \frac{216n}{\eta} \right)}{\sqrt{n}} + 4 \frac{(d^m)^{\frac{1}{2}} c_1^m(n)}{\sqrt{c_0^m}} \sqrt{\log \left( \frac{72n}{\eta}  \right)} \label{bound_A3}
    \end{align}
    holds with $\J^m$-probability at least $1-\frac{\eta}{3}$. Indeed, in (a) the definition of $\b A_3$ [see \eqref{eq:initial_decompose}] was used; (b) comes from $\left\|\b A \b B\right\| \le \left\|\b A\right\| \left\|\b B\right\|$. (c) follows from $ \left \| \b U^m_{\b K^m,d^m}\right \| \le 1$ and (d) is implied by \eqref{bound_R2} with $\J^m$-probability at least $1-\frac{\eta}{3}$ for any $\eta \in \left( 0,3 \right)$ and
    \begin{align}
        n & \ge \max \Bigg\{ \colAonei{ n^m_{0,(i)}\left( \frac{\eta}{72}\right)}, \colAonei{ n^m_{0,(i)}\left( \frac{\eta}{144}\right)}, \colAonei{ n^m_{0,(i)}\left( \frac{\eta}{216}\right)}, \colAoneii{n^m_{0,(ii)}\left( \frac{\eta}{72}\right)}, \colAoneii{n^m_{0,(ii)}\left( \frac{\eta}{216}\right)}, \colAoneii{n^m_{0,(ii)}\left( \frac{\eta}{36}\right)},\\
        & \hspace{1.5 cm} \colAi{n_{1,(i)}^m\left(\frac{\eta}{72}\right)}, \colAi{n^m_{1,(i)}\left( \frac{\eta}{36} \right)}, \colAii{n_{1,(ii)}^m \left(\frac{\eta}{18}\right)}, \colAii{n^m_{1,(ii)}\left(\frac{\eta}{36}\right)}, \colLfive{\left( \frac{\eta}{216} \right)^{-\frac{1}{c^m}}}, \colLsix{\frac{864}{\eta}}, \coletwo{e^{\frac{1}{\left( 2C_1^m - 4\sqrt{2} \right)^2}} } \Bigg\}.
    \end{align}
    
\end{itemize}
By using the notation 
\begin{align}
    \colAonei{\tilde{n}^m_{0,A1(i)}(\eta)} &:= \max \Bigg\{ n^m_{0,(i)}\left( \frac{\eta}{72}\right), n^m_{0,(i)}\left( \frac{\eta}{432}\right), n^m_{0,(i)}\left( \frac{\eta}{576}\right), n^m_{0,(i)}\left( \frac{\eta}{1152}\right), n^m_{0,(i)}\left( \frac{\eta}{48}\right),   \\
    & \hspace{1.7cm} n^m_{0,(i)}\left( \frac{\eta}{768}\right), n^m_{0,(i)}\left( \frac{\eta}{384}\right), n^m_{0,(i)}\left( \frac{\eta}{24}\right), n^m_{0,(i)}\left( \frac{\eta}{144}\right), n^m_{0,(i)}\left( \frac{\eta}{216}\right) \Bigg\}, \\
    & \hspace{1.7cm} \stackrel{(a)}{=} n^m_{0,(i)}\left( \frac{\eta}{1152}\right), \\
    \colAoneii{\tilde{n}^m_{0,A1(ii)}(\eta)} &:= \max \Bigg\{ n^m_{0,(ii)}\left( \frac{\eta}{72}\right), n^m_{0,(ii)}\left( \frac{\eta}{432}\right), n^m_{0,(ii)}\left( \frac{\eta}{576}\right), n^m_{0,(ii)}\left( \frac{\eta}{48}\right), n^m_{0,(ii)}\left( \frac{\eta}{384}\right), \\
    & \hspace{1.7cm} n^m_{0,(ii)}\left( \frac{\eta}{24}\right), n^m_{0,(ii)}\left( \frac{\eta}{216}\right), n^m_{0,(ii)}\left( \frac{\eta}{36}\right) \Bigg\} \stackrel{(a)}{=} n^m_{0,(ii)}\left( \frac{\eta}{576}\right), \\
    \colAi{\tilde{n}^m_{0,A2(i)}(\eta)} &:= \max \left\{ n^m_{1,(i)}\left( \frac{\eta}{576} \right), n^m_{1,(i)}\left( \frac{\eta}{384} \right), n_{1,(i)}^m\left(\frac{\eta}{72}\right), n^m_{1,(i)}\left( \frac{\eta}{36} \right) \right\} \stackrel{(a)}{=} n^m_{1,(i)}\left( \frac{\eta}{576} \right), \\
    \colAii{\tilde{n}^m_{0,A2(ii)}(\eta)}
    &:=\max \Bigg\{ n_{1,(ii)}^m\left(\frac{\eta}{36}\right), n_{1,(ii)}^m\left(\frac{\eta}{72}\right), n_{1,(ii)}^m \left( \frac{\eta}{24} \right), n_{1,(ii)}^m \left( \frac{\eta}{96} \right), n_{1,(ii)}^m \left( \frac{\eta}{16} \right), \\
    & \hspace{1.7cm} n_{1,(ii)}^m \left( \frac{\eta}{64} \right), n_{1,(ii)}^m\left( \frac{\eta}{6} \right), n^m_{1,(ii)}\left(\frac{\eta}{12}\right), n_{1,(ii)}^m \left(\frac{\eta}{18}\right) \Bigg\} \stackrel{(b)}{=} n_{1,(ii)}^m \left( \frac{\eta}{96} \right),\\
    \colLfive{\tilde{n}^m_{0,L5}(\eta)} &:= \max \left\{\left( \frac{\eta}{1152} \right)^{-\frac{1}{c^m}},\left( \frac{\eta}{72} \right)^{-\frac{1}{c^m}},\left( \frac{\eta}{216} \right)^{-\frac{1}{c^m}}\right\} \stackrel{\eqref{eq:x^a-monotonicity}}{=} \left( \frac{\eta}{1152} \right)^{-\frac{1}{c^m}}, \\
    \colLsix{\tilde{n}^m_{0,L6}(\eta)} &:= \max \left\{ \frac{2304}{\eta}, \frac{288}{\eta}, \frac{864}{\eta} \right\}= \frac{2304}{\eta}, \\
    \coleone{\tilde{n}_{0,\text{else}-1}^m(\eta)} & := \max\left\{32 \left(c_1^m(n)\right)^2 (c_0^m)^{-2} \log \left( \frac{1152 n}{\eta} \right), 16 \left(c_1^m(n) \right)^2 \log \left( \frac{24n}{\eta} \right) \right\}, \\
    &\hspace{0.18cm} \stackrel{(c)}{=} 32 \left(c_1^m(n)\right)^2 (c_0^m)^{-2} \log \left( \frac{1152 n}{\eta} \right), \\
    \coletwo{\tilde{n}_{0,\text{else}-2}^m} &:= e^{\frac{1}{\left( 2C_1^m - 4\sqrt{2} \right)^2}}, \label{eq:simplify_constriants}
\end{align}
where (a) holds as the constraints $n_{0,(i)}^m(\eta)$, $n_{0,(ii)}^m(\eta)$, and $n_{1,(i)}^m(\eta)$ ensure events to hold with probability at least $1-\eta$; hence, smaller $\eta$ imposes a stronger requirement and yields the most restrictive constraint. (b) follows from the fact that smaller $\eta$ gives the larger lower bound in Assumption~\ref{assum_lambda_delta}(ii) on $\delta^m_{d^m}$ to satisfy and hence stronger constraint to hold. (c) is implied by the inequality $16 \left(c_1^m(n)\right)^2 \log \left( \frac{24n}{\eta} \right) \le 32 \left(c_1^m(n)\right)^2 (c_0^m)^{-2} \log \left( \frac{1152 n}{\eta} \right)$ holding by $16 < 32$, $c_0^m \le 1$\footnote{By $0\stackrel{\ref{footnote:Km-has-positive-eigenvalues}}{<}\lambda_{d^m}^m (\b K^m) \le \lambda_1^m (\b K^m) \stackrel{\eqref{eig_delta_bound}}{\le} \Delta^m$, one has that $\lambda_{d^m}^m (\b K^m)/\Delta^m \le 1$. By Assumption~\ref{assum_lambda_delta}(i), we consider the event where $\lambda_{d^m}^m (\b K^m) / \Delta^m \ge c_0^m $ holds; hence $c_0^m \le 1$ holds.} Then by the results obtained above, we have
\begin{align}
     \MoveEqLeft\left\| \bm{\hat{\Phi}}^m_{d^m} - \bm{\Phi}^m_{d^m}\right\|_F \stackrel{\eqref{eq:initial_decompose-line1}}{=} \left\| \hat{\b \Phi}^m_{d^m} - \b U^m_{\b K^m,d^m} \left(\b S^m_{\b K^m,d^m} \right)^{\frac{1}{2}} \b W^m \right\|_F \stackrel{\eqref{eq:initial_decompose}}{=} \big\|\hspace{-0.25cm}\underbrace{\b A_1}_{\stackrel{\eqref{eq:A_1-decomposition}}{=}\b E_1-\b E_2}\hspace{-0.35cm} + \b A_2 + \b A_3\big\|_F \\
     & \hspace{-0.6cm}  \stackrel{(a)}{\le}  \left \| \b E_1 \right \|_F + \left \| \b E_2 \right \|_F + \left \| \b A_2 \right \|_F + \left \| \b A_3 \right \|_F \stackrel{(b)}{\le} \left \| \b E_1 \right \|_F + \left \| \b E_2 \right \|_F + \sqrt{d^m} \left \| \b A_2 \right \| + \left \| \b A_3 \right \|_F \\
     & \hspace{-0.6cm} \stackrel{(c)}{\le}  \left \| \left[ \b U^m_{d^m} \left( \b U^m_{d^m} \right)^{\top} \left | \b A^m \right |  - \b U^m_{\b K^m,d^m} \left( \b U^m_{\b K^m,d^m} \right)^\top \b K^m \right] \b U^m_{d^m} \left( \b S^m_{d^m} \right)^{-\frac{1}{2}} \right \|_F \\
     & \hspace{-0.6cm} \quad + 264 \sqrt{2} c_1^m(n) \left(d^m\right)^{\frac{1}{2}} \left( \delta^m_{d^m} \right)^{-\frac{1}{2}}  \sqrt{\log \left ( \frac{432 n}{\eta} \right ) } + 256 \sqrt{2} \left(d^m\right)^{\frac{1}{2}} \left(c_1^m(n)\right)^2 \left( \delta^m_{d^m} \right)^{-2} \frac{\log \left( \frac{72n}{\eta} \right)}{\sqrt{n}}\\
     & \hspace{-0.6cm}  \quad + 512 \frac{\left(d^m \right)^{\frac{1}{2}} \left(c_1^m(n)\right)^2}{\sqrt{c_0^m}} \left( \delta^m_{d^m} \right)^{-2} \frac{\log \left( \frac{216n}{\eta} \right)}{\sqrt{n}} + 1024 \frac{\left(d^m \right)^{\frac{1}{2}}\left(c_1^m(n)\right)^3}{\sqrt{c_0^m}} \left( \delta^m_{d^m} \right)^{-2} \frac{\log^{\frac{3}{2}} \left( \frac{216n}{\eta} \right)}{n} \\
     & \hspace{-0.6cm}  \quad + 64 \frac{\left(c_1^m(n)\right)^2}{\sqrt{c_0^m}} \left( \delta^m_{d^m} \right)^{-1} \frac{\left(d^m \right)^{\frac{1}{2}} \log \left( \frac{216n}{\eta} \right)}{\sqrt{n}} + 4 \frac{(d^m)^{\frac{1}{2}} c_1^m(n)}{\sqrt{c_0^m}} \sqrt{\log \left( \frac{72n}{\eta}  \right)}  \\
     &\hspace{-0.6cm}  \stackrel{(d)}{\le} \left \| \left[\b U^m_{d^m} \left( \b U^m_{d^m} \right)^{\top} \left | \b A^m \right |  - \b U^m_{\b K^m,d^m} \left( \b U^m_{\b K^m,d^m} \right)^\top \b K^m \right]\b U^m_{d^m} \left( \b S^m_{d^m} \right)^{-\frac{1}{2}} \right \|_F \\
     & \hspace{-0.6cm}  \quad + 264 \sqrt{2} c_1^m(n) d^m \left( \delta^m_{d^m} \right)^{-\frac{1}{2}}  \sqrt{\log \left ( \frac{432 n}{\eta} \right ) } + 896 \frac{\left(d^m \right)^{\frac{1}{2}}\left(c_1^m(n)\right)^2}{\sqrt{c_0^m}} \left( \delta^m_{d^m} \right)^{-2} \frac{\log \left( \frac{216 n}{\eta} \right)}{\sqrt{n}}\\
     & \hspace{-0.6cm}  \quad + 1024 \frac{\left(d^m \right)^{\frac{1}{2}}\left(c_1^m(n)\right)^3}{\sqrt{c_0^m}} \left( \delta^m_{d^m} \right)^{-2} \frac{\log^{\frac{3}{2}} \left( \frac{216 n}{\eta} \right)}{n} + 64 \frac{\left(c_1^m(n)\right)^2}{\sqrt{c_0^m}} \left( \delta^m_{d^m} \right)^{-1} \frac{\left(d^m \right)^{\frac{1}{2}} \log \left( \frac{216n}{\eta} \right)}{\sqrt{n}} \\
     & \hspace{-0.6cm} \quad + 4 \frac{(d^m)^{\frac{1}{2}} c_1^m(n) }{\sqrt{c_0^m}} \sqrt{\log \left( \frac{72n}{\eta}  \right)},
     \label{eq:final_bound}
\end{align}
holds with $\J^m$-probability at least $1-\eta$ for any $\eta \in \left(0,3 \right) \cap \left( 0,1 \right) = \left(0,1 \right)$ and
\begin{align}
    n  &\ge \max\left\{\tilde{n}^m_{0,A1(i)}(\eta), \tilde{n}^m_{0,A1(ii)}(\eta), \tilde{n}^m_{0,A2(i)}(\eta), \tilde{n}^m_{0,A2(ii)}(\eta), \tilde{n}^m_{0,L5}(\eta), \tilde{n}^m_{0,L6}(\eta), \tilde{n}_{0,\text{else-1}}^m(\eta),\tilde{n}_{0,\text{else-2}}^m \right\}.
\end{align}
(a) holds by the triangle inequality. In (b) one can notice that $\left\|\b A_2\right\|_F \le \sqrt{\rank(\b A_2)} \left\|\b A_2\right\| \le \sqrt{d^m} \left\|\b A_2\right\| $ since $\rank(\b A_2)\le d^m$ by $\b A_2 \in \R^{n\times d^m}$ (c) is implied by the definition of $\b E_1$ [see \eqref{eq:E1-def}], the bound obtained for $\left\|\b E_2\right\|_F$ in \eqref{bound_E2}, the bound derived for $\left\| \b A_2 \right\|$ in \eqref{bound_A2} and our bound on $\left\|\b A_3\right\|_F$ in \eqref{bound_A3}. (d) follows from $256 \sqrt{2} \log \left( \frac{72 n}{\eta} \right) \le \frac{256 \frac{3}{2}}{\sqrt{c_0^m}} \log \left( \frac{216 n}{\eta} \right)$ since $0<c_0^m \le 1$, $\sqrt{2} \le \frac{3}{2}$ and $\log \left( \frac{72 n}{\eta} \right) \le \log \left( \frac{216 n}{\eta} \right)$. 
\end{proof}

\textbf{Remark}: 
Lemma~\ref{decomposition_LPRG} generalizes Lemma~50 from \cite{athreya2018statistical} significantly. The original lemma assumes
\begin{enumerate}
\item  linear kernel in its RDPG model, while we allow an arbitrary (continuous and $[0,1]$-bounded) one when dealing with the LPRG model leading to a much broader class of random graph models.  
\item  the elements in $\mathbf{A}^m$ to be random only---as their goal is to describe the accuracy of the adjacency spectral decomposition---, in our extension both the entries of $\mathbf{A}^m$ and the sequence $\left( X_i^m \right)_{i=1}^n$ are random; this extension is key to the understanding of the performance of the proposed $\t{GHSIC}_k$ estimator relying on the random samples $\left( X_i \right)_{i=1}^n$.
\end{enumerate}

The following lemma is the continuation of Lemma~\ref{decomposition_LPRG}, bounding also the term 
\begin{align}
    \left \| \left(\b U^m_{d^m} \left( \b U^m_{d^m} \right)^{\top} \left | \b A^m \right | - \b U^m_{\b K^m,d^m} \left( \b U^m_{\b K^m,d^m} \right)^\top \b K^m\right)\b U^m_{d^m} \left( \b S^m_{d^m} \right)^{-\frac{1}{2}} \right \|_F.
\end{align}

The next lemma expands on the error term left open in Lemma~\ref{decomposition_LPRG}.

\begin{lemma}[Improved approximation of the spectral features of $\b K^m$ from the adjacency matrix]\label{concentration_adj} Let $\eta \in \left(0,1 \right)$, $m\in [M]$, where $M \in \Zp$ denotes the number of LPRGs and $c^m > 0$ be fixed.$^{\ref{footnote:cm-meaning}}$ Let $n$ denote the number of nodes for each LPRG. Suppose that Assumptions~\ref{assum_Delta}-\ref{assum_eigenval_hat_C_k} hold with $n_{0,(i)}^{m}(\eta)$, $n_{0,(ii)}^{m}(\eta)$, $n_{1,(i)}^{m}(\eta)$, $n_{1,(ii)}^{m}(\eta) \in \Zp$, $c^m_0>0$, $c_1^m (n) = c_2^m \log \left( 2 + \log (2n) \right)$, $c_2^m > 0$, $C_0^m>0$, and $C_1^m>0$ specified therein. Assume that $C_1^m > 2\sqrt{2}$ and $c_2^m \ge 2$. Let 
\begin{align}
    \colAonei{\tilde{n}^m_{0,A1(i)}(\eta)} &:= n^m_{0,(i)}\left( \frac{\eta}{2304}\right), \quad \colAoneii{\tilde{n}^m_{0,A1(ii)}(\eta)} := n^m_{0,(ii)}\left( \frac{\eta}{1152} \right), \quad \colAi{\tilde{n}^m_{0,A2(i)}(\eta)} := n^m_{1,(i)}\left( \frac{\eta}{1152} \right), \\
    \colAii{\tilde{n}^m_{0,A2(ii)}(\eta)}
    &:= n^m_{1,(ii)}\left( \frac{\eta}{192} \right), \quad \quad \colLfive{\tilde{n}^m_{0,L5}(\eta)} := \left( \frac{\eta}{2304} \right)^{-\frac{1}{c^m}}, \quad \quad \colLsix{\tilde{n}^m_{0,L6}(\eta)} := \frac{4608}{\eta} ,\\
    \coleone{\tilde{n}_{0,\text{else}-1}^m(\eta)} & := 32 \left(c_1^m(n)\right)^2 (c_0^m)^{-2} \log \left( \frac{2304n}{\eta} \right), \quad \quad \coletwo{\tilde{n}_{0,\text{else}-2}^m} := e^{\frac{1}{\left( 2C_1^m - 4\sqrt{2} \right)^2}},
\end{align}
Let $\left( X_i^m \right)_{i=1}^n$ be i.i.d.\ $\X^m$-valued random variables, where $\X^m$ is assumed to be a compact metric space.
Let the Gram matrix $\b K^m = [K^m_{i,j}]_{i,j=1}^{n} = \left [ k^m(X^m_i,X^m_j) \right ]_{i,j=1}^{n}\in \R^{n \times n}$ associated to a continuous kernel $k^m:\X^m \times \X^m \to [0,1]$ defined on $\X^m$ with the associated adjacency matrix $\b A^m \in \R^{n \times n}$ as specified according to \eqref{eq:A^m}. Let $d^m \in \Zp$, the eigenvalues (ordered in a non-increasing way) of the integral operator $T_{k^m}$ be $\left ( \lambda^m_i \right )_{i \in \Zp} $ and the gap of $d^m$-th and $(d^m+1)$-th eigenvalues is defined as $\delta^m_{d^m} = \lambda^m_{d^m} - \lambda^m_{d^m+1} \ne 0$ (by Assumption~\ref{assum_eigenval_C_k}).$^{\ref{footnote_delta_ne_0}}$ Let the spectral decomposition and its $d^m$-truncated approximation of $\left | \b A^m \right | := \left( (\b A^m)^\top \b A^m \right)^{\frac{1}{2}} \in \R^{n \times n}$ be denoted by
\begin{align}
    \left | \b A^m \right | = \b U^m \b S^m  (\b U^m)^{\top} \approx \b U^m_{d^m} \b S^m_{d^m} \left( \b U^m_{d^m} \right)^{\top}, \label{eq:Um-dm}
\end{align}
where $\b S^m_{d^m} \in \R^{d^m \times d^m}$ is the diagonal matrix composed of the $d^m$ largest eigenvalues of $\left | \b A^m \right |$ and $\b U^m_{d^m} \in \R^{n \times d^m}$ consists of the  corresponding orthogonal eigenvectors. Similarly, let the spectral decomposition and its $d^m$-truncated approximation of $\b K^m$ be denoted by 
\begin{align}
    \b K^m = \b U^m_{\b K^m} \b S^m_{\b K^m} (\b U^m_{\b K^m})^{\top} \approx \b U^m_{\b K^m,d^m} \b S^m_{\b K^m,d^m} (\b U^m_{\b K^m,d^m})^{\top}, \label{eq:Um-Km-dm}
\end{align}
where $\b S^m_{\b K^m,d^m} \in \R^{d^m \times d^m}$ and $\b U^m_{\b K^m,d^m} \in \R^{n \times d^m}$.
Let\footnote{We note that $\hat{\phi}^m_{k^m}(X^m_i)$ and $\phi^m_{k^m,d^m}(X^m_i)$ denote features associated to $X_i^m$ but it is not a functional dependence.\label{footnote:phi-X_i-dependence}}
\begin{align}
    \hat{\bm{\Phi}}^m_{d^m} = \b U^m_{d^m} \left( \b S^m_{d^m} \right)^{\frac{1}{2}} =:\left[ \left( \hat{\phi}^m_{k^m}(X^m_i) \right)^{\top} \right]_{i=1}^n \in \R^{n \times d^m}.  \label{eq:def_feature_est_by_adj}
\end{align}
Consider the orthogonal matrix $\b W^m \in \R^{d^m \times d^m}$ whose existence was shown in Lemma~\ref{decomposition_LPRG} [see \eqref{eq:Wm-def}], and let\textsuperscript{\ref{footnote:phi-X_i-dependence}}
\begin{align}
    \bm{\Phi}^m_{d^m} = \b U^m_{\b K^m,d^m} \left( \b S^m_{\b K^m,d^m} \right)^{\frac{1}{2}} \b W^m =:\left[ \left( \phi^m_{k^m,d^m}(X^m_i) \right)^{\top} \right]_{i=1}^n \in \R^{n \times d^m}. \label{eq:def_feature_est_by_gram}
\end{align}

Assume that $\lambda^m_{d^m+1} \le 8 \left( \log 2 \right) \sqrt{\frac{d^m \log \left( \frac{16n}{\eta}  \right)}{n}}$ holds for $n \ge \left( \frac{\eta}{8} \right)^{-\frac{1}{c^m}}$. Then for
\begin{align}
    n \ge \max & \Big\{\tilde{n}^m_{0,A1(i)}(\eta), \tilde{n}^m_{0,A1(ii)}(\eta), \tilde{n}^m_{0,A2(i)}(\eta), \tilde{n}^m_{0,A2(ii)}(\eta), \tilde{n}^m_{0,L5}(\eta), \tilde{n}^m_{0,L6}(\eta), \tilde{n}_{0,\text{else-1}}^m(\eta),\\
    & \quad \tilde{n}_{0,\text{else-2}}^m \Big\},
\end{align}
we have that
\begin{align}
    & \quad \left \| \hat{\b \Phi}^m_{d^m} - \b \Phi^m_{d^m}  \right \|_F \le \\
    & \quad 312 \sqrt{2} c_1^m(n) d^m \left( \delta^m_{d^m} \right)^{-\frac{1}{2}}  \sqrt{\log \left ( \frac{864 n}{\eta} \right ) } + 896 \frac{\left(d^m \right)^{\frac{1}{2}}\left(c_1^m(n)\right)^2}{\sqrt{c_0^m}} \left( \delta^m_{d^m} \right)^{-2} \frac{\log \left( \frac{432 n}{\eta} \right)}{\sqrt{n}} \\
     & \quad + 1024 \frac{\left(d^m \right)^{\frac{1}{2}}\left(c_1^m(n)\right)^3}{\sqrt{c_0^m}} \left( \delta^m_{d^m} \right)^{-2} \frac{\log^{\frac{3}{2}} \left( \frac{432 n}{\eta} \right)}{n} + 64 \frac{\left(c_1^m(n)\right)^2}{\sqrt{c_0^m}} \left( \delta^m_{d^m} \right)^{-1} \frac{\left(d^m \right)^{\frac{1}{2}} \log \left( \frac{432 n}{\eta} \right)}{\sqrt{n}} \\
     & \quad + 4 \frac{(d^m)^{\frac{1}{2}} c_1^m(n)}{\sqrt{c_0^m}} \sqrt{\log \left( \frac{144n}{\eta}  \right)}\label{eq:final_bound_mat_feature_map}
\end{align}
holds with $\J^m$-probability at least $1-\eta$, where $\J^m$ denotes the joint distribution of the entries of $\b A^m$ and $(X_i^m)_{i=1}^n$.
\end{lemma}

\begin{proof}(Lemma \ref{concentration_adj}) Throughout the proof, we use the same coloring convention as in Lemma~\ref{decomposition_LPRG}. For the considered orthogonal matrix $\b W^m$, by Lemma~\ref{decomposition_LPRG} one has 
\begin{align}
    \J^m & \Big(\left \| \hat{\b \Phi}^m_{d^m} - \b \Phi^m_{d^m}  \right \|_F \le \underbrace{ \left \|  \left[ \b U^m_{d^m} \left( \b U^m_{d^m} \right)^{\top} \left | \b A^m \right | - \b U^m_{\b K^m,d^m} \left( \b U^m_{\b K^m,d^m} \right)^\top \b K^m \right] \b U^m_{d^m} \left( \b S^m_{d^m} \right)^{-\frac{1}{2}} \right \|_F}_{\eqcolon R^{m} \eqcolon R}\\
    & \hspace {3.4cm} + \xi^m_n \Big) \ge 1-\eta, \label{bound_term2}
\end{align}
where $\J^m$ denotes the joint distribution of the entries of $\b A^m$ and $\left(X^m_i\right)_{i=1}^n$, and 
\begin{align}
    \xi^m_n &:= 264 \sqrt{2} c_1^m(n) d^m \left( \delta^m_{d^m} \right)^{-\frac{1}{2}}  \sqrt{\log \left ( \frac{432 n}{\eta} \right ) } + 896 \frac{\left(d^m \right)^{\frac{1}{2}} \left(c_1^m(n)\right)^2} {\sqrt{c_0^m}} \left( \delta^m_{d^m} \right)^{-2} \frac{\log \left( \frac{216 n}{\eta} \right)}{\sqrt{n}}\\
     & \quad + 1024 \frac{\left(d^m \right)^{\frac{1}{2}} \left(c_1^m(n)\right)^3}{\sqrt{c_0^m}} \left( \delta^m_{d^m} \right)^{-2} \frac{\log^{\frac{3}{2}} \left( \frac{216 n}{\eta} \right)}{n} + 64 \frac{ \left(c_1^m(n)\right)^2}{\sqrt{c_0^m}} \left( \delta^m_{d^m} \right)^{-1} \frac{\left(d^m \right)^{\frac{1}{2}} \log \left( \frac{216 n}{\eta} \right)}{\sqrt{n}} \\
     & \quad + 4 \frac{(d^m)^{\frac{1}{2}} c_1^m(n)}{\sqrt{c_0^m}} \sqrt{\log \left( \frac{72n}{\eta}  \right)},
\end{align}
for $n$ large enough [meant in the sense of \eqref{eq:L19:n-constraint}].

We proceed by upper bounding $R$ 
\begin{align}\label{eq:decompose_R}
    R \stackrel{(a)}{\le} \underbrace{\left\| \left[\b U^m_{d^m} \left( \b U^m_{d^m} \right)^{\top} \left | \b A^m \right | - \b U^m_{\b K^m,d^m} \left( \b U^m_{\b K^m,d^m} \right)^\top \b K^m \right]\b U^m_{d^m} \right \|}_{=:A_1} \underbrace{\left \| \left( \b S^m_{d^m} \right)^{-\frac{1}{2} } \right \|_F}_{=:A_2},
\end{align}
by doing so with  $A_1$ and $A_2$, where 
(a) follows from $\left \| \b A \b B \right \|_F \le \left \| \b A \right \| \left \| \b B \right \|_F$.

\begin{itemize}
    \item \textbf{Term $A_1$}: By recalling the definition of the projections $\mathcal{P}^m_{\left| \b A^m \right|}:= \b U^m_{d^m} \left( \b U^m_{d^m} \right)^{\top}$ and $\mathcal{P}^m_{\b K^m}:= \b U^m_{\b K^m,d^m} \left( \b U^m_{\b K^m,d^m} \right)^\top$, we have 
    \begin{align}
    A_1 &  \stackrel{(a)}{=} \left\| \left(\mathcal{P}^m_{\left| \b A^m \right|} \left | \b A^m \right | - \b K^m + \b K^m - \mathcal{P}^m_{\b K^m} \b K^m \right)\b U^m_{d^m} \right \| \\
    & \stackrel{(b)}{\le} \underbrace{\left \| \left( \mathcal{P}^m_{\left| \b A^m \right|} \left | \b A^m \right |  - \b K^m \right) \b U^m_{d^m} \right \|}_{=:B_1} + \underbrace{\left \| \left( \b I - \mathcal{P}^m_{\b K^m} \right)\b K^m \b U^m_{d^m} \right \|}_{=:B_2}, \label{eq:A1->B1,B2}
    \end{align} 
    where (a) is implied by adding terms $\pm \b K^m \b U^m_{d^m}$; in (b) the triangular inequality was applied.

    \begin{itemize}
        \item \textbf{Term $B_1$}: For $B_1$, for any $\eta \in \left(0,12\right) \cap (0,6) = (0,6)$ and 
        \begin{align}
            n \ge \max \left\{\colAonei{ n^m_{0,(i)}\left( \frac{\eta}{48}\right)}, \colAoneii{n^m_{0,(ii)}\left( \frac{\eta}{48}\right)}, \colLfive{ \left( \frac{\eta}{48} \right)^{-\frac{1}{c^m}}}, \colLsix{\frac{192}{\eta}}, \coletwo{e^{\frac{1}{\left( 2C_1^m - 4\sqrt{2} \right)^2}} }\right\}
        \end{align}
        we have
    \begin{align}
    B_1 & \stackrel{(a)}{=} \left \| \left( \mathcal{P}^m_{\left| \b A^m \right|}\left | \b A^m \right | - \left | \b A^m \right |  +\left | \b A^m \right |  - \b K^m \right) \b U^m_{d^m}  \right \| \\
    & \stackrel{(b)}{\le} \left \| \left( \mathcal{P}^m_{\left| \b A^m \right|} - \b I \right)\left | \b A^m \right |  \right \| \left \| \b U^m_{d^m} \right \| + \left \| \left | \b A^m \right |  - \b K^m \right \| \left \| \b U^m_{d^m} \right \| \\
    & \stackrel{(c)}{\le} \lambda^m_{d^m+1}(\left | \b A^m \right | ) + \left \| \left | \b A^m \right |  - \b K^m \right \| 
    \stackrel{(d)}{\le} \lambda^m_{d^m+1}(\left | \b A^m \right | ) + 4 c_1^m(n) \sqrt{n \log \left( \frac{48n}{\eta} \right)} \\
    & \stackrel{(e)}{\le} \lambda_{d^m+1}^m (\b K^m)+ 4 c_1^m(n) \sqrt{ n \log \left( \frac{48n}{\eta} \right)} + 4 c_1^m(n) \sqrt{ n \log \left( \frac{48n}{\eta} \right)} \label{eq:bound_lemm17_B1_sub}  
\end{align}
holds with $\J^m$-probability at least $1-\frac{\eta}{6}$, where restricting $\eta$ to $(0,6)$ guarantees that the probability is positive. (a) was obtained by adding $\pm \left | \b A^m \right | \b U^m_{d^m}$, (b) follows from the triangle inequality and the property $\left\|\b A \b B\right\| \le \left\|\b A\right\| \left\|\b B\right\|$. (c) holds by using that $\left \| \b U^m_{d^m} \right \| \le 1$ and that $\left \| \left( \mathcal{P}^m_{\left| \b A^m \right|} - \b I \right)\left | \b A^m \right | \right \| = \lambda_{d^m+1}^m \left( \left | \b A^m \right | \right)$. The latter can be seen by introducing the notation $\b M := \left( \b I - \mathcal{P}^m_{\left| \b A^m \right|} \right) \left | \b A^m \right | \in \R^{n \times n}$, observing that $\left\| \b M \right\| = \left\| \left( \mathcal{P}^m_{\left| \b A^m \right|} - \b I \right)\left | \b A^m \right | \right\|$ by the homogeneity of norms, and recalling that $\b M$ has $n$ eigenvalues, $\lambda_1^m(\b M) = \cdots = \lambda_{d^m}^m (\b M)=0$ and $\lambda_{d^m+1}^m (\b M) = \lambda_{d^m+1}^m \left( \left | \b A^m \right | \right) \ge \cdots \ge \lambda_{n}^m (\b M) = \lambda_{n}^m \left( \left | \b A^m \right | \right) \ge 0$ as discussed in \eqref{eq:M:eigs-1}-\eqref{eq:M:eigs-2} and that $\b M$ is symmetric as shown in \eqref{eq:M-mtx:symmetric}. (d) follows with $\J^m$-probability at least $1-\frac{\eta}{12}$ for any $\eta \in \left(0, 12 \right)$ and $n \ge \max \left\{ \colAonei{ n^m_{0,(i)}\left( \frac{\eta}{48}\right)}, \colAoneii{n^m_{0,(ii)}\left( \frac{\eta}{48}\right)}, \colLfive{ \left( \frac{\eta}{48} \right)^{-\frac{1}{c^m}}}, \colLsix{\frac{192}{\eta}}, \coletwo{e^{\frac{1}{\left( 2C_1^m - 4\sqrt{2} \right)^2}} } \right\}$ by \eqref{bound_diff_psd} with a substitution $\frac{\eta}{3} \leftarrow \frac{\eta}{12}$. (e) holds with $\J^m$-probability at least $1-\frac{\eta}{12}$ for any $\eta \in \left(0,12\right)$ and $n \ge \max \left\{\colAonei{ n^m_{0,(i)}\left( \frac{\eta}{48}\right)}, \colAoneii{n^m_{0,(ii)}\left( \frac{\eta}{48}\right)}, \colLfive{ \left( \frac{\eta}{48} \right)^{-\frac{1}{c^m}}}, \colLsix{\frac{192}{\eta}}, \coletwo{e^{\frac{1}{\left( 2C_1^m - 4\sqrt{2} \right)^2}} }\right\}$ by using \eqref{eq:eig_adj_sub1-pre} with a substitution $\frac{\eta}{8} \leftarrow \frac{\eta}{12}$. A union bound implies the claimed result in \eqref{eq:bound_lemm17_B1_sub}.

We continue bounding $B_1$ by bounding the r.h.s.\ of  \eqref{eq:bound_lemm17_B1_sub}. We first show that if for $\eta \in (0,12)$ it holds that $\lambda^m_{d^m+1} \le 8 \left( \log 2 \right) \sqrt{\frac{d^m \log \left( \frac{24n}{\eta} \right)}{n}}$ in case of $n \ge \colLfive{\left( \frac{\eta}{12} \right)^{-\frac{1}{c^m}}}$, then 
\begin{align}\label{eq:eig_d1_bound}
    \J^m\left(\lambda^m_{d^m+1}(\b K^m) \le 8 c_1^m(n) \sqrt{d^m n \log \left( \frac{24n}{\eta} \right) }\right) \ge 1-\frac{\eta}{12}.
\end{align}
To obtain \eqref{eq:eig_d1_bound}, we proceed by showing that for any $\eta \in (0,12)$ 
\begin{align}
    & \quad \frac{1}{n} \lambda^m_{d^m+1}(\b K^m) \stackrel{(a)}{\le} \left | \frac{1}{n} \lambda^m_{d^m+1}(\b K^m) \right | \stackrel{(b)}{=} \left | \frac{1}{n} \lambda^m_{d^m+1}(\b K^m) - \lambda^m_{d^m+1} + \lambda^m_{d^m+1} \right | \\
    & \stackrel{(c)}{\le}  \left | \frac{1}{n} \lambda^m_{d^m+1}(\b K^m) - \lambda^m_{d^m+1} \right | +  \left | \lambda^m_{d^m+1} \right |  \stackrel{(d)}{\le} 2 \sqrt{2} a_{k^m}^2 \sqrt{\frac{\log \left( \frac{24}{\eta} \right)}{n}} + \underbrace{\left | \lambda^m_{d^m+1} \right |}_{= \lambda^m_{d^m+1}} \label{bound_eig_d1}
\end{align}
holds with $\J^m$-probability at least $1-\frac{\eta}{12}$. (a) was obtained by $x \le \left | x \right | $, in (b) we added $\pm \lambda^m_{d^m+1}$,
(c) comes from the triangular inequality. The second part of (d) holds by $\lambda^m_{d^m+1}>0$ implied by Assumption~\ref{assum_eigenval_C_k}; the first part of (d) can be shown as follows.
\begin{itemize}
    \item According to \eqref{eq:gap_between_eigenvals} in Lemma~\ref{eigen_Tk}, by letting $2e^{-\tau} = \frac{\eta}{12}$ [in other words, $\tau=\log\left(\frac{24}{\eta}\right)$; the $\tau > 0$ requirement implies that $\eta \in (0,24)$] we have for any $\eta \in (0,24) \cap (0,12)=(0,12)$ that
    \begin{align}
        (\P^m)^{\otimes n} \left(\left | \frac{1}{n} \lambda^m_{d^m+1}(\b K^m) - \lambda^m_{d^m+1} \right | \le 2 \sqrt{2} a_{k^m}^2 \sqrt{\frac{\log \left( \frac{24}{\eta} \right)}{n}}\right) \ge 1-\frac{\eta}{12},
    \end{align}
    with $a_{k^m}=\sqrt{\sup_{x^m\in \X^m}k^m(x^m,x^m)}$, where restricting $\eta$ to $(0,12)$ guarantees the positivity of the probability.
    \item Define the following event
    \begin{align}
        C = \left\{ \left( x_i^m \right)_{i=1}^n \,:\, \left | \frac{1}{n} \lambda^m_{d^m+1}(\b K^m) - \lambda^m_{d^m+1} \right | \le 2 \sqrt{2} a_{k^m}^2 \sqrt{\frac{\log \left( \frac{24}{\eta} \right)}{n}}  \right\} \subseteq (\X^m)^n,
    \end{align}
    Recall that $\Y^m = \{ 0,1 \}^{\binom{n}{2} }$ denotes the sample space of r.v.s. from $\Q^m$, where $\Q^m$ is the distribution of entries in $\b A^m$ given the latent positions $(x_i^m)_{i=1}^m$. Then for any $\eta \in (0,12)$ we have
    \begin{align}
        (\J^m) (C \times \Y^m) =  (\P^m)^{\otimes n}(C) \ge 1-\frac{\eta}{12}.
    \end{align}
    
    Therefore, we have for any $\eta \in (0,12)$ that 
    \begin{align}
        \J^m\left(\left | \frac{1}{n} \lambda^m_{d^m+1}(\b K^m) - \lambda^m_{d^m+1} \right | \le 2 \sqrt{2} a_{k^m}^2 \sqrt{\frac{\log \left( \frac{24}{\eta} \right)}{n}}\right) \ge 1-\frac{\eta}{12},
    \end{align}
    which implies \eqref{bound_eig_d1}.
\end{itemize}
Having shown \eqref{bound_eig_d1}, let us now multiply both sides of it by $n$ to get that if for $\eta  \in (0,12)$ it holds that $\lambda^m_{d^m+1} \le 8 \left( \log 2 \right) \sqrt{\frac{d^m \log \left( \frac{24n}{\eta} \right)}{n}}$ in case of $n \ge \colLfive{\left( \frac{\eta}{12} \right)^{-\frac{1}{c^m}}}$, then
\begin{align}
    & \quad \lambda^m_{d^m+1}(\b K^m) \le 2 \sqrt{2} a_{k^m}^2 \sqrt{n \log \left( \frac{24}{\eta} \right)} + n \lambda^m_{d^m+1} \\
    & \stackrel{(a)}{\le} 2 \sqrt{2} a_{k^m}^2 \sqrt{n \log \left( \frac{24}{\eta} \right)} + 8\left( \log 2 \right) \sqrt{d^m n \log \left( \frac{24n}{\eta} \right) } \\
    & \stackrel{(b)}{\le} 4 c_1^m(n) \sqrt{d^m n \log \left( \frac{24n}{\eta} \right) } + 4 c_1^m(n) \sqrt{d^m n \log \left( \frac{24n}{\eta} \right) } = 8 c_1^m(n) \sqrt{d^m n \log \left( \frac{24n}{\eta} \right)}, 
\end{align}
holds with $\J^m$-probability at least $1-\frac{\eta}{12}$, where (a) follows from the assumption that for $n \ge \colLfive{\left( \frac{\eta}{12} \right)^{-\frac{1}{c^m}}}$ we have
\begin{align}
    \lambda^m_{d^m+1} \le 8 \left( \log 2 \right) \sqrt{\frac{d^m \log \left( \frac{24n}{\eta} \right)}{n}} \stackrel{(c)}{\le} 4 c_1^m(n) \sqrt{\frac{d^m \log \left( \frac{24n}{\eta} \right)}{n}},
\end{align}
where (c) is implied by $c_1^m(n):= c_2^m \log \left( 2 + \log(2n) \right) \ge 2 \log 2$ by $n \ge 1$, the monotonicity of the $\log$ function and $c_2^m \ge 2$, (b) comes from 
\begin{align}
    4 c_1^m(n) \sqrt{d^m n \log \left( \frac{24n}{\eta} \right)} \stackrel{(d)}{\ge} 4 c_1^m(n) \sqrt{d^m n \log \left( \frac{24}{\eta} \right)} \stackrel{(e)}{\ge} 2 \sqrt{2} a_{k^m}^2 \sqrt{n \log \left( \frac{24}{\eta} \right)},
\end{align}
where (d) is implied by $n \ge 2$ and (e) comes from $d^m \ge 1$, $a^2_{k^m} \le 1$, $2 \sqrt{2} \le 4$, and $c_1^m(n) \ge 2 \log 2 \ge 1$; this shows \eqref{eq:eig_d1_bound}.

Having shown \eqref{eq:eig_d1_bound}, and using it in \eqref{eq:bound_lemm17_B1_sub} with a union bound, one gets that if for $\eta \in \left(0,6\right) \cap (0,12) \cap (0,4) = \left(0,4\right)$ it holds that $\lambda^m_{d^m+1} \le 8 \left( \log 2 \right) \sqrt{\frac{d^m \log \left( \frac{24n}{\eta} \right)}{n}}$ in case of $n \ge \colLfive{\left( \frac{\eta}{12} \right)^{-\frac{1}{c^m}}}$, and 
\begin{align}
    n & \ge \max \left\{\colAonei{ n^m_{0,(i)}\left( \frac{\eta}{48}\right)},\colAoneii{n^m_{0,(ii)}\left( \frac{\eta}{48}\right)},  \colLfive{ \left( \frac{\eta}{48} \right)^{-\frac{1}{c^m}}}, \colLsix{\frac{192}{\eta}}, \coletwo{e^{\frac{1}{\left( 2C_1^m - 4\sqrt{2} \right)^2}} } \right\},
\end{align}
then
\begin{align}
    B_1 & \le 8 c_1^m(n) \sqrt{d^m n \log \left( \frac{24n}{\eta} \right) } + 8 c_1^m(n) \sqrt{ n \log \left( \frac{48n}{\eta} \right)} \\
    &\stackrel{(a)}{\le} 8 c_1^m(n) \sqrt{d^m n \log \left( \frac{48n}{\eta} \right) } + 8 c_1^m(n) \sqrt{ n \log \left( \frac{48n}{\eta} \right)} \stackrel{(b)}{\le} 16 c_1^m(n) \sqrt{d^m n \log \left( \frac{48n}{\eta} \right)} \label{eq:bound_lemm17_B1}
\end{align}
holds with $\J^m$-probability at least $1-(\frac{\eta}{6}+\frac{\eta}{12}) =1-\frac{\eta}{4}$, where (a) comes from $\log \left( \frac{24n}{\eta} \right) \le \log \left( \frac{48n}{\eta} \right) $, (b) is implied by $8 c_1^m(n) \sqrt{ n \log \left( \frac{48n}{\eta} \right)} \le 8 c_1^m(n) \sqrt{ d^m n \log \left( \frac{48n}{\eta} \right)}$ by $d^m \ge 1$, and $\eta$ was restricted to $(0,4)$ to ensure the positivity of the probability.

\item \textbf{Term $B_2$}: For $B_2$, if for $\eta \in \left(0,4 \right)$ one has that $\lambda^m_{d^m+1} \le 8 \left( \log 2 \right) \sqrt{\frac{d^m \log \left( \frac{8n}{\eta} \right)}{n}}$ in case of $n \ge \colLfive{\left( \frac{\eta}{4} \right)^{-\frac{1}{c^m}}}$, then the following inequality 
\begin{align}
    B_2 & \stackrel{(a)}{=} \left \| \left( \b I - \mathcal{P}^m_{\b K^m} \right)\b K^m \b U^m_{d^m}  \right \| \stackrel{(b)}{\le} \left \| \left( \b I - \mathcal{P}^m_{\b K^m} \right)\b K^m \right \| \left \| \b U^m_{d^m}  \right \| \stackrel{(c)}{\le} \left \| \left( \b I - \mathcal{P}^m_{\b K^m} \right)\b K^m \right \| \\
    &\stackrel{(d)}{=} \lambda^m_{d^m+1}(\b K^m)  \stackrel{(e)}{\le} 8 c_1^m(n) \sqrt{d^m n \log \left( \frac{8n}{\eta} \right)}, \label{eq:bound_lemm17_B2}
\end{align}
holds with $\J^m$-probability at least $1-\frac{\eta}{4}$. The definition of $B_2$ was applied in (a). (b) is implied by $\left \| \b A \b B \right \| \le \left \| \b A \right \| \left \| \b B \right \|$. (c) follows from $\left \| \b U^m_{d^m} \right \| \le 1$. For (d), let $\tilde{\b M} = \left( \b I - \mathcal{P}^m_{\b K^m} \right) \b K^m \in \R^{n \times n}$; with this notation $\left\| \tilde{\b M} \right\|=\lambda^m_{d^m+1}(\b K^m)$ holds similarly to (c) in \eqref{eq:bound_lemm17_B1_sub}. (e) holds with $\J^m$-probability at least $1-\frac{\eta}{4}$ by \eqref{eq:eig_d1_bound} using a substitution $\frac{\eta}{12} \leftarrow \frac{\eta}{4}$ when for $\eta \in \left(0, 4 \right)$ one has that $\lambda^m_{d^m+1} \le 8 \left( \log 2 \right) \sqrt{\frac{d^m \log \left( \frac{8n}{\eta} \right)}{n}}$ in case of $n \ge \colLfive{\left( \frac{\eta}{4} \right)^{-\frac{1}{c^m}}}$.
\end{itemize}

Combining the results obtained for $B_1$ and $B_2$ with a union bound, one gets that for any $\eta \in \left(0,4\right) \cap \left(0,2 \right) = (0,2)$ if $\lambda^m_{d^m+1} \le 8 \left( \log 2 \right) \sqrt{\frac{d^m \log \left( \frac{24n}{\eta} \right)}{n}}$ for $n \ge \colLfive{\left( \frac{\eta}{12} \right)^{-\frac{1}{c^m}}}$ and $\lambda^m_{d^m+1} \le 8 \left( \log 2 \right) \sqrt{\frac{d^m \log \left( \frac{8n}{\eta} \right)}{n}}$ for $n \ge \colLfive{\left( \frac{\eta}{4} \right)^{-\frac{1}{c^m}}}$ [in other words, if $\lambda^m_{d^m+1} \le 8 \left( \log 2 \right) \sqrt{\frac{d^m \log \left( \frac{8n}{\eta} \right)}{n}}$ in case of $n \ge \colLfive{\left( \frac{\eta}{4} \right)^{-\frac{1}{c^m}}}$ by using that $\log \left( \frac{8n}{\eta} \right) \le \log \left( \frac{24n}{\eta} \right)$ due to the monotonicity of the $\log$ function and that $\min\left( \colLfive{\left( \frac{\eta}{12} \right)^{-\frac{1}{c^m}}},\colLfive{\left( \frac{\eta}{4} \right)^{-\frac{1}{c^m}}}\right) =  \colLfive{\left( \frac{\eta}{4} \right)^{-\frac{1}{c^m}}}$ by \eqref{eq:x^a-monotonicity}], and 
\begin{align}
n &\ge \max \left\{ \colAonei{ n^m_{0,(i)}\left( \frac{\eta}{48}\right)}, \colAoneii{n^m_{0,(ii)}\left( \frac{\eta}{48}\right)}, \colLfive{ \left( \frac{\eta}{48} \right)^{-\frac{1}{c^m}}}, \colLsix{\frac{192}{\eta}}, \coletwo{e^{\frac{1}{\left( 2C_1^m - 4\sqrt{2} \right)^2}} } \right\}  
\end{align}
then 
    \begin{align}
        A_1 &\stackrel{\eqref{eq:A1->B1,B2}}{\le} B_1 + B_2 \stackrel{\eqref{eq:bound_lemm17_B1}, \eqref{eq:bound_lemm17_B2}}{\le} 16 c_1^m(n)\sqrt{d^m n \log \left( \frac{48n}{\eta} \right)} + 8 c_1^m(n) \sqrt{d^m n \log \left( \frac{8n}{\eta} \right)} \\
        & \stackrel{(a)}{\le} 24 c_1^m(n) \sqrt{d^m n \log \left( \frac{48n}{\eta} \right)} \label{eq:A1_lemm17}
    \end{align}
    holds with $\J^m$-probability at least $1-\frac{\eta}{2}$,  
    where $\eta$ was restricted to $(0,2)$ to ensure the positivity of the probability. (a) comes from $\log \left( \frac{8n}{\eta} \right) \le \log \left( \frac{48n}{\eta} \right)$ using the monotonicity of the $\log$ function.

    \item \textbf{Term $A_2$}: For $A_2$, we have that 
    \begin{align}\label{eq:A2_lemm17}
        A_2 \stackrel{(a)}{=} \left\| \left( \b S_{d^m}^m \right)^{-\frac{1}{2}} \right\|_F \stackrel{(b)}{\le} \sqrt{d^m} \left \| \left( \b S_{d^m}^m \right)^{-\frac{1}{2}} \right \| \stackrel{(c)}{\le} 
        2\sqrt{2 d^m} (n \delta^m_{d^m})^{-\frac{1}{2}},
    \end{align}
    holds with $\J^m$-probability at least $1-\frac{\eta}{2}$ for any $\eta \in \left(0,2\right)$ and
    \begin{align}
    n \ge \max &\Bigg\{ \colAonei{ n^m_{0,(i)}\left( \frac{\eta}{128}\right)}, \colAonei{ n^m_{0,(i)}\left( \frac{\eta}{64}\right)}, \colAoneii{n^m_{0,(ii)}\left( \frac{\eta}{64}\right)}, \colAi{n^m_{1,(i)}\left( \frac{\eta}{64} \right)}, \colAii{ n_{1,(ii)}^m \left( \frac{3\eta}{8} \right)}, \colAii{ n_{1,(ii)}^m \left( \frac{3\eta}{32} \right)}, \\
    & \hspace{0.3cm} \colLfive{\left( \frac{\eta}{128} \right)^{-\frac{1}{c^m}}}, \colLsix{\frac{256}{\eta}}, \coleone{32 \left(c_1^m(n)\right)^2 (c_0^m)^{-2} \log \left( \frac{128n}{\eta} \right)},\coletwo{e^{\frac{1}{\left( 2C_1^m - 4\sqrt{2} \right)^2}}} \Bigg\}. 
    \end{align}
    The definition of $A_2$ was used in (a), (b) was obtained by $\left \| \b A \right \|_F \le \sqrt{\rank(\b A)} \left \| \b A \right \| $ where $\rank \left( \left( \b S_{d^m}^m \right)^{-\frac{1}{2}} \right) = d^m$, (c) holds for any $\eta \in \left(0,2\right)$ with $\J^m$-probability at least $1-\frac{\eta}{2}$ by using \eqref{eq:bound_eig_A} with a substitution $\eta \leftarrow \frac{\eta}{2}$.
    \end{itemize}

Combining the bounds of $A_1$ and $A_2$, if for $\eta \in \left(0,2\right) \cap (0,1) = \left(0,1 \right)$ one has that $\lambda^m_{d^m+1} \le 8\left( \log 2 \right) \sqrt{\frac{d^m \log \left( \frac{8n}{\eta} \right)}{n}} $ holds in case of $n\ge \colLfive{\left( \frac{\eta}{4} \right)^{-\frac{1}{c^m}}}$, and
\begin{align}
    n \ge \max & \Bigg\{ \colAonei{ n^m_{0,(i)}\left( \frac{\eta}{48}\right)}, \colAonei{ n^m_{0,(i)}\left( \frac{\eta}{128}\right)}, \colAonei{ n^m_{0,(i)}\left( \frac{\eta}{64}\right)}, \colAoneii{n^m_{0,(ii)}\left( \frac{\eta}{48}\right)}, \colAoneii{n^m_{0,(ii)}\left( \frac{\eta}{64}\right)}, \colAi{n^m_{1,(i)}\left( \frac{\eta}{64} \right)},   \\
    & \hspace{0.3cm} \colAii{ n_{1,(ii)}^m \left( \frac{3\eta}{8} \right)}, \colAii{ n_{1,(ii)}^m \left( \frac{3\eta}{32} \right)}, \colLfive{\left( \frac{\eta}{48} \right)^{-\frac{1}{c^m}}}, \colLfive{\left( \frac{\eta}{128} \right)^{-\frac{1}{c^m}}},  \colLsix{\frac{192}{\eta}}, \colLsix{\frac{256}{\eta}},  \\
    & \hspace{0.3cm} \coleone{32 \left(c_1^m(n)\right)^2 (c_0^m)^{-2} \log \left( \frac{128n}{\eta} \right)},\coletwo{e^{\frac{1}{\left( 2C_1^m - 4\sqrt{2} \right)^2}}} \Bigg\} \\
    \stackrel{\eqref{eq:x^a-monotonicity}}{=} \max & \Bigg\{ \colAonei{ n^m_{0,(i)}\left( \frac{\eta}{48}\right)}, \colAonei{ n^m_{0,(i)}\left( \frac{\eta}{128}\right)}, \colAonei{ n^m_{0,(i)}\left( \frac{\eta}{64}\right)}, \colAoneii{n^m_{0,(ii)}\left( \frac{\eta}{48}\right)}, \colAoneii{n^m_{0,(ii)}\left( \frac{\eta}{64}\right)}, \colAi{n^m_{1,(i)}\left( \frac{\eta}{64} \right)}, \\
    & \hspace{0.3cm} \colAii{ n_{1,(ii)}^m \left( \frac{3\eta}{8} \right)}, \colAii{ n_{1,(ii)}^m \left( \frac{3\eta}{32} \right)}, \colLfive{\left( \frac{\eta}{128} \right)^{-\frac{1}{c^m}}}, \colLsix{\frac{256}{\eta}}, \coleone{32 \left(c_1^m(n)\right)^2 (c_0^m)^{-2} \log \left( \frac{128n}{\eta} \right)}, \\
    & \hspace{0.3cm} \coletwo{e^{\frac{1}{\left( 2C_1^m - 4\sqrt{2} \right)^2}}} \Bigg\},
\end{align}
then 
\begin{align}\label{bound_term1}
    R & \stackrel{\eqref{eq:decompose_R}}{\le} A_1 A_2 \stackrel{\eqref{eq:A1_lemm17}, \eqref{eq:A2_lemm17}}{\le} 48 \sqrt{2} c_1^m(n) d^m \left( \delta^m_{d^m} \right)^{-\frac{1}{2}} \sqrt{\log \left( \frac{48n}{\eta} \right)} 
\end{align}
holds with $\J^m$-probability at least $1-\eta$, where $\eta$ was restricted to $(0,1)$ to ensure the positivity of the probability.

Next we combine with a union bound \eqref{bound_term2} and \eqref{bound_term1} using a substitution $\eta \leftarrow \frac{\eta}{2}$. Particularly, let
\begin{align}
    \colAonei{\tilde{n}^m_{0,A1(i)}(\eta)} &:= \max \Bigg\{ n^m_{0,(i)}\left( \frac{\eta}{2304}\right), n^m_{0,(i)}\left( \frac{\eta}{96}\right), n^m_{0,(i)}\left( \frac{\eta}{256}\right), n^m_{0,(i)}\left( \frac{\eta}{128}\right) \Bigg\} \stackrel{(a)}{=} n^m_{0,(i)}\left( \frac{\eta}{2304}\right), \\
    \colAoneii{\tilde{n}^m_{0,A1(ii)}(\eta)} &:= \max \Bigg\{ n^m_{0,(ii)}\left( \frac{\eta}{1152} \right), n^m_{0,(ii)}\left( \frac{\eta}{96}\right), n^m_{0,(ii)}\left( \frac{\eta}{128}\right) \Bigg\} \stackrel{(a)}{=} n^m_{0,(ii)}\left( \frac{\eta}{1152} \right) , \\
    \colAi{\tilde{n}^m_{0,A2(i)}(\eta)} &:= \max \left\{ n^m_{1,(i)}\left( \frac{\eta}{1152} \right), n^m_{1,(i)}\left( \frac{\eta}{128} \right) \right\} \stackrel{(a)}{=} n^m_{1,(i)}\left( \frac{\eta}{1152} \right), \\
    \colAii{\tilde{n}^m_{0,A2(ii)}(\eta)}
    &:=\max \Bigg\{ n_{1,(ii)}^m \left( \frac{\eta}{192} \right), n_{1,(ii)}^m \left( \frac{3\eta}{16} \right), n_{1,(ii)}^m \left( \frac{3\eta}{64} \right) \Bigg\} \stackrel{(b)}{=} n^m_{1,(ii)}\left( \frac{\eta}{192} \right), \\
    \colLfive{\tilde{n}^m_{0,L5}(\eta)} &:= \max \left\{ \left( \frac{\eta}{2304} \right)^{-\frac{1}{c^m}}, \left( \frac{\eta}{256} \right)^{-\frac{1}{c^m}} \right\} \stackrel{\eqref{eq:x^a-monotonicity}}{=} \left( \frac{\eta}{2304} \right)^{-\frac{1}{c^m}}, \\
    \colLsix{\tilde{n}^m_{0,L6}(\eta)} &:= \max \left\{ \frac{4608}{\eta}, \frac{512}{\eta} \right\} = \frac{4608}{\eta} ,\\
    \coleone{\tilde{n}_{0,\text{else}-1}^m(\eta)} & := \max \left\{ 32 \left(c_1^m(n)\right)^2 (c_0^m)^{-2} \log \left( \frac{2304 n}{\eta} \right), 32 \left(c_1^m(n)\right)^2 (c_0^m)^{-2} \log \left( \frac{256 n}{\eta} \right) \right\} \\
    & \stackrel{(c)}{=} 32 \left(c_1^m(n)\right)^2 (c_0^m)^{-2} \log \left( \frac{2304 n}{\eta} \right), \\
    \coletwo{\tilde{n}_{0,\text{else}-2}^m} & := e^{\frac{1}{\left( 2C_1^m - 4\sqrt{2} \right)^2}},
\end{align}
where (a) and (b) follow from the same reasoning as (a) and (b) in \eqref{eq:simplify_constriants}, (c) is implied by the monotonicity of the log function. If for $\eta \in \left( 0,2 \right) \cap (0,1) = \left(0,1 \right)$ 
\begin{align}
    n \ge \max\left\{\tilde{n}^m_{0,A1(i)}(\eta), \tilde{n}^m_{0,A1(ii)}(\eta), \tilde{n}^m_{0,A2(i)}(\eta), \tilde{n}^m_{0,A2(ii)}(\eta), \tilde{n}^m_{0,L5}(\eta), \tilde{n}^m_{0,L6}(\eta), \tilde{n}_{0,\text{else-1}}^m(\eta),\tilde{n}_{0,\text{else-2}}^m \right\},
\end{align}
and if $\lambda^m_{d^m+1} \le 8 \left( \log 2 \right)\sqrt{\frac{d^m \log \left( \frac{16n}{\eta} \right)}{n}} $ in case of $n\ge \colLfive{\left( \frac{\eta}{8} \right)^{-\frac{1}{c^m}}}$, then
\begin{align}
    & \quad \left \| \hat{\b \Phi}^m_{d^m} - \b \Phi^m_{d^m}  \right \|_F \le 48 \sqrt{2} c_1^m(n) d^m \left( \delta^m_{d^m} \right)^{-\frac{1}{2}} \sqrt{\log \left( \frac{96n}{\eta} \right)} \\
    & \quad + 264 \sqrt{2} c_1^m(n) d^m \left( \delta^m_{d^m} \right)^{-\frac{1}{2}}  \sqrt{\log \left ( \frac{864 n}{\eta} \right ) } + 896 \frac{\left(d^m \right)^{\frac{1}{2}} \left(c_1^m(n)\right)^2} {\sqrt{c_0^m}} \left( \delta^m_{d^m} \right)^{-2} \frac{\log \left( \frac{432 n}{\eta} \right)}{\sqrt{n}}\\
     & \quad + 1024 \frac{\left(d^m \right)^{\frac{1}{2}} \left(c_1^m(n)\right)^3}{\sqrt{c_0^m}} \left( \delta^m_{d^m} \right)^{-2} \frac{\log^{\frac{3}{2}} \left( \frac{432 n}{\eta} \right)}{n} + 64 \frac{ \left(c_1^m(n)\right)^2}{\sqrt{c_0^m}} \left( \delta^m_{d^m} \right)^{-1} \frac{\left(d^m \right)^{\frac{1}{2}} \log \left( \frac{432 n}{\eta} \right)}{\sqrt{n}} \\
     & \quad + 4 \frac{(d^m)^{\frac{1}{2}} c_1^m(n)}{\sqrt{c_0^m}} \sqrt{\log \left( \frac{144n}{\eta}  \right)} \\
     & \stackrel{(a)}{\le} 312 \sqrt{2} c_1^m(n) d^m \left( \delta^m_{d^m} \right)^{-\frac{1}{2}}  \sqrt{\log \left ( \frac{864 n}{\eta} \right ) } + 896 \frac{\left(d^m \right)^{\frac{1}{2}}\left(c_1^m(n)\right)^2}{\sqrt{c_0^m}} \left( \delta^m_{d^m} \right)^{-2} \frac{\log \left( \frac{432 n}{\eta} \right)}{\sqrt{n}} \\
     & \quad + 1024 \frac{\left(d^m \right)^{\frac{1}{2}}\left(c_1^m(n)\right)^3}{\sqrt{c_0^m}} \left( \delta^m_{d^m} \right)^{-2} \frac{\log^{\frac{3}{2}} \left( \frac{432 n}{\eta} \right)}{n} + 64 \frac{\left(c_1^m(n)\right)^2}{\sqrt{c_0^m}} \left( \delta^m_{d^m} \right)^{-1} \frac{\left(d^m \right)^{\frac{1}{2}} \log \left( \frac{432 n}{\eta} \right)}{\sqrt{n}} \\
     & \quad + 4 \frac{(d^m)^{\frac{1}{2}} c_1^m(n)}{\sqrt{c_0^m}} \sqrt{\log \left( \frac{144n}{\eta}  \right)}  
\end{align}
holds with $\J^m$-probability at least $1-\eta$, where $\eta$ was restricted to $(0,1)$ to ensure the positivity of the probability. (a) holds by $\log \left( \frac{96n}{\eta} \right) \le \log \left( \frac{864n}{\eta} \right)$ which is implied by the monotonicity of the $\log$ function. This completes the proof.

\end{proof}

Having obtained the error bound between the $d^m$-truncated approximation of $\left| \b A^m \right|$ and the $d^m$-truncated approximation of $\b K^m$, we now extend our result to establish an error bound between the feature maps of the $d^m$-truncated approximation of $\left| \b A^m \right|$ and the $d^m$-truncated approximation of $\b K^m$. 

\begin{lemma}[approximation of the joint spectral features from the adjacency matrix]\label{concentration_joint_adj}
Let $\eta \in \left(0, 1 \right)$ be fixed. Let $c^m > 0$ be also fixed for all $m \in [M]$, where $M \in \Zp$ denotes the number of LPRGs.$^{\ref{footnote:cm-meaning}}$ Let $n$ denote the number of nodes for each LPRG. Suppose that for all $m \in [M]$ Assumption~\ref{assum_Delta}-\ref{assum_eigenval_hat_C_k} hold with $n_{0,(i)}^{m}(\eta), n_{0,(ii)}^{m}(\eta), n_{1,(i)}^{m}(\eta), n_{1,(ii)}^{m}(\eta) \in \Zp$, $c^m_0>0$, $c_1^m(n) = c_2^m \log \left( 2 + \log(2n)\right)$, $c_2^m > 0$, $C_0^m>0$, and $C_1^m>0$ specified therein. Assume that $C_1^m > 2\sqrt{2}$ and $c_2^m \ge 2$ for all $m \in [M]$.  For all $m \in [M]$, let
\begin{align}
    \colAonei{\tilde{n}^m_{0,A1(i)}(\eta)} &:= n^m_{0,(i)}\left( \frac{\eta}{2304}\right), \quad \colAoneii{\tilde{n}^m_{0,A1(ii)}(\eta)} := n^m_{0,(ii)}\left( \frac{\eta}{1152} \right), \quad \colAi{\tilde{n}^m_{0,A2(i)}(\eta)} := n^m_{1,(i)}\left( \frac{\eta}{1152} \right), \\
    \colAii{\tilde{n}^m_{0,A2(ii)}(\eta)}
    &:= n^m_{1,(ii)}\left( \frac{\eta}{192} \right), \quad \quad \colLfive{\tilde{n}^m_{0,L5}(\eta)} := \left( \frac{\eta}{2304} \right)^{-\frac{1}{c^m}}, \quad \quad \colLsix{\tilde{n}^m_{0,L6}(\eta)} := \frac{4608}{\eta} ,\\
    \coleone{\tilde{n}_{0,\text{else}-1}^m(\eta)} & := 32 \left(c_1^m(n)\right)^2 (c_0^m)^{-2} \log \left( \frac{2304n}{\eta} \right), \quad \quad \coletwo{\tilde{n}_{0,\text{else}-2}^m} := e^{\frac{1}{\left( 2C_1^m - 4\sqrt{2} \right)^2}}.
\end{align}
Let $d^m \in \Zp$ and $\hat{\phi}^m_{k^m}(X^m_i) \in\R^{d^m}$ defined as in \eqref{eq:def_feature_est_by_adj} and $\phi^m_{k^m,d^m}(X^m_i) \in\R^{d^m}$ defined as in \eqref{eq:def_feature_est_by_gram} for all $m \in [M]$. Let $\left( \lambda_i^m \right)_{i \in \Zp}$ denote the eigenvalues of the integral operator $T_{k^m}$, arranged in non-increasing order. The gap between the $d^m$-th and $(d^m+1)$-th eigenvalues is defined as $\delta_{d^m}^m = \lambda^m_{d^m} - \lambda^m_{d^m+1} \ne 0$ (by Assumption~\ref{assum_eigenval_C_k}) for all $m\in [M]$.$^{\ref{footnote_delta_ne_0}}$ Let 
\begin{align} 
\hat{\mu}_k(\hat{\P}_n) &=\frac{1}{n} \sum_{i=1}^n \otimes_{m=1}^M  \hat{\phi}_{k^m}^{m}(X_i^m) \in \R^{d^1 \times \cdots \times d^M}, & \breve{\mu}_{k}(\hat{\P}_n) = \frac{1}{n} \sum_{i=1}^n \otimes_{m=1}^M   \phi_{k^m,d^m}^{m} (X^m_i) \in \R^{d^1 \times \cdots \times d^M}. \label{eq:def_joint_feature_est}
\end{align}
Assume that $\lambda^m_{d^m+1} \le 8 \left( \log 2 \right) \sqrt{\frac{d^m \log \left( \frac{16Mn}{\eta}  \right)}{n}}$ holds for $n \ge \left( \frac{\eta}{8M} \right)^{-\frac{1}{c^m}}$ and all $m\in [M]$. Assume that for all $m \in [M]$ and all $\left(x_i^m \right)_{i=1}^n \subset \X^m$
\begin{align}
    \left\| \hat{\phi}_{k^m}^{m}(x_i^m) \right\|_2 \le d_{k^m},  %\label{assum:bounded_feature_map}
\end{align}
and let $d_k := \prod_{m=1}^M d_{k^m} $; w.l.o.g.\ we assume that $d_{k^m} \ge 1$ for all $m \in [M]$. Then for
\begin{align}
    n &\ge \max_{m \in [M]} \Big\{\tilde{n}^m_{0,A1(i)}\left( \frac{\eta}{M} \right), \tilde{n}^m_{0,A1(ii)}\left( \frac{\eta}{M} \right), \tilde{n}^m_{0,A2(i)}\left( \frac{\eta}{M} \right), \tilde{n}^m_{0,A2(ii)}\left( \frac{\eta}{M} \right), \tilde{n}^m_{0,L5}\left( \frac{\eta}{M} \right),  \\
    & \quad \hspace{1.4cm} \tilde{n}^m_{0,L6}\left( \frac{\eta}{M} \right), \tilde{n}_{0,\text{else-1}}^m\left( \frac{\eta}{M} \right), \tilde{n}_{0,\text{else-2}}^m \Big\},
\end{align}
we have that
\begin{align}
    \mathbb{H} \left( \left \| \breve{\mu}_{k}(\hat{\P}_n) -\hat{\mu}_k(\hat{\P}_n) \right \|_2 \le d_k \sum_{m=1}^M D_{k^m}\left( \frac{\eta}{M} \right) \right) \ge 1-\eta,
\end{align} 
where $\mathbb{H}$ denotes the joint distribution of the adjacency matrices $(\b A^m)_{m=1}^M$ and $(X_i)_{i=1}^n$, $\left\|\cdot\right\|_2$ on tensors in $\R^{d^1 \times \dots \times d^M}$ is defined according to \eqref{eq:2-norm-on-tensors}, and $D_{k^m}$ is defined as
\begin{align}
     D_{k^m}(\eta) & : = 312 \sqrt{2} c_1^m(n) d^m \left( \delta^m_{d^m} \right)^{-\frac{1}{2}}  \sqrt{\frac{\log \left ( \frac{864 n}{\eta} \right )}{n} } + 896 \frac{\left(d^m \right)^{\frac{1}{2}}\left(c_1^m(n)\right)^2}{\sqrt{c_0^m}} \left( \delta^m_{d^m} \right)^{-2} \frac{\log \left( \frac{432 n}{\eta} \right)}{n} \\
     & \quad + 1024 \frac{\left(d^m \right)^{\frac{1}{2}}\left(c_1^m(n)\right)^3}{\sqrt{c_0^m}} \left( \delta^m_{d^m} \right)^{-2} \frac{\log^{\frac{3}{2}} \left( \frac{432 n}{\eta} \right)}{n^{\frac{3}{2}}} + 64 \frac{\left(c_1^m(n)\right)^2}{\sqrt{c_0^m}} \left( \delta^m_{d^m} \right)^{-1} \frac{\left(d^m \right)^{\frac{1}{2}} \log \left( \frac{432 n}{\eta} \right)}{n} \\
     & \quad + 4 \frac{(d^m)^{\frac{1}{2}} c_1^m(n)}{\sqrt{c_0^m}} \sqrt{\frac{\log \left( \frac{144n}{\eta}  \right)}{n}}.  \label{eq:def_D_k^m}
\end{align}

\end{lemma}

\begin{proof}(Lemma \ref{concentration_joint_adj}):
We have
\begin{align}
    \left \| \breve{\mu}_k(\hat{\P}_n) -\hat{\mu}_k(\hat{\P}_n) \right \|^2_2 & = \left \| \frac{1}{n} \sum_{i=1}^n \otimes_{m=1}^M \phi_{k^m,d^m}^{m} (X^m_i) -\frac{1}{n} \sum_{i=1}^n \otimes_{m=1}^M  \hat{\phi}_{k^m}^{m}(X_i^m) \right \|^2_2 \\
    & \stackrel{(a)}{\le} \frac{1}{n^2} \left( \sum_{i=1}^n \left \| \otimes_{m=1}^M \phi_{k^m,d^m}^{m} (X^m_i) - \otimes_{m=1}^M  \hat{\phi}_{k^m}^{m}(X_i^m) \right \|_2 \right)^2,
\end{align}
where (a) follows from the homogeneity of norms, the triangle inequality, and the non-decreasing property of the function $z\mapsto z^2$ on non-negative reals. By taking the square root of both sides, for
\begin{align}
    n &\ge \max_{m \in [M]} \Big\{\tilde{n}^m_{0,A1(i)}\left( \frac{\eta}{M} \right), \tilde{n}^m_{0,A1(ii)}\left( \frac{\eta}{M} \right), \tilde{n}^m_{0,A2(i)}\left( \frac{\eta}{M} \right), \tilde{n}^m_{0,A2(ii)}\left( \frac{\eta}{M} \right), \tilde{n}^m_{0,L5}\left( \frac{\eta}{M} \right),  \\
    & \quad \hspace{1.4cm} \tilde{n}^m_{0,L6}\left( \frac{\eta}{M} \right), \tilde{n}_{0,\text{else-1}}^m\left( \frac{\eta}{M} \right), \tilde{n}_{0,\text{else-2}}^m \Big\},
\end{align}
we have
\begin{align}
    \left \| \breve{\mu}_k(\hat{\P}_n) -\hat{\mu}_k(\hat{\P}_n) \right \|_2 & \le \frac{1}{n} \sum_{i=1}^n \left \| \otimes_{m=1}^M \phi_{k^m,d^m}^{m} (X^m_i) - \otimes_{m=1}^M  \hat{\phi}_{k^m}^{m}(X_i^m) \right \|_2 \\
    & \stackrel{(a)}{\le} \frac{1}{n} \sum_{i=1}^n \sum_{m=1}^M \left( \prod_{j=m+1}^M d_{k^j} \right) \left \| \phi_{k^m,d^m}^{m}(X^m_i) -  \hat{\phi}_{k^m}^{m}(X^m_i) \right \|_2 \\
    & \stackrel{(b)}{=} \sum_{m=1}^M \left( \prod_{j=m+1}^M d_{k^j} \right) \frac{1}{n} \sum_{i=1}^n \left \| \phi_{k^m,d^m}^{m}(X^m_i) -  \hat{\phi}_{k^m}^{m}(X^m_i) \right \|_2 \\
    & \stackrel{(c)}{\le} \sum_{m=1}^M \left( \prod_{j=m+1}^M d_{k^j} \right) \left( \frac{1}{n} \sum_{i=1}^n \left \| \phi_{k^m,d^m}^{m}(X^m_i) -  \hat{\phi}_{k^m}^{m}(X^m_i) \right \|_2^2 \right)^{\frac{1}{2}}, \\
    & \stackrel{(d)}{=} \sum_{m=1}^M \left( \prod_{j=m+1}^M d_{k^j} \right) \frac{1}{\sqrt{n}} \left \| \hat{\bm{\Phi}}^m_{d^m} - \bm{\Phi}^m_{d^m} \right \|_F \\
    & \stackrel{(e)}{\le} d_k \sum_{m=1}^M \frac{1}{\sqrt{n}} \left \| \hat{\bm{\Phi}}^m_{d^m} - \bm{\Phi}^m_{d^m} \right \|_F \stackrel{(f)}{\le}d_k \sum_{m=1}^M D_{k^m} \left( \frac{\eta}{M} \right), \label{eq:ineq_tilde_mu}
\end{align}
holds with $\mathbb{H}$-probability at least $1-\eta$ for any $\frac{\eta}{M} \in \left(0,1 \right)$ [in other words, $\eta \in \left(0,M \right)$; using that $M\ge 2$, one can restrict $\eta$ to $(0,1)$ guaranteeing that the probability is positive], where $\mathbb{H}$ denotes the joint distribution of the adjacency matrices $(\b A^m)_{m=1}^M$ and $(X_i)_{i=1}^n$ and $D_{k^m}(\eta)$ is defined in \eqref{eq:def_D_k^m}.
(a) in \eqref{eq:ineq_tilde_mu} follows from Lemma~\ref{lemma:properties-of-tensor-products} [see \eqref{eq:diff-of-tensor-products_Am1}] by choosing $\X^m = \R^{d^m}$ and $k^m$ to be the linear kernel for $m \in [M]$, with a substitution of $a_m \leftarrow \phi_{k^m,d^m}^{m}(X^m_i)$, $b_m \leftarrow \hat{\phi}_{k^m}^{m}(X^m_i)$. 
Regarding the constants $B_m$ and $A_m$ ($m\in [M]$) in Lemma~\ref{lemma:properties-of-tensor-products}, by the assumption $\left\| \hat{\phi}_{k^m}^{m}(x_i^m) \right\|_2 \le d_{k^m}$ in \eqref{assum:bounded_feature_map} we have $B_m \leftarrow d_{k^m}$. Next we derive the value of $A_m$. Let
\begin{align}
\b B^m &:= \b U_{\b K^m}^m \left(\b S^m_{\b K^m} \right)^{\frac{1}{2}}=: \left[ \left( \b b^m_i \right)^\T  \right]_{i=1}^n \in \R^{n \times n},\\
\b C^m &:= \b U_{\b K^m,d^m}^m \left(\b S^m_{\b K^m,d^m} \right)^{\frac{1}{2}}=: \left[ \left( \b c^m_i \right)^\T  \right]_{i=1}^n \in \R^{n \times d^m}
\end{align}
where $\b U_{\b K^m}^m \in \R^{n \times n}$, $\b S^m_{\b K^m} \in \R^{n \times n}$, $\b U_{\b K^m,d^m}^m\in \R^{n\times d^m}$ and $\b S^m_{\b K^m,d^m} \in \R^{d^m\times d^m}$ are defined in \eqref{Km:SVD-SVDm}, $\b b_i^m \in \R^n$ denotes the transpose of the $i$-th row of $\b B^m$, $\b c_i^m \in \R^{d^m}$ stands for the transpose of the $i$-th row of $\b C^m$. Then we have for all $m\in [M]$ that
\begin{align}
\left\| \phi_{k^m,d^m}^{m} (x_i^m) \right\|_2^2  &\stackrel{(g)}{=} \left\| \left( \b c_i^m \right)^{\T} \b W^m \right\|_2^2 = \left( \left(\b c_i^m \right)^{\T} \b W^m \right)  \left( \left( \b c_i^m \right)^{\T} \b W^m \right)^{\T}\\
&= \left( \b c_i^m \right)^\T \overbrace{\b W^m\left( \b W^m \right)^\T}^{\stackrel{(h)}{=}\b I} \b c_i^m  = \left( \b c_i^m \right)^\T \b c_i^m = \left\| \b c_i^m \right\|_2^2 \stackrel{(i)}{\le} \left\| \b b^m_i \right\|_2^2 \\
&\stackrel{(j)}{=} K^m_{i,i} \stackrel{(k)}{\le} 1 \Rightarrow \left\| \phi_{k^m,d^m}^{m}(x_i^m)  \right\|_2 \le 1. 
\end{align}
(g) is implied by the definition of $\phi_{k^m,d^m}^{m} (x_i^m)$; (h) follows from the orthogonality of $\b W^m$; (i) holds as $\b c_i^m$ is a truncated version of $\b b_i^m$; (j) is implied by $\left\| \b b_i^m \right\|_2^2 =  \left \langle \b b_i^m, \b b_i^m \right \rangle  = K_{i,i}^m$ where $K_{i,i}^m$ denotes the $(i,i)$-th element of the Gram matrix $\b K^m$ in \eqref{Km:SVD-SVDm} and (k) holds by the boundedness assumption imposed on the kernel $k^m$ [$k^m\left(x^m,(x^m)'\right)\le 1$ for any $x^m,(x^m)'\in \X^m$]. Therefore, we have 
\begin{align}
\left \| \phi_{k^m,d^m}^{m}(x^m_i) \right \|_2 \le 1 \t{ for all }m\in [M],\label{eq:bound-on-truncated-feature-map}
\end{align}
and thus one can choose $A_m=1$. (b) in \eqref{eq:ineq_tilde_mu} is implied by exchanging the order of summation. Considering (c) in \eqref{eq:ineq_tilde_mu}, with $\b d_i^m:= \phi_{k^m,d^m}^{m}(X^m_i) - \hat{\phi}_{k^m}^{m}(X^m_i) \in \R^{d^m}$, we have
\begin{align}
    \frac{1}{n}\sum_{i=1}^n \left\| \b d_i^m \right\|_2 &= \frac{1}{n}\sum_{i=1}^n \left( \left\| \b d_i^m \right\|_2 \times 1 \right)  \stackrel{(l)}{\le} \frac{1}{n} \left(\sum_{i=1}^n \left\| \b d_i^m \right\|_2^2 \right)^{\frac{1}{2}} \left(\sum_{i=1}^n 1^{2} \right)^{\frac{1}{2}} \\
    & = \frac{1}{n} \left(\sum_{i=1}^n \left\| \b d_i^m \right\|_2^2 \right)^{\frac{1}{2}} \sqrt{n} = \left( \frac{1}{n} \sum_{i=1}^n \left\| \b d_i^m \right\|_2^2 \right)^{\frac{1}{2}},
\end{align}
where (l) comes from the CBS inequality. Regarding (d) in \eqref{eq:ineq_tilde_mu}, by defining $\b D :=[D_{i,j}]_{i=1,j=1}^{n,d^m}:=\hat{\bm{\Phi}}^m_{d^m} - \bm{\Phi}^m_{d^m}$ we have
\begin{align}
    & \left\| \hat{\bm{\Phi}}^m_{d^m} - \bm{\Phi}^m_{d^m} \right\|_F^2 = \sum_{i=1}^n \sum_{j=1}^{d^m} D_{i,j}^2 \stackrel{(m)}{=} \sum_{i=1}^n \left\| \b d_i^m \right\|_2^2 \\
    & \stackrel{(n)}{\Rightarrow} \left( \frac{1}{n}\sum_{i=1}^n \left\| \b d_i^m \right\|_2^2 \right)^{\frac{1}{2}} = \left( \frac{1}{n}\left\| \hat{\bm{\Phi}}^m_{d^m} - \bm{\Phi}^m_{d^m} \right\|_F^2 \right)^{\frac{1}{2}} = \frac{1}{\sqrt{n}} \left\| \hat{\bm{\Phi}}^m_{d^m} - \bm{\Phi}^m_{d^m} \right\|_F,
\end{align}
where (m) comes from the definition of $\b d_i^m$ and (n) follows from multiplying $\frac{1}{n}$ and then taking square root. (e) in \eqref{eq:ineq_tilde_mu} is obtained by the assumption that $d_{k^m} \ge 1$ for all $m \in [M]$ and that $d_k = \prod_{i=1}^M d_{k^i}$. For (f) in \eqref{eq:ineq_tilde_mu}, dividing by $\frac{1}{\sqrt{n}}$ both sides of \eqref{eq:final_bound_mat_feature_map} in Lemma~\ref{concentration_adj} we have
for $\eta \in (0,1)$ and $\lambda^m_{d^m+1} \le 8 \left( \log 2 \right)\sqrt{\frac{d^m \log \left( \frac{16n}{\eta} \right)}{n}} $ with $n\ge \colLfive{\left( \frac{\eta}{8} \right)^{-\frac{1}{c^m}}}$, and
\begin{align}
    n \ge \max & \Big\{\tilde{n}^m_{0,A1(i)}(\eta), \tilde{n}^m_{0,A1(ii)}(\eta), \tilde{n}^m_{0,A2(i)}(\eta), \tilde{n}^m_{0,A2(ii)}(\eta), \tilde{n}^m_{0,L5}(\eta), \tilde{n}^m_{0,L6}(\eta), \tilde{n}_{0,\text{else-1}}^m(\eta), \\
    & \hspace{0.3cm} \tilde{n}_{0,\text{else-2}}^m \Big\}
\end{align}
that 
\begin{align}
     & \frac{1}{\sqrt{n}} \left \|  \hat{\bm{\Phi}}^m_{d^m} - \bm{\Phi}^m_{d^m} \right \|_F  \\
     & \le  312 \sqrt{2} c_1^m(n) d^m \left( \delta^m_{d^m} \right)^{-\frac{1}{2}}  \sqrt{\frac{\log \left ( \frac{864 n}{\eta} \right )}{n} } + 896 \frac{\left(d^m \right)^{\frac{1}{2}}\left(c_1^m(n)\right)^2}{\sqrt{c_0^m}} \left( \delta^m_{d^m} \right)^{-2} \frac{\log \left( \frac{432 n}{\eta} \right)}{n} \\
     & \quad + 1024 \frac{\left(d^m \right)^{\frac{1}{2}}\left(c_1^m(n)\right)^3}{\sqrt{c_0^m}} \left( \delta^m_{d^m} \right)^{-2} \frac{\log^{\frac{3}{2}} \left( \frac{432 n}{\eta} \right)}{n^{\frac{3}{2}}} + 64 \frac{\left(c_1^m(n)\right)^2}{\sqrt{c_0^m}} \left( \delta^m_{d^m} \right)^{-1} \frac{\left(d^m \right)^{\frac{1}{2}} \log \left( \frac{432 n}{\eta} \right)}{n} \\
     & \quad + 4 \frac{(d^m)^{\frac{1}{2}} c_1^m(n)}{\sqrt{c_0^m}} \sqrt{\frac{\log \left( \frac{144n}{\eta}  \right)}{n}} = D_{k^m} (\eta)  
\end{align}
holds with $\J^m$-probability at least $1-\eta$.

Leveraging the obtained results for $m\in [M]$ with a substitution $\eta \leftarrow \frac{\eta}{M}$ in a union bound, we got that \eqref{eq:ineq_tilde_mu} holds for $\lambda^m_{d^m+1} \le 8 \left( \log 2 \right) \sqrt{\frac{d^m \log \left( \frac{16Mn}{\eta}  \right)}{n}}$ with $n \ge \left( \frac{\eta}{8M} \right)^{-\frac{1}{c^m}}$ and
\begin{align}
    n& \ge \max_{m \in [M]} \Big\{ \tilde{n}^m_{0,A1(i)}\left( \frac{\eta}{M} \right), \tilde{n}^m_{0,A1(ii)}\left( \frac{\eta}{M} \right), \tilde{n}^m_{0,A2(i)}\left( \frac{\eta}{M} \right), \tilde{n}^m_{0,A2(ii)}\left( \frac{\eta}{M} \right), \tilde{n}^m_{0,L5}\left( \frac{\eta}{M} \right), \\
    & \quad \hspace{1.4cm} \tilde{n}^m_{0,L6}\left( \frac{\eta}{M} \right), \tilde{n}_{0,\text{else-1}}^m\left( \frac{\eta}{M} \right), \tilde{n}_{0,\text{else-2}}^m \Big\}. \label{eq:A2_joint_feature_map-n_part}
\end{align}
\end{proof}

The following statement is used in the discussion of the convergence rate after Theorem~\ref{errorbound_GHSIC1}.

\begin{lemma}[properties of $f$ in \eqref{eq:f:rate}]\label{lemma:f-properties}
For the function
\begin{align}
        f: \alpha \in \left( \frac{1+ \sqrt{6}}{2},\infty\right) \mapsto \frac{2\alpha +1}{6 \alpha+5} - \frac{1}{2\alpha} \in \R,
        \end{align}
(i) $f(\alpha) \in \Rp$ for $\alpha \in \left(\frac{1+ \sqrt{6}}{2},\infty\right)$.
(ii) $\lim_{\alpha\to \infty}f(\alpha)=\frac{1}{3}$,
(iii) $\lim_{\alpha \to \frac{1+ \sqrt{6}}{2} } f(\alpha)=0$, and (iv) $f$ is strictly increasing.
\end{lemma}
\begin{proof}(Lemma~\ref{lemma:f-properties}):
\begin{itemize}
\item[(i)] Positivity: Regarding the positivity of the function, using that $\alpha>0$, one has that 
\begin{align}
\frac{2\alpha +1}{6 \alpha+5} - \frac{1}{2 \alpha} > 0 
\Leftrightarrow \frac{2\alpha +1}{6 \alpha+5} > \frac{1}{2 \alpha}  
\Leftrightarrow 4 \alpha^2 + 2 \alpha > 6 \alpha + 5 \Leftrightarrow  4 \alpha^2 - 4 \alpha - 5 > 0
\end{align}
which holds for $\alpha \in \left(\frac{1+ \sqrt{6}}{2},\infty\right)$ as checked after \eqref{eq:alpha-constraint2}. 

\item[(ii)] $\lim_{\alpha \to \infty} f(\alpha)$:
\begin{align}
\lim_{\alpha \to \infty} \frac{2\alpha +1}{6 \alpha+5} - \frac{1}{2\alpha} = 
\lim_{\alpha \to \infty} \frac{4\alpha^2 + 2 \alpha-6\alpha-5}{(6\alpha+5)(2\alpha)} = 
\lim_{\alpha \to \infty}\frac{4 \alpha^2 - 4 \alpha - 5}{12 \alpha^2 + 10 \alpha} = 
\lim_{\alpha \to \infty} \frac{4\alpha^2}{12\alpha^2} = \frac{1}{3}.
\end{align}
\item[(iii)] $\lim_{\alpha \to \frac{1+ \sqrt{6}}{2}}  f(\alpha)$: One has
\begin{align}
\lim_{\alpha \to \frac{1+ \sqrt{6}}{2}} \frac{2\alpha +1}{6 \alpha+5} - \frac{1}{2\alpha} \stackrel{(a)}{=}\lim_{\alpha \to \frac{1+ \sqrt{6}}{2}}  \frac{4 \alpha^2 - 4 \alpha - 5}{12 \alpha^2 + 10 \alpha} \stackrel{(b)}{=} 0.
\end{align}
(a) follows from the reasoning in item (ii), (b) holds as the numerator goes to zero by item (i) [see the discussion after \eqref{eq:alpha-constraint2}] and the denominator goes to a positive constant when $\alpha \to \frac{1+ \sqrt{6}}{2} $.
\item[(iv)] Strictly increasing property: The statement is equivalent to showing that $f'(\alpha)>0$ for all $\alpha \in \left(\frac{1+ \sqrt{6}}{2},\infty\right)$. Considering this derivative, one has
\begin{align}
f'(\alpha)&= \left( \frac{4 \alpha^2 - 4 \alpha - 5}{12 \alpha^2 + 10 \alpha} \right)' = \frac{(8 \alpha-4)(12 \alpha^2 + 10 \alpha) - (4 \alpha^2 - 4 \alpha - 5)(24 \alpha + 10)}{\left( 12 \alpha^2 + 10 \alpha \right)^2}\\
&=\frac{96\alpha^3+80\alpha^2-48\alpha^2-40\alpha-96\alpha^3-40\alpha^2+96\alpha^2+40\alpha+120\alpha+50}{\left( 12 \alpha^2 + 10 \alpha \right)^2}\\
&=\frac{(96-96)\alpha^3+(80-48-40+96)\alpha^2+(-40+40+120)\alpha+50}{\left( 12 \alpha^2 + 10 \alpha \right)^2}\\
&= \frac{88\alpha^2+120\alpha+50}{\left(12 \alpha^2 + 10 \alpha \right)^2 }=2\frac{44\alpha^2+ 60\alpha+25}{\left(12 \alpha^2 + 10 \alpha \right)^2}.
\end{align}
The denominator of $f'(\alpha)$ is positive. The numerator $\alpha \mapsto 44\alpha^2+ 60\alpha+25$ is convex; its discriminant is $\Delta= 60^2 - 4 \times 44 \times 25=-800<0$, hence the numerator of $f'(\alpha)$ is also positive. Therefore, $f'(\alpha)$ is strictly positive for all $\alpha \in \left(\frac{1+ \sqrt{6}}{2},\infty\right)$.

\end{itemize}
\end{proof}

The following lemma provides some properties of tensor product in Hilbert space.

\begin{lemma}[Properties of tensor product in a Hilbert space]\label{lemma:properties-of-tensor-products}
Let $M\in \Z^+$. Let $(\mH_m)_{m=1}^M$ be Hilbert spaces; their tensor product is denoted by $\mH := \otimes_{m=1}^M \mH_m$. 
\begin{itemize}
    \item[(i)] Let $a_m \in \mH_m$ for $m\in [M]$. Then 
    \begin{align}
        \left\| \otimes_{m=1}^m a_m \right\|_\mH^2 & = \prod_{m=1}^M \left\| a_m \right\|_{\mH_m}^2. \label{eq:norm-of-tensor-products}
    \end{align}
    Note: Let $M\ge 2$. Let $a_m, b_m \in \mH_m$ for $m\in [M-1]$, and let $c_M \in \mH_M$. Then, applying \eqref{eq:norm-of-tensor-products} with a substitution $M=2$, $\mH_1 \leftarrow \otimes_{m=1}^{M-1} \mH_m$ and $\mH_2 \leftarrow \mH_M$, one gets that 
    \begin{align}
    \left\|\left(\otimes_{m=1}^{M-1} a_m - \otimes_{m=1}^{M-1} b_m\right) \otimes c_M\right\|_{\mH} & = \left\|\otimes_{m=1}^{M-1}  a_m - \otimes_{m=1}^{M-1} b_m\right\|_{\otimes_{m=1}^{M-1}\mH_m} \left\| c_M \right\|_{\mH_M}. \label{eq:norm-of-tensor-products-spec}
    \end{align}
    \item[(ii)] Let $a_m, b_m \in \mH_m$, $\left\| a_m \right\|_{\mH_m} \le A_m$ and $\left\| b_m \right\|_{\mH_m} \le B_m$ for $m \in [M]$. Then\footnote{The products $\prod_j A_j$ and $\prod_i B_i$ with empty index sets are defined to be $1$; specifically for $M=1$ \eqref{eq:diff-of-tensor-products} holds and reads as $\left\| a_1 - b_1\right\|_{\mH_1} \le \left\| a_1 - b_1\right\|_{\mH_1}$. \label{footnote:statement-for-M=1}}
     \begin{align}
    \left\| \otimes_{m=1}^M a_m - \otimes_{m=1}^M b_m \right\|_{\mH} \le \sum_{m=1}^M \left( \prod_{j=1}^{m-1} A_j \right) \left( \prod_{i=m+1}^M B_i \right) \left\| a_m - b_m \right\|_{\mH_m}. \label{eq:diff-of-tensor-products}
    \end{align}
    Note: Specifically, if $A_m=1$ for $m\in [M]$, then one gets
    \begin{align}
    \left\| \otimes_{m=1}^M a_m - \otimes_{m=1}^M b_m \right\|_{\mH} \le \sum_{m=1}^M \left( \prod_{i=m+1}^M B_i \right) \left\| a_m -  b_m \right\|_{\mH_m}. \label{eq:diff-of-tensor-products_Am1}
    \end{align}
\end{itemize}
\end{lemma}

\begin{proof}(of Lemma~\ref{lemma:properties-of-tensor-products})
\begin{itemize}
    \item[(i)] The statement can be shown as follows.
    \begin{align}
    \left\| \otimes_{m=1}^M a_m \right\|_\mH^2 & \stackrel{(a)}{=} \langle \otimes_{m=1}^M a_m,\otimes_{m=1}^M a_m \rangle_{\mH}  \stackrel{\eqref{eq:def:tensor-product}}{=} \prod_{m=1}^M \langle a_m, a_m\rangle_{\mH_m}\stackrel{(a)}{=}\prod_{m=1}^M \left\| a_m \right\|_{\mH_m}^2.
    \end{align}
    In (a) we used that in a Hilbert space the norm is induced by the inner product.
    
    \item[(ii)] We proceed by induction; for $M=1$ the assumption holds as discussed in footnote~\ref{footnote:statement-for-M=1}; hence we assume that the statement holds for $M-1$, and show it for $M$. For the l.h.s.\ of \eqref{eq:diff-of-tensor-products}, we can consider the following decomposition
\begin{align}
    \MoveEqLeft \left\| \otimes_{m=1}^M a_m - \otimes_{m=1}^M b_m \right\|_{\mH} \\
    & \stackrel{(a)}{=} \left\| \otimes_{m=1}^M a_m - \left( \otimes_{m=1}^{M-1} a_m \right) \otimes b_M + \left( \otimes_{m=1}^{M-1} a_m \right) \otimes b_M - \otimes_{m=1}^M b_m \right\|_{\mH} \\
    & \stackrel{(b)}{=} \left\| \left( \otimes_{m=1}^{M-1} a_m \right) \otimes \left( a_M - b_M \right) + \left( \otimes_{m=1}^{M-1} a_m - \otimes_{m=1}^{M-1} b_m \right) \otimes b_M \right\|_{\mH} \\
    & \stackrel{(c)}{\le} \underbrace{\left\| \left( \otimes_{m=1}^{M-1} a_m \right) \otimes \left( a_M - b_M \right)  \right\|_{\mH} }_{=:t_1} + \underbrace{\left\| \left( \otimes_{m=1}^{M-1} a_m - \otimes_{m=1}^{M-1} b_m \right) \otimes b_M \right\|_{\mH} }_{=:t_2},
\end{align}
where (a) was obtained by adding $\pm \left( \otimes_{m=1}^{M-1} a_m \right) \otimes b_M$, in (b) we pulled out $\otimes_{m=1}^{M-1} a_m$ and $b_m$ from the 1st and the 2nd term, respectively; (c) follows from the triangle inequality. Next we bound $t_1$ and $t_2$.
\begin{itemize}
    \item \textbf{Term $t_1$}: For $t_1$ we have
    \begin{align}
        t_1 & \stackrel{(a)}{=} \left( \prod_{m=1}^{M-1} \left\| a_m \right\|_{\mH_m} \right) \left\| a_M - b_M \right\|_{\mH_M} \stackrel{(b)}{\le}  \left( \prod_{m=1}^{M-1} A_m \right) \left\| a_M - b_M \right\|_{\mH_M} ,
    \end{align}
    where (a) holds by Lemma~\ref{lemma:properties-of-tensor-products} [see \eqref{eq:norm-of-tensor-products}], (b) is implied by $\left\| a_m \right\|_{\mH_m} \le A_m$ holding according to our assumption. 

    \item \textbf{Term $t_2$}: For $t_2$ we have
    \begin{align}
        t_2 \stackrel{(a)}{=} \left\| \otimes_{m=1}^{M-1} a_m - \otimes_{m=1}^{M-1} b_m \right\|_{\otimes_{m=1}^{M-1}\mH_m} \left\| b_M \right\|_{\mH_M} \stackrel{(b)}{\le} B_M \left\| \otimes_{m=1}^{M-1} a_m - \otimes_{m=1}^{M-1} b_m \right\|_{\otimes_{m=1}^{M-1}\mH_m},
    \end{align}
    where (a) is implied by Lemma~\ref{lemma:properties-of-tensor-products} [see \eqref{eq:norm-of-tensor-products-spec}] and (b) follows from our assumption $\left\| b_M \right\|_{\mH_M} \le B_M$.
\end{itemize}
Combining the results obtained for $t_1$ and $t_2$, we got that
\begin{align}
    \MoveEqLeft \left\| \otimes_{m=1}^M a_m - \otimes_{m=1}^M b_m \right\|_{\mH} \\
    & \le  \left( \prod_{m=1}^{M-1} A_m \right) \left\| a_M - b_M \right\|_{\mH_M} + B_M \left\| \otimes_{m=1}^{M-1} a_m - \otimes_{m=1}^{M-1} b_m \right\|_{\otimes_{m=1}^{M-1}\mH_m} \\
    & \stackrel{(a)}{\le} \left( \prod_{m=1}^{M-1} A_m \right) \left\| a_M - b_M \right\|_{\mH_M} + B_M \left[\sum_{m=1}^{M-1} \left( \prod_{j=1}^{m-1} A_j \right) \left( \prod_{i=m+1}^{M-1} B_i \right) \left\| a_m - b_m \right\|_{\mH_m} \right]\\
    & \stackrel{(b)}{=} \sum_{m=1}^M \left( \prod_{j=1}^{m-1} A_j \right) \left( \prod_{i=m+1}^M B_i \right) \left\| a_m - b_m \right\|_{\mH_m}.
\end{align}
In (a) we used the validity of \eqref{eq:diff-of-tensor-products} (holding by the induction assumption); (b) holds by taking $B_M$ inside the sum, using footnote~\ref{footnote:statement-for-M=1} for the first term, and merging the two terms.

\end{itemize}

\end{proof}

The following lemma is about the isometric property of the truncated feature maps.

\begin{lemma}[Isometric property of the truncated feature maps]\label{lemma:isometric_embedding}
Let
\begin{align}
\tilde{\mu}_{k^m,d^m} \left(\hat{\P}^m_n\right) &= \frac{1}{n} \sum_{i=1}^n k^m_{d^m} \left(\cdot, X^m_i\right) \in \mH_{k^m}, &
\tilde{\mu}_{k,\b d}\left(\hat{\P}_n\right) &= \frac{1}{n} \sum_{i=1}^n \otimes_{m=1}^M  k^m_{d^m} (\cdot,X^m_i)  \in \mH_k,\\
\breve{\mu}_{k^m}\left(\hat{\P}^m_n\right) &= \frac{1}{n} \sum_{i=1}^n \phi_{k^m,d^m}^m (X_i^m) \in \R^{d^m}, & \breve{\mu}_{k}\left(\hat{\P}_n\right) &= \frac{1}{n} \sum_{i=1}^n \otimes_{m=1}^M   \phi_{k^m,d^m}^{m} (X^m_i) \in \R^{d^1 \times \cdots \times d^M}\label{eq:def_breve_mu_km}
\end{align}
with (i) $\tilde{\mu}_{k^m,d^m}\left(\hat{\P}^m_n\right)$, $\tilde{\mu}_{k,\b d}\left(\hat{\P}_n\right)$, and $\breve{\mu}_k(\hat{\P}_n)$ recalled from \eqref{eq:mu-tilde-and-truncated}, \eqref{eq:def_joint_tilde_mu} and \eqref{eq:def_joint_feature_est}, respectively, (ii) $k^m_{d^m}$ specified in \eqref{eq:def:km} [$k^m_{d^m}\left(\cdot,z\right) := P^{m}_{d^m} \left(\hat{C}_{k^m}\right)k^m(\cdot,z)\in \mH_{k^m}$ with $P^m_{d^m}\left(\hat{C}_{k^m}\right):=\sum_{i=1}^{d^m} \hat{f}^m_i \otimes \hat{f}^m_i$ defined in \eqref{eq:projection-op}, $z\in \X^m$], and (iii) $\phi_{k^m,d^m}^m (X_i^m) \in \R^{d^m}$ defined in \eqref{eq:def_feature_est_by_gram}. Then we have \begin{align}\label{isometric_embedding}
        \left \| \tilde{\mu}_{k,\b d}(\hat{\P}_n)  - \otimes_{m=1}^M  \tilde{\mu}_{k^m,d^m}(\hat{\P}^m_n) \right \|_{\mH_k} = \left \| \breve{\mu}_k (\hat{\P}_n) - \otimes_{m=1}^M  \breve{\mu}_{k^m} (\hat{\P}^m_n) \right \|_2 .
    \end{align}
\end{lemma}

\begin{proof}(of Lemma~\ref{lemma:isometric_embedding}):
For all $m\in [M]$, using Lemma~3.4 by \citet{tang2013universally}, with the substitution
\begin{align}
\Big(\mathcal{K}_{\mathcal{H},n} &= \frac{1}{n}\sum_{i=1}^n \kappa(\cdot,X_i) \otimes \kappa(\cdot,X_i), \hat{\mathcal{P}}_d = \sum_{s=1}^d v^{(s)} \otimes v^{(s)}, \Phi(X_i) \Big)\leftarrow\\
\Big(\hat{C}_{k^m} &= \frac{1}{n} \sum_{i=1}^n k^m\left(\cdot,X^m_i\right) \otimes k^m\left(\cdot,X^m_i\right), P^m_{d^m}\left(\hat{C}_{k^m}\right) = \sum_{i=1}^{d^m} \hat{f}^m_i \otimes \hat{f}^m_i, k^m(\cdot,X_i)\Big),
\end{align}
one gets that there exist isometric isomorphisms 
\begin{align}
\iota^m&: S^m:=\Span\left(\hat{f}_i^m\,:\,i\in \left[d^m\right]\right) \subset \mH_{k^m} \to \R^{d^m} \quad (m\in [M])
\end{align}
such that
\begin{align}
    \phi^m_{k^m,d^m}(X_i^m) = \iota^m \left( k^m_{d^m}(\cdot, X_i^m) \right) \t{ for all $i\in [n]$, and $m\in [M]$}. \label{eq:isom}
\end{align}
The sequence $(\iota^m)_{m=1}^M$ induces an isometric isomorphism
\begin{align}
\iota &:= \otimes_{m=1}^M \iota^m: S:=\otimes_{m=1}^M S^m:=\Span\left\{\otimes_{m=1}^M s^m\,:\, s^m \in S^m, m\in [M]\right\}\\
& \hspace{5.2cm}\subset \otimes_{m=1}^M \mH_{k^m} \to \R^{d^1 \times \ldots \times d^M}.
\end{align}
Particularly, as $\left(S^m,\langle \cdot,\cdot\rangle_{\mH_{k^m}}\right)$-s ($m\in [M]$) are Hilbert spaces, their tensor product $S$ is also a Hilbert space. Using the fact that a linear operator is uniquely determined by its values on the basis elements of its domain ($S$), and that the elements $\left(\otimes_{m=1}^M \hat{f}_{i_m}^m\right)_{i_1=1,\ldots,i_M=1}^{d^1,\ldots,d^M}$ form a basis of $S$ [specifically, this implies that $\Dim(S) = \prod_{m=1}^M d^m$], we prescribe $\iota$ as
\begin{align}
\iota\left(\otimes_{m=1}^M \hat{f}_{i_m}^m\right):= \otimes_{m=1}^M \iota^m\left(\hat{f}_{i_m}^m\right) \quad \left(i_m \in [d^m], m\in M\right). \label{eq:iota-definition}
\end{align}
This means that for an arbitrary point $s = \sum_{i_1=1}^{d^1} \ldots \sum_{i_M=1}^{d^M} a_{i_1,\ldots,i_M}\otimes_{m=1}^M \hat{f}_{i_m}^m \in S$, one has 
\begin{align}
\MoveEqLeft
\iota\left(\sum_{i_1=1}^{d^1} \ldots \sum_{i_M=1}^{d^M} a_{i_1,\ldots,i_M}\otimes_{m=1}^M \hat{f}_{i_m}^m \right) = \\
&\hspace{-0.6cm} = \sum_{i_1=1}^{d^1} \ldots \sum_{i_M=1}^{d^M} a_{i_1,\ldots,i_M}\iota\left(\otimes_{m=1}^M \hat{f}_{i_m}^m \right) \stackrel{\eqref{eq:iota-definition}}{=}  \sum_{i_1=1}^{d^1} \ldots \sum_{i_M=1}^{d^M} a_{i_1,\ldots,i_M} \otimes_{m=1}^M \iota^m\left(\hat{f}_{i_m}^m\right), \label{eq:iota-definition-conseq}
\end{align}
and specifically for an element $s = \otimes_{m=1}^M s^m \in S$ with $s^m = \sum_{i_m=1}^{d^m} a_{i_m} \hat{f}_{i_m}^m$ ($m\in [M]$),  
\begin{align}
\iota(s) & \stackrel{(a)}{=} \iota\left(\otimes_{m=1}^M s^m\right)
\stackrel{(b)}{=} \iota\left(\otimes_{m=1}^M \left(\sum_{i_m=1}^{d^m} a_{i_m} \hat{f}_{i_m}^m \right) \right)
 \stackrel{(c)}{=} \iota\left(\sum_{i_1=1}^{d^1} \ldots \sum_{i_M=1}^{d^M}\left(\prod_{m=1}^M a_{i_m}\right) \otimes_{m=1}^M \hat{f}_{i_m}^m \right)\\
 &\stackrel{(d)}{=} \sum_{i_1=1}^{d^1} \ldots \sum_{i_M=1}^{d^M}\left(\prod_{m=1}^M a_{i_m}\right) \otimes_{m=1}^M \iota^m\left(\hat{f}_{i_m}^m \right) \stackrel{(e)}{=} \otimes_{m=1}^M \left[ \sum_{i^m=1}^{d^m}a_{i_m}\iota^m\left(\hat{f}_{i_m}^m \right)\right]\\
 &\stackrel{(f)}{=} \otimes_{m=1}^M \left[ \iota^m\left(\sum_{i^m=1}^{d^m}a_{i_m}\hat{f}_{i_m}^m \right)\right] 
 \stackrel{(g)}{=} \otimes_{m=1}^M \iota^m\left(s^m\right).
\end{align}
In (a) we used the definition of $s$, (b) follows from the basis expansion of $\left(s^m\right)_{m=1}^M$-s, in (c) the multi-linearity of $\otimes$ was leveraged, (d) comes from \eqref{eq:iota-definition-conseq}, in (e) the the multi-linearity of $\otimes$ was used again, (f) follows from the linearity of $\left(\iota^m\right)_{m=1}^M$-s, in (g) the basis expansions of $\left(s^m\right)_{m=1}^M$-s were used.

Next we show that 
$\iota$ is an \tb{isometric isomorphism}: beyond the linearity this means that 
    \begin{enumerate}
        \item $\iota$ is an isometry, in other words
        $\left\|\iota(s)\right\|_2 = \left\|s\right\|_{\otimes_{m=1}^M \mH_{k^m}}$ for all $s\in S$.
        Indeed, let us consider an arbitrary point $s\in S$ and let its basis expansion be $s = \sum_{i_1=1}^{d^1} \ldots \sum_{i_M=1}^{d^M} a_{i_1,\ldots,i_M}\otimes_{m=1}^M \hat{f}_{i_m}^m$. In this case, 
        \begin{align}
        \left\| \iota(s) \right\|_2^2 & \stackrel{\eqref{eq:iota-definition-conseq}}{=} 
        \left\| \sum_{i_1=1}^{d^1} \ldots \sum_{i_M=1}^{d^M} a_{i_1,\ldots,i_M} \otimes_{m=1}^M \iota^m\left(\hat{f}_{i_m}^m\right) \right\|^2\\
        & \stackrel{\eqref{eq:2-norm-on-tensors}}{=} \left\langle \sum_{i_1=1}^{d^1} \ldots \sum_{i_M=1}^{d^M} a_{i_1,\ldots,i_M} \otimes_{m=1}^M \iota^m\left(\hat{f}_{i_m}^m\right), \sum_{j_1=1}^{d^1} \ldots \sum_{j_M=1}^{d^M} a_{j_1,\ldots,j_M} \otimes_{m=1}^M \iota^m\left(\hat{f}_{j_m}^m\right) \right \rangle \\
        &\stackrel{(a)}{=}
        \sum_{i_1=1}^{d^1} \ldots \sum_{i_M=1}^{d^M}  \sum_{j_1=1}^{d^1} \ldots \sum_{j_M=1}^{d^M} a_{i_1,\ldots,i_M} a_{j_1,\ldots,j_M}  \left \langle \otimes_{m=1}^M\iota^m\left(\hat{f}_{i_m}^m\right), \otimes_{m=1}^M \iota^m\left(\hat{f}_{j_m}^m\right)\right \rangle\\
        &\stackrel{\eqref{eq:def:tensor-product}}{=}
        \sum_{i_1=1}^{d^1} \ldots \sum_{i_M=1}^{d^M}  \sum_{j_1=1}^{d^1} \ldots \sum_{j_M=1}^{d^M} a_{i_1,\ldots,i_M} a_{j_1,\ldots,j_M} \prod_{m=1}^M \left \langle \iota^m\left(\hat{f}_{i_m}^m\right), \iota^m\left(\hat{f}_{j_m}^m\right)\right \rangle\\
        & \stackrel{(b)}{=} \sum_{i_1=1}^{d^1} \ldots \sum_{i_M=1}^{d^M}  \sum_{j_1=1}^{d^1} \ldots \sum_{j_M=1}^{d^M} a_{i_1,\ldots,i_M} a_{j_1,\ldots,j_M} \prod_{m=1}^M \left \langle \hat{f}_{i_m}^m, \hat{f}_{j_m}^m \right\rangle_{\mH_{k^m}}\\
        &\stackrel{\eqref{eq:def:tensor-product}}{=}
        \sum_{i_1=1}^{d^1} \ldots \sum_{i_M=1}^{d^M}  \sum_{j_1=1}^{d^1} \ldots \sum_{j_M=1}^{d^M} a_{i_1,\ldots,i_M} a_{j_1,\ldots,j_M}  \left \langle \otimes_{m=1}^M \hat{f}_{i_m}^m, \otimes_{m=1}^M\hat{f}_{j_m}^m \right\rangle_{\otimes_{m=1}^M\mH_{k^m}}\\
        &\stackrel{(a)}{=}
          \left \langle \sum_{i_1=1}^{d^1} \ldots \sum_{i_M=1}^{d^M}  a_{i_1,\ldots,i_M} \otimes_{m=1}^M \hat{f}_{i_m}^m, \sum_{j_1=1}^{d^1} \ldots \sum_{j_M=1}^{d^M}  a_{j_1,\ldots,j_M}\otimes_{m=1}^M\hat{f}_{j_m}^m \right\rangle_{\otimes_{m=1}^M\mH_{k^m}}\\
          &\stackrel{(c)}{=} 
          \left\| \sum_{i_1=1}^{d^1} \ldots \sum_{i_M=1}^{d^M}  a_{i_1,\ldots,i_M} \otimes_{m=1}^M \hat{f}_{i_m}^m \right\|^2_{\otimes_{m=1}^M\mH_{k^m}} \stackrel{(d)}{=} \left\|s\right\|^2_{\otimes_{m=1}^M\mH_{k^m}}.
        \end{align}  
        (a) follows from the linearity of the inner product, (b) comes from the isometric property of $\left(\iota^m\right)_{m=1}^M$-s, in (c) we used the fact that in a Hilbert space the norm is induced by the inner product, the basis expansion of $s$ was leveraged in (d).
        \item $\iota$ is injective. For the linear $\iota$ this means that $\Kernel(\iota)=\{0\}$, in other words $\iota(s) = 0 \Rightarrow s=0$.  Indeed, assume that $\iota(s) = 0$. By the previous point, this implies that $0 = \left\|\iota(s)\right\| = \left\|s\right\|_{\otimes_{m=1}^M \mH_{k^m}}$; hence $s=0$.
        \item $\iota$ is surjective: This means that 
        for all $\mathcal{A} \in \R^{d^1\times \ldots \times d^M}$, there exists $s \in S$ such that $\iota(s) = \mathcal{A}$. Indeed, by (i) using the fact that $\Dim(S) = \prod_{m=1}^M d^m$, (ii) the isomorpism theorem $\Image(\iota) \cong S / \Kernel(\iota)$, and (iii) $\Kernel(\iota) = \{0\}$ (shown in the previous bullet), one gets that $\Image(\iota) = \R^{d^1\times \ldots \times d^M}$.
    \end{enumerate}

With the shorthand $\phi_i^m := \phi^m_{k^m,d^m}(X_i^m)$ and $k_i^m := k^m(\cdot,X_i^m)$, our goal now is to prove that 
\begin{align}
\hspace{-0.5cm}
\left\|
\frac{1}{n}\sum_{i=1}^n \otimes_{m=1}^M k_i^m - \otimes_{m=1}^M \left[\frac{1}{n}\sum_{i=1}^n k_i^m\right]\right\|_{\mH_k} & \hspace{-0.2cm} = \left\| \frac{1}{n}\sum_{i=1}^n \otimes_{m=1}^M\phi_i^m - \otimes_{m=1}^M \left[\frac{1}{n} \sum_{i=1}^n \phi_i^m \right]\right\|_2. \label{eq:isom-goal-in-short}
\end{align} Focusing in the respective norms on the 
\begin{itemize}
    \item \tb{first terms}:
        \begin{align}
                \frac{1}{n}\sum_{i=1}^n \otimes_{m=1}^M\phi_i^m& \stackrel{(a)}{=} \frac{1}{n}\sum_{i=1}^n \otimes_{m=1}^M \left[\iota^m\left(k_i^m\right)\right]
                \stackrel{(b)}{=} \frac{1}{n}\sum_{i=1}^n \iota \left(\otimes_{m=1}^Mk_i^m\right)\\
                &\stackrel{(c)}{=} \iota \left(\frac{1}{n}\sum_{i=1}^n \otimes_{m=1}^Mk_i^m\right). \label{eq:isom:1st-terms}
        \end{align}
        In (a) we used the relation $\phi_i^m=\iota^m\left(k_i^m\right)$ of \eqref{eq:isom}, (b) follows from $\iota = \otimes_{m=1}^M \iota_m$, the linearity of $\iota$ implies (c).
    \item \tb{second terms}:
        \begin{align}
            \otimes_{m=1}^M \left[\frac{1}{n} \sum_{i=1}^n \phi_i^m\right]& \stackrel{(a)}{=} \otimes_{m=1}^M \left[\frac{1}{n} \sum_{i=1}^n \iota^m\left(k_i^m\right)\right]
            \stackrel{(b)}{=} \otimes_{m=1}^M \left[ \iota^m\left(\frac{1}{n} \sum_{i=1}^nk_i^m\right)\right]\\
            &\stackrel{(c)}{=} \iota\left(\otimes_{m=1}^M \left[ \frac{1}{n} \sum_{i=1}^nk_i^m\right]\right).
            \label{eq:isom:2nd-terms}
        \end{align}
        In (a) we used the relation $\phi_i^m=\iota^m\left(k_i^m\right)$ of \eqref{eq:isom}, (b) is implied by the linearity of $\iota^m$-s, (c) follows from $\iota = \otimes_{m=1}^M \iota^m$.
\end{itemize}
Substituting the obtained results \eqref{eq:isom:1st-terms} and \eqref{eq:isom:2nd-terms} to \eqref{eq:isom-goal-in-short}, we got that 
\begin{align}
\MoveEqLeft \left\| \frac{1}{n}\sum_{i=1}^n \otimes_{m=1}^M\phi_i^m - \otimes_{m=1}^M \left[\frac{1}{n} \sum_{i=1}^n \phi_i^m \right]\right\|_2 = \left\| \iota \left(\frac{1}{n}\sum_{i=1}^n \otimes_{m=1}^Mk_i^m\right) - \iota\left(\otimes_{m=1}^M \left[ \frac{1}{n} \sum_{i=1}^nk_i^m\right]\right)\right\|_2 \\
&\stackrel{(a)}{=} \left\| \iota \left(\frac{1}{n}\sum_{i=1}^n \otimes_{m=1}^Mk_i^m - \otimes_{m=1}^M \left[ \frac{1}{n} \sum_{i=1}^nk_i^m\right]\right)\right\|_2\\
& \stackrel{(b)}{=} \left\| \frac{1}{n}\sum_{i=1}^n \otimes_{m=1}^Mk_i^m - \otimes_{m=1}^M \left[ \frac{1}{n} \sum_{i=1}^nk_i^m\right]\right\|_{\mH_k}.
\end{align}
In (a) we used the linearity of $\iota$, (b) follows from the isometric property of $\tau$. 
\end{proof}

The following lemma provides upper and lower bound of the Lipschitz constant in Lemma~\ref{bound_psd_trans}.

\begin{lemma}[Upper and lower bound of the Lipschitz constant in Lemma~\ref{bound_psd_trans}]\label{lemma:bound_Cm} Let $(X^m_i)_{i=1}^n$ be i.i.d.\ random variables on a compact metric space $\X^m$. Let the Gram matrix $\b K^m = [K^m_{i,j}]_{i,j=1}^{n} = \left [ k^m(X^m_i,X^m_j) \right ]_{i,j=1}^{n}\in \R^{n \times n}$ be associated to a continuous kernel $k^m:\X^m \times \X^m \to [0,1]$ defined on  $\X^m$, with corresponding adjacency matrix $\b A^m \in \R^{n \times n}$ as specified in \eqref{eq:A^m}. Let
\begin{align}
    C^m= c_2^m \log \left( 2 + \log\left ( \frac{\left \| \b A^m \right \| + \left \| \b K^m \right \| }{\left \| \b A^m - \b K^m \right \| }  \right )  \right)
\end{align}
for a constant $c_2^m > 0$. 
\begin{itemize}
    \item Lower bound: Assume that $c_2^m \ge 2$. Then
          \begin{align}
            2 \log 2 \le C^m, \label{eq:bound_Cm:lower}
          \end{align}
    \item Upper bound: Let $c_1^m(n)=c_2^m \log \left( 2 + \log(2n) \right)$, in line with Assumption~\ref{assum_lambda_delta}(ii). Suppose that Assumption~\ref{assum_Delta}(ii) holds with $C_1^m > 0$ specified therein. Assume that $C_1^m \ge 2 \sqrt{2}$. Then,  for any $\eta \in (0,1)$ and $n \ge \max \left\{ n^m_{0,(ii)}\left( \frac{\eta}{2}\right), \frac{8}{\eta}, e^{\frac{1}{\left( 2C_1^m - 4\sqrt{2} \right)^2}} \right\}$ with $n^m_{0,(ii)}\left(\eta\right)$ defined in Assumption~\ref{assum_Delta}(ii), it holds that 
    \begin{align}
        \J^m\left(C^m \le c_1^m(n)\right) \ge 1-\eta, \label{eq:bound_Cm:upper}
    \end{align}
    where $\J^m$ stands for the joint distribution of $(X_i^m)_{i=1}^n$ and the entries of $\b A^m$. 
\end{itemize}
\end{lemma}

\begin{proof}(of Lemma~\ref{lemma:bound_Cm}): 

\begin{itemize}
    \item \textbf{Lower bound}: For the lower bound of $C^m$, we have
    \begin{align}
        C^m \ge c_2^m \log 2 \ge 2 \log 2\label{eq:C^m-bound}
    \end{align}
    by $\left \| \b A^m- \b K^m \right \| \le \left \| \b A^m \right \| + \left \| \b K^m \right \|$ and the assumption that $c_2^m \ge 2$. 

    \item \textbf{Upper bound}: For the upper bound, we have
    \begin{align}\label{upper_bound_Cm}
        C^m &\stackrel{(a)}{=} c_2^m \log \left( 2 + \log \left( \left \| \b A^m \right \| + \left \| \b K^m \right \|  \right) - \log \left( \left \| \b A^m - \b K^m \right \| \right) \right) \\
        & \stackrel{(b)}{\le} c_2^m \log \left( 2 + \log \left( \left \| \b A^m \right \| + \left \| \b K^m \right \|  \right) \right) \stackrel{(c)}{\le} c_2^m \log\left( 2 + \log (2n) \right) \stackrel{(d)}{=} c_1^m(n), 
    \end{align}
    where we applied the definition of $C^m$ and $c_1^m(n)$ in (a) and (d), respectively.
    Regarding (b), we want to show $\left \| \b A^m - \b K^m \right \| \ge 1$, which implies that $\log\left(\left \| \b A^m - \b K^m \right \|\right)\ge 0$ and hence the claimed inequality holds by the monotonicity of the $\log$ function; we will discuss (c) after showing (b). 
    
    To prove the inequality $\left \| \b A^m - \b K^m \right \| \ge 1$, let $\b D^{m} := \b A^m - \b K^m = \left[ D_{i,j}^{m} \right]_{i,j=1}^n \in \R^{n \times n}$. 
        \begin{itemize}
            \item \textbf{Fixed $(X^m_i)_{i=1}^n=(x^m_i)_{i=1}^n$, randomness in $\b A^m$}: The matrix $\b D^{m}$ satisfies the following three properties:  (i) $D_{i,j}^m$ are independent, (ii) $\left| D_{i,j}^{m} \right| \le \left | A_{i,j}^m \right | + \left | K_{i,j}^m \right | \le 2$, (iii) $\E \left(D_{i,j}^{m}\right) = 0$ for all $i,j\in [n]$. Hence, by Lemma~\ref{lowerbound_A_diff_K} [see \eqref{eq:lowerbound_A_diff_K}] for any $\eta \in \left(0,2 \right)$ and $n\ge \frac{8}{\eta}$, we get that with $\Q^m$-probability at least $1-\frac{\eta}{2}$ (with $\Q^m$ denoting the distribution of the entries in $\b A^m$ given the latent positions $(X_i^m)_{i=1}^n = (x_i^m)_{i=1}^n$) (a) in \eqref{eq:lowerbound_Dm} holds, and
            \begin{align}\label{eq:lowerbound_Dm}
            \left \| \b D^{m} \right \|  \stackrel{(a)}{\ge}  \E \left( \left \| \b D^{m} \right \| \right) - \sqrt{32 \log(n)} \stackrel{(b)}{\ge} 2\bar{\sigma}^m \sqrt{n} - \sqrt{32 \log(n)}.
            \end{align} 
            Indeed, (b) in \eqref{eq:lowerbound_Dm} can be obtained as follows:
            \begin{align}
            \E \left\| \b D^m \right\| & \stackrel{(c)}{\ge}  \frac{1}{\sqrt{n}} \E \left\| \b D^m \right\|_F \stackrel{(d)}{=} \E \sqrt{\frac{1}{n} \sum_{i=1}^n \sum_{j=1}^n \left( A_{i,j}^m -K_{i,j}^m \right)^2}\\
            &\hspace{-0.7cm} = \E \sqrt{n \frac{1}{n^2} \sum_{i=1}^n \sum_{j=1}^n \left( A_{i,j}^m -K_{i,j}^m \right)^2} 
            \stackrel{(e)}{\ge} \E \left( \sqrt{n} \frac{1}{n^2} \sum_{i=1}^n \sum_{j=1}^n \left| A_{i,j}^m -K_{i,j}^m \right| \right) \\
            &\hspace{-0.7cm}\stackrel{(f)}{=} \sqrt{n} \frac{1}{n^2} \sum_{i=1}^n \sum_{j=1}^n \E \left( \left| A_{i,j}^m -K_{i,j}^m \right| \right)
            \stackrel{(g)}{=} 2 \sqrt{n} \underbrace{\frac{1}{n^2} \sum_{i=1}^n \sum_{j=1}^n K_{i,j}^m(1-K_{i,j}^m)}_{=:\bar{\sigma}^m= \bar{\sigma}^m(x_1^m,x_2^m,\ldots,x_n^m)} = 2\bar{\sigma}^m \sqrt{n},
            \end{align}
        where (c) is implied by $\left\| \b D^m \right\| \ge \frac{1}{\sqrt{n}} \left\| \b D^m \right\|_F$ \citep[(2.3.7) on page~72]{golub2013matrix}, (d) comes from the definition of the Frobenius norm and that of $\b D^m$, (e) follows from $\sqrt{\frac{1}{n} \sum_{i=1}^n x_i^2} \ge \frac{1}{n} \sum_{i=1}^n x_i$ for $x_i \ge 0$ \citep[(2.36) on page~36]{steele2004cauchy}, (f) is implied by the additivity of the expectation and (g) is obtained by
        \begin{align}
            \E \left( \left| A_{i,j}^m -K_{i,j}^m \right| \right) \stackrel{(i)}{=} K_{i,j}^m(1-K_{i,j}^m) + (1-K_{i,j}^m)K_{i,j}^m = 2K_{i,j}^m(1-K_{i,j}^m),
        \end{align}
        where (i) follows by using that $A^m_{i,j} \sim \mathrm{Bernoulli}\left(K_{i,j}^m\right)$.

        \item \textbf{Randomness in $(X_i^m)_{i=1}^n$ and $\b A^m$}: In the previous bullet, we showed that in case of fixed $(X_i^m)_{i=1}^n=(x_i^m)_{i=1}^n$, for any $\eta \in (0,2)$ with $n\ge \frac{8}{\eta}$, the inequality \eqref{eq:lowerbound_Dm} holds with $\Q^m$-probability at least $1-\frac{\eta}{2}$. By Assumption~\ref{assum_Delta}(ii) we have that for any $\eta \in (0,2)$ and $n \ge n^m_{0,(ii)} \left( \frac{\eta}{2}\right)$, $\bar{\sigma}^m > C_1^m \sqrt{\frac{\log (n)}{n}}$ holds with $\left(\P^m\right)^{\otimes n}$-probability at least $1-\frac{\eta}{2}$. Define the following events:
        \begin{align}
            A & = \left \{ \left( (x^m_i)_{i=1}^n, \b A^m \right): - \left \| \b D^m \right \| - \sqrt{32 \log(n)} \le -2C_1^m \sqrt{\log(n)}  \right \}\\
             &= \left \{ \left( (x^m_i)_{i=1}^n, \b A^m \right):  \underbrace{- \left \| \b D^m \right \| - \sqrt{32 \log(n)}}_{=:f(x,y)} \le \underbrace{2\sqrt{n}}_{=:g(x)} \underbrace{\left( -C_1^m \sqrt{\frac{\log(n)}{n}}\right)}_{=:b}  \right \},\\
            B' & = \left \{ (x^m_i)_{i=1}^n : \bar{\sigma}^m > C_1^m \sqrt{\frac{\log (n)}{n}} \right \} = \left \{ (x^m_i)_{i=1}^n : -\bar{\sigma}^m < -C_1^m \sqrt{\frac{\log (n)}{n}} \right \} \\
            & \subseteq B = \left \{ (x^m_i)_{i=1}^n : -\bar{\sigma}^m \le -C_1^m \sqrt{\frac{\log (n)}{n}} \right \} \\
            & \quad \quad = \left \{ (x^m_i)_{i=1}^n : \underbrace{-\infty}_{=:a} < \underbrace{-\bar{\sigma}^m}_{=:S_0(x)} \le \underbrace{-C_1^m \sqrt{\frac{\log (n)}{n}}}_{=b} \right \}, \\
            C & = \left \{ \left( (x^m_i)_{i=1}^n, \b A^m \right): \left \| \b D^m  \right \|  \ge 2\bar{\sigma}^m \sqrt{n}-\sqrt{32 \log(n)}, (x^m_i)_{i=1}^n \in B \right\}\\
            & =\left \{ \left( (x^m_i)_{i=1}^n, \b A^m \right): -\left \| \b D^m  \right \| - \sqrt{32 \log(n)}  \le -2\bar{\sigma}^m \sqrt{n}, (x^m_i)_{i=1}^n \in B \right\}\\
            & = \left \{ \left( (x^m_i)_{i=1}^n, \b A^m \right): \underbrace{-\left \| \b D^m  \right \| - \sqrt{32 \log(n)} }_{=f(x,y)} \le \underbrace{2\sqrt{n}}_{=g(x)} \underbrace{\left(-\bar{\sigma}^m \right)}_{=S_0(x)} , (x^m_i)_{i=1}^n \in B \right\},
        \end{align}
        where notice that in the transition from $B'$ to $B$ we changed '$<$' to $\le$',
        and $C \subseteq A$. With these notations at hand, 
        \begin{enumerate}
            \item For $n \ge \frac{8}{\eta}$, we have
                \begin{align*}
                \Q^m \left(C \mid \left(x_i^m \right)_{i=1}^n \right) & \ge 1-\frac{\eta}{2},
                \end{align*}
                and
            \item for $n \ge n^m_{0,(ii)} \left( \frac{\eta}{2}\right)$, we have 
            \begin{align}
            \left(\P^m \right)^{\otimes n}(B) &  \stackrel{(a)}{\ge} \left(\P^m \right)^{\otimes n}(B') \stackrel{(b)}{\ge} 1-\frac{\eta}{2}.
        \end{align}
        In (a) we used that $B' \subseteq B$ and in (b) the previously established probabilistic bound on $B'$.
            \end{enumerate}
        Therefore, applying  Lemma~\ref{lemm:combine_probs}\ref{group:combine_probs1} with the notations $f$, $g$, $S_0$, $a$ and $b$ introduced above, and with $\delta_1 = \delta_2 = \frac{\eta}{2}$, we get that  for any $\eta \in (0,2)$ and $n\ge \max \left\{ n^m_{0,(ii)}\left( \frac{\eta}{2}\right), \frac{8}{\eta} \right\}$
        \begin{align}
            (\J^m)(A) \ge  1-\left(\frac{\eta}{2} + \frac{\eta}{2} \right) = 1-\eta, 
        \end{align} 
        where $\J^m$ denotes the joint distribution of $(X_i^m)_{i=1}^n$ and the entries of $\b A^m$. 
        This means that for any $\eta \in \left(0,1 \right)$ and $n\ge \max \left\{ n^m_{0,(ii)} \left( \frac{\eta}{2}\right), \frac{8}{\eta} \right\}$ one has 
        \begin{align}
            \J^m \left(\left \| \b D^{m} \right \| \ge 2 C_1^m \sqrt{\log(n)} -  \sqrt{32\log(n)}\right) \ge 1-\eta, \label{eq:D^m-lower-bound}
        \end{align}
        
        \end{itemize}
        Hence, the inequality $\left \| \b D^m \right \|\ge 1$ for (b) in \eqref{upper_bound_Cm} can be guaranteed with $\J^m$-probability at least $1-\eta$ by the assumption $C_1^m > 2 \sqrt{2}$ and $n$ such that for the obtained lower bound of  $\left \| \b D^m \right \|$ in \eqref{eq:D^m-lower-bound}, beyond $n\ge \max \left\{ n^m_{0,(ii)}\left( \frac{\eta}{2}\right), \frac{8}{\eta} \right\}$ used in \eqref{eq:lowerbound_Dm}
        \begin{align}
        \left(2C_1^m - 4\sqrt{2}\right) \sqrt{ \log (n)} \ge 1  \Leftrightarrow n \ge e^{\frac{1}{\left( 2C_1^m - 4\sqrt{2} \right)^2}} 
        \label{eq:D^m-lower-bound->n-constraint}
        \end{align}
        also holds.
        
        Regarding (c) in \eqref{upper_bound_Cm}, we have 
        \begin{align}
            \left \| \b A^m \right \| \stackrel{(a)}{\le} \left \| \b A^m \right \|_F = \sqrt{\sum_{i=1}^n \sum_{j=1}^n \left | A^m_{i,j} \right |^2  } \le \sqrt{\sum_{i=1}^n \sum_{j=1}^n 1 } = n,
        \end{align}
        where (a) is implied by \citep[(2.3.7) on page~72]{golub2013matrix} and similarly $\left \| \b K^m \right \| \le n$; hence the claimed inequality (c) in \eqref{upper_bound_Cm} follows from the monotonicity of the $\log$ function.
\end{itemize}
\end{proof}

\begin{table}%[h!]
    \caption{Combination of probability events using Lemma~\ref{lemm:combine_probs}. For all probability events we use the subtitution $(\X, \Y) \leftarrow \left(\X^m,\{0,1\}^{\binom{n}{2} }\right)$.
    }
    \label{prob_evnt_table}
    \begin{center}
    \begin{tabular}{@{}l@{\hspace{0.3cm}}l@{\hspace{0.3cm}}l@{\hspace{0.3cm}}l@{\hspace{0.3cm}}l@{\hspace{0.3cm}}l@{\hspace{0.3cm}}l@{}}
    \toprule 
     & Case &$f(x,y)$ & $g(x)$ & $S_0(x)$ & $a$ & $b$\\
    \midrule
    \eqref{eq:prob_event1} & \ref{group:combine_probs1} & $\left\| \left| \b A^m \right| - \b K^m  \right\|$ & $4\tilde{C}^m \sqrt{\log \left( \frac{12n}{\eta}\right)}$ & $ \sqrt{\Delta^m}$ & $\sqrt{C_0^m \log(n)}$ & $\sqrt{n}$ \\
    \eqref{eq:prob_event2} & \ref{group:combine_probs2} & $\left\| \left| \b A^m \right| - \b K^m \right\|$ & $4\tilde{C}^m \sqrt{\log \left( \frac{144n}{\eta}\right)}$ & $\sqrt{\Delta^m}$ & $\sqrt{C_0^m \log(n)}$ & \ \\
    \eqref{eq:prob_event3} & \ref{group:combine_probs2} & $\left\| \left| \b A^m \right| - \b K^m \right\|$ & $4\tilde{C}^m \sqrt{\log \left( \frac{144n}{\eta}\right)}$ & $\sqrt{\Delta^m}$ & $\sqrt{C_0^m \log(n)}$ & \ \\
    \eqref{eq:prob_event4} & \ref{group:combine_probs2} & $\left\| \left| \b A^m \right| - \b K^m \right\|$ & $4\tilde{C}^m \sqrt{\log \left( \frac{24n}{\eta}\right)}$ & $\sqrt{\Delta^m}$ & $\sqrt{C_0^m \log(n)}$ & \ \\
    \eqref{eq:D^m-lower-bound} & \ref{group:combine_probs1} & $-\left \| \b D^{m} \right \|-\sqrt{32 \log(n)}$ & $2 \sqrt{n}$ & $-\bar{\sigma}^m $ & $-\infty$ & $-C_1^m \sqrt{\frac{\log(n)}{n}} $ \\
    \bottomrule
    \end{tabular}
    \par\end{center}
\end{table}

\subsection{Proofs}\label{appendix_proof}
In this section we prove Lemma~\ref{lemm:combine_probs}, Lemma~\ref{lemma:HSIC2:Gram} (Section~\ref{prove_non-negative}) and Theorem~\ref{errorbound_GHSIC1} (Section~\ref{prove_error_bound1}). Lemma~\ref{lemm:combine_probs} is about combination of probability events used in two of our proofs. We verify various assumptions for the  Laplacian kernel and the uniform distribution: (i) in Lemma~\ref{lemma:universality_verify_laplacian} we check the universality of the Laplacian kernel (Section~\ref{sec:universality_verify_laplacian}); (ii) Assumptions~\ref{eq:poly-decay:lambda^m-decay}–\ref{eq:poly-decay:c(i)_assumption2} are verified in Lemma~\ref{lemma:assumption_verify_laplacian0} (Section~\ref{sec:assumption_verify_laplacian0}), (iii) Assumptions~\ref{assum_Delta}–\ref{assum_eigenval_hat_C_k} are verified in Lemma~\ref{lemma:assumption_verify_laplacian} relying on  Lemma~\ref{lemma:exp_triangular_dist} about the expectation of the triangular distribution (Section~\ref{sec:assumption_verify_laplacian}).

\begin{lemma}[Combination of probability events]\label{lemm:combine_probs}
Let $\X$ and $\Y$ be measurable spaces, and let $\X\times \Y$ be enriched with the product $\sigma$-algebra; let $\R$ and $\Rnn$ be endowed with the Borel $\sigma$-field. Let $f:\X  \times \Y \to \R$ be measurable, $g: \X \to \Rnn$ be measurable, $S_0: \X \to \R$ be measurable, $a \le b $ and $a, b \in \bar{\R} \coloneq \R \cup \{\pm \infty\}$. Define the following two groups of events 
\begin{enumerate}[label=(\roman*)]
    \item \label{group:combine_probs1}
          \begin{align}
            A &= \{(x,y)\in \X \times \Y\,:\, f(x,y) \le g(x) b \} \subseteq \X \times \Y,\\
            B &= \{x\in \X\,:\, a < S_0(x) \le b \}\subseteq \X,\\
            C &= \{(x,y) \in \X \times \Y\,:\, f(x,y) \le g(x) S_0(x), x\in B\} \subseteq A \subseteq \X \times \Y.
          \end{align}
    \item \label{group:combine_probs2} 
          \begin{align}
            A &= \{(x,y)\in \X \times \Y\,:\, f(x,y) \le g(x) S_0(x) \} \subseteq \X \times \Y,\\
            B &= \{x\in \X\,:\, a < S_0(x) \}\subseteq \X,\\
            C &= \{(x,y) \in \X \times \Y\,:\, f(x,y) \le g(x) S_0(x), x\in B\} \subseteq A \subseteq \X \times \Y.
          \end{align}
          
\end{enumerate}

Then, the followings hold. 
\begin{enumerate}
\item For both groups, $A$, $B$ and $C$ are measurable sets.
\item Let $\delta_1, \delta_2 \in (0,1)$.  If $\P_X(B) \ge 1-\delta_1$ and $\P(C|X=x) \ge 1-\delta_2$ for all $x\in \X$, then $\P_{XY}(A) \ge  1-(\delta_1+\delta_2)$.
\end{enumerate}

\end{lemma}

\begin{proof}(of Lemma~\ref{lemm:combine_probs}): 
    \begin{itemize}
        \item Measurability of $A$, $B$, $C$ in \ref{group:combine_probs1}:
           \begin{itemize}
                \item 
                Set $A$: Let $h_1(x,y):= f(x,y) - b g(x)$ with $(x,y) \in \X \times \Y$. The function $h_1$ is measurable by the measurability of $f$ and $g$. Moreover, we have $A = \{(x,y)\in \X \times \Y\,:\, h_1(x,y) \le 0 \} = h_1^{-1}\left( \left(-\infty, 0 \right] \right)$; as $\left(-\infty, 0 \right]$ is Borel, $A$ is measurable by the measurability of $h_1$.

                \item Set $B$: We have $B=\{ x \in \X: S_0(x) \in (a,b] \} = S_0^{-1}\left( (a,b] \right)$. Therefore, the set $B$ is measurable by that of $(a,b]$ and  the measurability of $S_0$. 

                \item Set $C$: Let $h_2(x,y):= f(x,y) - S_0(x) g(x)$ with $(x,y) \in \X \times \Y$. The function $h_2$ is measurable by that of $f$, $g$ and $S_0$. Then, we have $C = \{(x,y)\in \X \times \Y\,:\, h_2(x,y) \le 0 \} \cap (B \times \Y) = h_2^{-1}\left( \left(-\infty, 0 \right] \right) \cap (B \times \Y)$. As $\left(-\infty, 0 \right]$ is Borel, $C$ is measurable by that of $h_2$ and the measurability of $B \times \Y$, using the fact that the intersection of measurable sets is measurable.
        \end{itemize}    
               
        \item Measurability of $A$, $B$, $C$ in \ref{group:combine_probs2}: 
        \begin{itemize}
            \item Set $A$: Let $h_3(x,y):= f(x,y) - S_0(x) g(x)$ with $(x,y) \in \X \times \Y$. The function $h_3$ is measurable by that of $f$, $g$ and $S_0$. Then, we have $A = \{(x,y)\in \X \times \Y\,:\, h_3(x,y) \le 0 \} = h_3^{-1}\left( \left(-\infty, 0 \right] \right)$; as $\left(-\infty, 0 \right]$ is Borel, $A$ is measurable by the measurability of $h_3$.
            \item Set $B$: We have $B=\{ x \in \X: S_0(x) \in (a,+\infty) \} = S_0^{-1}\left( (a,+\infty) \right)$. Therefore, the set $B$ is measurable by that of $(a,+\infty)$ and the measurability of $S_0$. 
            \item Let $h_4(x,y):= f(x,y) - S_0(x) g(x)$ with $(x,y) \in \X \times \Y$. The function $h_4$ is measurable by that of $f$, $g$ and $S_0$. Then we have $C = \{(x,y)\in \X \times \Y\,:\, h_4(x,y) \le 0 \} \cap (B \times \Y) = h_4^{-1}\left( \left(-\infty, 0 \right] \right) \cap (B \times \Y) $. $C$ is measurable by that of $h_4$ and the measurability of $B \times \Y$, using the fact that the intersection of measurable sets is measurable.
        \end{itemize}

        \item Bound on $\P_{XY}(A)$:
                    We have
                \begin{align}
                    \P_{XY}(A)  & = \int_{\X} \P(A|X=x) \d \P_X (x) \stackrel{(a)}{\ge} \int_{B} \P(A|X=x) \d \P_X (x) \stackrel{(b)}{\ge} \int_{B} \P(C|X=x) \d \P_X (x) \\
                    & \stackrel{(c)}{\ge} \left( 1-\delta_2 \right) \P_X (B) \stackrel{(d)}{\ge} (1-\delta_2)(1-\delta_1) 
                     \stackrel{(e)}{=} 1-(\delta_1 + \delta_2) + \delta_1 \delta_2 \stackrel{(f)}{\ge} 1-(\delta_1 + \delta_2),
                \end{align}
                where (a) follows from the non-negative of probabilities, $B\subseteq \X$, and the monotonicity of integration, (b) comes from $C\subseteq A$, and the monotonicity of probability measures and that of the integration, (c) is implied by $\P(C|X=x) \ge 1-\delta_2$ for all $x \in \X$, the monotonicity of the integration and the integration of a constant function, (d) is implied by $\delta_2 \in (0,1)$ and $\P_X(B) \ge 1-\delta_1$, in (e) the multiplication was carried, (f) follows from discarding the term $\delta_1 \delta_2 \ge 0$.         
    \end{itemize}    
\end{proof}

\subsubsection{Proof of Lemma~\ref{lemma:HSIC2:Gram}}\label{prove_non-negative}
By replacing the $\hat{\mu}_k(\hat{\P}_n)$ and $\hat{\mu}_{k^m}(\hat{\P}^m_n)$ by the form of $ \hat{\phi}_{k^m}$, we have
\begin{align}
    \widehat{\t{GHSIC}^2_{k}}& = \left \|  \frac{1}{n} \sum_{i=1}^n \otimes_{m=1}^M  \hat{\phi}_{k^m}(x_i^m)-\otimes_{m=1}^M \frac{1}{n} \sum_{i=1}^n  \hat{\phi}_{k^m}(x_i^m)  \right \|_2^2  \\
    & = \underbrace{\left \|  \frac{1}{n} \sum_{i=1}^n \otimes_{m=1}^M  \hat{\phi}_{k^m}(x_i^m)  \right \|_2^2}_{=: A_1} + \underbrace{\left \| \otimes_{m=1}^M \frac{1}{n} \sum_{i=1}^n  \hat{\phi}_{k^m}(x_i^m)  \right \|_2^2}_{=: A_2} \\
    & \quad \quad - 2 \underbrace{\left \langle \frac{1}{n} \sum_{i=1}^n \otimes_{m=1}^M  \hat{\phi}_{k^m}(x_i^m),\otimes_{m=1}^M \frac{1}{n} \sum_{i=1}^n  \hat{\phi}_{k^m}(x_i^m) \right \rangle}_{=: A_3}.
\end{align}

For $A_1$ we have
\begin{align}
    A_1 & = \left \langle \frac{1}{n} \sum_{i=1}^n \otimes_{m=1}^M  \hat{\phi}_{k^m}(x_i^m),\frac{1}{n} \sum_{i=j}^n \otimes_{m=1}^M  \hat{\phi}_{k^m}(x_j^m) \right \rangle \\
    & \stackrel{(a)}{=} \frac{1}{n^2} \sum_{i=1}^n \sum_{j=1}^n \left \langle \otimes_{m=1}^M  \hat{\phi}_{k^m}(x_i^m), \otimes_{m=1}^M  \hat{\phi}_{k^m}(x_j^m) \right \rangle \\
    &\stackrel{(b)}{=}\frac{1}{n^2} \sum_{i=1}^n \sum_{j=1}^n \prod_{m=1}^M \left \langle  \hat{\phi}_{k^m}(x_i^m),  \hat{\phi}_{k^m}(x_j^m) \right \rangle \stackrel{(c)}{=} \frac{1}{n^2} \sum_{i=1}^n \sum_{j=1}^n \prod_{m=1}^M \hat{K}^m_{i,j} = \frac{1}{n^2} \bm{1}_n^{\top}\left ( \odot_{m\in[M]} \hat{\b K}^m \right )\bm{1}_n,
\end{align}
where $(a)$ follows from the linearity of the inner product, $(b)$ obtained by the property of inner product for tensor product and $(c)$ is obtained by $\hat{K}^m_{i,j} = \left \langle  \hat{\phi}_{k^m}(x_i^m),  \hat{\phi}_{k^m}(x_j^m) \right \rangle$.

For $A_2$ we have
\begin{align}
    A_2 &= \left \langle \otimes_{m=1}^M \frac{1}{n} \sum_{i=1}^n  \hat{\phi}_{k^m}(x_i^m),\otimes_{m=1}^M \frac{1}{n} \sum_{j=1}^n  \hat{\phi}_{k^m}(x_j^m) \right \rangle = \prod_{m=1}^M \frac{1}{n^2} \sum_{i=1}^n \sum_{j=1}^n \left \langle   \hat{\phi}_{k^m}(x_i^m),  \hat{\phi}_{k^m}(x_j^m) \right \rangle \\
    &= \frac{1}{n^{2M}} \prod_{m=1}^M \sum_{i=1}^n \sum_{j=1}^n \hat{K}^m_{i,j} = \frac{1}{n^{2M}}\prod_{m=1}^{M} \bm{1}_n^{\top} \hat{\b K}^m \bm{1}_n.
\end{align}

For $A_3$ we have:
\begin{align}
    A_3 & = \frac{1}{n} \sum_{i=1}^n \left \langle \otimes_{m=1}^M  \hat{\phi}_{k^m}(x_i^m),\otimes_{m=1}^M \frac{1}{n} \sum_{j=1}^n  \hat{\phi}_{k^m}(x_j^m) \right \rangle\\
    & = \frac{1}{n} \sum_{i=1}^n \prod_{m=1}^M \left \langle  \hat{\phi}_{k^m}(x_i^m), \frac{1}{n} \sum_{j=1}^n  \hat{\phi}_{k^m}(x_j^m) \right \rangle 
    = \frac{1}{n} \sum_{i=1}^n \prod_{m=1}^M \frac{1}{n} \sum_{j=1}^n \left \langle  \hat{\phi}_{k^m}(x_i^m),  \hat{\phi}_{k^m}(x_j^m) \right \rangle\\
    & = \frac{1}{n^{M+1}} \sum_{i=1}^n \prod_{m=1}^M \sum_{j=1}^n \hat{K}^m_{ij} = \frac{1}{n^{M+1}} \bm{1}_n^{\top} \left ( \odot_{m\in[M]} \hat{\b K}^m \bm{1}_n \right ).
\end{align}

\subsubsection{Proof of Theorem~\ref{errorbound_GHSIC1}}\label{prove_error_bound1}
Recall that by \eqref{eq:def_hat_mu_k_hat_mu_k^m}, one has
\begin{align}
\hat{\mu}_k\left(\hat{\P}_n\right) &= \frac{1}{n} \sum_{i=1}^n \otimes_{m=1}^M  \hat{\phi}_{k^m}\left(X_i^m\right) \in \R^{d^1 \times \cdots \times d^M}, \\
\hat{\mu}_{k^m}\left(\hat{\P}^m_n\right) & = \frac{1}{n} \sum_{i=1}^n  \hat{\phi}_{k^m}\left(X_i^m\right) \in \R^{d^m} \quad (m \in [M])
\end{align}
with $\hat{\phi}_{k^m}\left(X_i^m\right)$ ($i\in [n]$, $m\in [M]$) defined according to \eqref{eq:hat-Phi-m}.
To obtain the error bound of $\left | \mathrm{GHSIC}_{k} (\P) - \widehat{\mathrm{GHSIC}_{k}}  \right |$, we have
\begin{align}
        \MoveEqLeft \left | \mathrm{GHSIC}_{k} (\P) - \widehat{\mathrm{GHSIC}_{k}}  \right |  \stackrel{(a)}{=} \left | \left\| \mu_k(\P) - \mu_k\left(\otimes_{m=1}^M \P^m\right) \right\|_{\mH_k} - \left \| \hat{\mu}_k \left(\hat{\P}_n\right) - \otimes_{m=1}^M  \hat{\mu}_{k^m} \left(\hat{\P}^m_n\right) \right \|_2 \right | \\
        & \stackrel{(b)}{\le}  \underbrace{\left | \left\| \mu_k(\P) - \mu_k\left(\otimes_{m=1}^M \P^m\right) \right\|_{\mH_k} - \left \| \tilde{\mu}_{k,\b d}\left(\hat{\P}_n\right) - \otimes_{m=1}^M  \tilde{\mu}_{k^m,d^m}\left(\hat{\P}^m_n\right) \right \|_{\mH_k} \right |}_{=:A_1} \\
        & \quad + \underbrace{\left| \left \| \tilde{\mu}_{k,\b d}\left(\hat{\P}_n\right) - \otimes_{m=1}^M  \tilde{\mu}_{k^m,d^m}\left(\hat{\P}^m_n\right) \right \|_{\mH_k} - \left \| \hat{\mu}_k \left(\hat{\P}_n\right) - \otimes_{m=1}^M  \hat{\mu}_{k^m}\left(\hat{\P}^m_n\right) \right \|_2 \right |}_{=:A_2},
\end{align}
where (a) follows from the definition of $\mathrm{GHSIC}_{k} (\P)$ and $\widehat{\mathrm{GHSIC}_{k}}$ in \eqref{eq:target} and \eqref{est_HSIC2:v0}, respectively. (b) is implied by adding terms $\pm \left \| \tilde{\mu}_{k,\b d}\left(\hat{\P}_n\right) - \otimes_{m=1}^M  \tilde{\mu}_{k^m,d^m}\left(\hat{\P}^m_n\right) \right \|_{\mH_k} $ and the triangle inequality, where 
\begin{align}
\tilde{\mu}_{k^m,d^m} \left(\hat{\P}^m_n\right) &= \frac{1}{n} \sum_{i=1}^n k^m_{d^m} \left(\cdot, X^m_i\right) \in \mH_{k^m},& 
\tilde{\mu}_{k,\b d}\left(\hat{\P}_n\right) &= \frac{1}{n} \sum_{i=1}^n \otimes_{m=1}^M  k^m_{d^m} \left(\cdot,X^m_i\right) \in \mH_k 
\end{align}
are recalled from \eqref{eq:mu-tilde-and-truncated} and \eqref{eq:def_joint_tilde_mu}, respectively, and  $k^m_{d^m}\left(\cdot,z\right)$ is specified according to  \eqref{eq:def:km}.

We proceed by bounding $A_1$ and $A_2$. 
\begin{itemize}
    \item \textbf{Term $A_1$}: For $A_1$ we have
    \begin{align}
        A_1 & = \left | \left\| \mu_k(\P) - \mu_k\left(\otimes_{m=1}^M \P^m\right) \right\|_{\mH_k} - \left \| \tilde{\mu}_{k,\b d}\left(\hat{\P}_n\right) - \otimes_{m=1}^M  \tilde{\mu}_{k^m,d^m}\left(\hat{\P}^m_n\right) \right \|_{\mH_k} \right | \\
        & \stackrel{(a)}{\le }  \left\| \mu_k(\P) - \mu_k\left(\otimes_{m=1}^M \P^m\right) - \tilde{\mu}_{k,\b d}\left(\hat{\P}_n\right) + \otimes_{m=1}^M  \tilde{\mu}_{k^m,d^m}\left(\hat{\P}^m_n\right) \right \|_{\mH_k} \\
        & \stackrel{(b)}{\le }  \underbrace{\left\| \mu_k(\P)  -  \tilde{\mu}_{k,\b d}\left(\hat{\P}_n\right) \right \|_{\mH_k}}_{=: t_1} +  \underbrace{\left\| \mu_k\left(\otimes_{m=1}^M \P^m\right)  -  \otimes_{m=1}^M  \tilde{\mu}_{k^m,d^m}\left(\hat{\P}^m_n\right) \right \|_{\mH_k}}_{=: t_2},
    \end{align}
    where (a) was obtained from the reverse triangle inequality, (b) follows from the triangle inequality. 
    \begin{itemize}
        \item \textbf{Term $t_1$}: For $t_1$ we have that for any $\eta \in (0,1)$
        \begin{align}
            t_1 & = \left\| \mu_k(\P)  -  \tilde{\mu}_{k,\b d}\left(\hat{\P}_n\right) \right \|_{\mH_k} \stackrel{(a)}{=} \left\| \mu_k(\P)  -  \tilde{\mu}_{k}\left(\hat{\P}_n\right) + \tilde{\mu}_{k}\left(\hat{\P}_n\right) - \tilde{\mu}_{k,\b d}\left(\hat{\P}_n\right) \right \|_{\mH_k} \\
            & \hspace{-0.5cm}\stackrel{(b)}{\le} \left\| \mu_k(\P)  -  \tilde{\mu}_{k}\left(\hat{\P}_n\right) \right \|_{\mH_k} + \left\| \tilde{\mu}_{k}\left(\hat{\P}_n\right) - \tilde{\mu}_{k,\b d}\left(\hat{\P}_n\right) \right \|_{\mH_k} \\
            & \hspace{-0.5cm}\stackrel{(c)}{\le } 2 a_k \sqrt{\frac{2 \log \left(\frac{16}{\eta}\right)}{n} } + 3 \sqrt{\frac{3}{2}} \sum_{m=1}^M b_{k^m} \sqrt{\mathcal{N}_{C_{k^m}}\left(t_{k^m}\left(n, \frac{\eta}{8M}\right)\right)}\left(\lambda^m_{d^m+1} + t_{k^m} \left(n, \frac{\eta}{8M}\right) \right),
        \end{align}
        holds with $\P^{\otimes n}$-probability at least $1-\frac{\eta}{4}$ provided that $t_{k^m}\left( n, \frac{\eta}{8M} \right) \le \left\| C_{k^m} \right\|$ for all $m\in [M]$, where 
        \begin{align}
        \tilde{\mu}_{k}\left(\hat{\P}_n\right) &=\frac{1}{n} \sum_{i=1}^n k(\cdot,X_i) \in \mH_k
        \end{align}
        is defined according to \eqref{eq:def_joint_tilde_mu} and $t_{k^m}$ is defined in \eqref{eq:def_t_km}. (a) is implied by adding terms $\pm \tilde{\mu}_{k}\left(\hat{\P}_n\right)$; (b) follows from the triangle inequality. The first term in (c) was obtained by Lemma~\ref{concentration_ineq1} with a substitution $\eta \leftarrow \frac{\eta}{8}$ and $A = 2 a_k$
        ; the second term is implied by Lemma~\ref{truncated_ineq2} with a substitution $\eta \leftarrow \frac{\eta}{8}$ assuming that Assumption~\ref{assum_eigenval_C_k}-\ref{assum_eigenval_hat_C_k} hold, the kernel $k^m: \X^m \times \X^m \to [0,1]$ is continuous on the compact metric space $\X^m$ and $\mathrm{supp}(\P^m)=\X^m$ for all $m\in [M]$. 
        \item \textbf{Term $t_2$}: For $t_2$, we have for any $\eta \in (0,1)$
        \begin{align}
            t_2 &= \left\| \mu_k\left(\otimes_{m=1}^M \P^m\right)  -  \otimes_{m=1}^M  \tilde{\mu}_{k^m,d^m}\left(\hat{\P}^m_n\right) \right \|_{\mH_k} \\&\stackrel{(a)}{=} \left\| \otimes_{m=1}^M \mu_{k^m}\left(\P^m\right)  -  \otimes_{m=1}^M  \tilde{\mu}_{k^m,d^m}\left(\hat{\P}^m_n\right) \right \|_{\mH_k} \\
            & \stackrel{(b)}{\le} \sum_{m=1}^M \left( \prod_{i=m+1}^M B_i \right) \left\|  \mu_{k^m}\left(\P^m\right) - \tilde{\mu}_{k^m,d^m}\left(\hat{\P}^m_n\right) \right\|_{\mH_{k^m}}\\
            &\stackrel{(c)}{\le} \sum_{m=1}^M \left\|  \mu_{k^m}\left(\P^m\right) - \tilde{\mu}_{k^m,d^m}\left(\hat{\P}^m_n\right) \right\|_{\mH_{k^m}} \\
            & \stackrel{(d)}{=} \sum_{m=1}^M \left\|  \mu_{k^m}\left(\P^m\right) - \tilde{\mu}_{k^m}\left(\hat{\P}^m_n\right) + \tilde{\mu}_{k^m}\left(\hat{\P}^m_n\right) - \tilde{\mu}_{k^m,d^m}\left(\hat{\P}^m_n\right) \right\|_{\mH_{k^m}} \\
            & \stackrel{(e)}{\le} \sum_{m=1}^M \left\|  \mu_{k^m}\left(\P^m\right) - \tilde{\mu}_{k^m}\left(\hat{\P}^m_n\right) \right\|_{\mH_{k^m}} + \sum_{m=1}^M \left\| \tilde{\mu}_{k^m}\left(\hat{\P}^m_n\right) - \tilde{\mu}_{k^m,d^m}\left(\hat{\P}^m_n\right) \right\|_{\mH_{k^m}} \\
            & \stackrel{(f)}{\le} \sum_{m=1}^M 2 a_{k^m} \sqrt{\frac{2 \log \left(\frac{16M}{\eta}\right)}{n} }\\
            & \quad + \sum_{m=1}^M 3 \sqrt{\frac{3}{2}} b_{k^m} \sqrt{\mathcal{N}_{C_{k^m}}\left(t_{k^m} \left(n, \frac{\eta}{8M}\right) \right)}\left(\lambda^m_{d^m+1} + t_{k^m} \left(n, \frac{\eta}{8M}\right) \right),
        \end{align}
        holds with $\P^{\otimes n}$-probability at least $1-\frac{\eta}{4}$ provided that $t_{k^m}\left(n, \frac{\eta}{8M}\right) \le \left\| C_{k^m} \right\|$ for all $m\in [M]$, where $t_{k^m}$ is defined in \eqref{eq:def_t_km}. In (a) we used the fact that $\mu_k\left(\otimes_{m=1}^M \P^m\right) = \otimes_{m=1}^M \mu_{k^m}\left(\P^m\right)$, (b) is implied by Lemma~\ref{lemma:properties-of-tensor-products} [see \eqref{eq:diff-of-tensor-products_Am1}] with $A_m=1$ and $\left\|\tilde{\mu}_{k^m,d^m}\left(\hat{\P}_n^m\right)\right\|_{\mH_{k^m}}\le B_m$ ($m\in [M]$), in (c) we used that $B_{m} \le 1$ following from our assumption $k^m: \X^m \times \X^m \to [0,1]$ for $m\in [M]$, (d) comes from adding terms $\pm \tilde{\mu}_{k^m}\left(\hat{\P}^m_n\right)$ with 
        \begin{align}
        \tilde{\mu}_{k^m} \left(\hat{\P}^m_n\right) &= \frac{1}{n} \sum_{i=1}^n k^m \left(\cdot, X^m_i\right)\in \mH_{k^m}
        \end{align}
        defined as in \eqref{eq:mu-tilde-and-truncated}, (e) follows from the triangle inequality. The first term in (f) was obtained by Lemma~\ref{concentration_ineq1} with a substitution $\eta \leftarrow \frac{\eta}{8M}$ and $A = 2 a_{k^m}$, the second term is implied by Lemma~\ref{truncated_ineq} with a substitution $\eta \leftarrow \frac{\eta}{8M}$ supposing that Assumption~\ref{assum_eigenval_C_k}-\ref{assum_eigenval_hat_C_k} hold, the kernel $k^m: \X^m \times \X^m \to [0,1]$ is continuous on the compact metric space $\X^m$ and $\mathrm{supp}(\P^m)=\X^m$ for all $m\in [M]$. 
    \end{itemize}

    Combining the results obtained for $t_1$ and $t_2$, for any $\eta\in (0,1)$ we have
    that 
    \begin{align}
        A_1 &\le 2 a_k \sqrt{\frac{2 \log \left( \frac{16}{\eta} \right)}{n} } + \sum_{m=1}^M 2 a_{k^m} \sqrt{\frac{2 \log \left( \frac{16M}{\eta} \right)}{n} } \\
        & \quad + 6 \sqrt{\frac{3}{2}} \sum_{m=1}^M b_{k^m} \sqrt{\mathcal{N}_{C_{k^m}}\left(t_{k^m} \left(n, \frac{\eta}{8M}\right) \right)} \left(\lambda^m_{d^m+1} + t_{k^m} \left(n, \frac{\eta}{8M}\right) \right)
    \end{align}
    holds with $\P^{\otimes n}$-probability at least $1-\frac{\eta}{2}$ provided that (i) Assumption~\ref{assum_eigenval_C_k}-\ref{assum_eigenval_hat_C_k} hold, and for all $m\in [M]$ (ii) the kernel $k^m: \X^m \times \X^m \to [0,1]$ is continuous on the compact metric space $\X^m$ and $\mathrm{supp}(\P^m)=\X^m$, (iii) $t_{k^m}\left(n, \frac{\eta}{8M} \right) \le \left\| C_{k^m} \right\|$ with $t_{k^m}$ defined in \eqref{eq:def_t_km}. Define the following event
    \begin{align}
        A = \Big\{ (x_i)_{i=1}^n\, &: \, A_1 \le 2 a_k \sqrt{\frac{2 \log \left( \frac{16}{\eta} \right)}{n} } + \sum_{m=1}^M 2 a_{k^m} \sqrt{\frac{2 \log \left(\frac{16M}{\eta} \right)}{n} } \\
        & \quad + 6 \sqrt{\frac{3}{2}} \sum_{m=1}^M b_{k^m} \sqrt{\mathcal{N}_{C_{k^m}}\left(t_{k^m} \left(n, \frac{\eta}{8M}\right) \right)} \left(\lambda^m_{d^m+1} + t_{k^m} \left(n, \frac{\eta}{8M}\right) \right) \Big\} \subseteq \X^n.
    \end{align}
    Let $\Y := \times_{m=1}^M \{ 0, 1\}^{\binom{n}{2} }$ be the sample space of r.v.s from $\Q$, where $\Q$ denotes the joint distribution of entries in $\left( \b A^m \right)_{m=1}^M$, given $(X_i)_{i=1}^n = (x_i)_{i=1}^n$. Recall that $\mathbb{H}$ stands for the joint distribution of the adjacency matrices $(\b A_m)_{m=1}^M$ and $(X_i)_{i=1}^n$. Then we have
    \begin{align}
            (\mathbb{H})(A \times \Y) 
            & = \P^{\otimes n} (A) \ge 1-\frac{\eta}{2}.
    \end{align} 
    Therefore, we have that for any $\eta \in (0,1)$
    \begin{align}
        A_1 &\le 2 a_k \sqrt{\frac{2 \log \left( \frac{16}{\eta} \right)}{n} } + \sum_{m=1}^M 2 a_{k^m} \sqrt{\frac{2 \log \left( \frac{16M}{\eta} \right)}{n} } \\
        & \quad + 6 \sqrt{\frac{3}{2}} \sum_{m=1}^M b_{k^m} \sqrt{\mathcal{N}_{C_{k^m}}\left(t_{k^m} \left(n, \frac{\eta}{8M}\right) \right)} \left(\lambda^m_{d^m+1} + t_{k^m} \left(n, \frac{\eta}{8M}\right) \right) \label{eq:bound-A1}
    \end{align}
    holds with $\mathbb{H}$-probability at least $1-\frac{\eta}{2}$ provided that (i) Assumption~\ref{assum_eigenval_C_k}-\ref{assum_eigenval_hat_C_k} hold and for all $m\in [M]$ (ii) the kernel $k^m: \X^m \times \X^m \to [0,1]$ is continuous on the compact metric space $\X^m$ and $\mathrm{supp}(\P^m)=\X^m$, (iii) $t_{k^m}\left( n, \frac{\eta}{8M}\right) \le \left\| C_{k^m} \right\|$ with $t_{k^m}$ defined in \eqref{eq:def_t_km}.
    \item \textbf{Term $A_2$}: Regarding $A_2$, we have
    \begin{align}
        A_2 & = \left| \left \| \tilde{\mu}_{k,\b d} \left( \hat{\P}_n \right) - \otimes_{m=1}^M  \tilde{\mu}_{k^m,d^m}\left( \hat{\P}^m_n \right) \right \|_{\mH_k} - \left \| \hat{\mu}_k \left(\hat{\P}_n \right) - \otimes_{m=1}^M  \hat{\mu}_{k^m} \left(\hat{\P}^m_n \right) \right \|_2 \right| \\
        & \stackrel{(a)}{=} \left| \left \| \breve{\mu}_k \left(\hat{\P}_n \right) - \otimes_{m=1}^M  \breve{\mu}_{k^m} \left(\hat{\P}^m_n \right) \right \|_2 - \left \| \hat{\mu}_k \left(\hat{\P}_n \right) - \otimes_{m=1}^M  \hat{\mu}_{k^m} \left(\hat{\P}^m_n \right) \right \|_2 \right| \\
        & \stackrel{(b)}{\le} \left\| \breve{\mu}_k \left(\hat{\P}_n \right) - \otimes_{m=1}^M  \breve{\mu}_{k^m} \left( \hat{\P}^m_n \right) - \hat{\mu}_k \left( \hat{\P}_n \right) +\otimes_{m=1}^M  \hat{\mu}_{k^m} \left( \hat{\P}^m_n \right) \right\|_2 \\
        & \stackrel{(c)}{\le} \underbrace{\left \| \breve{\mu}_k \left( \hat{\P}_n \right)  - \hat{\mu}_k \left( \hat{\P}_n \right) \right \|_{2}}_{=:C_1} + \underbrace{\left \| \otimes_{m=1}^M  \breve{\mu}_{k^m} \left( \hat{\P}^m_n \right) -  \otimes_{m=1}^M  \hat{\mu}_{k^m} \left( \hat{\P}^m_n \right) \right \|_{2}}_{=:C_2} ,
    \end{align}
    where (a) is implied by Lemma~\ref{lemma:isometric_embedding} with 
    \begin{align}
   \breve{\mu}_{k}\left(\hat{\P}_n\right) &= \frac{1}{n} \sum_{i=1}^n \otimes_{m=1}^M   \phi_{k^m,d^m}^{m} (X^m_i) \in \R^{d^1 \times \cdots \times d^M}, &  \breve{\mu}_{k^m}\left(\hat{\P}^m_n\right) &= \frac{1}{n} \sum_{i=1}^n \phi_{k^m,d^m}^m (X_i^m) \in \R^{d^m},
    \end{align}
    recalled from \eqref{eq:def_breve_mu_km}, 
    and  $\phi_{k^m,d^m}^m (X_i^m) \in \R^{d^m}$ defined in \eqref{eq:def_feature_est_by_gram}.
    (b) comes from the reverse triangle inequality and (c) follows from the triangle inequality. 
    \begin{itemize}
        \item \textbf{Term $C_1$}: Let us now use  Lemma~\ref{concentration_joint_adj} with a substitution $\eta \leftarrow \frac{\eta}{4}$. Particularly, let $c^m>0$ be fixed for any $m\in [M]$; assuming that (i) Assumptions~\ref{assum_Delta}-\ref{assum_eigenval_hat_C_k} hold, with $n_{0,(i)}^{m}(\eta), n_{0,(ii)}^{m}(\eta), n_{1,(i)}^{m}(\eta), n_{1,(ii)}^{m}(\eta) \in \Zp$, $c^m_0>0$, $c_1^m(n) = c_2^m \log \left( 2 + \log(2n) \right)$, $c_2^m > 0$, $C_0^m>0$, and $C_1^m>0$ specified therein ($m \in [M]$), (ii) $C_1^m > 2\sqrt{2}$ and $c_2^m \ge 2$ for all $m \in [M]$, (iii) $\delta_{d^m}^m = \lambda^m_{d^m} - \lambda^m_{d^m+1} \ne 0$ ($m\in [M]$) with $\left( \lambda_i^m \right)_{i \in \Zp}$ denoting the eigenvalues of the integral operator $T_{k^m}$ arranged in non-increasing order, and
        (iv) $\lambda^m_{d^m+1} \le 8 \left( \log 2 \right) \sqrt{\frac{d^m \log \left( \frac{64Mn}{\eta}  \right)}{n}}$ holds for $n \ge \left( \frac{\eta}{32M} \right)^{-\frac{1}{c^m}}$ and all $m\in [M]$, the result for  
        \begin{align}
            n & \ge \max_{m \in [M]} \left\{ \tilde{n}^m_{0,A1(i)}\left( \frac{\eta}{4M} \right), \tilde{n}^m_{0,A1(ii)}\left( \frac{\eta}{4M} \right),\tilde{n}^m_{0,A2(i)}\left( \frac{\eta}{4M} \right), \tilde{n}^m_{0,A2(ii)}\left( \frac{\eta}{4M} \right),   \right. \\
            &\hspace{1.5cm}\left.\tilde{n}^m_{0,L5}\left( \frac{\eta}{4M} \right),\tilde{n}^m_{0,L6}\left( \frac{\eta}{4M} \right),\tilde{n}_{0,\text{else-1}}^m\left( \frac{\eta}{4M} \right),\tilde{n}_{0,\text{else-2}}^m \right\},\t{ with }\\   
    \tilde{n}^m_{0,A1(i)}(\eta) &:= n^m_{0,(i)}\left( \frac{\eta}{2304}\right), \tilde{n}^m_{0,A1(ii)}(\eta) := n^m_{0,(ii)}\left( \frac{\eta}{1152} \right), \tilde{n}^m_{0,A2(i)}(\eta) := n^m_{1,(i)}\left( \frac{\eta}{1152} \right), \\
    \tilde{n}^m_{0,A2(ii)}(\eta)
    &:= n^m_{1,(ii)}\left( \frac{\eta}{192} \right), \tilde{n}^m_{0,L5}(\eta) := \left( \frac{\eta}{2304} \right)^{-\frac{1}{c^m}}, \tilde{n}^m_{0,L6}(\eta) := \frac{4608}{\eta}, \\
    \tilde{n}_{0,\text{else}-1}^m(\eta) & := 32 \left(c_1^m(n)\right)^2 (c_0^m)^{-2} \log \left( \frac{2304 n}{\eta} \right), \tilde{n}_{0,\text{else}-2}^m := e^{\frac{1}{\left( 2C_1^m - 4\sqrt{2} \right)^2}},
        \end{align}
        implies that for any $\eta \in (0,1)$  
        \begin{align}
            C_1 \le d_k \sum_{m=1}^M D_{k^m}\left( \frac{\eta}{4M} \right) 
        \end{align}
        holds with $\mathbb{H}$-probability at least $1-\frac{\eta}{4}$, where $\mathbb{H}$ denotes the joint distribution of the adjacency matrices $(\b A^m)_{m=1}^M$ and $(X_i)_{i=1}^n$, $d_k:= \prod_{m=1}^M d_{k^m}$ with $d_{k^m}$ is defined in \eqref{assum:bounded_feature_map}, and $D_{k^m}$ is specified according to \eqref{eq:def_D_k^m}.     
        \item \textbf{Term $C_2$}: Considering $C_2$, we have \begin{align}
            C_2 & \stackrel{(a)}{\le} \sum_{m=1}^M \left( \prod_{i=m+1}^M \hat{B}_i \right) \left\| \breve{\mu}_{k^m} \left(\hat{\P}^m_n \right) - \hat{\mu}_{k^m} \left( \hat{\P}^m_n \right) \right\|_2 \\
            & \stackrel{(b)}{\le} \sum_{m=1}^M \left( \prod_{i=m+1}^M d_{k^i} \right) \left\| \breve{\mu}_{k^m} \left( \hat{\P}^m_n \right) - \hat{\mu}_{k^m} \left( \hat{\P}^m_n \right) \right\|_2 \\
            & \stackrel{(c)}{\le} d_k \sum_{m=1}^M \left\| \breve{\mu}_{k^m} \left( \hat{\P}^m_n \right) - \hat{\mu}_{k^m} \left( \hat{\P}^m_n \right) \right\|_2\label{eq:prop2_C2}
        \end{align}
        where (a) follows from Lemma~\ref{lemma:properties-of-tensor-products} [see \eqref{eq:diff-of-tensor-products_Am1}] by choosing $\X^m = \R^{d^m}$, $k^m$ to be the linear kernel, $\left\| \hat{\mu}_{k^m} \left( \hat{\P}^m_n \right) \right\|_2 \le \hat{B}_m $ for $m \in [M]$ and using that
        \begin{align}
            \left\| \breve{\mu}_{k^m} \left( \hat{\P}^m_n \right) \right\|_2 \stackrel{(d)}{=} \left\| \frac{1}{n} \sum_{i=1}^n \phi^m_{k^m, d^m}\left( X_i^m \right) \right\|_2 \stackrel{(e)}{\le} \frac{1}{n} \sum_{i=1}^n \left\| \phi^m_{k^m,d^m}\left( X_i^m \right) \right\|_2 \stackrel{(f)}{\le} 1
        \end{align}
        for $m \in [M]$, where (d) comes from the definition of $\breve{\mu}_{k^m} \left( \hat{\P}^m_n \right)$, (e) is implied by the homogeneity of norms and the triangle inequality, (f) follows from $\left\| \phi^m_{k^m,d^m}\left( X_i^m \right) \right\|_2 \le 1$ for all $i \in [n]$ by \eqref{eq:bound-on-truncated-feature-map}.
        For (b) in \eqref{eq:prop2_C2}, we show that one can choose $\hat{B}_m := d_{k^m}$ ($m\in [M]$); indeed, one has
        \begin{align}
            \left\| \hat{\mu}_{k^m} \left( \hat{\P}^m_n \right) \right\|_2 \stackrel{(g)}{=} \left\| \frac{1}{n} \sum_{i=1}^n \hat{\phi}^m_{k^m}\left( X_i^m \right) \right\|_2 \stackrel{(h)}{\le} \frac{1}{n} \sum_{i=1}^n \left\| \hat{\phi}^m_{k^m}\left( X_i^m \right) \right\|_2 \stackrel{(i)}{\le} d_{k^m},
        \end{align}
        where (g) is implied by the definition of $\hat{\mu}_{k^m} \left( \hat{\P}^m_n \right)$ and $\hat{\phi}^m_{k^m}(X_i^m) \in \R^{d^m}$ is specified in \eqref{eq:def_feature_est_by_adj}, (h) comes from the homogeneity of norms and the triangle inequality, (i) follows from $\left\| \hat{\phi}^m_{k^m}(X_i^m) \right\|_2 \le d_{k^m} $ for all $i \in [n]$. (c) in \eqref{eq:prop2_C2} was obtained by the assumption that $d_{k^m} \ge 1$ for all $m \in [M]$ and that $d_k = \prod_{m=1}^M d_{k^m}$.

        We continue the proof by bound the r.h.s.\ of \eqref{eq:prop2_C2}. We control $\left\| \breve{\mu}_{k^m} \left( \hat{\P}^m_n \right) - \hat{\mu}_{k^m} \left( \hat{\P}^m_n \right) \right\|_2$ by Lemma~\ref{concentration_joint_adj} for each fixed $m \in [M]$, using $M=1$ in Lemma~\ref{concentration_joint_adj} and a substitution $\eta \leftarrow \frac{\eta}{4M}$, followed by a union bound. In this case, under the same notations and assumptions as in the control of the term $C_1$, 
        we have for any $\eta \in (0,1)$
        \begin{align}\label{eq:prop2_diff_vector}
            \left\| \breve{\mu}_{k^m} \left( \hat{\P}^m_n \right) - \hat{\mu}_{k^m} \left( \hat{\P}^m_n \right) \right\|_2 \stackrel{(a)}{\le}  D_{k^m}\left(\frac{\eta}{4M} \right)
        \end{align}
        holds with $\mathbb{H}$-probability at least $1-\frac{\eta}{4M}$,\footnote{The factor $d_{k^m}$ can be removed in the second term on the r.h.s.\ of \eqref{eq:prop2_diff_vector} because, when $M=1$, in \eqref{eq:ineq_tilde_mu} the product $\prod_{i=2}^1 d_{k^i}=1$.} and hence for any $\eta \in (0,1)$ one gets that
        \begin{align}
            C_2 \le d_k \sum_{m=1}^M D_{k^m}\left(\frac{\eta}{4M} \right)
        \end{align}
        holds with $\mathbb{H}$-probability at least $1-\frac{\eta}{4}$.
        
    \end{itemize}

    Combining the bounds obtained for $C_1$ and $C_2$ while assuming (i) that Assumptions~\ref{assum_Delta}-\ref{assum_eigenval_hat_C_k} hold and for all $m\in [M]$ (ii) $C_1^m > 2\sqrt{2}$ and $c_2^m \ge 2$, (iii) $\delta_{d^m}^m \ne 0$, and (iv) $\lambda^m_{d^m+1} \le 8 \left( \log 2 \right) \sqrt{\frac{d^m \log \left( \frac{64Mn}{\eta}  \right)}{n}}$  holds for $n \ge \left( \frac{\eta}{32M} \right)^{-\frac{1}{c^m}}$, we got that for any $\eta \in (0,1)$ and for 
    \begin{align}
        n &\ge \max_{m \in [M]} \Big\{ \tilde{n}^m_{0,A1(i)}\left( \frac{\eta}{4M} \right), \tilde{n}^m_{0,A1(ii)}\left( \frac{\eta}{4M} \right), \tilde{n}^m_{0,A2(i)}\left( \frac{\eta}{4M} \right), \tilde{n}^m_{0,A2(ii)}\left( \frac{\eta}{4M} \right), \tilde{n}^m_{0,L5}\left( \frac{\eta}{4M} \right), \\
        &\quad \hspace{1.5cm} \tilde{n}^m_{0,L6}\left( \frac{\eta}{4M} \right), \tilde{n}_{0,\text{else-1}}^m\left( \frac{\eta}{4M} \right), \tilde{n}_{0,\text{else-2}}^m \Big\},
    \end{align}
    the bound 
    \begin{align}
        A_2 \le 2 d_k \sum_{m=1}^M D_{k^m}\left( \frac{\eta}{4M} \right) \label{eq:bound-A2}
    \end{align}
    holds with $\mathbb{H}$-probability at least $1-\frac{\eta}{2}$.
    
\end{itemize}

Finally, combining the bounds derived for $A_1$ and $A_2$ [\eqref{eq:bound-A1} and \eqref{eq:bound-A2}] while supposing that (i) Assumptions~\ref{assum_Delta}-\ref{assum_eigenval_hat_C_k} hold, and for all $m\in [M]$ (ii) the kernel $k^m: \X^m \times \X^m \to [0,1]$ is continuous on the compact metric space $\X^m$ and $\mathrm{supp}(\P^m)=\X^m$, (iii) $t_{k^m}\left(n, \frac{\eta}{8M}\right) \le \left\| C_{k^m} \right\|$ with $t_{k^m}$ defined in \eqref{eq:def_t_km}, (iv) $C_1^m > 2\sqrt{2}$ and $c_2^m \ge 2$, (v) $\delta_{d^m}^m \ne 0$, and (v) $\lambda^m_{d^m+1} \le 8 \left( \log 2 \right) \sqrt{\frac{d^m \log \left( \frac{64Mn}{\eta}  \right)}{n}}$ holds for $n \ge \left( \frac{\eta}{32M} \right)^{-\frac{1}{c^m}}$, we got that for any $\eta \in (0,1)$
    \begin{align}
        n &\ge \max_{m \in [M]} \Big\{\tilde{n}^m_{0,A1(i)}\left( \frac{\eta}{4M} \right), \tilde{n}^m_{0,A1(ii)}\left( \frac{\eta}{4M} \right), \tilde{n}^m_{0,A2(i)}\left( \frac{\eta}{4M} \right), \tilde{n}^m_{0,A2(ii)}\left( \frac{\eta}{4M} \right), \tilde{n}^m_{0,L5}\left( \frac{\eta}{4M} \right), \\
        &\quad \hspace{1.5cm} \tilde{n}^m_{0,L6}\left( \frac{\eta}{4M} \right), \tilde{n}_{0,\text{else-1}}^m\left( \frac{\eta}{4M} \right), \tilde{n}_{0,\text{else-2}}^m \Big\},
    \end{align}
we have
\begin{align}
    \MoveEqLeft \left| \mathrm{GHSIC}_{k} (\P) - \widehat{\mathrm{GHSIC}_{k}}  \right| \le 2 a_k \sqrt{\frac{2 \log \left(\frac{16}{\eta}\right)}{n} } + \sum_{m=1}^M 2 a_{k^m} \sqrt{\frac{2 \log \left(\frac{16M}{\eta}\right)}{n} } \\
    &\quad  + 6 \sqrt{\frac{3}{2}} \sum_{m=1}^M b_{k^m} \sqrt{\mathcal{N}_{C_{k^m}}\left(t_{k^m} \left(n, \frac{\eta}{8M}\right) \right)} \left(\lambda^m_{d^m+1} + t_{k^m} \left(n, \frac{\eta}{8M}\right) \right) \\
    & \quad + 2 d_k \sum_{m=1}^M D_{k^m}\left( \frac{\eta}{4M} \right) 
\end{align}
holds with $\mathbb{H}$-probability at least $1-\eta$.

\subsubsection{Validating universality of the Laplacian kernel}\label{sec:universality_verify_laplacian}

\begin{lemma}[$c$-universality of the Laplacian kernel]\label{lemma:universality_verify_laplacian} Let $m\in [M]$ be arbitrary and fixed. Below we prove (recall) that the kernel $k^m(x,y) \coloneq  e^{-\left| x-y \right|}$ is universal.

\end{lemma}

\begin{proof}(of Lemma~\ref{lemma:universality_verify_laplacian}):  The Laplacian kernel (even) on $\R^d$ is known to be $cc$-universal (\citealt[Proposition~14, Theorem~17]{micchelli06universal}; \citealt[page~2396]{sriperumbudur11universality}). $cc$-universality is equivalent to $c$-universality on every compact subset of the space (\citealt[Remark~4.1]{carmeli10vector}).  Since $\X^m=[0,1]\subset\R$ is compact, the restriction of the $cc$-universal Laplacian kernel to $[0,1]$ is $c$-universal on $\X^m$.
\end{proof}

\subsubsection{Validation of Assumptions~\ref{eq:poly-decay:lambda^m-decay}-\ref{eq:poly-decay:c(i)_assumption2} for the Laplacian kernel and the uniform distribution}\label{sec:assumption_verify_laplacian0}

\begin{lemma}[Assumptions~\ref{eq:poly-decay:lambda^m-decay}-\ref{eq:poly-decay:c(i)_assumption2} for the Laplacian kernel and the  uniform distribution]\label{lemma:assumption_verify_laplacian0}
Let $m \in [M]$ be arbitrary and fixed. For $X^m \sim U[0,1] =:\P^m$ and $k^m(x,y) \coloneq  e^{-\left| x-y \right|}$. Let $(\lambda_i^m)_{i\in I^m}$ denote the eigenvalues of the integral operator $T_{k^m}$. Then the followings hold.
\begin{itemize}
    \item \textbf{(Appropriate) polynomial spectral decay, as in Assumption~\ref{eq:poly-decay:lambda^m-decay}}: There exists an $i^m_0 \in \Zp$ such that for all $i \ge i^m_0$ 
    \begin{align}
    \lambda^m_i &= c^m(i) i^{-\alpha_m}, \label{eq:lambda-decay-Lapl-uniform}
    \end{align}
    where $c^m(i) \in [ c_l^m, c_u^m ]$ for some constants $0 < c_l^m < c_u^m $.
    \item \textbf{Assumption~\ref{eq:poly-decay:c(i)_assumption1}:} $\exists \lim_{i \to \infty} c^m(i) \eqcolon \bar{c}^m_1 > c_l^m$. 
    \item \textbf{Assumption~\ref{eq:poly-decay:c(i)_assumption2}:} There exists $i^m_1 \in \Zp$ and a constant $K^m$ with $0 \le K^m < \alpha_m c_l^m$ such that for all $i \ge i^m_1$ it holds that $\left| c^m(i) - c^m(i+1) \right| \le \frac{K^m}{i}$. 
\end{itemize}  

\end{lemma}

\begin{proof}(of Lemma~\ref{lemma:assumption_verify_laplacian0}): Let us fix an arbitrary $m \in [M]$. Throughout the proof, we choose $\X^m=[0,1]$, the Laplacian kernel $k^m(x,y) = e^{- \left| x - y \right|}$, and $X^m \sim U[0,1]$. In the following, we show that (i) the eigenvalue decay of $T_{k^m}$ (or equivalently that of $C_{k^m}$) is polynomial, and (ii) Assumptions~\ref{eq:poly-decay:lambda^m-decay}-\ref{eq:poly-decay:c(i)_assumption2} hold.

\begin{itemize}
    \item  \tb{Eigenspectrum decay of $T_{k^m}$ (checking of Assumption~\ref{eq:poly-decay:lambda^m-decay}: part-1}): 
    The task boils down \citep[Section~4]{hawkins1989some} to the eigenproblem  $\int_0^1 e^{-\left|s-t \right|} \phi(t)\d t = \lambda \phi(s)$, and the resulting eigenvalues are
    \begin{align}
        \lambda^m_i = \frac{2}{1+ \left(\omega^m_i\right)^2}, \quad i \in \Zp,\label{eq:laplacian_kernel_eigval}
    \end{align}
    where $\omega^m_i $ solves the equation 
    \begin{align}
        \tan (\omega_i^m) = \frac{2\omega_i^m}{\left(\omega_i^m \right)^2-1} \label{eq:omega_i_solution}
    \end{align}
    over $\omega^m_i \in \left( (2i-3)\frac{\pi}{2}, (2i-1)\frac{\pi}{2} \right)$\footnote{The sequence $\omega_i^m$ satisfies \citep[(4.8)]{hawkins1989some} $\omega^m_i \in \left( (2i-1)\frac{\pi}{2}, (2i+1)\frac{\pi}{2} \right)$ for $i =0,1,\ldots$. In our setting, however, the index starts from $i=1$. Re-indexing accordingly, we obtain $\omega^m_i \in \left( \left(2(i-1)-1 \right)\frac{\pi}{2}, \left(2(i-1)+1\right)\frac{\pi}{2} \right)$.} 
    while choosing $\omega_i^m$ to be the positive solution in each interval  \cite[page~13]{cai2020eigenvalue}. The decay rate is known to be $\lambda^m_i = \O(i^{-2})$ \citep[page~14]{cai2020eigenvalue}, giving a polynomial decay and $\alpha_m=2$ in Assumption~\ref{eq:poly-decay:lambda^m-decay}.

    \item \tb{Threshold $i^m_0$, bounds $c^m_l$ and $c^m_u$ (checking of Assumption~\ref{eq:poly-decay:lambda^m-decay}: part~2)}: To establish that there exists $i_0^m \in \Zp$ such that for all $i \ge i_0^m$, one has $\lambda^m_i = c^m(i) i^{-2}$ with $c^m(i) \in [c^m_l, c^m_u]$ for some constants $0 < c^m_l < c^m_u$, it suffices to define $c^m(i) := \lambda_i^m i^2$ and show that there exist $i_0^m\in\Zp$ and constants $0 < c_l^m < c_u^m$ such that for all $i \ge i_0^m$ one has $c^m(i) \in [c^m_l, c^m_u]$. With this definition at hand, we have
    \begin{align}
       c^m(i) & \stackrel{(a)}{=} \lambda_i^m i^2 \stackrel{(b)}{=} \frac{2 i^2}{1+ \left(\omega^m_i\right)^2} \stackrel{(c)}{\in} \left( \frac{2 i^2}{1 + \frac{\pi^2}{4} \left( 2 i-1 \right)^2}, \frac{2 i^2}{1 + \frac{\pi^2}{4} \left( 2 i-3 \right)^2} \right) \\
       & \stackrel{(d)}{\Rightarrow} \frac{2 i^2}{1 + \pi^2 \left( i- \frac{1}{2} \right)^2} < c^m(i) < \frac{2 i^2}{1 + \pi^2 \left( i- \frac{3}{2} \right)^2} , \label{eq:range_c^m_i}
    \end{align}
    where (a) is implied by the definition of $c^m(i)$ and (b) follows from \eqref{eq:laplacian_kernel_eigval}. For (c), since we have $\omega^m_i \in \left( (2i-3)\frac{\pi}{2}, (2i-1)\frac{\pi}{2} \right)$ (by the definition of $\omega^m_i$-s), we get that
    \begin{align}
        \MoveEqLeft (2i-3)\frac{\pi}{2} < \omega^m_i < (2i-1)\frac{\pi}{2} \stackrel{(e)}{\Rightarrow} 1 + \frac{\pi^2}{4}(2i-3)^2 < 1 + \left( \omega^m_i \right)^2 < 1 + \frac{\pi^2}{4}(2i-1)^2 \\
        & \stackrel{(f)}{\Rightarrow} \frac{1}{1 + \frac{\pi^2}{4}(2i-1)^2} < \frac{1}{1 + \left( \omega^m_i \right)^2} < \frac{1}{1 + \frac{\pi^2}{4}(2i-3)^2} \\
        & \stackrel{(g)}{\Rightarrow} \frac{2 i^2}{1 + \frac{\pi^2}{4}(2i-1)^2} < \frac{2 i^2}{1 + \left( \omega^m_i \right)^2} < \frac{2 i^2}{1 + \frac{\pi^2}{4}(2i-3)^2}. \label{eq:ineq_c^m_i_bound}
    \end{align}
    Here, (e) uses the fact that $2i-3>0$ and $2i-1>0$ for $i \ge i_0^m:=2$, 
    so applying the transformation $z \mapsto z^2+1$ preserves the order; (f) follows by taking reciprocals, which reverses the inequalities since all terms are positive; and (g) follows by multiplying with $2i^2>0$.
    (d) in \eqref{eq:range_c^m_i} is obtained by extracting $2$ from $(2i-1)$ and $(2i-3)$, followed by simplification.
    Let 
    \begin{align}
        a_i & \coloneq \frac{3i^2}{2 \left[ 1+\pi^2 \left( i-\frac{1}{2} \right)^2 \right]}, & b_i &\coloneq \frac{2i^2}{1+\pi^2 \left( i-\frac{3}{2} \right)^2},  \label{eq:def_ai_bi}\\
        c_l^m & \coloneq \inf_{i \ge i^m_0} a_i, & c_u^m & \coloneq \sup_{i \ge i^m_0} b_i.      
    \end{align}
    With these notations, by \eqref{eq:range_c^m_i} and the choice of $i_0^m$, we showed that for all $i\ge i_0^m$, $c_l^m\le c^m(i)\le c_u^m$. Next we prove the required properties of $c_l^m$ and $c_u^m$ by analyzing the $(a_i)_{i\ge i_0^m}$ and $(b_i)_{i\ge i_0^m}$ sequences.\footnote{Notice that a factor of 2 was replaced by $\frac{3}{2}$ at \eqref{eq:range_c^m_i} in the choice of $a_i$ in \eqref{eq:def_ai_bi}; this will ensure that the limit of $c^m(i)$ is strictly larger than $c_l^m$.} 
    \begin{itemize}
        \item \tb{Sequence $(a_i)_{i\ge i_0^m}$, $0<c_l^m$}: Taking any $i\ge i_0^m$, for $a_i$ we have
            \begin{align}
                \frac{2}{3} a_i - \frac{1}{\pi^2} & \stackrel{(a)}{=} \frac{i^2}{1+\pi^2\left(i-\frac{1}{2}\right)^2} - \frac{1}{\pi^2} \stackrel{(b)}{=} \frac{\pi^2 i^2  - \left[ 1+\pi^2\left(i-\frac{1}{2}\right)^2 \right]}{\pi^2 \left[ 1+\pi^2\left(i-\frac{1}{2}\right)^2 \right]} \\
                & \stackrel{(c)}{=} \frac{\pi^2 i^2 - 1 - \pi^2 \left( i^2 - i + \frac{1}{4} \right)}{\pi^2 \left[ 1+\pi^2\left(i-\frac{1}{2}\right)^2 \right]} = \frac{\pi^2 i -1-\frac{\pi^2}{4}}{\pi^2 \left[ 1+\pi^2\left(i-\frac{1}{2}\right)^2 \right]} \stackrel{(d)}{>} 0, \label{eq:compare_ai_1/pi^2}
            \end{align}
            where (a) comes from the definition of $a_i$, (b) follows from combining the two fractions, (c) is implied by expanding $\left(i-\frac{1}{2}\right)^2$ in the numerator, (d) follows from $i\ge 2$. Since $\lim_{i \to \infty}\frac{2}{3} a_i=\lim_{i \to \infty} \frac{i^2}{1+\pi^2\left(i-\frac{1}{2}\right)^2}=\frac{1}{\pi^2}$    
            and $\frac{2}{3} a_i > \frac{1}{\pi^2}$ for all $i \ge i_0^m$ by \eqref{eq:compare_ai_1/pi^2}, we have
            \begin{align}
                c_l^m = \inf_{i \ge i^m_0} a_i = \frac{3}{2} \inf_{i \ge i^m_0} \frac{2}{3} a_i = \frac{3}{2} \frac{1}{\pi^2} (\approx 0.15)>0. \label{eq:c_l^m_express}
            \end{align}
        \item \tb{Sequence $(b_i)_{i\ge i_0^m}$, $c_l^m < c_u^m$}: To understand the behavior of the sequence $(b_i)_{i\ge i_0^m}$ from monotonicity perspective, consider the function 
        \begin{align}
        B: [i_0^m,\infty) &\to \R, &x &\mapsto \frac{2x^2}{1+\pi^2 \left( x-\frac{3}{2} \right)^2};
        \end{align}
        which when evaluated on integers $i\ge i_0^m$ gives $b_i$ [$B(i) = b_i$ for all $i\ge i_0^m$].
        Differentiating $B$ yields
                \begin{align}
                    B'(x) & \stackrel{(a)}{=} \frac{2 \cdot 2 x \left[ 1+\pi^2 \left( x -\frac{3}{2} \right)^2 \right] - 2x^2  2 \pi^2 \left(x - \frac{3}{2} \right)}{\left[ 1+\pi^2 \left( x-\frac{3}{2} \right)^2 \right]^2} \\&\stackrel{(b)}{=} \frac{4x \left[ 1+ \pi^2 \left( x-\frac{3}{2} \right)^2 - \pi^2 x \left( x-\frac{3}{2} \right) \right]}{\left[ 1+\pi^2 \left( x-\frac{3}{2} \right)^2 \right]^2}\\
                    &\stackrel{(c)}{=} \frac{4x \left[ 1 + \pi^2 \left(x^2 - 3x + \frac{9}{4}\right) - \pi^2 x^2 +\frac{3}{2} \pi^2 x  \right]}{\left[ 1+\pi^2 \left( x-\frac{3}{2} \right)^2 \right]^2} \\
                    &\stackrel{(d)}{=} \frac{4x \left[ \left( \frac{3}{2}\pi^2 - 3 \pi^2 \right)x + \frac{9}{4} \pi^2 + 1 \right]}{\left[ 1+\pi^2 \left( x-\frac{3}{2} \right)^2 \right]^2},
                \end{align}      
                where (a) follows from the quotient rule, (b) is obtained by factoring out $4x$, and (c) follows from expanding $\left(x-\frac{3}{2}\right)^2$, the resulting expression was simplified in (d). For $x=i_0^m$ (for the derivative from the right), we have 
                \begin{align}
                    B^{'}\left(i_0^m\right) = \frac{8 \left[ \left( \frac{3}{2} \pi^2 - 3 \pi^2 \right)2 + \frac{9}{4} \pi^2 + 1 \right]}{\left[ 1+\pi^2 \left( 2-\frac{3}{2} \right)^2 \right]^2} = \frac{8 \left[ -\frac{3}{4} \pi^2 + 1 \right]}{\left[ 1+\pi^2 \left( 2-\frac{3}{2} \right)^2 \right]^2} < 0.
                \end{align}   
                Since $x \mapsto 1+\frac{9}{4}\pi^2-\frac{3}{2}\pi^2 x$ is decreasing, it follows that $1+\frac{9}{4}\pi^2-\frac{3}{2}\pi^2 x <0$ for all $x\ge 2$, and hence by using the positivity of the other 2 terms in $B'(x)$, we get that $B'(x) < 0$ for all $x \ge i_0^m$. This property specifically implies that $b_i$ is non-increasing for $i\ge i_0^m$. Therefore, we have 
                \begin{align}
                    c_u^m = \sup_{i \ge i^m_0} b_i = b_2 = \frac{32}{4 + \pi^2} (\approx 2.31). \label{eq:c_u^m_express}
                \end{align}     
                Combining \eqref{eq:c_l^m_express} and \eqref{eq:c_u^m_express} we obtained that $c_l^m$ and $c_u^m$ are finite and $0 < c_l^m < c_u^m$.
                    Therefore, there exist constants $0 < c_l^m < c_u^m $ such that $c^m(i) \in [ c_l^m, c_u^m ]$ for all $i \ge i_0^m$. 
            \end{itemize}
    \item \tb{Checking of Assumption~\ref{eq:poly-decay:c(i)_assumption1}}: We proceed by proving the two parts of Assumption~\ref{eq:poly-decay:c(i)_assumption1} separately.
    \begin{itemize}
        \item $\exists \bar{c}_1^m = \lim_{i \to \infty} c^m(i)$: In the following, we show that the sequences $\left(\frac{4}{3}a_i\right)_{i\ge i^m_0}$ and $(b_i)_{i\ge i^m_0}$ defined at \eqref{eq:def_ai_bi} converge to the same limits, hence according to  \eqref{eq:range_c^m_i} and the sandwich theorem $c^m(i) \in \left(\frac{4}{3} a_i,b_i \right)$ [holding by \eqref{eq:range_c^m_i}] also converges to the same limit. Indeed, we have 
            \begin{align}
              \lim_{i \to \infty} b_i& =  \lim_{i \to \infty} \frac{2i^2}{1+\pi^2 \left( i-\frac{1}{2} \right)^2} \stackrel{(a)}{=} \lim_{i \to \infty} \frac{2}{\frac{1}{i^2} + \pi^2 \left( 1-\frac{1}{2i} \right)^2}
                \stackrel{(b)}{=} \frac{2}{0 + \pi^2 \cdot 1}= \frac{2}{\pi^2},
            \end{align}
            where (a) comes from dividing both the numerator and the denominator by $i^2$, (b) hold as $\frac{1}{i^2} \to 0$ and $\left( 1-\frac{1}{2i} \right)^2 \to 1$ as $i \to \infty$. Similarly, we have 
            \begin{align}
            \lim_{i \to \infty} \frac{4}{3} a_i& =\lim_{i \to \infty} \frac{2i^2}{1+\pi^2 \left( i-\frac{3}{2} \right)^2} = \frac{2}{\pi^2} .
            \end{align}
            Therefore, we have that 
            \begin{align}\exists \bar{c}_1^m = \lim_{i \to \infty} c^m(i) = \frac{2}{\pi^2}. \label{eq:c_1^m-bar-exists}
            \end{align}
        \item $\bar{c}_1^m > c_l^m$: We have $\bar{c}_1^m > c_l^m$ by combining \eqref{eq:c_1^m-bar-exists} and \eqref{eq:c_l^m_express}. This settles Assumption~\ref{eq:poly-decay:c(i)_assumption1}.
        \end{itemize}
    \item \tb{Checking of Assumption~\ref{eq:poly-decay:c(i)_assumption2}}: We next show that the absolute value of the consecutive difference of $c^m(i)$ is $\O(i^{-2})$. 
    Since 
     \begin{align}
         \omega_i^m \in \left(\left(2i-3\right)\frac{\pi}{2}, \left(2i-1\right)\frac{\pi}{2}\right) = \left(\left(i - \frac{3}{2} \right) \pi, \left(i - \frac{1}{2} \right)\pi \right)
     \end{align}
     with the midpoint $(i-1)\pi$, we can write $\omega_i^m$ as $\omega_i^m=(i-1)\pi+\xi_i^m$ for $\xi_i^m \in \left( - \frac{\pi}{2}, \frac{\pi}{2} \right)$. Therefore, for $i \ge i_0^m=2$, we have $\omega_i^m > \left(i-\frac{3}{2}\right)\pi \ge \frac{\pi}{2} > 1$, and hence $\frac{2\omega_i^m}{(\omega_i^m)^2-1}>0$, and $\tan(\omega_i^m) \stackrel{\eqref{eq:omega_i_solution}}{=} \frac{2\omega_i^m}{\left(\omega_i^m\right)^2-1} > 0$. Hence, by the properties of the $\tan$ function, the solution for \eqref{eq:omega_i_solution} must lie to the right of the midpoint $(i-1)\pi$, in other words we have 
     \begin{align}
         \omega_i^m &= (i-1)\pi+\xi_i^m \quad \t{for } \xi_i^m \in \left( 0, \frac{\pi}{2} \right). \label{eq:rewrite_omega_i}
     \end{align}
    With this rewriting of $\omega^m_i$ at hand, consider
    \begin{align}
    \left|c^m(i) - c^{m}(i+1)\right| &\stackrel{(a)}{=} \left| \frac{2 i^2}{1+ \left(\omega^m_i\right)^2}  - \frac{2 (i+1)^2}{1+ \left(\omega^m_{i+1}\right)^2} \right| \\&\stackrel{(b)}{=} 2 \left| \frac{i^2 \left[1+ \left(\omega^m_{i+1}\right)^2\right] - (i+1)^2\left[1+ \left(\omega^m_{i}\right)^2\right]}{\left[1+ \left(\omega^m_{i}\right)^2\right] \left[1+ \left(\omega^m_{i+1}\right)^2\right]} \right| \\
    &\stackrel{(c)}{=} 2  \frac{\left|i^2-(i+1)^2 + i^2\left(\omega^m_{i+1}\right)^2 - (i+1)^2 \left(\omega^m_{i}\right)^2 \right|} {\left[1+\left(\omega^m_{i}\right)^2 \right] \left[ 1+\left(\omega^m_{i+1}\right)^2\right]} \\
    &\stackrel{(d)}{\le } 2 \frac{\left|-2i-1+ i^2\left(\omega^m_{i+1}\right)^2 - (i+1)^2 \left(\omega^m_{i}\right)^2 \right|}{ \left[1+(i-1)^2\pi^2\right] \left( 1+i^2\pi^2\right)}\\
    &\stackrel{(e)}{\le} 2 \frac{2i+1+\left| i^2\left(\omega^m_{i+1}\right)^2 - (i+1)^2 \left(\omega^m_{i}\right)^2 \right|}{ \left[1+(i-1)^2\pi^2\right] \left(1+i^2\pi^2\right)} \label{eq:c_i-diff}.
    \end{align}
    (a) follows from (b) in \eqref{eq:range_c^m_i}, in (b) we pulled out $2$ and combined the two fractions, in (c) terms were multiplied out in the numerator and the absolute value was discarded in the non-negative denominator, in (d) we combined the first 2 terms in the numerator after expanding $(i+1)^2$ and used \eqref{eq:rewrite_omega_i} in the denominator, in (e) we applied the triangle inequality. By \eqref{eq:rewrite_omega_i} one has
    \begin{align}
    (i-1)\pi&< \omega^m_i< \left(i-\frac{1}{2}\right)\pi,& i\pi < \omega^m_{i+1}< \left(i+\frac{1}{2}\right)\pi,
    \end{align}
    and hence one can bound the term in the absolute value of the numerator of \eqref{eq:c_i-diff} as 
    \begin{align}
    L_i &<
    i^2\left(\omega^m_{i+1}\right)^2 - (i+1)^2 \left(\omega^m_{i}\right)^2 < U_i,\\
    L_i &= i^2 (i\pi)^2 - (i+1)^2 \left[\left(i-\frac{1}{2}\right)\pi \right]^2, &
    U_i & = i^2 \left[\left(i+\frac{1}{2}\right)\pi\right]^2 - (i+1)^2 \left[(i-1)\pi\right]^2.
    \end{align}
    Rewriting the obtained lower and upper bounds, we get that
    \begin{align}
    L_i & \stackrel{(a)}{=} \pi^2 \left[i^4  - (i+1)^2 \left(i-\frac{1}{2}\right)^2 \right] \stackrel{(b)}{=} \pi^2 \left[i^4 - \left(i^2+2i+1\right) \left(i^2 - i + \frac{1}{4}\right)\right],\\
    &\stackrel{(c)}{=} \pi^2 \left[ i^4 - i^4 - i^3 + \frac{3}{4}i^2 + \frac{1}{2}i - \frac{1}{4}\right] \stackrel{(d)}{=} \pi^2 \left[ - i^3 + \frac{3}{4}i^2  + \frac{1}{2}i - \frac{1}{4} \right]. \\
    U_i & \stackrel{(e)}{=} \pi^2 \left[i^2 \left(i+\frac{1}{2}\right)^2 - (i+1)^2 (i-1)^2 \right] \stackrel{(f)}{=} \pi^{2} \left[i^2 \left( i^2 + i + \frac{1}{4} \right)  - \left(i^2 -1\right)^2 \right] \\
    &\stackrel{(g)}{=}\pi^{2}\left( i^4 + i^3 + \frac{1}{4}i^2 -i^4 + 2i^2 -1 \right) \stackrel{(h)}{=} \pi^{2} \left( i^3 + \frac{9}{4}i^2 - 1 \right).
    \end{align}
    In (a) we pulled out $\pi^{2}$, $(i+1)^2$ and $\left(i-\frac{1}{2}\right)^2$ were expanded in (b), terms were multiplied out in (c) and combined in (d), (e) is obtained by extracting $\pi^{2}$, (f) is implied by expanding $\left(i-\frac{1}{2}\right)^2$ and using the identity $(i+1)(i-1) = i^2-1$, terms were multiplied out in (g), terms were combined in (h). 
    Observing that the dominant term in $U_i$ and in $L_i$ is $\pi^2 i^3$ and $-\pi^2 i^3$, respectively, we have that 
    \begin{align}
        \left| i^2\left(\omega^m_{i+1}\right)^2 - (i+1)^2 \left(\omega^m_{i}\right)^2 \right| \le \max\{|U_i|,|L_i|\}  = \pi^2 i^3 + \O(i^2).
    \end{align}  
    Using that the largest term is $\pi^4 i^4$ in the denominator at the r.h.s.\ of \eqref{eq:c_i-diff}, we got that
    \begin{align}
        \limsup_{i \to \infty} i \left|c^m(i) - c^{m}(i+1)\right| \le \frac{2 \pi^2}{\pi^4} = \frac{2}{\pi^2}. 
    \end{align}
    This means that for any $\epsilon>0$, there exists $i_1^m \in \Zp$ such that $i \left|c^m(i) - c^{m}(i+1)\right| \le \frac{2}{\pi^2} + \epsilon$ for all $i \ge i_1^m$. 
    Using that $\alpha^m = 2$ and $c_l^m=\frac{3}{2\pi^2}$ [as shown in \eqref{eq:c_l^m_express}], one has that $\alpha^m c_l^m=\frac{3}{\pi^2}>0$. Since $\frac{2}{\pi^2}<\frac{3}{\pi^2}$, we can choose 
    \begin{align}
        K^m \in \left(\frac{2}{\pi^2},\frac{3}{\pi^2}\right). \label{eq:chose_K^m}
    \end{align}
    Taking $\epsilon=K^m-\frac{2}{\pi^2}>0$, we obtain that for all $i\ge i_1^m$, $i\left|c^m(i)-c^m(i+1)\right|\le K^m$. Equivalently, we have $\left|c^m(i+1)-c^m(i)\right| \le K^m i^{-1}$ for all $i \ge i_1^m$.
    This shows that Assumption~\ref{eq:poly-decay:c(i)_assumption2} holds.
\end{itemize}
\end{proof}

\subsubsection{Validation of Assumption~\ref{assum_Delta}-\ref{assum_eigenval_hat_C_k} for the Laplacian kernel and the uniform distribution}\label{sec:assumption_verify_laplacian}

To verify Assumptions~\ref{assum_Delta}–\ref{assum_eigenval_hat_C_k} for the Laplacian kernel with respect to the uniform distribution, we first establish the following lemma on expectations of the triangular distribution on $[-1,1]$.

\begin{lemma}[Expectations for the triangular distribution]\label{lemma:exp_triangular_dist}
    Let $Z$ follow a triangular distribution on $[-1, 1]$ with probability density function $f(z) = 1 - \left|z \right|$. Then, (i) $\E \left( e^{- \left| Z \right|} \right) = 2e^{-1}$, and (ii) $\E \left( e^{- 2\left| Z \right|} \right) = \frac{1+e^{-2}}{2}$. 
\end{lemma}

\begin{proof}(of Lemma~\ref{lemma:exp_triangular_dist}): We prove the two conclusions as follows.
\begin{itemize}
        \item Considering $\E \left( e^{- \left| Z \right|} \right)$, we have
        \begin{align}
            \MoveEqLeft \E \left( e^{- \left| Z \right|} \right) \stackrel{(a)}{=} \int_{-1}^1 e^{-\left| z \right|} \left( 1-\left|z \right| \right) \d z \stackrel{(b)}{=} 2 \int_{0}^1 e^{-z} \left( 1-z \right) \d z\\
            & \hspace{-0.5cm}\stackrel{(c)}{=} 2 \left( \int_{0}^1 e^{-z} \d z - \int_{0}^1 z e^{-z} \d z \right) 
            \stackrel{(d)}{=} 2 \left[ -e^{-z}\right]_{z=0}^{z=1} -2\left[ -ze^{-z} - e^{-z} \right]_{z=0}^{z=1} \\
            &\hspace{-0.5cm}\stackrel{(e)}{=}  \left[2ze^{-z}\right]_{z=0}^{z=1} \stackrel{(f)}{=}  2e^{-1}-0 \stackrel{(g)}{=} 2e^{-1}, \label{eq:exp_z}
        \end{align}
        where (a) follows from $\E \left[ f(Z) \right] = \int_{\R} f(z) p(z) \d z$ with $p$ denoting the pdf of $Z$, (b) holds as the integrand is an even function, (c) comes from carrying out the multiplication and the linearity of integration, (d) follows from the Newton-Leibniz formula applied with $\left( - e^{-z} \right)'= e^{-z}$ and $\left( -ze^{-z} - e^{-z} \right)' = ze^{-z}$, in (e) we combined the terms, followed by carrying out the substitutions in (f) and simplification in (g).
        \item Regarding $\E \left( e^{- 2 \left| Z \right|} \right)$, we have
        \begin{align}
            \E \left( e^{- 2\left| Z \right|} \right) & \stackrel{(a)}{=} \int_{-1}^1 e^{-2 \left| z \right|} \left( 1-\left|z \right| \right) \d z \stackrel{(b)}{=} 2 \int_{0}^1 e^{-2z} \left( 1-z \right) \d z \\
            & \stackrel{(c)}{=} 2 \left( \int_{0}^1 e^{-2z} \d z - \int_{0}^1 z e^{-2z} \d z \right)\\
            &\stackrel{(d)}{=} 2 \left[ -\frac{1}{2}e^{-2z}\right]_{z=0}^{z=1} -2 \left[-\frac{1}{2}ze^{-2z} - \frac{1}{4}e^{-2z} \right]_{z=0}^{z=1} \\
            & \stackrel{(e)}{=} \left[ze^{-2z} -\frac{1}{2}e^{-2z}\right]_{z=0}^{z=1}
            \stackrel{(f)}{=} \left(e^{-2} - \frac{1}{2}e^{-2}\right) - \left(-\frac{1}{2}\right) \stackrel{(g)}{=} \frac{1+e^{-2}}{2}, \label{eq:exp_2z}
        \end{align}
        where (a) follows from $\E \left[ f(Z) \right] = \int_{\R} f(z) p(z) \d z$ with $p$ denoting the pdf of $Z$, (b) holds as the integrand is an even function, (c) comes from carrying out the multiplication and the linearity of integration, (d) follows from the Newton-Leibniz formula applied with $\left( - \frac{1}{2}e^{-2z} \right)'= e^{-2z}$ and $\left( -\frac{1}{2}ze^{-2z} - \frac{1}{4}e^{-2z} \right)' = ze^{-2z}$, in (e) we merged the terms, followed by carrying out the substitutions in (f) and  simplification in (g).
    \end{itemize}   
\end{proof}

\begin{lemma}[Assumptions~\ref{assum_Delta}-\ref{assum_eigenval_hat_C_k} for the Laplacian kernel and the uniform distribution]\label{lemma:assumption_verify_laplacian}
    Let $m\in [M]$ be arbitrary and fixed. Below we prove that Assumptions~\ref{assum_Delta}-\ref{assum_eigenval_hat_C_k} hold for $X^m \sim U[0,1] =:\P^m$ and $k^m(x,y) \coloneq e^{-\left| x-y \right|}$.
    \begin{itemize}
        \item \tb{Assumption~\ref{assum_Delta}}: There exist universal constants $C_0^m > 0$ and $C_1^m > 0$, and for any $\eta \in (0,1)$ there exist $n_{0,(i)}^{m}(\eta) \in \Zp$ and $ n_{0,(ii)}^{m}(\eta) \in \Zp$ with the following properties. (i) For all $n \ge n_{0,(i)}^m(\eta)$ and for any Gram matrix $\b K^m = \left[ K^m_{i,j} \right]_{i,j=1}^n =\left[ k^m(X_i^m,X_j^m) \right]_{i,j=1}^n \in \R^{n \times n}$ associated to some sequence $\left( X_i^m \right)_{i=1}^n \sim (\P^m)^{n}$, $\Delta^m > C_0^m \log(n)$ holds with $\left(\P^m\right)^{\otimes n}$-probability at least $1-\eta$, where 
        $\Delta^m = \Delta^m \left(X_1^m,\dots, X_n^m;n\right):= \max_{i \in [n]} \sum_{j \in [n]} K^m_{i,j}$. (ii) For all $n \ge n_{0,(ii)}^m(\eta)$ and for any Gram matrix $\b K^m = \left[ K^m_{i,j} \right]_{i,j=1}^n =\left[ k^m(X_i^m,X_j^m) \right]_{i,j=1}^n \in \R^{n \times n}$ associated to some sequence $\left( X_i^m \right)_{i=1}^n \sim (\P^m)^{n}$, $\bar{\sigma}^m > C_1^m \sqrt{\frac{\log(n)}{n}}$ holds with $\left(\P^m\right)^{\otimes n}$-probability at least $1-\eta$, where $\bar{\sigma}^m = \bar{\sigma}^m \left(X_1^m,\dots, X_n^m;n\right) := \frac{1}{n^2} \sum_{i,j\in [n]} K^m_{i,j}\left( 1- K^m_{i,j} \right)$.               
        \item \tb{Assumption~\ref{assum_lambda_delta}}: There exist universal constants $c_0^m > 0$ and $c_2^m > 0$, and for any $\eta \in (0,1)$ there exist $n_{1,(i)}^{m}(\eta)\in \Zp$ and $n_{1,(ii)}^{m}(\eta) \in \Zp$ with the following properties. (i) For all $n \ge n_{1,(i)}^{m}(\eta)$ and for any Gram matrix $\b K^m = \left[ K^m_{i,j} \right]_{i,j=1}^n =\left[ k^m_{i,j}(X_i^m,X_j^m) \right]_{i,j=1}^n \in \R^{n \times n}$ associated to some sequence $\left( X_i^m \right)_{i=1}^n \sim (\P^m)^{n}$, $\frac{\lambda^m_{d^m}(\b K^m)}{\Delta^m} \ge c^m_0$ holds with $\left(\P^m\right)^{\otimes n}$-probability at least $1-\eta$. (ii) For all $n \ge n_{1,(ii)}^{m}(\eta)$, $\delta^m_{d^m} \ge 16 c_1^m(n) \sqrt{\frac{\log \left( \frac{6n}{\eta} \right)}{n}}$ holds, where $c_1^m(n) := c_2^m \log\left( 2 + \log (2n) \right)$.
        
        \item \tb{Assumption~\ref{assum_eigenval_C_k}}: The eigenvalues $(\lambda_i^m)_{i \in I^m}$ of $T_{k^m}$ are simple and positive. 

        \item \tb{Assumption~\ref{assum_eigenval_hat_C_k}}: For any realization, the eigenvalues $\left ( \hat{\lambda}^m_i \right )_{i \in \hat{I}^m} $ of $\hat{C}_{k^m}$ are simple and positive. 
    \end{itemize}
    
\end{lemma}

\begin{proof}(of Lemma~\ref{lemma:assumption_verify_laplacian}):
Let us fix an arbitrary $m\in [M]$. Throughout the proof, we choose $\X^m=[0,1]$, the Laplacian kernel $k^m(x,y) = e^{- \left| x - y \right|}$, and $X^m \sim U[0,1]$.

We proceed by checking Assumptions~\ref{assum_Delta}-\ref{assum_eigenval_hat_C_k}; to do so we calculate the expected value and the second moment of the off-diagonal entries of the Gram matrix generated by the assumed Laplacian kernel and uniform distribution. 
With $X_i^m \sim U[0,1]$ and $X_j^m \sim U[0,1]$ independent, $Z_{i,j}^m:= X_i^m - X_j^m$ follows a triangular distribution on $[-1, 1]$ with probability density function $f_Z(z) = 1 - \left|z \right|$ ($z\in [0,1]$). With this triangular distribution at hand, we compute the mentioned expectations. 
\begin{itemize}
    \item $\E \left( K^m_{i,j} \right)$ for $i \ne j$: we have
    \begin{align}
        \MoveEqLeft \E \left( K^m_{i,j} \right) \stackrel{(a)}{=} \E \left[ k^m(X_i^m, X_j^m) \right] \stackrel{(b)}{=} \E \left( e^{- \left| X_i^m - X_j^m \right|}\right) \stackrel{(c)}{=} \E \left( e^{- \left| Z_{i,j}^m \right|}\right) \stackrel{(d)}{=} 2e^{-1} ,\label{eq:exp_k_ij}
    \end{align}
    where (a), (b) and (c) follows from the definition of $K_{i,j}^m$, $k^m$ and $Z_{i,j}^m$, respectively. (d) is implied by Lemma~\ref{lemma:exp_triangular_dist}(i).
    \item $\E \left[ \left(K^m_{i,j}\right)^2 \right]$ for $i \ne j$: We have
    \begin{align}
        \E \left[ \left(K^m_{i,j}\right)^2 \right] & \stackrel{(a)}{=} \E \left[ \left[ k^m(X_i^m, X_j^m)\right]^2 \right] \stackrel{(b)}{=} \E \left( e^{- 2\left| X_i^m - X_j^m \right|}\right) \stackrel{(c)}{=} \E \left( e^{- 2\left| Z_{i,j}^m \right|}\right) \\
        & \stackrel{(d)}{=} \frac{1+e^{-2}}{2},\label{eq:exp_k^2_ij}
    \end{align}
    where (a), (b) and (c) follows from the definition of $K_{i,j}^m$, $k^m$ and $Z_{i,j}^m$, respectively. (d) is implied by Lemma~\ref{lemma:exp_triangular_dist}(ii).
\end{itemize}

Next we verify Assumption~\ref{assum_Delta}-\ref{assum_eigenval_hat_C_k}.

\begin{itemize}
    \item \textbf{Assumption~\ref{assum_Delta}}:
    \begin{itemize}
        \item[(i)] We verify that there exists a universal constant $C_0^m > 0$ and for any $\eta \in (0,1)$ there exists $n_{0,(i)}^m(\eta) \in \Zp$ such that for all $n \ge n_{0,(i)}^m(\eta)$ and for any Gram matrix $\b K^m =\left[ k^m(X_i^m,X_j^m) \right]_{i,j=1}^n \in \R^{n \times n}$,
        \begin{align}
         \MoveEqLeft \Delta^m = \max_{i \in [n]} \sum_{j=1}^n K^m_{i,j} > C_0^m \log (n) \Leftrightarrow \max_{i \in [n]} \frac{1}{n} \sum_{j=1}^n K^m_{i,j} > C_0^m \frac{\log(n)}{n} \label{eq:Delta^m_constraint}
        \end{align}
        holds with $\left(\P^m\right)^{\otimes n}$-probability at least $1-\eta$. Let $C_0^m$ be any positive real number. 
        Since we have 
        \begin{align}
            \max_{i \in [n]} \frac{1}{n} \sum_{j=1}^n K^m_{i,j} \stackrel{(a)}{\ge} \frac{1}{n} \sum_{i=1}^n \left( \frac{1}{n} \sum_{j=1}^n K^m_{i,j} \right),
        \end{align}
        where (a) follows from $\max_{i \in [n]} a_i \ge \frac{1}{n} \sum_{i=1}^n a_i$ holding for any sequence $(a_i)_{i=1}^n\in \R^n$. Hence, it is sufficient to show  that
        \begin{align}
            \frac{1}{n} \sum_{i=1}^n \left( \frac{1}{n} \sum_{j=1}^n K^m_{i,j} \right) = \frac{1}{n^2} \sum_{i,j=1}^{n} K^m_{i,j} > C_0^m \frac{\log(n)}{n}
        \end{align}
        holds with $\left(\P^m\right)^{\otimes n}$-probability at least $1-\eta$. Let 
        \begin{align}
        Y^{m} &= \frac{1}{n^2} \sum_{i,j=1}^{n} K^m_{i,j} = \frac{1}{n^2} \sum_{i,j=1}^n k^m (X_i^m, X_j^m),\\
            y^m &= f(x_1^m,\dots,x_n^m) := \frac{1}{n^2} \sum_{i,j=1}^n k^m (x_i^m, x_j^m)
        \end{align}
        with $(x_i^m)_{i=1}^n \in \left(\X^m\right)^n$ arbitrary and fixed.
        Next we verify the bounded difference property of $f$, as described in \eqref{eq:bounded-diff-property}. For any $\tilde{x}_l^m \in \X^m$, we have
            \begin{align}
                \MoveEqLeft \left| f(x_1^m,\ldots,x_l^m, \ldots, x_n^m) - f(x_1^m,\ldots, \tilde{x}_l^m, \ldots, x_n^m) \right| \\
                & \hspace{-0.8cm}  
                \stackrel{(a)}{=} \Bigg| \frac{1}{n^2} \Big( \sum_{i \in [n]\backslash\{l\}} \left[ k^m(x_i^m, x_l^m) - k^m(x_i^m, \tilde{x}_l^m) \right]\\
                & \hspace{-0.8cm} 
                \quad  \hspace{0.2cm} + \sum_{j \in [n]\backslash \{l\}} \left[ k^m(x_l^m, x_j^m) - k^m(\tilde{x}_l^m, x_j^m) \right] + \left[ k^m(x_l^m, x_l^m) - k^m(\tilde{x}_l^m, \tilde{x}_l^m) \right] \Big) \Bigg| \\
                & \hspace{-0.8cm} 
                \stackrel{(b)}{\le} \frac{1}{n^2} \sum_{i \in [n]\backslash \{l\}} \left| k^m(x_i^m, x_l^m) - k^m(x_i^m, \tilde{x}_l^m) \right| \\
                &\hspace{-0.8cm} \quad \hspace{0.2cm} + \frac{1}{n^2} \sum_{j \in [n] \backslash \{l\}} \left| k^m(x_l^m, x_j^m) - k^m(\tilde{x}_l^m, x_j^m) \right| + \frac{1}{n^2}\left| k^m(x_l^m, x_l^m) - k^m(\tilde{x}_l^m, \tilde{x}_l^m) \right| \\
                & \hspace{-0.8cm}
                \stackrel{(c)}{\le} \frac{1}{n^2} \left[(n-1) + (n-1) +  1 \right] \stackrel{(d)}{\le} \frac{2}{n} =: c_l,\label{eq:bound_diff_avg_K}
            \end{align}
            where (a) is obtained by considering the replacement of the $l$-th column except for its $l$-th element, the $l$-th row except for its $l$-th element, and the ($l$,$l$)-th element, respectively, in (b) we applied the triangle inequality, (c) is obtained by the assumption that $k^m: \X^m \times \X^m \to [0,1]$, (d) follows from $2n-1 \le 2n$ and simplification of the fraction. Applying the McDiarmid’s inequality (Lemma~\ref{mcdiarmid_ineq}) to control $\left| Y^m - \E(Y^m) \right|$, we have for any $t >0$
            \begin{align}
                \left(\P^m\right)^{\otimes n} \left( \left| Y^m - \E \left( Y^m \right) \right| \ge t \right) \le 2 \exp \left( - \frac{2 t^2}{\sum_{i=1}^n \left( \frac{2}{n}\right)^2 } \right) = 2 \exp \left( - \frac{n t^2}{2} \right).
            \end{align}
            By the reparameterization $\eta = 2 \exp \left( -\frac{n t^2}{2} \right)$, we have that
            \begin{align}
                \left(\P^m\right)^{\otimes n}\left(\left| Y^m  - \E \left( Y^m \right) \right| < \sqrt{\frac{2 \log \left(\frac{2}{\eta} \right)}{n}}\right) &\ge 1-\eta \Rightarrow \\
                 \left(\P^m\right)^{\otimes n}\left(Y^m > \E \left( Y^m \right) - \sqrt{\frac{2 \log \left(\frac{2}{\eta} \right)}{n}}\right) &\ge 1-\eta. \label{eq:Y^m-lower-bound_with_exp}
            \end{align}
            Then, we have for $\eta \in (0,1)$ that 
            \begin{align}
                \frac{1}{n^2} \sum_{i,j=1}^n K^m_{i,j} & = Y^m > \E \left( Y^m \right) - \sqrt{\frac{2 \log \left(\frac{2}{\eta}\right)}{n}} \\
                & \stackrel{(a)}{\Leftrightarrow} Y^m  > \frac{n^2-n}{n^2} 2e^{-1} - \sqrt{\frac{2 \log \left(\frac{2}{\eta}\right)}{n}} \label{eq:Ym:lower-bound}
            \end{align}
            holds with $\left(\P^m\right)^{\otimes n}$-probability at least $1-\eta$, where (a)  follows from
            \begin{align}
                & \quad \E \left( Y^m \right) = \E \left( \frac{1}{n^2} \sum_{i,j=1}^n K^m_{i,j} \right) \stackrel{(b)}{=} \E \left( \frac{1}{n^2} \sum_{i,j=1, i \ne j}^n K^m_{i,j} \right) \\
                & \stackrel{(c)}{=} \E \left( \frac{n^2-n}{n^2} \frac{1}{n^2-n} \sum_{i,j=1, i \ne j}^n K^m_{i,j} \right) \stackrel{(d)}{=} \frac{n^2-n}{n^2} \E \left( \frac{1}{n^2-n} \sum_{i,j=1, i \ne j}^n K^m_{i,j} \right) \\
                & \stackrel{(e)}{=} \frac{n^2-n}{n^2} \E (K^m_{i,j}) \stackrel{(f)}{=} \frac{n^2-n}{n^2} 2e^{-1} \label{eq:EY^m-rewriting}
            \end{align}
            holding for any $i\ne j \in [n]$, where (b) follows from the fact that $K^m_{i,j}=0$ when $i=j$ [by \eqref{eq:def_K^m_ij}], in (c) we multiplied and divided by the number of off-diagonal entries ($n^2-n$), (d) follows from the linearity of expectation, (e) follows from the fact that all off-diagonal entries $K^m_{i,j}$, $i\neq j$, are identically distributed, and (f) follows from \eqref{eq:exp_k_ij}.  
            Then we have
            \begin{align}
                \max_{i \in [n]} \frac{1}{n} \sum_{j=1}^n K^m_{i,j} > C_0^m \frac{\log(n)}{n} & \Leftarrow 
               \underbrace{\frac{1}{n^2} \sum_{i,j=1}^n K^m_{i,j}}_{=Y^m} > C_0^m \frac{\log(n)}{n} \\
               &\stackrel{\eqref{eq:Y^m-lower-bound_with_exp}}{\Leftarrow} \E \left( Y^m \right) - \sqrt{\frac{2 \log \left(\frac{2}{\eta} \right)}{n}} > C_0^m \frac{\log(n)}{n} \\
               & \Leftrightarrow \E \left( Y^m \right) > \sqrt{\frac{2 \log \left(\frac{2}{\eta} \right)}{n}} + C_0^m \frac{\log(n)}{n}\\
               & \stackrel{\eqref{eq:EY^m-rewriting}}{\Leftrightarrow} \frac{n^2-n}{n^2} 2e^{-1} > \sqrt{\frac{2\log \left(\frac{2}{\eta} \right)}{n}} + C_0^m \frac{\log(n)}{n} \label{eq:A1(i):sufficient}
            \end{align}
            holds with $\left(\P^m\right)^{\otimes n}$-probability at least $1-\eta$ for $\eta \in (0,1)$. Since the l.h.s.\ of \eqref{eq:A1(i):sufficient} goes to $2e^{-1}$ as $n \to \infty$ and the r.h.s.\ of \eqref{eq:A1(i):sufficient} depends on $\eta$ and for any fixed $\eta\in (0,1)$ it goes to zero as $n \to \infty$, we got that there exists a universal constant $C_0^m >0$ and for any $\eta \in (0,1)$ there exists $n_{0,(i)}^m(\eta) \in \Zp$ such that for all $n \ge n_{0,(i)}^m(\eta)$ we have that \eqref{eq:A1(i):sufficient} holds, and hence $\P^m\left(\Delta^m > C_0^m \log(n)\right)\ge 1-\eta$ is also satisfied.           
        
        \item[(ii)] We show that there exists a universal constant $C_1^m > 0$ and for any $\eta \in (0,1)$ there exists $n_{0,(ii)}^m(\eta) \in \Zp$ such that for all $n \ge n_{0,(ii)}^m(\eta)$, and for any Gram matrix $\b K^m =\left[ k^m(X_i^m,X_j^m) \right]_{i,j=1}^n \in \R^{n \times n}$, we have that 
        \begin{align}
         \bar{\sigma}^m &= \frac{1}{n^2} \sum_{i,j\in [n]} K^m_{i,j} \left( 1- K^m_{i,j} \right)\\
         &=  \underbrace{\frac{1}{n^2} \sum_{i,j \in [n]} \left[ K^m_{i,j} - \left(K^m_{i,j}\right)^2 \right]}_{= \underbrace{\frac{1}{n^2} \sum_{i,j \in [n]}  K^m_{i,j}}_{=Y_m} -  \underbrace{\frac{1}{n^2} \sum_{i,j \in [n]}  \left(K^m_{i,j}\right)^2}_{=:Z_m}} > C_1^m \sqrt{\frac{\log (n)}{n}} \label{eq:sigma-hat-control}
        \end{align} 
        holds with $\left(\P^m\right)^{\otimes n}$-probability at least $1-\eta$. Next we control the two terms on the l.h.s.\ of \eqref{eq:sigma-hat-control}.
        
        \begin{itemize}
            \item For $Y_m=\frac{1}{n^2} \sum_{i,j \in [n]} K^m_{i,j}$: According to \eqref{eq:Ym:lower-bound} we have for $\eta \in (0,1)$ that 
            \begin{align}
                Y^m  > \frac{n^2-n}{n^2} 2e^{-1} - \sqrt{\frac{2 \log \left(\frac{2}{\eta}\right)}{n}} \label{eq:Y^m-lower-bound}
            \end{align}
            holds with $\left(\P^m\right)^{\otimes n}$-probability at least $1-\eta$.
            
            \item For $Z_m=\frac{1}{n^2} \sum_{i,j \in [n]} \left(K^m_{i,j} \right)^2 $: Let us define
            \begin{align}
                z^m &:= \tilde{f}\left(x_1^m,\dots,x_n^m\right) := \frac{1}{n^2} \sum_{i,j=1}^n \left[ k^m (x_i^m, x_j^m) \right]^2,
            \end{align}
            with $(x_i^m)_{i=1}^n \in \left(\X^m\right)^n$ arbitrary and fixed. Since $\left[ k^m (x_i^m, x_j^m) \right]^2 \le k^m (x_i^m, x_j^m) \le 1$ using that $k^m(x^m,\tilde{x}^m)\in [0,1]$ for any $x^m, \tilde{x}^m \in \X^m$, \eqref{eq:bound_diff_avg_K} still holds; therefore by the McDiarmid's inequality (Lemma~\ref{mcdiarmid_ineq}) one has that 
            \begin{align}
                \left(\P^m\right)^{\otimes n}\left(\left| Z^m  - \E \left( Z^m \right) \right| < \sqrt{\frac{2 \log \left(\frac{2}{\eta} \right)}{n}}\right) &\ge 1-\eta \Rightarrow \\
                \left(\P^m\right)^{\otimes n}\left(Z^m < \E \left( Z^m \right) + \sqrt{\frac{2 \log \left(\frac{2}{\eta} \right)}{n}}\right) &\ge 1-\eta.
            \end{align}
            Therefore, we have for any $\eta \in (0,1)$ that
            \begin{align}
             \frac{1}{n^2} \sum_{i,j=1}^n \left(K^m_{i,j}\right)^2 = Z^m &< \E \left( Z^m \right) + \sqrt{\frac{2 \log \left(\frac{2}{\eta}\right)}{n}} \\
             & \stackrel{(a)}{=} \frac{n^2-n}{n^2} \frac{1+e^{-2}}{2} + \sqrt{\frac{2 \log \left(\frac{2}{\eta}\right)}{n}} \label{eq:Zm:upper-bound}
            \end{align}
            holds with $\left(\P^m\right)^{\otimes n}$-probability at least $1-\eta$, where (a)  follows from
            \begin{align}
                \E \left( Z^m \right) &= \E \left[ \frac{1}{n^2} \sum_{i,j=1}^n \left(K^m_{i,j}\right)^2 \right] \stackrel{\eqref{eq:def_K^m_ij}}{=} \E \left[ \frac{1}{n^2} \sum_{i,j=1, i \ne j}^n \left(K^m_{i,j}\right)^2 \right] \\
                & = \E \left[ \frac{n^2-n}{n^2} \frac{1}{n^2-n} \sum_{i,j=1, i \ne j}^n \left(K^m_{i,j}\right)^2 \right] = \frac{n^2-n}{n^2} \E \left[ \frac{1}{n^2-n} \sum_{i,j=1, i \ne j}^n \left(K^m_{i,j}\right)^2 \right] \\
                & = \frac{n^2-n}{n^2} \E \left[ (K^m_{i,j})^2 \right] \stackrel{(b)}{=} \frac{n^2-n}{n^2} \frac{1+e^{-2}}{2}, 
            \end{align}
            with (b) coming from \eqref{eq:exp_k^2_ij}.
            
        \end{itemize}
        Combining the results obtained with  probability $1-\frac{\eta}{2}$ and  $1-\frac{\eta}{2}$ using a union bound, it is sufficient that
        \begin{align}
            \MoveEqLeft \frac{1}{n^2} \sum_{i,j=1}^n \left[ K^m_{i,j} - \left(K^m_{i,j}\right)^2 \right] = Y^m-Z^m \\
            & \stackrel{(a)}{>} \frac{n^2-n}{n^2}2e^{-1} - \sqrt{\frac{2 \log \left(\frac{4}{\eta} \right)}{n}} - \left( \frac{n^2-n}{n^2} \frac{1+e^{-2}}{2} + \sqrt{\frac{2 \log \left(\frac{4}{\eta} \right)}{n}} \right) \\
            & = \frac{n^2-n}{n^2} 2e^{-1} - \frac{n^2-n}{n^2} \frac{1+e^{-2}}{2} - 2 \sqrt{\frac{2 \log \left(\frac{4}{\eta} \right)}{n}} \stackrel{?}{>} C_1^m \sqrt{\frac{\log(n)}{n}} \\
            & \Leftrightarrow \left( \frac{n^2-n}{n^2}2e^{-1} - \frac{n^2-n}{n^2} \frac{1+e^{-2}}{2} \right) > 2 \sqrt{\frac{2 \log \left(\frac{4}{\eta} \right)}{n}} + C_1^m \sqrt{\frac{\log(n)}{n}} \label{eq:A2(ii):to-satisfy}
        \end{align}
        holds with $\left(\P^m\right)^{\otimes n}$-probability at least $1-\eta$ for $\eta \in \left(0, 1 \right)$, where (a) follows from a substitution $\eta \leftarrow \frac{\eta}{2}$ in \eqref{eq:Y^m-lower-bound} and \eqref{eq:Zm:upper-bound}. Since the r.h.s.\ of \eqref{eq:A2(ii):to-satisfy} goes to $0$ as $n \to \infty$ and the l.h.s.\ of \eqref{eq:A2(ii):to-satisfy} goes to $2e^{-1} - \frac{1+e^{-2}}{2}\approx 0.17>0$ as $n \to \infty$, there exist constant $C_1^m > 0$ and for any $\eta \in (0,1)$ a threshold $n_{0,(ii)}^m(\eta) \in \Zp$ such that for all $n \ge n_{0,(ii)}^m(\eta)$ we have that \eqref{eq:A2(ii):to-satisfy} and hence $\bar{\sigma}^m > C_1^m \sqrt{\frac{\log(n)}{n}}$ holds with $\left(\P^m\right)^{\otimes n}$-probability at least $1-\eta$.
    \end{itemize}
    \item \textbf{Assumption~\ref{assum_lambda_delta}}:
    \begin{itemize}
        \item[(i)] We verify that there exists a universal constant $c_0^m > 0$ and $n_{1,(i)}^m(\eta) \in \Zp$ such that for all $n \ge n_{1,(i)}^m(\eta) $ and for any Gram matrix $\b K^m =\left[ k^m(X_i^m,X_j^m) \right]_{i,j=1}^n \in \R^{n \times n}$, we have
        \begin{align}
            \frac{\lambda_{d^m}^m (\b K^m)}{\Delta^m} \ge c_0^m \Leftrightarrow \frac{\lambda_{d^m}^m (\b K^m)}{n} \ge \frac{c_0^m \Delta^m}{n} \label{eq:eigen_lower_bound1}
        \end{align}
        holds with $\left(\P^m\right)^{\otimes n}$-probability at least $1-\eta$.  For $\lambda_{d^m}^m (\b K^m)$, we have by Lemma~\ref{eigen_Tk} [see \eqref{eq:gap_between_eigenvals}] that for any $\tau >0$
        \begin{align}
           \left(\P^{m}\right)^{\otimes n} \Bigg( \sup_{i \in \Zp} \left| \lambda_i^m - \frac{\lambda_i^m \left( \b K^m \right)}{n} \right| \le \underbrace{2\sqrt{2} a_{k^m}^2 \sqrt{\frac{\tau}{n}}}_{\le 2\sqrt{2} \sqrt{\frac{\tau}{n}}} \Bigg) \ge 1-2e^{-\tau},
        \end{align}
        where $a_{k^m}^2 = \sup_{x^m \in \X^m} k^m(x^m, x^m)$ and by the assumption that $k^m$ takes values in  $[0,1]$, one has $a_{k^m}^2 \le 1$. By setting $\eta = 2e^{-\tau} \Leftrightarrow \tau = \log \left( \frac{2}{\eta} \right)$, this means that for any fixed $i\in \Zp$ we have for any $\eta \in (0,1)$ that
        \begin{align}
            \hspace{-0.5cm} 
            \left| \lambda_i^m - \frac{\lambda_i^m \left( \b K^m \right)}{n} \right| \le 2\sqrt{2} \sqrt{\frac{\log \left( \frac{2}{\eta} \right)}{n}} \Rightarrow \frac{\lambda_i^m \left( \b K^m \right)}{n} \ge \lambda_i^m - 2\sqrt{2} \sqrt{\frac{\log \left( \frac{2}{\eta} \right)}{n}} \label{eq:eigen_lower_bound2}
        \end{align}
        holds with $\left(\P^m\right)^{\otimes n}$-probability at least $1-\eta$. 
         
        Next we prove that with $i=d^m$, there exists a constant $\tilde{c}_0^m > 0$ and for any $\eta \in (0,1)$ there exists a threshold $n_{1,(i)}^m(\eta) \in \Zp$ such that for all $n \ge n_{1,(i)}^m(\eta)$
        \begin{align}
            \lambda_{d^m}^m - 2\sqrt{2} \sqrt{\frac{\log \left( \frac{2}{\eta} \right)}{n}} \ge \tilde{c}_0^m. \label{eq:eigen_lower_bound3}
        \end{align}        
        Therefore, we want to show that there exists a constant $\tilde{c}_0^m > 0$ and for any $\eta \in (0,1)$ there exists a threshold $n_{1,(i)}^m(\eta) \in \Zp$ such that for all $n \ge n_{1,(i)}^m(\eta)$ and an $a (\coloneq c_l^m) > 0$
        \begin{align}
            \lambda_{d^m}^m \ge \tilde{c}_0^m+ 2\sqrt{2} \sqrt{\frac{\log \left( \frac{2}{\eta} \right)}{n}} \stackrel{(a)}{\Leftarrow} a \left(d^m\right)^{-2} \ge \tilde{c}_0^m+ 2\sqrt{2} \sqrt{\frac{\log \left( \frac{2}{\eta} \right)}{n}},
        \end{align}
        where (a) is implied by \eqref{eq:lambda-decay-Lapl-uniform}. For the uniform $X^m \sim U[0,1]$ and Laplacian kernel $k^m(x,y) = e^{-\left|x-y \right|}$, we have
        \begin{align}
            d^m  &\stackrel{\eqref{eq:poly-decay:d^m-choice}}{=} \left. n^{\frac{\theta_m}{\alpha_m}} \right|_{\alpha_m=2}=n^{\frac{\theta_m}{2}},& \theta_m  & \stackrel{\eqref{eq:poly-decay:theta^m-choice}}{=} \left. \frac{2\alpha_m +1}{6 \alpha_m + 5 }\right|_{\alpha_m = 2} = \frac{5}{17}. \label{eq:dm-thetam-value} 
        \end{align}
        Recall that we want that  
        \begin{align}
            \MoveEqLeft a\left( d^m \right)^{-2} \ge \tilde{c}_0^m + 2\sqrt{2} \sqrt{\frac{\log \left( \frac{2}{\eta} \right)}{n}} \stackrel{\eqref{eq:dm-thetam-value}}{\Leftrightarrow} a \left( n^{\frac{\theta_m}{2}} \right)^{-2} \ge \tilde c_0^m+ 2\sqrt{2} \sqrt{\frac{\log \left( \frac{2}{\eta} \right)}{n}} \\
            & \stackrel{(a)}{\Leftrightarrow} an^{-\theta_m} \ge \tilde c_0^m + 2\sqrt{2} \sqrt{\frac{\log \left( \frac{2}{\eta} \right)}{n}}\\
            &\stackrel{\eqref{eq:dm-thetam-value}}{\Leftrightarrow} a n^{-\frac{5}{17}} \ge \tilde c_0^m + 2\sqrt{2} \sqrt{\log \left( \frac{2}{\eta} \right)}n^{-\frac{1}{2}} , \label{eq:A2(i)-sufficient-condition}
        \end{align}
        where (a) holds by simplification. Since $\frac{5}{17} < \frac{1}{2}$, taking any fixed constant $\tilde{c}_0^m > 0$, for any $\eta \in (0,1)$ there exists a threshold $n_{1,(i)}^m(\eta) \in \Zp$ such that for all $n \ge n_{1,(i)}^m(\eta)$ \eqref{eq:A2(i)-sufficient-condition} holds. 

        Therefore by \eqref{eq:eigen_lower_bound2}, we got that there exists a constant $\tilde{c}_0^m > 0$, and for any $\eta \in (0,1)$ there exists a threshold $n_{1,(i)}^m(\eta) \in \mathbb{Z}_+$ such that for all $n \ge n_{1,(i)}^m(\eta)$, we have that $\frac{\lambda_{d^m}^m (\b K^m)}{n} \ge \tilde{c}_0^m $ holds with $\left(\P^m\right)^{\otimes n}$-probability at least $1-\eta$. Considering \eqref{eq:eigen_lower_bound1}, since we have 
        \begin{align}
            \frac{c_0^m \Delta^m}{n} \stackrel{(a)}{\le} c_0^m,
        \end{align}
        where (a) is implied by $\Delta^m \le n$ and by choosing $c_0^m \coloneq \tilde{c}_0^m$, we got that there exists a constant $c_0^m > 0$ and for any $\eta \in (0,1)$ there exists a threshold $n_{1,(i)}^m(\eta) \in \Zp$ such that for all $n \ge n_{1,(i)}^m(\eta)$, we have that $\frac{\lambda_{d^m}^m (\b K^m)}{\Delta^m} \ge c_0^m $ holds with $\left(\P^m\right)^{\otimes n}$-probability at least $1-\eta$.
        
        \item[(ii)] Our aim is to prove that there exists a universal constant $c_2^m > 0$ and for any $\eta \in (0,1)$ there exists a threshold $n_{1,(ii)}^m(\eta) \in \Zp$ such that for all $n \ge n_{1,(ii)}^m(\eta) $, $\delta^m_{d^m} \ge 16c_1^m(n)\sqrt{\frac{\log \left( \frac{6n}{\eta} \right)}{n}}$ holds with $c_1^m(n) = c_2^m \log \left( 2 + \log(2n) \right)$. For $d^m = n^{\frac{\theta_m}{\alpha_m}}$, \eqref{eq:lambda-decay-Lapl-uniform} with $\alpha_m=2$ implies that there exists $n_0^m \in \Zp$ such that for all $n \ge n_0^m$, 
        \begin{align}
        \lambda^m_{d^m} &= \underbrace{c^m(d^m)}_{\eqcolon c^m_{d^m}} \left( d^m \right)^{-2},&
        \lambda^m_{d^m+1} &= \underbrace{c^m(d^m+1)}_{\eqcolon c^m_{d^m+1}} \left( d^m+1 \right)^{-2}.
        \end{align}
         Then, for any $\epsilon>0$, there exist $n_1^m,n_2^m=n_2^m(\epsilon) \in \Zp$ (defined below) such that for all $n \ge \max \{n_0^m, n_1^m, n_2^m \}$, we have
        \begin{align}
            \delta_{d^m}^m &= \lambda^m_{d^m} - \lambda^m_{d^m+1} = c^m_{d^m} \left( d^m \right)^{-2} - c^m_{d^m+1} \left( d^m+1 \right)^{-2} \\
            & \stackrel{(a)}{=} \frac{c^m_{d^m} \left( d^m+1 \right)^2 -  c^m_{d^m+1} \left( d^m \right)^2}{\left( d^m \right)^2 \left( d^m+1 \right)^2} \stackrel{(b)}{=} \frac{c^m_{d^m} \left( d^m \right)^2 - c^m_{d^m+1} \left( d^m \right)^2 + 2 c^m_{d^m} d^m + c^m_{d^m}}{\left( d^m \right)^2 \left( d^m+1 \right)^2} \\
            & \stackrel{(c)}{=} \frac{\left(c^m_{d^m}- c^m_{d^m+1} \right) \left( d^m \right)^2 + 2 c^m_{d^m} d^m + c^m_{d^m}}{\left( d^m \right)^2 \left( d^m+1 \right)^2} \stackrel{(d)}{\ge} \frac{-K^m \left( d^m \right)^{-1} \left( d^m \right)^2 + 2 c^m_{d^m} d^m + c^m_{d^m}}{\left( d^m \right)^2 \left( d^m+1 \right)^2} \\
            & = \frac{ \left(2 c^m_{d^m} - K^m \right) d^m + c^m_{d^m}}{\left( d^m \right)^2 \left( d^m+1 \right)^2} \stackrel{(e)}{\ge} \frac{ \left[2 \left( \bar{c}_1^m - \epsilon \right) - K^m \right] d^m + c^m_{d^m}}{\left( d^m \right)^2 \left( d^m+1 \right)^2}\\
            & = \frac{\left[2 \bar{c}_1^m - \left(2 \epsilon + K^m \right) \right] d^m + c^m_{d^m}}{\left( d^m \right)^2 \left( d^m+1 \right)^2},
        \end{align}
        where (a) follows from combining the fractions, (b) is obtained by expanding $(d^m+1)^2$, and (c) follows by pulling out $\left(d^m\right)^2$ from the first two terms. Regarding (d), Assumption~\ref{eq:poly-decay:c(i)_assumption2} (holding in our case by Lemma~\ref{lemma:assumption_verify_laplacian0}) implies that there exists $n_1^m \in \Zp$ such that, for all $n \ge n_1^m$, 
        \begin{align}
            \left| c^m_{d^m} - c^m_{d^m+1} \right| \le \frac{K^m}{d^m} \Rightarrow c^m_{d^m} - c^m_{d^m+1} \ge -\frac{K^m}{d^m}.
        \end{align}
        Considering (e), Assumption~\ref{eq:poly-decay:c(i)_assumption1} (holding in our case by Lemma~\ref{lemma:assumption_verify_laplacian0}) implies that for all $\epsilon >0$ there exists $n_2^m =n_2^m(\epsilon) \in \Zp$ such that for all $n \ge n_2^m$  we have 
        \begin{align}
            \left| c^m_{d^m} - \bar{c}_1^m \right| \le \epsilon \Rightarrow c^m_{d^m} \ge \bar{c}_1^m - \epsilon.
        \end{align}
        Let us now choose $\epsilon \in \left( 0, \frac{1}{2\pi^2} \right)$. Using the facts that $K^m \in \left(\frac{2}{\pi^2}, \frac{3}{\pi^3} \right)$ by \eqref{eq:chose_K^m} (holding in our case by Lemma~\ref{lemma:assumption_verify_laplacian0}) and $2 \bar{c}_1^m = \frac{4}{\pi^2}$ by \eqref{eq:c_1^m-bar-exists} (holding in our case by Lemma~\ref{lemma:assumption_verify_laplacian0}), we conclude that $2 \bar{c}_1^m - \left(2 \epsilon + K^m \right) > \frac{4}{\pi^2} - \frac{1}{\pi^2} - \frac{3}{\pi^2} = 0 $. 
        Thus we have 
        \begin{align}
            \MoveEqLeft \delta_{d^m}^m = \Omega \left( \left(d^m \right)^{-3} \right) \stackrel{\eqref{eq:dm-thetam-value} }{=} \Omega \left( \left(n^{\frac{\theta_m}{\alpha_m}}\right)^{-3} \right) \stackrel{\eqref{eq:dm-thetam-value} }{=} \Omega \left( n^{-\frac{15}{34}} \right).
        \end{align}
        Compared with $16c_1^m(n)\sqrt{\frac{\log \left( \frac{6n}{\eta} \right)}{n}}$, as we have $16c_1^m(n)\sqrt{\frac{\log \left( \frac{6n}{\eta} \right)}{n}} = \tilde{\O}\left(n^{-\frac{1}{2}}\right)$ and $\frac{15}{34} < \frac{1}{2}$, thus there exists $n_3^m = n_3^m(\eta) \in \Zp$ such that for all $n\ge n_{1,(ii)}^m(\eta)\coloneq \max\{n_0^m,n_1^m,n_2^m,n_3^m\}$ one has $\delta_{d^m}^m \ge 16c_1^m(n)\sqrt{\frac{\log \left( \frac{6n}{\eta} \right)}{n}}$.
        Therefore, for any fixed constant $c_2^m >0$ and for any $\eta \in (0,1)$ there exists a threshold $n_{1,(ii)}^m(\eta) \in \Zp$ such that for all $n \ge n_{1,(ii)}^m(\eta)$ the inequality $\delta^m_{d^m} \ge 16c_1^m(n)\sqrt{\frac{\log \left( \frac{6n}{\eta} \right)}{n}}$ holds.  
        \end{itemize}    
        \item \textbf{Assumption~\ref{assum_eigenval_C_k}}: 
        According to \eqref{eq:laplacian_kernel_eigval} (holding in our case by Lemma~\ref{lemma:assumption_verify_laplacian0}), we have $\lambda_i^m = \frac{2}{1+\left(\omega_i^m\right)^2} > 0$ for all $i\in \Zp$. By \eqref{eq:rewrite_omega_i} (holding in our case by Lemma~\ref{lemma:assumption_verify_laplacian0}) we have $\omega_1^m \in \left( 0, \frac{\pi}{2} \right)$, $\omega_2^m \in \left(\pi, \frac{3\pi}{2} \right)$, $\omega_3^m \in \left(2\pi, \frac{5\pi}{2} \right), \ldots$; therefore, for $i<j$, we have $\omega_i^m < \omega_j^m$, which implies $\omega_i^m \neq \omega_j^m$ whenever $i \neq j$. Since the map $x \mapsto \frac{2}{1+x^2}$ is strictly decreasing for $x>0$, it follows that $\lambda_i^m$ are pairwise distinct, and hence simple.
        \item \textbf{Assumption~\ref{assum_eigenval_hat_C_k}}: In the following, we establish that the eigenvalues of the Gram matrix $\b K^m$ are positive and simple using properties of oscillation matrices, which by the relation $\lambda^m_i(\b K^m) = n \hat{\lambda}^m_i$ \citep[page~3]{sterge2020gain} implies that  Assumption~\ref{assum_eigenval_hat_C_k} holds. Let $X_1^m,\ldots,X_n^m \stackrel{\mathrm{i.i.d.}}{\sim} U[0,1]$, and consider the Gram matrix $\b K^m := \left[ k^m(X_i^m,X_j^m)\right]_{i,j=1}^n$ associated to the Laplace kernel $k^m(x,y)=e^{-\left|x-y\right|}$ on $[0,1]$. Since $U[0,1]$ is absolutely continuous w.r.t.\ the Lebesgue measure on $\R$, we have that $X_i^m \neq X_j^m$ almost surely for any $i \neq j$. Hence, almost surely, the sample points are pairwise distinct, and there exists a unique (realization-dependent) permutation $\pi$ such that $X_{\pi(1)}^m < X_{\pi(2)}^m < \cdots < X_{\pi(n)}^m$. Define $\tilde X_i^m := X_{\pi(i)}^m$, and let $\b K^m_{\mathrm{sort}} := \left[\tilde K^m_{i,j}\right]_{i,j=1}^n \in \R^{n \times n}$ be the Gram matrix induced by the Laplace kernel and the sorted sample $\left\{\tilde{X}_i^m\right\}_{i=1}^n$. For $i \le j$, we have $\tilde{X}_i^m \le \tilde{X}_j^m$; therefore
        \begin{align}
            \tilde{K}^m_{i,j} = e^{-\left|\tilde{X}_i^m - \tilde{X}_j^m \right|} = e^{-\left(\tilde{X}_j^m - \tilde{X}_i^m \right)} = e^{-\tilde{X}_j^m + \tilde{X}_i^m} = e^{\tilde{X}_i^m} e^{-\tilde{X}_j^m}.
        \end{align}
        Similarly, for $i > j$, 
        \begin{align}
            \tilde{K}^m_{i,j} = e^{-\left|\tilde{X}_i^m - \tilde{X}_j^m \right|} = e^{-\left(\tilde{X}_i^m - \tilde{X}_j^m \right)} = e^{-\tilde{X}_i^m + \tilde{X}_j^m} = e^{\tilde{X}_j^m} e^{-\tilde{X}_i^m}.
        \end{align}
        Thus, $\tilde{K}^m_{i,j} = e^{\tilde{X}^m_{\min\{i,j\}}} e^{-\tilde{X}^m_{\max\{i,j\}}}$. We now verify the three conditions in Lemma~\ref{lemm:oscillation_matrix}.
        \begin{itemize}
            \item \textbf{Totally positivity of $\b K^m_{\mathrm{sort}}$:} The $\tilde{K}^m_{i,j}$ can be written of the form $a_{ij} = c_{\min\{i,j\}} d_{\max\{i,j\}}$ with $c_i := e^{\tilde{X}_i^m} > 0$ and $d_i := e^{-\tilde{X}_i^m} > 0$. Moreover, $\frac{c_i}{d_i}=e^{2\tilde X_i^m}>0$ and $\frac{c_i}{d_i}<\frac{c_{i+1}}{d_{i+1}}$ for $i=1,\ldots,n-1$ since $\tilde{X}_i^m<\tilde{X}_{i+1}^m$. Hence $0<\frac{c_1}{d_1}<\cdots<\frac{c_n}{d_n}$, and Lemma~\ref{lemm:totally_positive_matrix} implies that $\b K^m_{\mathrm{sort}}$ is totally positive.
            \item \textbf{Nonsingularity of $\b K^m_{\mathrm{sort}}$}: 
            By Lemma~\ref{lemma:universality_verify_laplacian}, the Laplacian kernel $k^m$ is $cc$-universal on $\R$. By Corollary 4.3 of \citet{carmeli10vector}, $cc$-universality\footnote{$cc$-universality is referred to as compact-universality by  \citet{carmeli10vector}, which according to their paragraph after Definition~4.1 is equivalent to cc-universality as defined by \cite{caponnetto2008universal}.} of the kernel implies the positive definiteness of the associated Gram matrix for pairwise different samples ($X_i^m \neq X_j^m$; this holds in our case according to the discussion above). The positive definiteness of $\b K^m_{\t{sort}}$ guarantees its invertibility. 
            
            \item \textbf{Positive off-diagonal entries of $\b K^m_{\mathrm{sort}}$}: We have $\tilde{K}^m_{i,i+1} = e^{\tilde{X}^m_i} e^{-\tilde{X}^m_{i+1}} > 0$ and $\tilde{K}^m_{i+1,i} = e^{\tilde{X}^m_{i+1}} e^{-\tilde{X}^m_{i}} > 0$ according to the positivity of the exponential function.
        \end{itemize}
        Therefore, by Lemma~\ref{lemm:oscillation_matrix}, $\b K^m_{\mathrm{sort}}$ is an oscillation matrix. Lemma~\ref{lemm:oscillation_matrix_eigval} then implies that $\b K^m_{\mathrm{sort}}$ has positive and simple eigenvalues. Finally, since $\b K^m$ and $\b K^m_{\mathrm{sort}}$ differ only by a simultaneous permutation of rows and columns, there exists a (realization-dependent) permutation matrix $\b P \in \R^{n \times n}$ such that $\b K^m=\b P \b K^m_{\mathrm{sort}} \b P^\top$. Hence $\b K^m$ and $\b K^m_{\mathrm{sort}}$ share the eigenvalues, and therefore the eigenvalues of $\b K^m$ are also positive and simple, which finishes the proof.
        
\end{itemize}
\end{proof}

\section{Independence Testing with the Permutation Approach}\label{Testing_Procedure}

In this section, we leverage the proposed GHSIC$_k$ estimator in independence testing of LPRGs  
$(G^m)_{m=1}^M$: 
\begin{align}
    H_0: (G^m)_{m=1}^M \ \t{are independent}, \quad 
    H_1: (G^m)_{m=1}^M \ \t{are not independent}.
\end{align}
using the permutation approach. Particularly, to our permutation test, let $\varphi^m: [n]  \rightarrow [n] $ denote a permutation of $[n]$ for each $m \in [M]$. Recall that in our estimator we have $\hat{\b K}^m = \left(\hat{\bm{\Phi}}^m\right)^\T\hat{\bm{\Phi}}^m \in \R^{n \times n}$, where $\hat{\bm{\Phi}}^m = \left [  \hat{\phi}_{k^m}\left(x^m_1\right), \ldots,\hat{\phi}_{k^m}\left(x^m_n\right)   \right ]_{i=1}^n \in \R^{d^m \times n}$ and $m\in [M]$. For each $m \in [M]$, shuffling the columns of $\hat{\bm{\Phi}}^m$ according to $\varphi^m$, one gets with $\X=\times_{m=1}^M\X_m$ and $d=\sum_{m=1}^M d^m$:
\begin{align}
\underbrace{\begin{bmatrix}
x_1^1  & \cdots & x_n^1 \\
\vdots &        & \vdots \\
x_1^M  & \cdots & x_n^M 
\end{bmatrix}}_{(x^m_i)_{m \in [M],i \in [n]} \in \X^n}
\xrightarrow{\left(\hat{\phi}_{k^m}\right)_{m=1}^M}
\underbrace{\begin{bmatrix}
\hat{\phi}_{k^1}\left(x_1^1\right)  & \cdots & \hat{\phi}_{k^1}\left(x_n^1\right) \\
\vdots &        & \vdots \\
\hat{\phi}_{k^M}\left(x_1^M\right)  & \cdots & \hat{\phi}_{k^M}\left(x_n^M\right)
\end{bmatrix}}_{\left( \hat{\phi}_{k^m}\left(x_i^m\right) \right)_{ m \in [M],i \in [n]}\in \R^{d\times n}}
\xrightarrow{\left(\varphi^m\right)_{m=1}^M}
\underbrace{\begin{bmatrix}
\hat{\phi}_{k^1}\left(x_{\varphi^1(1)}^1\right)  & \cdots & \hat{\phi}_{k^1}\left(x_{\varphi^1(n)}^1\right) \\
\vdots &        & \vdots \\
\hat{\phi}_{k^M}\left(x_{\varphi^M(1)}^M\right)  & \cdots & \hat{\phi}_{k^M}\left(x_{\varphi^M(n)}^M\right)
\end{bmatrix}}_{\bm{\varphi}\left(\hat{\bm{\Phi}}\right):=\left( \hat{\phi}_{k^m}\left(x_{\varphi^m(i)}^m\right) \right)_{m \in [M],i \in [n]} \in \R^{d\times n}},
\end{align}
where  $\bm{\varphi} = \left(\varphi^m\right)_{m=1}^M$ represents the collection of the $M$ permutations, this leads to the Gram matrix estimate $\varphi^m \left(\hat{\b K}^m\right):=\left(\varphi^m \left(\hat{\bm{\Phi}}^m\right)\right)^\T \varphi^m \left(\hat{\bm{\Phi}}^m\right) \in \R^{n\times n}$ which can be substituted to \eqref{est_HSIC2} instead of $\hat{\b K}^m$ to get the estimate $\widehat{\mathrm{GHSIC}^2_k}\left(\bm{\varphi}\right)$. Repeating the process $B \in \Z^+$ times with random permutations, one arrives at the estimates $\left(\widehat{\mathrm{GHSIC}^2_k}\left(\bm{\varphi}_b\right)\right)_{b=1}^B$, which gives rise to the  empirical cumulative distribution estimate:
\begin{align}
    \hat{R}(t) = \frac{1}{B}\sum_{b=1}^B \mathbbm{1}_{\left \{ \widehat{\mathrm{GHSIC}^2_k}\left ( \bm{\varphi}_b \right ) \le t \right \} } \in [0,1], \quad t \in \R.
\end{align}

Then for a level $\alpha \in (0,1)$, let $\hat{R}^{-1}(1-\alpha)$ be the $(1-\alpha)$ quantile of $\hat{R}$. Then we reject the null hypothesis iff $\widehat{\mathrm{GHSIC}^2_k} \ge \hat{R}^{-1}(1-\alpha)$.

\section{Optimal Embedding Dimensions}\label{diss_opt_rate}

Throughout the paper, we assume that the embedding dimensions $\left(d^m \right)_{m=1}^M$-s can depend on the node size $n$: $d^m=d^m(n)$ for $m\in [M]$. We consider and discuss two examples: polynomial and exponential decay of the spectrum of $\left(T_{k^m}\right)_{m=1}^M$-s (or equivalently that of $\left(C_{k^m}\right)_{m=1}^M$-s). Below we analyze the convergence rates attainable (using the bound in Theorem~\ref{errorbound_GHSIC1}; optimality is meant in this sense) with the appropriate choice of the embedding dimensions $\left(d^m \right)_{m=1}^M$.

Let us recall the derived error bound of Theorem~\ref{errorbound_GHSIC1}, and introduce the notations
\begin{align}
        \MoveEqLeft \left | \mathrm{GHSIC}_{k} (\P) - \widehat{\mathrm{GHSIC}_{k}} \right | \le \underbrace{2 a_k \sqrt{\frac{2\log \left( \frac{16}{\eta} \right)}{n}}}_{=: U_1} + 2 \sum_{m=1}^M \underbrace{a_{k^m} \sqrt{\frac{2\log \left( \frac{16M}{\eta} \right)}{n}} }_{=: U_{2,m}} \\
        & \hspace{-1.2cm} + \sum_{m=1}^M \underbrace{\left[ \overbrace{6 \sqrt{\frac{3}{2}} b_{k^m} \sqrt{\mathcal{N}_{C_{k^m}}\left(t_{k^m} \left(n, \frac{\eta}{8M}\right) \right)} \left(\lambda^m_{d^m+1} + t_{k^m} \left(n, \frac{\eta}{8M}\right) \right)}^{=:A_m} + \overbrace{2 d_k D_{k^m}\left( \frac{\eta}{4M} \right)}^{=:B_m} \right] }_{=:U_{3,m}}, \label{eq:U1-U2m-U3m}
\end{align}
and
\begin{align}
    D_{k^m}(\eta) & = \underbrace{312 \sqrt{2} c_1^m(n) d^m \left( \delta^m_{d^m} \right)^{-\frac{1}{2}}  \sqrt{\frac{\log \left ( \frac{864 n}{\eta} \right ) }{n}}}_{=:B_{m,1}} + \underbrace{896 \frac{\left(d^m \right)^{\frac{1}{2}} \left(c_1^m(n) \right)^2}{\sqrt{c_0^m}} \left( \delta^m_{d^m} \right)^{-2} \frac{\log \left( \frac{432 n}{\eta} \right)}{n}}_{=:B_{m,2}} \\
    & + \underbrace{1024 \frac{\left(d^m \right)^{\frac{1}{2}}\left(c_1^m(n) \right)^3}{\sqrt{c_0^m}} \left( \delta^m_{d^m} \right)^{-2} \frac{\log^{\frac{3}{2}} \left( \frac{432 n}{\eta} \right)}{n^{\frac{3}{2}}}}_{=:B_{m,3}} + \underbrace{64 \frac{\left(c_1^m(n) \right)^2}{\sqrt{c_0^m}} \left( \delta^m_{d^m} \right)^{-1} \frac{\left(d^m \right)^{\frac{1}{2}} \log \left( \frac{432n}{\eta} \right)}{n}}_{=:B_{m,4}} \\
    & + \underbrace{4 \frac{(d^m)^{\frac{1}{2}} c_1^m(n)}{\sqrt{c_0^m}} \sqrt{\frac{\log \left( \frac{144n}{\eta}  \right)}{n}}}_{=:B_{m,5}}.
\end{align}

\paragraph{Polynomial spectral decay of $\left(T_{k^m}\right)_{m=1}^M$-s.}    The trace class property and the non-negativity of $T_{k^m}$ mean that $\sum_{i\in I} \lambda^m_i < \infty$. Let $m\in [M]$ be arbitrary and fixed. We assume that the eigenvalues of $T_{k^m}$ decay polynomially; particularly, we suppose that 
    \begin{itemize}
        \item[] \begin{assumption}[eigenvalue decay]\label{eq:poly-decay:lambda^m-decay}
            There exists an $i^m_0 \in \Zp$ such that for all $i \ge i^m_0$ 
            \begin{align}
            \lambda^m_i &= c^m(i) i^{-\alpha_m},
            \end{align}
            where $c^m(i) \in [ c_l^m, c_u^m ]$ for some constants $0 < c_l^m < c_u^m $,\footnote{Assumption~\ref{eq:poly-decay:lambda^m-decay} implies that $\alpha_m > 1$.}  
        \end{assumption}
        \item[]\begin{assumption}[convergence]\label{eq:poly-decay:c(i)_assumption1}
            $\exists \lim_{i \to \infty} c^m(i) \eqcolon \bar{c}^m_1 > c_l^m$. 
        \end{assumption}
        \item[]
        \begin{assumption}[gap control]\label{eq:poly-decay:c(i)_assumption2}
        There exists $i^m_1 \in \Zp$ and a constant $K^m$ with $0 \le K^m < \alpha_m c_l^m$ such that for all $i \ge i^m_1$ it holds that $\left| c^m(i) - c^m(i+1) \right| \le \frac{K^m}{i}$.\footnote{Assumptions~\ref{eq:poly-decay:lambda^m-decay}-\ref{eq:poly-decay:c(i)_assumption2} are satisfied for instance for an identically constant function $c^{m}>0$; in this case $\bar{c}_1^m=c^m$, and one can choose $c_l^m = c^m / 2$, $c_u^m = 2 c^m$, $K^{m}=0$.}
        \end{assumption}
    \end{itemize}
    We also suppose that the embedding dimension is chosen as
    \begin{align}
        d^m &= n^{\frac{\theta_m}{\alpha_m}} \label{eq:poly-decay:d^m-choice}
    \end{align}
    for $0 < \theta_m \le \alpha_m$.$^{\ref{footnote:reduction}}$ 
        \begin{itemize}[labelindent=0em,leftmargin=*,topsep=0cm,partopsep=0cm,parsep=0cm,itemsep=0cm]
            \item Term $U_1$: The first term is $U_1 = \O\left(n^{-\frac{1}{2}}\right)$.
            \item Term $U_{2,m}$ (with $m\in [M]$ fixed): The order of this term is $U_{2,m} = \O\left(n^{-\frac{1}{2}}\right)$.

            \item Term $U_{3,m}$ (with $m\in [M]$ fixed):
            \begin{itemize}
            
                \item Term $A_m$:  Under the imposed polynomial spectral decay, one can show (Lemma~\ref{eigenvalue_decay}) that $\mathcal{N}_{C_{k^m}}\left(t_{k^m}\left(n,\frac{\eta}{8M}\right)\right) = \Theta \left(\left[t_{k^m} \left(n,\frac{\eta}{8M}\right)\right]^{-\frac{1}{\alpha_m}}\right) = \Theta \left(  \frac{n^{\frac{1}{\alpha_m}}}{\log^{\frac{1}{\alpha_m}} (n)}  \right)$ in the 2nd equality using that $t_{k^m}\left(n,\frac{\eta}{8M}\right) = \Theta \left(\frac{\log(n)}{n}\right)$.  Then we have $\lambda^m_{d^m+1} = \Theta \left((d^m+1)^{-\alpha_m}\right)=\Theta\left(n^{- \theta_m}\right)$ leveraging Assumption~\ref{eq:poly-decay:lambda^m-decay} and \eqref{eq:poly-decay:d^m-choice}, and therefore  by using that $t_{k^m}\left(n,\frac{\eta}{8M}\right) =\Theta\left(\frac{\log(n)}{n}\right)$, one gets 
        \begin{align}
        A_m = \Theta \left( \frac{n^{\frac{1}{2\alpha_m}}}{\log^{\frac{1}{2\alpha_m}} (n)}  \left(\frac{1}{n^{\theta_m}} + \frac{\log (n)}{n} \right)\right ) = \left\{ \begin{array}{cl}
        \Theta \left (\frac{n^{\frac{1}{2\alpha_m}-1}}{\log^{\frac{1}{2\alpha_m}-1}(n)} \right),&  \ \theta_m \geq 1 \\
        \Theta \left (\frac{n^{\frac{1}{2\alpha_m}-\theta_m}}{\log ^{\frac{1}{2\alpha_m}} (n)} \right) ,&  \ \theta_m < 1
        \end{array} \right. .
        \end{align}
        On the 
        \begin{itemize}
            \item  first branch ($\theta_m \ge 1$): $A_m\xrightarrow{n\to \infty} 0$ iff.\ $\frac{1}{2\alpha_m}-1<0 \Leftrightarrow \frac{1}{2} < \alpha_m$; this holds since $1<\alpha_m$.
            \item second branch ($\theta_m <1$): $A_m\xrightarrow{n\to \infty} 0$ iff.\ $\frac{1}{2\alpha_m}-\theta_m \le 0$, i.e.\ $\frac{1}{2\alpha_m} \le \theta_m$, leading to two sub-cases.
            \begin{itemize}

                \item When $\theta_m > \frac{1}{2 \alpha_m}$, one has $\frac{1}{2\alpha_m}-\theta_m < 0$. Therefore, $n^{\frac{1}{2\alpha_m}-\theta_m} \to 0$ as $n \to \infty$, and $A_m = \Theta \left(\frac{1}{n^{\theta_m - \frac{1}{2\alpha_m}} \log ^{\frac{1}{2\alpha_m}} (n)}\right) \xrightarrow{n\to \infty}0$.
                \item When $\theta_m = \frac{1}{2 \alpha_m}$, $n^{\frac{1}{2\alpha_m}-\theta_m} = 1$, and hence $A_m = \Theta \left(\frac{1}{\log ^{\frac{1}{2\alpha_m}} (n)}\right) \xrightarrow{n\to \infty}0$. 
            \end{itemize}
            The orders of $A_m$ and $B_m$ determine that of $U_{3,m}$. Next we will show that 
            \begin{align}
            B_m &= \tilde{\O}(n^{-a}) \quad \t{ with }a>0. \label{eq:B_m:promised-rate}
            \end{align}
            This means that it is sufficient to focus on the case where $\theta_m > \frac{1}{2 \alpha_m}$; the case of $\theta_m = \frac{1}{2 \alpha_m}$ would give a slower convergence rate. 
        \end{itemize}
        
                \item Term $B_m$: $B_m$ is determined by $D_{k^m}$, which itself is controlled by (the negative exponents of) $\delta_{d^m}^m$; see $B_{m,1}$ [with $(\delta^m_{d^m})^{-1/2}$], $B_{m,2}$ [with $(\delta^m_{d^m})^{-2}$], $B_{m,3}$ [with $(\delta^m_{d^m})^{-2}$] and $B_{m,4}$ [with $(\delta^m_{d^m})^{-1}$]. Therefore, to obtain an upper bound $B_m$, it is sufficient to derive a lower bound on the eigengap $\delta_{d^m}^m=\lambda_{d^m}^m - \lambda_{d^m+1}^m$. Let $a^m := \bar{c}^m_1 - c_l^m >0$ (the positivity of $a^m$ holds by Assumption~\ref{eq:poly-decay:c(i)_assumption1}). Let $\epsilon \in (0, a^m)$ be fixed. By the convergence of the sequence $c^m(i)$, there exists a $i_2^m=i_2^m(\epsilon) \in \Zp$ such that for all $i \ge i_2^m$, one has $\left| c^m(i+1) - \bar{c}_1^m \right| \le \epsilon$. Let  
                \begin{align}
                i \ge  L^m\coloneq L^m(\epsilon)\coloneq \max \left\{i_0^m, i_1^m , i_2^m(\epsilon) \right\}\label{eq:i-constraint}
                \end{align}
                be arbitrary but fixed. According to Assumption~\ref{eq:poly-decay:lambda^m-decay}, in this case, we have $\lambda_i^m = c^m(i) i^{-\alpha_m}$. Let $C^m_i = c^m(i)$ and $C^m_{i+1} = c^m(i+1)$; with this notation, by  definition $\lambda_i^m = C^m_i i^{-\alpha_m}$ and $\lambda_{i+1}^m = C^m_{i+1} (i+1)^{-\alpha_m}$, and $C^m_i, C^m_{i+1} \in [c_l^m,c_u^m]$. Therefore, we have
                \begin{align}
                    \delta_{i}^m &= \lambda_i^m - \lambda_{i+1}^m = C^m_i i^{-\alpha_m} - C^m_{i+1} (i+1)^{-\alpha_m} \stackrel{(a)}{=} C^m_i i^{-\alpha_m} - C^m_{i+1} \left[ i \left( 1+\frac{1}{i} \right) \right]^{-\alpha_m} \\
                    & \stackrel{(b)}{=} C^m_i i^{-\alpha_m} - C^m_{i+1} i^{-\alpha_m} \left( 1+\frac{1}{i} \right)^{-\alpha_m} \stackrel{(c)}{\ge} C^m_i i^{-\alpha_m} - C^m_{i+1} i^{-\alpha_m}\left[ 1-\frac{\alpha_m}{i} + \frac{\alpha_m(\alpha_m+1)}{2 i^2} \right] \\
                    & \stackrel{(d)}{=} C^m_i i^{-\alpha_m} - C^m_{i+1} i^{-\alpha_m} + \alpha_m C^m_{i+1} i^{-\left(\alpha_m +1 \right)} - \frac{C^m_{i+1} \alpha_m(\alpha_m+1)}{2} i^{-\left(\alpha_m + 2 \right)}\\
                    & \stackrel{(e)}{=} \left( C^m_i - C^m_{i+1} \right)i^{-\alpha_m} + \alpha_m C^m_{i+1} i^{-\left(\alpha_m +1 \right)} - \frac{C^m_{i+1} \alpha_m(\alpha_m+1)}{2} i^{-\left(\alpha_m + 2 \right)},\label{eq:lowerbound_delta_i}
                \end{align}
                where (a) was obtained by using the fact that $i+1 = i\left( 1+\frac{1}{i} \right)$; in (b) we pulled out $i^{-\alpha_m}$ from the 2nd term. (c) can be shown as follows. Given a real-valued twice-differentiable function $f$, the Taylor expansion of $f$ at $x_0$ with Lagrange remainder gives that for some $\theta \in (0, 1)$, 
                \begin{align}
                    f(x) = f(x_0) + f'(x_0)(x-x_0) + \frac{f''\left( x_0 + \theta(x-x_0) \right)}{2}(x-x_0)^2. \label{eq:Taylor_expansion}
                \end{align}
                For $f(x)=(1+x)^a$, we have $f'(x)=a(1+x)^{a-1}$ and $f''(x)= a(a-1)(1+x)^{a-2}$. With this choice of $f$ and taking $x_0=0$ in \eqref{eq:Taylor_expansion}, we obtain that for some $\theta \in (0,1)$,
                \begin{align}
                    (1+x)^a = 1+ ax + \frac{a(a-1)}{2}(1+\theta x)^{a-2} x^2 \stackrel{(f)}{\le} 1+ax + \frac{a(a-1)}{2}x^2, \label{eq:compare_fx}
                \end{align}
                where (f) holds since for $a < -1$ we have $a(a-1) > 0$ and $a-2 \le 0$, hence for $x >0$ we have $0<(1+\theta x)^{a-2} \le 1$.
                With a substitution $x \leftarrow \frac{1}{i}$ and $a \leftarrow -\alpha_m$, one gets
                \begin{align}
                    \left( 1+\frac{1}{i} \right)^{-\alpha_m} \le  1-\frac{\alpha_m}{i} +\frac{\alpha_m(\alpha_m+1)}{2 i^2}. 
                \end{align}     
                In (d) we carried out the multiplication, in (e) we pulled out the term $i^{-\alpha_m}$.
                
                Now substituting $i = d^m =n^{\frac{\theta_m}{\alpha_m}}$ into \eqref{eq:lowerbound_delta_i}---which satisfies \eqref{eq:i-constraint} when $n^{\frac{\theta_m}{\alpha_m}} \ge L^m$, in other words, $n \ge \left( L^m \right)^{\frac{\alpha_m}{\theta_m}}$---with the notations $b^m_n:= C^m_i$ and $\tilde{b}^m_n:=C^m_{i+1}$, one gets that
                \begin{align}
                    \MoveEqLeft \delta_{d^m}^m \ge \left( b^m_n - \tilde{b}^m_n \right)n^{-\theta_m} + \alpha_m \tilde{b}^m_n  n^{-\theta_m \left(1 + \frac{1}{\alpha_m} \right)} - \frac{\tilde{b}^m_n \alpha_m(\alpha_m+1)}{2} n^{-\theta_m \left(1 + \frac{2}{\alpha_m} \right)} \\
                    & \stackrel{(a)}{\ge} - K^m n^{-\frac{\theta_m}{\alpha_m}} n^{-\theta_m} + \alpha_m \tilde{b}^m_n n^{-\theta_m \left(1 + \frac{1}{\alpha_m} \right)} - \frac{\tilde{b}^m_n \alpha_m(\alpha_m+1)}{2} n^{-\theta_m \left(1 + \frac{2}{\alpha_m} \right)} \\
                    & \stackrel{(b)}{=} \left( \alpha_m \tilde{b}^m_n - K^m \right) n^{-\theta_m \left(1 + \frac{1}{\alpha_m} \right)}  - \frac{\tilde{b}^m_n \alpha_m(\alpha_m+1)}{2} n^{-\theta_m \left(1 + \frac{2}{\alpha_m} \right)} \\
                    & \stackrel{(c)}{>} \underbrace{\left( \alpha_m \tilde{b}^m_n -\alpha_m c^m_l \right)}_{=\alpha_m\left( \tilde{b}^m_n - c^m_l \right)} n^{-\theta_m \left(1 + \frac{1}{\alpha_m} \right)}  - \frac{\tilde{b}^m_n \alpha_m(\alpha_m+1)}{2} n^{-\theta_m \left(1 + \frac{2}{\alpha_m} \right)} \\
                    & \stackrel{(d)}{\ge} \alpha_m \left( \bar{c}^m_1 - \epsilon - c_l^m \right) n^{-\theta_m \left(1 + \frac{1}{\alpha_m} \right)}  - \frac{\tilde{b}^m_n \alpha_m(\alpha_m+1)}{2} n^{-\theta_m \left(1 + \frac{2}{\alpha_m} \right)} \\
                    & \stackrel{(e)}{\ge} \alpha_m \left( \bar{c}^m_1 - \epsilon - c_l^m \right) n^{-\theta_m \left(1 + \frac{1}{\alpha_m} \right)}  - \frac{c_u^m \alpha_m(\alpha_m+1)}{2} n^{-\theta_m \left(1 + \frac{2}{\alpha_m} \right)}.  \label{eq:lowerbound_delta_d^m}
                \end{align}
                Using Assumption~\ref{eq:poly-decay:c(i)_assumption2} with \eqref{eq:i-constraint} and by substitution of $c^m(i) \leftarrow b^m_n$ and $c^m(i+1) \leftarrow \tilde{b}^m_n$, (a) was obtained by
                \begin{align}
                    \left| b^m_n - \tilde{b}^m_n \right| \le \frac{K^m}{i} \Rightarrow  
                    - \frac{K^m}{i} \le b^m_n - \tilde{b}^m_n \le \frac{K^m}{i} & \Rightarrow 
                    b^m_n - \tilde{b}^m_n \ge - \frac{K^m}{i} = - K^m n^{-\frac{\theta_m}{\alpha_m}}.
                \end{align}
                In (b) we combined the first two terms. In (c), we used the fact that by Assumption~\ref{eq:poly-decay:c(i)_assumption2} with \eqref{eq:i-constraint}, the inequality $0\le K^m < \alpha_m c^m_l$ holds, which implies that $-K^m > -\alpha_m c^m_l$. 
                In (d), by \eqref{eq:i-constraint} and the definition of $\tilde{b}^m_n$, one has that $\left| \tilde{b}^m_n - \bar{c}^m_1 \right| \le \epsilon$, which implies that $-\epsilon \le \tilde{b}^m_n - \bar{c}^m_1$ i.e.\ $\tilde{b}^m_n \ge \bar{c}^m_1 - \epsilon$.
                Recall that $a^m = \bar{c}^m_1 - c_l^m >0$. Since $\epsilon < a^m$ (by assumption), we have 
                \begin{align}
                    \bar{c}^m_1 - \epsilon - c_l^m > \bar{c}^m_1 - a^m - c_l^m = 0.
                \end{align}
                (e) follows from the fact that $\tilde{b}^m_n \in [ c_l^m, c_u^m ]$. Recall that at the beginning we fixed $i$ according to \eqref{eq:i-constraint}; the reasoning holds for any such $i$, in other words for any $n \ge \left( L^m \right)^{\frac{\alpha_m}{\theta_m}}$. Thus, \eqref{eq:lowerbound_delta_d^m} holds for all $n \ge \left( L^m \right)^{\frac{\alpha_m}{\theta_m}}$.      
                Moreover, since $n^{-\theta_m \left(1 + \frac{2}{\alpha_m} \right)} \le n^{-\theta_m \left(1 + \frac{1}{\alpha_m} \right)}$ for sufficiently large $n$, the r.h.s.\ of \eqref{eq:lowerbound_delta_d^m} is dominated by the term $n^{-\theta_m \left(1 + \frac{1}{\alpha_m} \right)}$. Therefore, we got the lower bound
                \begin{align}
                \delta_{d^m}^m = \Omega \left( n^{-\theta_m \left(1 + \frac{1}{\alpha_m} \right)} \right). \label{eq:lower-bound-delta}
                \end{align} 
                Having obtained the lower bound in \eqref{eq:lower-bound-delta} on $\delta^m_{d^m}$, we continue with the bounding of the individual terms of $B_m$.
                \begin{itemize}
                    \item Term $B_{m,1}$:  
                    \begin{align}
                    B_{m,1} & \stackrel{(a)}{=} 312 \sqrt{2} c_1^m(n) d^m \left( \delta^m_{d^m} \right)^{-\frac{1}{2}}  \sqrt{\frac{\log \left ( \frac{864 n}{\eta} \right ) }{n}} \\
                    &\stackrel{(b)}{=} 
                    \O\left(\log(\log(n)) n^{\frac{\theta_m}{\alpha_m}} n^{\frac{\theta_m}{2} \left(1+\frac{1}{\alpha_m}\right)} \sqrt{\frac{\log (n)} {n}} \right)\\
                    &=\O \left(\log(\log(n)) n^{\frac{\theta_m}{2} \left(1+\frac{1}{\alpha_m}\right)} \sqrt{\frac{\log (n)} {n^{1-\frac{2\theta_m}{\alpha_m}}}} \right)
                    = \O\left(\log(\log(n))  \sqrt{\frac{\log (n)} {n^{1-\frac{3\theta_m}{\alpha_m}- \theta_m }}} \right).
                    \end{align}
                    In (a) we used the definition of $B_{m,1}$; (b) follows from the fact that $c_1^m(n)=\O(\log(\log(n)))$, $d^m = n^{\frac{\theta_m}{\alpha_m}}$ [by  \eqref{eq:poly-decay:d^m-choice}] and that $\left( \delta^m_{d^m} \right)^{-\frac{1}{2}}=\O \left(n^{\frac{\theta_m}{2} \left(1+\frac{1}{\alpha_m}\right)} \right)$ [by  \eqref{eq:lower-bound-delta}].
                    This means that $B_{m,1} \xrightarrow{n \to \infty} 0$ iff.\ $0 < 1-\frac{3 \theta_m}{\alpha_m}- \theta_m $, in other words $\theta_m < \frac{1}{\frac{3}{\alpha_m}+1}$. Using that $\frac{1}{\frac{3}{\alpha_m}+1} < 1$ as $0<\alpha_m$, this implies that $\theta_m < 1$ and hence in $A_m$ the lower branch ($\theta_m < 1$) will be active.

                    \item Term $B_{m,2}$: 
                    \begin{align}
                    B_{m,2} & \stackrel{(a)}{=} 896 \frac{\left(d^m \right)^{\frac{1}{2}} \left(c_1^m(n) \right)^2}{\sqrt{c_0^m}} \left( \delta^m_{d^m} \right)^{-2} \frac{\log \left( \frac{432 n}{\eta} \right)}{n} \\
                    &\stackrel{(b)}{=}\O \left( n^{\frac{\theta_m}{2 \alpha_m}}[\log(\log(n))]^2n^{2\theta_m \left(1+\frac{1}{\alpha_m}\right)} \frac{\log (n)}{n} \right)\\
                    &= \O\left( [\log(\log(n))]^2 \frac{\log (n)} {n^{1-\frac{5\theta_m}{2\alpha_m}- 2\theta_m}} \right).
                    \end{align}
                    In (a) we used the definition of $B_{m,2}$; (b) follows from  $d^m = n^{\frac{\theta_m}{\alpha_m}}$ [by  \eqref{eq:poly-decay:d^m-choice}], $c_1^m(n)=\O(\log(\log(n)))$, and that $\left( \delta_{d^m}^m \right)^{-2} = \O \left( n^{2\theta_m \left(1 + \frac{1}{\alpha_m} \right)} \right)$ [by \eqref{eq:lower-bound-delta}].
                    This means that $B_{m,2} \xrightarrow{n \to \infty} 0$ iff.\ $0 < 1-\frac{5 \theta_m}{2\alpha_m}- 2\theta_m$, in other words $\theta_m < \frac{1}{\frac{5}{2\alpha_m}+2}$. Using that $\frac{1}{\frac{5}{2\alpha_m}+2} < \frac{1}{2}$ as $0<\alpha_m$, this implies that $\theta_m < \frac{1}{2}$ and hence in $A_m$ the lower branch ($\theta_m < 1$) will be active.

                    \item Term $B_{m,3}$:  
                    \begin{align}
                    B_{m,3} &\stackrel{(a)}{=} 1024 \frac{\left(d^m \right)^{\frac{1}{2}}\left(c_1^m(n) \right)^3}{\sqrt{c_0^m}} \left( \delta^m_{d^m} \right)^{-2} \frac{\log^{\frac{3}{2}} \left( \frac{432 n}{\eta} \right)}{n^{\frac{3}{2}}} \\
                    &\stackrel{(b)}{=}\O \left(n^{\frac{\theta_m}{2 \alpha_m}} [\log(\log(n))]^3 n^{2\theta_m \left(1+\frac{1}{\alpha_m}\right) }\frac{ \log^{\frac{3}{2}} (n)}{n^{\frac{3}{2}}} \right)\\ 
                    & = \O\left([\log(\log(n))]^3 \frac{\log^{\frac{3}{2}} (n)} {n^{\frac{3}{2}-\frac{5 \theta_m}{2\alpha_m}- 2\theta_m}} \right).
                    \end{align}
                    (a) holds by the definition of $B_{m,3}$; (b) follows from $d^m = n^{\frac{\theta_m}{\alpha_m}}$ [by  \eqref{eq:poly-decay:d^m-choice}], $c_1^m(n)=\O(\log(\log(n)))$, and that $\left( \delta_{d^m}^m \right)^{-2} = \O \left( n^{2\theta_m \left(1 + \frac{1}{\alpha_m} \right)} \right)$ [by \eqref{eq:lower-bound-delta}].
                    This means that $B_{m,3} \xrightarrow{n \to \infty} 0$ iff.\ $0 < \frac{3}{2}-\frac{5 \theta_m}{2\alpha_m}- 2\theta_m$, in other words $\theta_m < \frac{\frac{3}{2}}{\frac{5}{2\alpha_m}+2}$. Using that $\frac{1}{\frac{5}{2\alpha_m}+2} < \frac{1}{2}$ as $0<\alpha_m$, this implies that $\theta_m < \frac{3}{2} \frac{1}{2} = \frac{3}{4}$ and hence in $A_m$ the lower branch ($\theta_m < 1$) will be active.

                    \item Term $B_{m,4}$:
                    \begin{align}
                    B_{m,4} & \stackrel{(a)}{=} 64 \frac{\left(c_1^m(n) \right)^2}{\sqrt{c_0^m}} \left( \delta^m_{d^m} \right)^{-1} \frac{\left(d^m \right)^{\frac{1}{2}} \log \left( \frac{432n}{\eta} \right)}{n}
                    \\ 
                    &\stackrel{(b)}{=}\O \left([\log(\log(n))]^2 n^{\theta_m \left(1+\frac{1}{\alpha_m}\right) }\frac{n^{\frac{\theta_m}{2 \alpha_m}} \log (n)}{n} \right) = \O\left([\log(\log(n))]^2 \frac{\log(n)} {n^{1-\frac{3\theta_m}{2\alpha_m}- \theta_m}} \right).
                    \end{align}
                    (a) holds by the definition of $B_{m,4}$; (b) follows from  $c_1^m(n)=\O(\log(\log(n)))$, $\left( \delta_{d^m}^m \right)^{-1} = \O \left( n^{\theta_m \left(1 + \frac{1}{\alpha_m} \right)} \right)$ [by \eqref{eq:lower-bound-delta}], and that $d^m = n^{\frac{\theta_m}{\alpha_m}}$ [by  \eqref{eq:poly-decay:d^m-choice}].
                    This means that $B_{m,4} \xrightarrow{n \to \infty} 0$ iff.\ $0 < 1-\frac{3 \theta_m}{2\alpha_m}- \theta_m$, in other words $\theta_m < \frac{1}{\frac{3}{2\alpha_m}+1}$. Using that $\frac{1}{\frac{3}{2\alpha_m}+1} < 1$ as $0<\alpha_m$, this implies that $\theta_m < 1$ and hence in $A_m$ the lower branch ($\theta_m < 1$) will be active.

                    \item Term $B_{m,5}$: 
                    \begin{align}
                    B_{m,5} & \stackrel{(a)}{=} 4 \frac{(d^m)^{\frac{1}{2}} c_1^m(n)}{\sqrt{c_0^m}} \sqrt{\frac{\log \left( \frac{144n}{\eta}  \right)}{n}} \stackrel{(b)}{=} \O\left(\log(\log(n)) \sqrt{\frac{\log (n)}{n^{1-\frac{\theta_m}{\alpha_m}}} } \right).
                    \end{align}
                    In (a) we used the definition of $B_{m,5}$; (b) follows from $d^m = n^{\frac{\theta_m}{\alpha_m}}$ [by  \eqref{eq:poly-decay:d^m-choice}] and $c_1^m(n)=\O(\log(\log(n)))$.
                    This means that $B_{m,5} \xrightarrow{n \to \infty} 0$ iff.\ $0 < 1-\frac{\theta_m}{\alpha_m}$, which holds when $\theta_m < \alpha_m$.
                \end{itemize}
            Combining these results, we specifically got that 
            \begin{align}
            \theta_m<\min\left\{1,\frac{1}{2},\frac{3}{4},1,\alpha_m\right\}=\frac{1}{2} \label{eq:theta_m<1/2}
            \end{align}
            needs to hold (by $\alpha_m>1$) and therefore we conclude that the lower branch of $A_m$ $(\theta_m <1)$ will be activated. 
            
            Before combining $A_m$ and $B_m$, we first compare the terms of $B_m$.
            \begin{itemize} 
                \item Comparing $B_{m,1}$ and $B_{m,2}$: 
                \begin{itemize}
                    \item $B_{m,1}$ converges faster than $B_{m,2}$ (with non-equal exponents in $n$): This holds iff.              \begin{align}
                    &\frac{1}{2} - \frac{3\theta_m}{2 \alpha_m} - \frac{\theta_m}{2} > 1-\frac{5 \theta_m}{2\alpha_m} - 2\theta_m  \Leftrightarrow \frac{3\theta_m}{2} + \frac{\theta_m}{\alpha_m} - \frac{1}{2} > 0\\
                    &\stackrel{2\alpha_m}{\Leftrightarrow} 3 \theta_m \alpha_m + 2 \theta_m - \alpha_m >0
                     \Leftrightarrow \theta_m (3\alpha_m +2)>\alpha_m
                    \Leftrightarrow \theta_m > \frac{\alpha_m}{3 \alpha_m + 2 }. \label{con:Bm1_faster-v0}
                \end{align}
                \item  $B_{m,1}$ converges slower than $B_{m,2}$ (with non-equal exponents in $n$): This holds iff.                \begin{align}\label{con:Bm2_faster-v0}
                    \theta_m < \frac{\alpha_m}{3 \alpha_m + 2 }.
                \end{align}
                \item Equal exponents: With $\frac{1}{2} - \frac{3 \theta_m}{2\alpha_m} - \frac{\theta_m}{2} = 1-\frac{5\theta_m}{2\alpha_m} - 2\theta_m \Leftrightarrow \theta_m = \frac{\alpha_m}{3 \alpha_m + 2 }$, the question boils down to the comparison of $a_n = \log(\log(n)) \sqrt{\log(n)}$ and 
                $b_n = [\log(\log(n))]^2 \log(n) =a_n^2$. Noticing that $\frac{a_n}{b_n} =\frac{a_n}{a_n^2} = \frac{1}{a_n} \to 0$ as $n \to \infty$, in other words, $a_n =o(b_n)$. This means that in this case $B_{m,1}$ converges faster than $B_{m,2}$.
                
                \end{itemize} 
                To sum up, we got that $B_{m,1}$ converges faster than $B_{m,2}$ iff.  
                \begin{align}
                \theta_m & \ge  \frac{\alpha_m}{3 \alpha_m + 2 };
                \label{con:Bm1_faster}
                \end{align}
                otherwise, i.e.\ for \begin{align}\label{con:Bm2_faster}
                    \theta_m < \frac{\alpha_m}{3 \alpha_m + 2 },
                \end{align}
                $B_{m,2}$ converges faster.
                
                \item Comparing $B_{m,2}$ and $B_{m,3}$: one has that $\frac{3}{2}-\frac{5\theta_m}{2\alpha_m}- 2\theta_m > 1-\frac{5 \theta_m}{2\alpha_m}- 2\theta_m \Leftrightarrow \frac{3}{2}>1$ holds, therefore $B_{m,3}$ converges faster than $B_{m,2}$.
                
                \item Comparing $B_{m,2}$ and $B_{m,4}$: one has that 
                \begin{align}1-\frac{3 \theta_m}{2\alpha_m}- \theta_m > 1-\frac{5 \theta_m}{2\alpha_m}- 2\theta_m \Leftrightarrow \frac{2\theta_m}{2\alpha_m} + \theta_m >0 \Leftrightarrow \theta_m\left(\frac{1}{\alpha_m}+1\right) \Leftrightarrow \theta_m>0
                \end{align}
                holds since $\alpha_m >0$ and $\theta_m > 0$, therefore $B_{m,4}$ converges faster than $B_{m,2}$.

                \item Comparing $B_{m,1}$ and $B_{m,5}$: one has that $1-\frac{\theta_m}{\alpha_m} > 1-\frac{3 \theta_m}{\alpha_m}- \theta_m \Leftrightarrow \frac{2 \theta_m}{\alpha_m} + \theta_m =\theta_m\left(\frac{2}{\alpha_m}+1\right) > 0$ holds since $\theta_m>0$ and $\alpha_m>0$, therefore $B_{m,5}$ converges faster than $B_{m,1}$.
            \end{itemize}
                
            \item Order of the term $U_{3,m}$: By collecting the terms in $U_{3,m}$, one arrives at  
             \begin{align}
            U_{3,m}=A_m + B_{m,1} + B_{m,2} + B_{m,3} + B_{m,4} + B_{m,5}. 
            \end{align}
            Since $B_{m,3}$, $B_{m,4}$ and $B_{m,5}$ converge faster than $B_{m,1}$ or $B_{m,2}$, the final convergence rate is determined $A_m + B_{m,1}$ or $A_m + B_{m,2}$. We therefore discuss these two cases separately in the following.
            \begin{itemize}
                \item \tb{Case of \eqref{con:Bm2_faster} holding}: In this case the order is that of $A_m + B_{m,1}$, and we have
                \begin{align}
                    A_m + B_{m,1} = \O\Bigg( \underbrace{\frac{n^{\frac{1}{2\alpha_m}-\theta_m}}{\log ^{\frac{1}{2\alpha_m}} (n)}}_{=:a_{m}(n)} \Bigg)  + \O\Bigg(\underbrace{\log(\log(n)) \frac{\log^{\frac{1}{2}} (n)} {n^{\frac{1}{2}-\frac{3\theta_m}{2 \alpha_m}- \frac{\theta_m}{2}}}}_{=:b_{m,1}(n)}\Bigg)
                \end{align}
                under the conditions that $\theta_m < \frac{\alpha_m}{3 \alpha_m + 2}$, $1 < \alpha_m$, $0 < \theta_m \le \alpha_m$, $\frac{1}{2\alpha_m} < \theta_m$ and  $\theta_m < \frac{1}{\frac{3}{\alpha_m}+1}$, which can be written as 
                \begin{align}
                1 &< \alpha_m,& \underbrace{\max\left\{0,\frac{1}{2\alpha_m}\right\}}_{=\frac{1}{2\alpha_m}\Leftarrow \alpha_m>0} < \theta_m &< \min\left\{\frac{\alpha_m}{3 \alpha_m +2} , \frac{1}{\frac{3}{\alpha_m}+1} \right\},& \theta_m& \le \alpha_m.
                \end{align}
                Noticing that (i)
                \begin{align}
                \frac{\alpha_m}{3 \alpha_m +2} \le \frac{1}{\frac{3}{\alpha_m}+1}   \Leftrightarrow  3 +  \alpha_m \le 3\alpha_m +2 \Leftrightarrow 1 \le 2 \alpha_m \Leftrightarrow \frac{1}{2} \le \alpha_m 
                \end{align}
                holds (by $\alpha_m>1$) and (ii)
                 $\frac{1}{\frac{3}{\alpha_m}+1} \le \alpha_m \Leftrightarrow 1\le 3 + \alpha_m \Leftrightarrow -2 \le \alpha_m$ holds (by $\alpha_m>1$), our constraints are 
                \begin{align}
                1 &< \alpha_m,& \frac{1}{2\alpha_m} < \theta_m < \frac{\alpha_m}{3\alpha_m+2}.
                \end{align}
                To be able to choose $\theta_m$ from a non-empty set, one needs that 
                \begin{align}
                \frac{1}{2\alpha_m} < \frac{\alpha_m}{3\alpha_m+2} \Leftrightarrow 3\alpha_m +2 < 2 \alpha_m^2 \Leftrightarrow 0 < 2\alpha_m^2 - 3 \alpha_m - 2, \label{eq:A_m+B_{m,1/2}}
                \end{align}
                by using that $\alpha_m>0$.
                Solving for $(\alpha_m)_{1,2}= \frac{3 \pm \sqrt{9+16}}{4} = \frac{3\pm 5}{4}$. Using that $\alpha_m \mapsto 2\alpha_m^2 - 3 \alpha_m - 2$ is convex, this means that it is needed that either $\alpha_m < \frac{3-5}{4}=-\frac{1}{2}$ or  $\frac{3+5}{4}=2 < \alpha_m$. Given that $\alpha_m>1$ needs to hold, our final constraints are
                \begin{align}
                2 &< \alpha_m,& \frac{1}{2\alpha_m} < \theta_m < \frac{\alpha_m}{3\alpha_m+2}. \label{eq:poly:A_m+B_{m,1}:final-constraints}
                \end{align}
                Considering the order of
                \begin{itemize}
                    \item $a_{m}(n)$: One has
                    \begin{align}
                            \hspace{-0.5cm}a_{m}(n)  & = \frac{1}{n^{\theta_m-\frac{1}{2\alpha_m}} \log ^{\frac{1}{2\alpha_m}} (n)} = \Omega\left( \frac{1}{n^{\frac{1}{2}}}  \right) \label{eq:am(n):order} 
                    \end{align}
                    using that $\theta_m <\frac{1}{2}$ by \eqref{eq:theta_m<1/2} and $0 < \alpha_m$, hence $\theta_m - \frac{1}{2\alpha_m} < \frac{1}{2}$. 
                    \item $b_{m,1}(n)$: One has
                    \begin{align}
                            b_{m,1}(n)  & = \log(\log(n)) \frac{\log^{\frac{1}{2}} (n)} {n^{\frac{1}{2}-\frac{3\theta_m}{2\alpha_m}- \frac{\theta_m}{2}}} = \Omega\left(\frac{1}{n^{\frac{1}{2}-\frac{3\theta_m}{2\alpha_m}- \frac{\theta_m}{2}}}\right) = \Omega\left(\frac{1}{n^{\frac{1}{2}}}\right),\label{eq:poly:bm1(n):order}
                    \end{align}    
                    using that $0<\alpha_m$ and $0 <\theta_m$, hence $\frac{1}{2}-\frac{3\theta_m}{2\alpha_m}- \frac{\theta_m}{2}<\frac{1}{2}$. 
                \end{itemize}
                 With $\theta_m$ increasing, $a_m(n)$ converges to zero faster and $b_{m,1}(n)$ goes to zero slower as $n\to \infty$. Hence, we match the orders of the (polynomial) terms in $a_m(n)$ and $b_{m,1}(n)$ to get 
                 \begin{align}
                 \theta_m - \frac{1}{2 \alpha_m} = \frac{1}{2}-\frac{3\theta_m}{2\alpha_m}- \frac{\theta_m}{2} & \Leftrightarrow \frac{3}{2} \theta_m = \frac{1}{2}-\frac{3\theta_m}{2\alpha_m} + \frac{1}{2 \alpha_m} \\
                 &\stackrel{2\alpha_m}{\Leftrightarrow} 3 \alpha_m \theta_m = \alpha_m - 3\theta_m +1 \\
                 &\hspace{0.2cm} \Leftrightarrow 3\theta_m (\alpha_m+1)= \alpha_m+1 \Leftrightarrow \theta_m=\frac{1}{3}, 
                 \end{align}
                 by using that $\alpha_m+1 \ne 0$ (by $\alpha_m >1$).
                 With this $\theta_m$ choice, in \eqref{eq:poly:A_m+B_{m,1}:final-constraints}
                 \begin{itemize}
                     \item the constraint $\frac{1}{2\alpha_m} < \theta_m$ holds:
                     $\frac{1}{2\alpha_m} <  \frac{1}{3} \Leftrightarrow \alpha_m > \frac{3}{2}$, 
                     as guaranteed in \eqref{eq:poly:A_m+B_{m,1}:final-constraints} by $2<\alpha_m$, 
               \item  the constraint $\theta_m < \frac{\alpha_m}{3 \alpha_m + 2}$ doesn't holds since
               \begin{align}
                   \frac{1}{3} < \frac{\alpha_m}{3 \alpha_m + 2}
                   \Leftrightarrow 3 \alpha_m+2 < 3 \alpha_m
                   \Leftrightarrow 2 < 0.
               \end{align} 
               Thus \eqref{con:Bm2_faster} doesn't hold, hence the $A_m + B_{m,1}$ thread can be discarded.
               \end{itemize}   

               \item Case of \eqref{con:Bm1_faster} holding: In this case the order of $A_m+B_m$ is that of $A_m + B_{m,2}$, for which we have
                \begin{align}
                    A_m + B_{m,2} = \O\Bigg( \underbrace{\frac{n^{\frac{1}{2\alpha_m}-\theta_m}}{\log^{\frac{1}{2\alpha_m}} (n)}}_{=a_m(n)} \Bigg)  + \O\Bigg(\underbrace{[\log(\log(n))]^2\frac{\log (n)} {n^{1-\frac{5 \theta_m}{2\alpha_m}- 2\theta_m}} }_{=:b_{m,2}(n)}\Bigg)
                \end{align}
                under the conditions that $\theta_m \ge  \frac{\alpha_m}{3 \alpha_m + 2}$, $1 < \alpha_m$, $0 < \theta_m \le \alpha_m$, $\frac{1}{2\alpha_m} < \theta_m$ and  $\theta_m < \frac{1}{\frac{5}{2\alpha_m}+2}$, which
                can be written as
                \begin{align}
                1 &< \alpha_m & \underbrace{\max\left\{0,\frac{1}{2\alpha_m}\right\}}_{=\frac{1}{2\alpha_m} \Leftarrow \alpha_m>0} &< \theta_m < \frac{1}{\frac{5}{2\alpha_m}+2},& \frac{\alpha_m}{3 \alpha_m + 2} &\le \theta_m \le \alpha_m.
                \end{align}
                Noticing that
                $\frac{1}{\frac{5}{2\alpha_m}+2} \le \alpha_m \Leftrightarrow 1 \le \frac{5}{2}+2\alpha_m \Leftrightarrow -\frac{3}{2}\le 2 \alpha_m \Leftrightarrow -\frac{3}{4} \le \alpha_m$ holds (by $\alpha_m>1$) our constraints simplify to
                \begin{align}
                 1 &< \alpha_m, & \frac{1}{2\alpha_m} < \theta_m &< \frac{1}{\frac{5}{2\alpha_m}+2}, & \frac{\alpha_m}{3 \alpha_m + 2} &\le \theta_m.
                \end{align}
                
                By using that $\alpha_m>0$,
                \begin{align}
                    \frac{\alpha_m}{3 \alpha_m + 2} > \frac{1}{2 \alpha_m} \Leftrightarrow 2 \alpha_m^2 - 3 \alpha_m - 2 >0
                \end{align}
                which holds when $2< \alpha_m$ [by \eqref{eq:A_m+B_{m,1/2}}]. Therefore, we consider two sub-cases.               \begin{enumerate}
                    \item Case of $\alpha_m>2$: In this case, we have $\frac{\alpha_m}{3 \alpha_m + 2} > \frac{1}{2 \alpha_m}$, and our constraints are
                    \begin{align}
                        2 &< \alpha_m,& \frac{\alpha_m}{3 \alpha_m + 2} \le \theta_m < \frac{1}{\frac{5}{2\alpha_m}+2}. \label{eq:poly:A_m+B_{m,2}:final-constraints1}
                    \end{align}
                    To be able to choose $\theta_m$ from a non-empty set, one needs that  
                    \begin{align}
                      \frac{\alpha_m}{3 \alpha_m + 2} &< \frac{1}{\frac{5}{2\alpha_m}+2}
                     \Leftrightarrow \frac{5}{2} + 2 \alpha_m < 3 \alpha_m + 2 \Leftrightarrow \frac{1}{2} < \alpha_m
                     \end{align}
                     by using that $\alpha_m>0$; this is guaranteed by $\alpha_m > 1$.
                    \item Case of $1<\alpha_m \le 2$: In this case, we have $\frac{\alpha_m}{3 \alpha_m + 2} \le \frac{1}{2 \alpha_m}$, therefore the constraints are
                    \begin{align}
                        1 &< \alpha_m \le 2,& \frac{1}{2 \alpha_m} < \theta_m < \frac{1}{\frac{5}{2\alpha_m}+2}. 
                    \end{align}
                    To be able to choose $\theta_m$ from a non-empty set, one needs that  
                    \begin{align}
                      \frac{5}{2\alpha_m}+2  &< 2\alpha_m
                     \Leftrightarrow 0 < 4\alpha_m^2-4\alpha_m-5 \label{eq:alpha-constraint2}
                     \end{align}
                     by using that $\alpha_m>0$.
                     Solving $4\alpha_m^2-4\alpha_m-5=0$, one gets that $(\alpha_m)_{1,2} = \frac{4 \pm \sqrt{16+80}}{8}=\frac{1\pm \sqrt{6}}{2}$. Hence by using the convexity of $\alpha_m \mapsto 4\alpha_m^2-4\alpha_m-5$, it is needed that either $\alpha_m < \frac{1-\sqrt{6}}{2}$ or $\frac{1+\sqrt{6}}{2} (\approx 1.72) < \alpha_m$. Combined with the condition $1 < \alpha_m \le 2$, we got that the final constraints are 
                     \begin{align}
                     \frac{1+\sqrt{6}}{2} &< \alpha_m \le 2, &\frac{1}{2\alpha_m} &< \theta_m < \frac{1}{\frac{5}{2\alpha_m}+2}. \label{eq:poly:A_m+B_{m,2}:final-constraints2}
                    \end{align}
                \end{enumerate} 
                With $\theta_m$ increasing, $a_m(n)$ converges to zero faster and $b_{m,2}(n)$ goes to zero slower as $n\to \infty$.
                Hence, we match the orders of the (polynomial) terms in $a_m(n)$ and $b_{m,2}(n)$ to get 
                 \begin{align}
                 \theta_m - \frac{1}{2 \alpha_m} &= 1-\frac{5 \theta_m}{2\alpha_m}- 2\theta_m  \Leftrightarrow 3 \theta_m = 1 -\frac{5 \theta_m}{2 \alpha_m} + \frac{1}{2 \alpha_m} \\
                 &\stackrel{2\alpha_m}{\Leftrightarrow} 6 \alpha_m \theta_m = 2 \alpha_m - 5 \theta_m +1 \Leftrightarrow \theta_m = \frac{2\alpha_m +1}{6 \alpha_m + 5}. \label{eq:poly-decay:theta^m-choice}
                 \end{align}
                With this choice of $\theta_m$, assuming that
                \begin{enumerate}
                    \item $2 <\alpha_m$: In \eqref{eq:poly:A_m+B_{m,2}:final-constraints1}, the constraint
                \begin{itemize}
                  \item $\theta_m < \frac{1}{\frac{5}{2\alpha_m}+2}$ holds: using that $\alpha_m>0$,
                  \begin{align}
                  \frac{2\alpha_m +1}{6\alpha_m+5} &< \frac{1}{\frac{5}{2\alpha_m}+2} \Leftrightarrow (2\alpha_m+1)\left(\frac{5}{2\alpha_m}+2\right) < 6\alpha_m+5 \Leftrightarrow \\
                  5+4\alpha_m + \frac{5}{2\alpha_m}+2 &  < 6\alpha_m+5 \Leftrightarrow \frac{5}{2\alpha_m}+2 < 2 \alpha_m \stackrel{2\alpha_m} {\Leftrightarrow} 5 + 4\alpha_m < 4 \alpha_m^2 \Leftrightarrow \\
                   0 & < 4\alpha_m^2-4\alpha_m - 5
                  \end{align}
                        as ensured by \eqref{eq:alpha-constraint2} (since $\alpha_m > 2$),
                  \item $\frac{\alpha_m}{3 \alpha_m +2} \le \theta_m$ holds: using that $\alpha_m>0$, 
                           \begin{align}
                               \frac{2\alpha_m +1}{6 \alpha_m + 5} \ge \frac{\alpha_m}{3 \alpha_m+2} &\Leftrightarrow (2\alpha_m+1)(3\alpha_m+2)\ge (6\alpha_m+5) \alpha_m\\   & \Leftrightarrow 6\alpha_m^2 + 4 \alpha_m + 3 \alpha_m +2 \ge 6 \alpha_m^2 + 5 \alpha_m \\
                               &\Leftrightarrow  2 \alpha_m + 2 \ge 0 \Leftrightarrow \alpha_m\ge -1,
                           \end{align}
                           which holds by $\alpha_m>1$.
               \end{itemize} 
                    \item $\frac{1+\sqrt{6}}{2} < \alpha_m \le 2$: In \eqref{eq:poly:A_m+B_{m,2}:final-constraints2} the constraint
                    \begin{itemize}
                     \item $\frac{1}{2\alpha_m} < \theta_m$ holds: using that $\alpha_m>0$, $\frac{1}{2\alpha_m} <  \frac{2\alpha_m +1}{6 \alpha_m+5} \Leftrightarrow  6 \alpha_m+5 < 4 \alpha_m^2+2\alpha_m \Leftrightarrow 0 < 4\alpha_m^2-4\alpha_m - 5$, as guaranteed in \eqref{eq:alpha-constraint2};
                  \item $\theta_m < \frac{1}{\frac{5}{2\alpha_m}+2}$ holds: using that $\alpha_m>0$, 
                        $\frac{2\alpha_m +1}{6\alpha_m+5} < \frac{1}{\frac{5}{2\alpha_m}+2} \Leftrightarrow  (2\alpha_m+1)\left(\frac{5}{2\alpha_m}+2\right)  < 6\alpha_m+5 \Leftrightarrow  5 + 4\alpha_m + \frac{5}{2\alpha_m}+2 < 6 \alpha_m + 5
                        \Leftrightarrow \frac{5}{2\alpha_m} +2 <2\alpha_m \Leftrightarrow  0  < 4\alpha_m^2-4\alpha_m - 5$,
                        as ensured again by \eqref{eq:alpha-constraint2}.
                \end{itemize} 
                \end{enumerate}
               Therefore, we got that with $\frac{1+\sqrt{6}}{2} <\alpha_m$, the resulting embedding dimension is 
               \begin{align}
               d^m &= n^{\frac{2\alpha_m+1}{(6\alpha_m+5)\alpha_m}} 
               \end{align}
               by \eqref{eq:poly-decay:d^m-choice}. By noticing that $a_m(n) = o(b_{m,2}(n))$ by $\alpha_m>0$ [since $\frac{a_m(n)}{b_{m,2}(n)} = \frac{1}{\log^{\frac{1}{2\alpha_m}}(n) [\log(\log(n))]^2 \log(n)}\xrightarrow{n \to \infty} 0$], 
            \begin{align}
            A_m + B_{m,2} &= \O(b_{m,2}(n)),\\
            b_{m,2}(n) &= \Theta \left( \frac{[\log(\log(n))]^2 \log(n)}{n^{\frac{2\alpha_m +1}{6 \alpha_m+5} - \frac{1}{2\alpha_m}}}\right). \label{eq:bm2-order}
            \end{align}
            We note that  $B_m = \Theta(B_{m,2})$, $B_{m,2} = \O(b_{m,2}(n))$;
            this specifically shows by \eqref{eq:bm2-order} the $B_m=\tilde{\O}(n^{-a})$ bound (with $a>0$) claimed in \eqref{eq:B_m:promised-rate}.
            \end{itemize}  
            
            \end{itemize}        
        \item Final rate: By \eqref{eq:U1-U2m-U3m}, one can see that the order of the error is that of $U_1 + \sum_{m\in [M]}U_{2,m} + \sum_{m\in [M]}U_{3,m}$, which is the same order as that of $\sum_{m\in [M]}U_{3,m}$ since (i) $U_1 = \O\left(n^{-1/2}\right)$, $U_{2,m} = \O\left(n^{-1/2}\right)$, (ii) $U_{3,m}=\Theta\left(A_m + B_{m,2}\right)$, $A_m + B_{m,2} =\O\left(b_{m,2}(n)\right)$ and (iii) $b_{m,2}(n)=\Omega\left(n^{-1/2}\right)$. Indeed, the last property follows from
        \begin{align}
        \frac{2\alpha_m +1}{6\alpha_m+5} - \frac{1}{2\alpha_m} < \frac{1}{2} &\stackrel{(a)}{\Leftrightarrow} (2\alpha_m)(2\alpha_m+1) - (6 \alpha_m+5) < (6 \alpha_m+5) \alpha_m \\
        & \Leftrightarrow 4 \alpha_m^2 + 2 \alpha_m - 6 \alpha_m -5 < 6 \alpha_m^2 + 5 \alpha_m \\
        & \Leftrightarrow 0 < 2 \alpha_m^2 + 9 \alpha_m + 5 \Leftarrow \alpha_m >0, 
        \end{align}
        where (a) follows from multiplying with $(6\alpha_m+5)(2\alpha_m)>0$ and using the positivity of $\alpha_m$ implying that of $6\alpha_m+5$ and $2 \alpha_m$, hence that of their product.
        According to our obtained result, we focus on $\alpha_m > \frac{1+\sqrt{6}}{2}$. By introducing the notation 
        \begin{align}
        f: \alpha \in \left(\frac{1+ \sqrt{6}}{2},\infty\right) \mapsto \frac{2\alpha +1}{6 \alpha+5} - \frac{1}{2\alpha} \in \R,
        \label{eq:f:rate}
        \end{align}
        combining the results obtained for $(U_{3,m})_{m=1}^M$, 
        we got, assuming that $\alpha_m > \frac{1+\sqrt{6}}{2}$ for all $m \in [M]$, that the error of the proposed GHSIC estimator is 
        \begin{align}
        &\O\left(\frac{[\log(\log(n))]^2\log(n)}{n^\beta}\right) & \beta &= \beta(\alpha_1,\ldots,\alpha_M) = \min\left\{ f(\alpha_1),\ldots,f(\alpha_M)\right\}.
        \end{align}
        One can show (Lemma~\ref{lemma:f-properties}) that (i) $f$ is positive, hence $\beta>0$,  (ii) $\lim_{\alpha\to \infty}f(\alpha)=\frac{1}{3}$, (iii) $\lim_{\alpha \to \frac{1+ \sqrt{6}}{2}} f(\alpha)=0$, and (iv) $f$ is strictly increasing; see Fig.~\ref{fig2_alpha} for a visual illustration. This means that (i) as the problem becomes easier (i.e.\ $\alpha_m$-s increase) the rate of the GHSIC estimator improves, (iii) this rate can be arbitrarily close to $\O\left(\frac{[\log(\log(n))]^2\log(n)}{n^{\frac{1}{3}}}\right)$, and (iii) in the limit of the other extreme as one considers the hardest possible problem ($\alpha \to \frac{1+ \sqrt{6}}{2}$) the rate can become arbitrary slow. 
        \item \textbf{$d^m=\Theta \left( \log^{\theta_m} (n) \right)$ is not enough for consistency}. One might raise the question of whether the error in \eqref{eq:U1-U2m-U3m} can be driven to zero with an embedding dimension $\left(d^m\right)_{m=1}^M$-s which is only poly-logarithmic function of the node size $n$. The answer is negative (given the current error bound). Indeed, assume that $d^m=\Theta \left( \log^{\theta_m} (n) \right)$ for some $\theta_m>0$. In this case, $\lambda^m_{d^m+1} = \Theta((d^m+1)^{-\alpha_m}) = \log^{-\theta_m \alpha_m}(n)$, and 
        \begin{align}
        A_m & = \Theta \left( \frac{n^{\frac{1}{2\alpha_m}}}{\log^{\frac{1}{2\alpha_m}} (n)}  \left(\log^{-\theta_m \alpha_m}(n) + \frac{\log (n)}{n} \right) \right) = \Theta\left(\frac{n^{\frac{1}{2\alpha_m}}}{\log^{\frac{1}{2\alpha_m}+\theta_m \alpha_m}(n)} \right)\xrightarrow{n\to \infty} \infty;
        \end{align}
        specifically $A_m$ does \emph{not} go to zero as $n \to \infty$. 
        \end{itemize}  
        \paragraph{Exponential spectral decay of $\left(T_{k^m}\right)_{m=1}^M$-s.} We assume now that the eigenvalues of $T_{k^m}$ decay exponentially; furthermore, we suppose that 
        \begin{itemize}
        \item[] \begin{assumption}[eigenvalue decay]\label{eq:exp-decay:lambda^m-decay}
            There exists an $\tilde{i}_0^m \in \Zp$ such that for all $i \ge \tilde{i}_0^m$
            \begin{align}
            \lambda^m_i &= \tilde{c}^m(i) e^{- \tau_m i}, 
            \end{align}
            for $\tau_m >0$, where $\tilde{c}^m(i) \in \left[ \tilde{c}^m_l, \tilde{c}^m_u \right]$ for some constants $0 < \tilde{c}^m_l < \tilde{c}^m_u$.
        \end{assumption}
        \item[]\begin{assumption}[convergence]\label{eq:exp-decay:c(i)_assumption1}
            $\exists \lim_{i \to \infty} \tilde{c}^m(i) \eqcolon \bar{c}^m_2$.\footnote{We note that $\bar{c}^m_2>0$ by combining $0 < \tilde{c}^m_l < \tilde{c}^m(i)$ and $\bar{c}^m_2 = \lim_{i \to \infty} \tilde{c}^m(i)$.\label{footnote:c2m>0}}
        \end{assumption}
        \end{itemize}
        We also suppose that the embedding dimension is chosen as 
        \begin{align}
            d^m &=\frac{1}{\tau_m}\log \left(n^{\theta_m}\right) +c \label{eq:exp-decay:d^m-choice}
        \end{align}
        with $\theta_m >0$ and $c\ge 0$.

        \begin{itemize}[labelindent=0em,leftmargin=*,topsep=0cm,partopsep=0cm,parsep=0cm,itemsep=0cm]
            \item Term $U_1$: The first term is $U_1 = \O\left(n^{-\frac{1}{2}}\right)$.

            \item Term $U_{2,m}$ (with $m \in [M]$ fixed): The order of this term $U_{2,m} = \O\left(n^{-\frac{1}{2}}\right)$.
            
            \item Term $U_{3,m}$ (with $m\in [M]$ fixed):  
            \begin{itemize}
                \item Term $A_m$:
                By using Lemma~\ref{eigenvalue_decay}, under the imposed exponential spectral decay $\mathcal{N}_{C_{k^m}}\left(t_{k^m}\left(n,\frac{\eta}{8M}\right)\right) = \Theta \left( \log \left(\frac{1}{t_{k^m}\left(n,\frac{\eta}{8M}\right)} \right) \right) = \Theta \left(\log\left(\frac{n}{\log(n)}\right)\right)$,  $\log\left(\frac{n}{\log(n)}\right) = \O(\log (n))$ leveraging that $t_{k^m}\left(n,\frac{\eta}{8M}\right) = \Theta \left(\frac{\log(n)}{n}\right)$.  By Assumption~\ref{eq:exp-decay:lambda^m-decay} and \eqref{eq:exp-decay:d^m-choice}, one has  $\lambda^m_{d^m+1}=\O\left(n^{-\theta_m}\right)$ and 
            \begin{align}
            A_m = \O\left ( \log^{\frac{1}{2}} (n) \left( \frac{1}{n^{\theta_m}} + \frac{\log(n)}{n} \right)\right ) = \left\{ \begin{array}{cl}
            \O\left ( \frac{\log^{\frac{3}{2}}(n)}{n}\right ) ,&  \ \theta_m \geq 1 \\
            \O\left ( \frac{\log^{\frac{1}{2}} (n)}{n^{\theta_m}} \right ) ,&  \ \theta_m < 1
            \end{array} \right. .
            \end{align}
            In both cases, $A_m \stackrel{n \to \infty}{\longrightarrow} 0$. 
            
                \item Term $B_m$: To obtain the upper bound for $B_m$, it is sufficient to upper bound $\left( \delta_{d^m}^m \right)^{-1}$, which is equivalent to deriving a lower bound on the eigengap $\delta_{d^m}^m = \lambda_{d^m}^m - \lambda_{d^m+1}^m$. Let $\epsilon \in \left(0, \bar{c}^m_2 \frac{e^{\tau_m} -1}{e^{\tau_m}+1} \right)$.\footnote{We note that $\bar{c}^m_2 \frac{e^{\tau_m} -1}{e^{\tau_m}+1}>0$ (hence one can choose the prescribed $\epsilon$) by (i) $\bar{c}_2^m>0$ (following from footnote~\ref{footnote:c2m>0}), (ii) $e^{\tau_m}+1> e^{\tau_m}>0$ (by the positivity of the exponential function), (iii) $e^{\tau_m} -1 >0 \Leftrightarrow e^{\tau_m} > 1 \Leftrightarrow \tau_{m}>0$ which holds by assumption.} By the convergence of the sequence $\tilde{c}^m(i)$, there exists a $\tilde{i}_1^m = \tilde{i}_1^m(\epsilon) \in \Zp$ such that for all $i \ge \tilde{i}_1^m$, one has 
                \begin{align}
                    \left| \tilde{c}^m(i) - \bar{c}^m_2 \right| & \le \epsilon, & \left| \tilde{c}^m(i+1) - \bar{c}^m_2 \right| & \le \epsilon. \label{eq:converge_ci_ci+1}
                \end{align}
                Let 
                \begin{align}
                    i \ge \tilde{L}^m := \tilde{L}^m(\epsilon) := \max \{ \tilde{i}_0^m, \tilde{i}_1^m(\epsilon) \}
                \end{align}
                be arbitrary but fixed. According to Assumption~\ref{eq:exp-decay:lambda^m-decay}, we have $\lambda_i^m = \tilde{c}^m(i) e^{-\tau_m i}$. Let $\tilde{C}_i^m = \tilde{c}^m(i)$ and $\tilde{C}_{i+1}^m = \tilde{c}^m(i+1)$, and (by definition) $\tilde{C}_i^m, \tilde{C}_{i+1}^m \in [ \tilde{c}_l^m, \tilde{c}_u^m]$.
                Therefore we have
                \begin{align}
                    \delta_{i}^m &= \lambda_i^m - \lambda_{i+1}^m = \tilde{C}_i^m e^{-\tau_m i} - \tilde{C}_{i+1}^m e^{-\tau_m (i+1)} = \tilde{C}_i^m e^{-\tau_m i} - \tilde{C}_{i+1}^m e^{-\tau_m i - \tau_m} \\
                    & = \tilde{C}_i^m e^{-\tau_m i} - \frac{\tilde{C}_{i+1}^m}{e^{\tau_m}} e^{-\tau_m i} = \left( \tilde{C}_i^m - \frac{\tilde{C}_{i+1}^m}{e^{\tau_m}} \right) e^{-\tau_m i} \\
                    & \stackrel{(a)}{\ge} \left[ \left( \bar{c}^m_2 - \epsilon \right) - \frac{\bar{c}^m_2 + \epsilon}{e^{\tau_m}} \right] e^{-\tau_m i} = \left[ \bar{c}^m_2 \left( 1-\frac{1}{e^{\tau_m}} \right) - \epsilon \left( 1+\frac{1}{e^{\tau_m}} \right) \right] e^{-\tau_m i}, \label{eq:exp_delta_i}
                \end{align}
                where (a) comes from \eqref{eq:converge_ci_ci+1} implying that $\tilde{C}_i^m \ge \bar{c}^m_2 - \epsilon $ and $\tilde{C}_{i+1}^m \le \bar{c}^m_2 + \epsilon $ with substitutions $\tilde{c}^m(i) \leftarrow \tilde{C}_i^m$ and $\tilde{c}^m(i+1) \leftarrow \tilde{C}_{i+1}^m$. For $\tau_m >0$ we have $e^{\tau_m} > 1$ therefore $\frac{1}{e^{\tau_m}} < 1$, this implies the positivity of $\left( 1-\frac{1}{e^{\tau_m}} \right)$. Since $\epsilon < \bar{c}^m_2 \frac{e^{\tau_m} -1}{e^{\tau_m}+1}$ (by definition), we have
                \begin{align}
                    \bar{c}^m_2 \left( 1-\frac{1}{e^{\tau_m}} \right) - \epsilon \left( 1+\frac{1}{e^{\tau_m}} \right) &> \bar{c}^m_2 \Bigg( \underbrace{1-\frac{1}{e^{\tau_m}}}_{\frac{e^{\tau_m}-1}{e^{\tau_m}} } \Bigg) - \bar{c}^m_2 \frac{e^{\tau_m} -1}{e^{\tau_m}+1} \Bigg(\underbrace{1+\frac{1}{e^{\tau_m}}}_{\frac{e^{\tau_m}+1}{e^{\tau_m} }} \Bigg)\\
                    &=0. \label{eq:coeff>0}
                \end{align}
                Therefore, \eqref{eq:exp_delta_i} holds for all $i \ge \tilde{L}^m$. Consequently, $\delta_i^m = \Omega \left( e^{-\tau_m i} \right)$ according to \eqref{eq:exp_delta_i} and \eqref{eq:coeff>0}. In the  asymptotic order the additive constant $c$ in \eqref{eq:exp-decay:d^m-choice} is not relevant, hence we set $c=0$: 
                \begin{align}
                    d^m = \frac{1}{\tau_m} \log \left( n^{\theta_m} \right). \label{eq:exp-decay:d^m-choice1}
                \end{align}
                Substituting $i \leftarrow d^m$, with $d^m$ defined in \eqref{eq:exp-decay:d^m-choice1}, into $\delta_i^m = \Omega \left( e^{-\tau_m i} \right)$, we obtained that 
                \begin{align}
                    \delta_{d^m}^m = \Omega \left( n^{-\theta_m} \right) . \label{eq:lower-bound-delta_exp}
                \end{align}
                Therefore we have\footnote{Throughout the analysis we also keep track of $\tau_m$ to be able to discuss its role.}
                \begin{itemize}
                    \item Term $B_{m,1}$: 
                    \begin{align}
                    B_{m,1} & \stackrel{(a)}{=} 312 \sqrt{2} c_1^m(n) d^m \left( \delta^m_{d^m} \right)^{-\frac{1}{2}}  \sqrt{\frac{\log \left ( \frac{864 n}{\eta} \right ) }{n}} \\
                    &\stackrel{(b)}{=} \O\left(\log(\log(n))  \frac{\log\left(n^{\theta_m}\right)}{\tau_m} n^{\frac{\theta_m}{2}}  \sqrt{\frac{\log(n)}{n}}\right) \\
                    & = \O\left(\frac{1}{\tau_m} \log(\log(n)) n^{\frac{\theta_m}{2} - \frac{1}{2}} \log^{\frac{3}{2}}(n) \right),
                    \end{align}
                    where (a) is implied by the definition of $B_{m,1}$; (b) follows from the fact that $c_1^m(n) = \O \left( \log \left( \log(n) \right) \right)$, $d^m = \frac{1}{\tau_m} \log \left( n^{\theta_m} \right)$ [by \eqref{eq:exp-decay:d^m-choice1}] and $\left( \delta^m_{d^m} \right)^{-\frac{1}{2}} = \O \left( n^{\frac{\theta_m}{2}} \right)$ [by \eqref{eq:lower-bound-delta_exp}]. This means that $B_{m,1} \xrightarrow{n \to \infty} 0$ iff.\ $\frac{\theta_m}{2} - \frac{1}{2} < 0$, in other words $\theta_m < 1$, hence in $A_m$ the lower branch ($\theta_m < 1$) will be active.

                    \item Term $B_{m,2}$: 
                    \begin{align}
                    B_{m,2} & \stackrel{(a)}{=} 896 \frac{\left(d^m \right)^{\frac{1}{2}} \left(c_1^m(n) \right)^2}{\sqrt{c_0^m}} \left( \delta^m_{d^m} \right)^{-2} \frac{\log \left( \frac{432 n}{\eta} \right)}{n} \\
                    &\stackrel{(b)}{=} \O \left( \frac{ \log^{\frac{1}{2}}\left(n^{\theta_m}\right)}{\sqrt{\tau_m}} [\log(\log(n))]^2 n^{2\theta_m}\frac{\log (n)}{n} \right) \\
                    & = \O\left( [\log(\log(n))]^2 \frac{\log^{\frac{3}{2}} (n)} {\sqrt{\tau_m} n^{1- 2\theta_m}} \right).
                    \end{align}
                    In (a) we used the definition of $B_{m,2}$; (b) follows from  $d^m = \frac{1}{\tau_m} \log \left( n^{\theta_m} \right)$ [by \eqref{eq:exp-decay:d^m-choice1}], $c_1^m(n)=\O(\log(\log(n)))$, and that  $\left( \delta^m_{d^m} \right)^{-2} = \O \left( n^{2\theta_m} \right)$ [by \eqref{eq:lower-bound-delta_exp}].
                    This means that $B_{m,2} \xrightarrow{n \to \infty} 0$ iff.\ $0 < 1 - 2\theta_m$, in other words $\theta_m < \frac{1}{2}$, hence in $A_m$ the lower branch ($\theta_m < 1$) will be active.

                    \item Term $B_{m,3}$: 
                    \begin{align}
                    B_{m,3} & \stackrel{(a)}{=} 1024 \frac{\left(d^m \right)^{\frac{1}{2}}\left(c_1^m(n) \right)^3}{\sqrt{c_0^m}} \left( \delta^m_{d^m} \right)^{-2} \frac{\log^{\frac{3}{2}} \left( \frac{432 n}{\eta} \right)}{n^{\frac{3}{2}}} \\
                    &\stackrel{(b)}{=} \O \left( \frac{\log^{\frac{1}{2}}\left(n^{\theta_m}\right)}{\sqrt{\tau_m}} \left[\log(\log(n))\right]^3 n^{2\theta_m}\frac{\log^{\frac{3}{2}} (n)}{ n^{\frac{3}{2}}} \right) \\
                    &= \O\left( [\log(\log(n))]^3  \frac{\log^2 (n)} {\sqrt{\tau_m} n^{\frac{3}{2}- 2\theta_m}} \right).
                    \end{align}
                    In (a) we used the definition of $B_{m,3}$; (b) follows from  $d^m = \frac{1}{\tau_m} \log \left( n^{\theta_m} \right)$ [by \eqref{eq:exp-decay:d^m-choice1}], $c_1^m(n)=\O(\log(\log(n)))$, and that  $\left( \delta^m_{d^m} \right)^{-2} = \O \left( n^{2\theta_m} \right)$ [by \eqref{eq:lower-bound-delta_exp}].
                    This means that $B_{m,3} \xrightarrow{n \to \infty} 0$ iff.\ $0 < \frac{3}{2} - 2\theta_m$, in other words $\theta_m < \frac{3}{4}$, hence in $A_m$ the lower branch ($\theta_m < 1$) will be active.

                    \item Term $B_{m,4}$: 
                    \begin{align}
                    B_{m,4} & \stackrel{(a)}{=} 64 \frac{\left(c_1^m(n) \right)^2}{\sqrt{c_0^m}} \left( \delta^m_{d^m} \right)^{-1} \frac{\left(d^m \right)^{\frac{1}{2}} \log \left( \frac{432n}{\eta} \right)}{n} 
                    \\ 
                    &\stackrel{(b)}{=} \O \left( [\log(\log(n))]^2 n^{\theta_m}\frac{\log^{\frac{1}{2}}\left(n^{\theta_m}\right) \log (n)}{\sqrt{\tau_m} n} \right) \\
                    &= \O\left( [\log(\log(n))]^2 \frac{\log^{\frac{3}{2}}(n)} {\sqrt{\tau_m} n^{1- \theta_m}} \right), 
                    \end{align}
                    where (a) is implied by the definition of $B_{m,4}$; (b) follows from the fact that $c_1^m(n) = \O \left( \log \left( \log(n) \right) \right)$, $d^m = \frac{1}{\tau_m} \log \left( n^{\theta_m} \right)$ [by \eqref{eq:exp-decay:d^m-choice1}] and $\left( \delta^m_{d^m} \right)^{-1} = \O \left( n^{\theta_m} \right)$ [by \eqref{eq:lower-bound-delta_exp}].
                    This means that $B_{m,4} \xrightarrow{n \to \infty} 0$ iff.\ $0 < 1 - \theta_m$, in other words $\theta_m < 1$, hence in $A_m$ the lower branch ($\theta_m < 1$) will be active.

                    \item Term $B_{m,5}$: 
                    \begin{align}
                    B_{m,5} & \stackrel{(a)}{=} 4 \frac{(d^m)^{\frac{1}{2}} c_1^m(n)}{\sqrt{c_0^m}} \sqrt{\frac{\log \left( \frac{144n}{\eta}  \right)}{n}} \stackrel{(b)}{=} \O \left( \frac{\log^{\frac{1}{2}}\left(n^{\theta_m}\right)}{\sqrt{\tau_m}} \log(\log(n)) \sqrt{\frac{\log (n)}{n} } \right) \\
                    & = \O \left( \frac{1}{\sqrt{\tau_m}} \frac{\log(\log(n)) \log(n)}{\sqrt{n}} \right).
                    \end{align}
                    In (a) we used the definition of $B_{m,5}$; (b) follows from $d^m = \frac{1}{\tau_m} \log \left( n^{\theta_m} \right)$ [by \eqref{eq:exp-decay:d^m-choice1}] and $c_1^m(n)=\O(\log(\log(n)))$. We have $B_{m,5} \xrightarrow{n \to \infty} 0$.
                \end{itemize}
            Combining these results, the final range of $\theta_m$ should be $0 < \theta_m < \min\left\{1,\frac{1}{2},\frac{3}{4},1\right\}=\frac{1}{2}$, and we can conclude that the lower branch of $A_m$ $(\theta_m <1)$ will be activated. Before combining $A_m$ and $B_{m}$, we first compare the terms of $B_m$.
            \begin{itemize} 
                \item Comparing $B_{m,1}$ and $B_{m,2}$: 
                \begin{itemize}
                \item $B_{m,1}$ converges faster than $B_{m,2}$: This holds by using that $\log(\log(n)) \log^{3/2}(n) = \O\left([\log(\log(n))]^2 \log^{3/2}(n)\right)$ [hence the question boils down to the comparison of the polynomial terms] iff.      \begin{align}\label{con:exp_Bm1_faster}
                    \frac{1}{2} - \frac{\theta_m}{2} \ge  1- 2\theta_m 
                    \Leftrightarrow 
                    \frac{3}{2}\theta_m \ge \frac{1}{2}\Leftrightarrow \theta_m \ge \frac{1}{3}.
                \end{align}
                \item $B_{m,1}$ converges slower than $B_{m,2}$:
                This holds iff.\             \begin{align}\label{con:exp_Bm2_faster}
                    \theta_m < \frac{1}{3}.
                \end{align}
                \end{itemize}    
                \item Comparing $B_{m,2}$ and $B_{m,3}$: one has that $\frac{3}{2}- 2\theta_m > 1- 2\theta_m \Leftrightarrow \frac{3}{2}>1$ holds, therefore $B_{m,3}$ converges faster than $B_{m,2}$.
                
                \item Comparing $B_{m,2}$ and $B_{m,4}$: one has  that $1- \theta_m > 1- 2\theta_m \Leftrightarrow \theta_m>0$ holds since $\theta_m > 0$, therefore $B_{m,4}$ converges faster than $B_{m,2}$.

                \item Compare $B_{m,1}$ and $B_{m,5}$: one has that $1 > 1- \theta_m \Leftrightarrow \theta_m > 0$ holds since $\theta_m > 0$, therefore $B_{m,5}$ converges faster than $B_{m,1}$.
            \end{itemize}
            \item Order of the term $U_{3,m}$: By collecting the terms in $U_{3,m}$, one arrives at  
            \begin{align}
            U_{3,m}=A_m + B_{m,1} + B_{m,2} + B_{m,3} + B_{m,4} + B_{m,5}. 
            \end{align}
            Since $B_{m,3}$, $B_{m,4}$ and $B_{m,5}$ converge faster than $B_{m,1}$ or $B_{m,2}$, the final convergence rate is determined $A_m + B_{m,1}$ or $A_m + B_{m,2}$. We therefore discuss these two cases in the following.
            \begin{itemize}
                \item \tb{Case of \eqref{con:exp_Bm2_faster} holding}: In this case the order is that of $A_m + B_{m,1}$, for which we have
                \begin{align}
                    A_m + B_{m,1} = \O\Bigg( \underbrace{\frac{\log^{\frac{1}{2}} (n)}{n^{\theta_m}}}_{=:a_{m}(n)} \Bigg) + \O\Bigg(\underbrace{\frac{1}{\tau_m} \frac{\log(\log(n)) \log^{\frac{3}{2}}(n)}{n^{\frac{1}{2}-\frac{\theta_m}{2}}}}_{=:b_{m,1}(n)} \Bigg)
                \end{align}
                under the condition that $0<\theta_m < \min \left(\frac{1}{2},\frac{1}{3}\right) =\frac{1}{3}$. We have 
                \begin{align}
                 a_m(n) &= \Omega \left( \frac{1}{n^{\frac{1}{3}}} \right),&   
                 b_{m,1}(n) &= \Omega\left( \frac{1}{n^{\frac{1}{2}}} \right) 
                 \label{eq:exp:bm1(n):order}
                \end{align}
                as $\theta_m \in \left(0,\frac{1}{3}\right)$. With $\theta_m$ increasing, $a_m(n)$ converges to zero faster and $b_{m,1}(n)$ goes to zero slower as $n\to \infty$.
                Matching the orders of the (polynomial) terms in $a_m(n)$ and $b_{m,1}(n)$, one gets that 
                \begin{align}
                    \theta_m = \frac{1}{2} - \frac{\theta_m}{2} \Leftrightarrow 2\theta_m = 1 - \theta_m \Leftrightarrow  \theta_m = \frac{1}{3}.
                \end{align}
                 With the choice $\theta_m \to \frac{1}{3}$ (allowed by $\theta_m \in \left(0,\frac{1}{3}\right)$), one gets that  the embedding dimension can be arbitrary close to
                \begin{align}
                    d^m &=\frac{1}{\tau_m} \log \left(n^{\frac{1}{3}}\right) +c,
                \end{align}
                and the order of the obtained upper bound on $U_{3,m}$ can be arbitrarily close to $\O\left (\frac{1}{\tau_m} \log(\log(n))\frac{\log^{\frac{3}{2}}(n)}{n^{\frac{1}{3}}} \right )$ using that $\log^{\frac{1}{2}}(n)=\O\left(\log(\log(n)) \log^{\frac{3}{2}}(n)\right)$.
            \item \tb{Case of \eqref {con:exp_Bm1_faster} holding}:
            In this case the order is that of $A_m+B_{m,2}$, for which we have 
            \begin{align}
            A_m + B_{m,2} & = \O\Bigg( \underbrace{\frac{\log^{\frac{1}{2}} (n)}{n^{\theta_m}}}_{=a_{m}(n)} \Bigg) +  \O\Bigg( \underbrace{ [\log(\log(n))]^2 \frac{\log^{\frac{3}{2}} (n)}{\sqrt{\tau_m} n^{1- 2\theta_m}}}_{=:b_{m,2}(n)} \Bigg)
            \end{align}
            under the condition that $\frac{1}{3}\le \theta_m < \frac{1}{2}$. We have 
            \begin{align}
            a_m(n) &= \Omega\left(\frac{1}{n^{\frac{1}{2}}}\right),&
            b_{m,2}(n) &= \Omega\left(\frac{1}{n^{1-2\frac{1}{3}}}\right) = \Omega\left(\frac{1}{n^{\frac{1}{3}}}\right).
            \label{eq:exp:bm2(n):order}
            \end{align}
            With $\theta_m$ increasing, $a_m(n)$ converges to zero faster and $b_{m,2}(n)$ goes to zero slower as $n\to \infty$. Matching the orders of the (polynomial) terms in $a_m(n)$ and $b_{m,2}(n)$, one arrives at
            \begin{align}
            \theta_m = 1-2\theta_m \Leftrightarrow \theta_m = \frac{1}{3}.
            \end{align}
            With this choice of $\theta_m = \frac{1}{3}$, one gets the embedding dimension is 
            \begin{align}
                    d^m &=\frac{1}{\tau_m} \log \left(n^{\frac{1}{3}}\right) +c,
                \end{align}
            and
            \begin{align}
            A_m + B_{m,2} & = \O\left([\log(\log(n))]^2 \frac{\log^{\frac{3}{2}}(n)} {\sqrt{\tau_m} n^{\frac{1}{3}}}\right)
            \end{align}
            using that $\log^{\frac{1}{2}}(n)=\O\left([\log(\log(n))]^2 \log^{\frac{3}{2}}(n)\right)$.                
            \end{itemize} 

        \end{itemize}
                                
        \item Final rate: By \eqref{eq:U1-U2m-U3m}, one can see that the order of the error is that of $U_1 + \sum_{m\in [M]}U_{2,m} + \sum_{m\in [M]}U_{3,m}$, which given the shown bounds is the same order as $\sum_{m\in [M]}U_{3,m}$ by (i) $U_1 = \O\left(n^{-1/2}\right)$, $U_{2,m} = \O\left(n^{-1/2}\right)$, and
        \begin{align}
        U_{3,m}  & = 
            \begin{cases}
              \Theta\left(A_m+B_{m,1}\right), \quad A_m+B_{m,1} = \O(b_{m,1}(n)), \quad b_{m,1}(n)=\Omega\left(n^{-1/2}\right)      & \t{ if }\theta_m \to \frac{1}{3}-,\\
              \Theta\left(A_m+B_{m,2}\right), \quad A_m+B_{m,2} = \O(b_{m,2}(n)), \quad b_{m,2}(n)=\Omega\left(n^{-1/3}\right)      & \t{ if }\theta_m = \frac{1}{3}
            \end{cases}
        \end{align}
        following from \eqref{eq:exp:bm1(n):order} and \eqref{eq:exp:bm2(n):order}. Combining the results obtained for $(U_{3,m})_{m=1}^M$, one obtains that the error of the GHSIC estimator is 
        \begin{align}
        \begin{cases}
         \t{arbitrarily close to } \O\left (\frac{1}{\tau_m} \log(\log(n))\frac{\log^{\frac{3}{2}}(n)}{n^{\frac{1}{3}}} \right )& \theta_m \to \frac{1}{3}-,\\
         \O\left(\frac{1}{\sqrt{\tau_m}}\frac{[\log(\log(n))]^2 \log^{\frac{3}{2}}(n)}{n^{\frac{1}{3}}}\right)& \t{ if }\theta_m = \frac{1}{3}.
        \end{cases}
        \end{align} 
        This means that for $\theta_m = 1/3$ we got that the GHSIC estimation error is
        \begin{align}
        \O\left(\frac{1}{\sqrt{\tau_m}}\frac{[\log(\log(n))]^2 \log^{\frac{3}{2}}(n)}{n^{\frac{1}{3}}}\right) = \tilde{\O}\left(\frac{1}{\sqrt{\tau_m}}\frac{1}{n^{\frac{1}{3}}}\right).
        \end{align}
        
        \end{itemize}

\graphicspath{{figs/}}
	\begin{figure}[htbp!]
		\scriptsize
		\begin{center}
			\begin{tabular}{c}
				\includegraphics[width=0.45\textwidth]{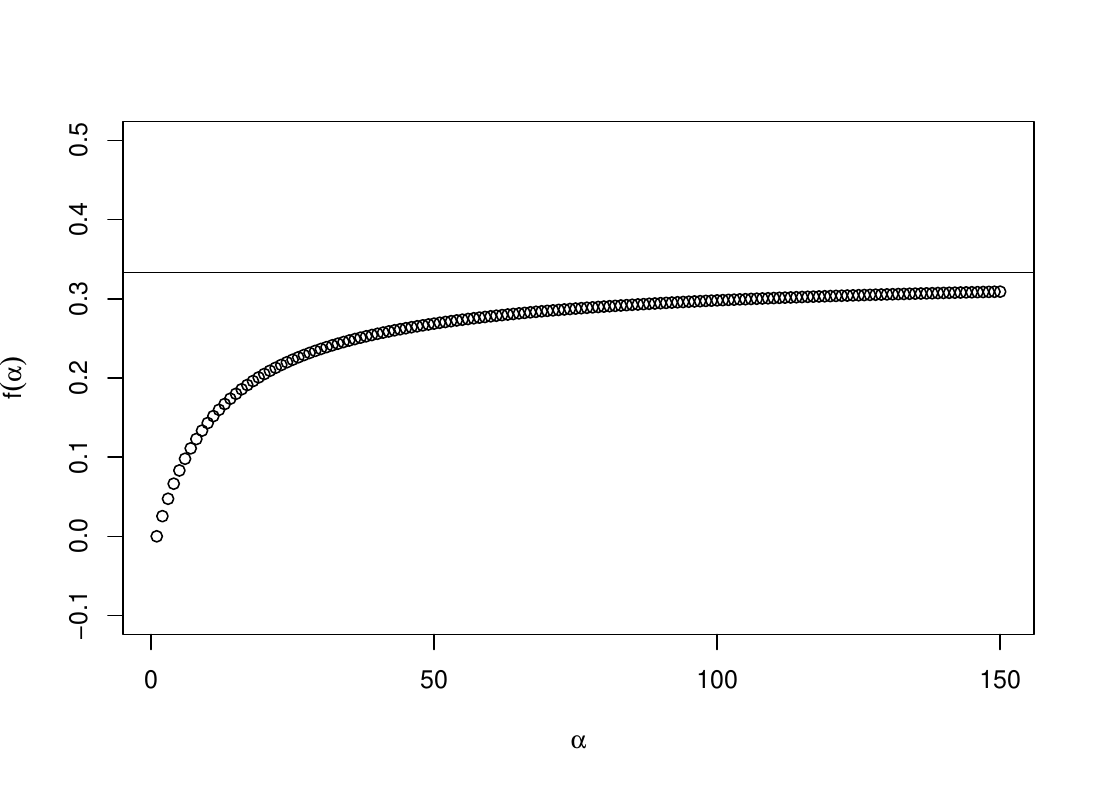}
				\\
			\end{tabular}
		\end{center}
		\caption{Illustration of $f$ defined in \eqref{eq:f:rate}. The horizontal line is $y=\frac{1}{3}$.}
		\label{fig2_alpha}
\end{figure}

\bibliography{BIB/collected,BIB/collected_plus,BIB/collected_plusZ,BIB/collected_own}
\end{document}